\documentclass[%
aps,
prx,
twocolumn,
superscriptaddress,
longbibliography,
10pt,
amsfonts,
amssymb,
amsmath,
floatfix]{revtex4-2}
\usepackage[
    colorlinks=true,
    linkcolor=blue!90!black,
    filecolor=magenta!90!black,      
    urlcolor=magenta!90!black,
    citecolor = green!60!black
    ]{hyperref}
\usepackage{textcomp}
\usepackage[autostyle]{csquotes}
\usepackage{amsthm}
\usepackage{bm}
\usepackage{footnote}
\usepackage{graphicx}
\usepackage{notations}
\usepackage{subfiles}
\usepackage{mathtools}
\usepackage{pgfplots}
\usepackage{lipsum}
\usepackage{enumitem}
\pgfplotsset{/pgf/number format/use comma,compat=newest}
\usepackage{url}
\usepackage{tikz}
\usepackage{bbm}
\usepackage{floatflt}
\usepackage{soul} 
\usetikzlibrary{arrows,decorations.pathmorphing,backgrounds,positioning,fit,petri}
\usepackage{comment}
\usepgfplotslibrary{fillbetween}
\usepackage{booktabs}
\usepackage{capt-of}

\usepackage{float}
\makeatletter
\let\newfloat\newfloat@ltx
\makeatother

\usepackage{algorithm,algorithmic}

\newenvironment{talign}
 {\align}
 {\endalign}
\newenvironment{talign*}
 {\csname align*\endcsname}
 {\endalign}

\makeatletter
\newenvironment{sqcases}{%
  \matrix@check\sqcases\env@sqcases
}{%
  \endarray\right.%
}
\def\env@sqcases{%
  \let\@ifnextchar\new@ifnextchar
  \left\lbrack
  \def\arraystretch{1.2}%
  \array{@{}l@{\quad}l@{}}%
}
\makeatother

\newtheorem{theorem}{Theorem}

\newtheorem{proposition}{Proposition}
\newtheorem{result}[theorem]{Result}
\newtheorem{assumption}{Assumption}

\theoremstyle{remark} 
\newtheorem{remark}{\bf Remark}

\begin{document}
\title{\texorpdfstring{Statistical physics of deep learning:\\ Optimal learning of a multi-layer perceptron near interpolation}{Statistical physics of deep learning: Optimal learning of a multi-layer perceptron near interpolation}}

\author{Jean Barbier}
\affiliation{The Abdus Salam International Centre for Theoretical Physics\\Strada Costiera 11, 34151 Trieste, Italy}

\author{Francesco Camilli}
\affiliation{Alma Mater Studiorum - Università di Bologna, Dipartimento di Matematica\\Piazza di Porta S. Donato 5, 40126 Bologna, Italy}

\author{Minh-Toan Nguyen}
\affiliation{The Abdus Salam International Centre for Theoretical Physics\\Strada Costiera 11, 34151 Trieste, Italy}

\author{Mauro Pastore}

\affiliation{The Abdus Salam International Centre for Theoretical Physics\\Strada Costiera 11, 34151 Trieste, Italy}

\author{Rudy Skerk}
\thanks{All authors contributed equally and names are ordered alphabetically. R. Skerk and M.-T. Nguyen have carried out the numerical experiments, in addition to theoretical work. Corresponding authors: \url{rskerk@sissa.it}, \url{jbarbier@ictp.it}}

\affiliation{International School for Advanced
Studies\\ Via Bonomea 265, 34136 Trieste, Italy}


\begin{abstract}
For four decades statistical physics has been providing a framework to analyse neural
networks. A long-standing question remained on its capacity to tackle deep learning models capturing rich feature learning effects, thus going beyond the narrow networks or kernel methods analysed until now. We positively answer through the study of the supervised learning of a multi-layer perceptron.
Importantly, $(i)$ its width scales as the input dimension, making it more prone to feature learning than ultra wide networks, and more expressive than narrow ones or ones with fixed embedding layers; and $(ii)$ we focus on the challenging interpolation regime where the number of trainable parameters and data are comparable, which forces the model to adapt to the task. We consider the matched teacher-student setting. Therefore, we provide the fundamental limits of learning random deep neural network targets and identify the sufficient statistics describing what is learnt by an optimally trained network as the data budget increases. A rich phenomenology emerges with various learning
transitions. With enough data, optimal performance is attained through the model's ``specialisation'' towards the target, but it can be hard to reach for training algorithms which get attracted by sub-optimal solutions predicted by the theory. Specialisation occurs inhomogeneously across layers, propagating from shallow towards deep ones, but also across neurons in each layer. Furthermore, deeper targets are harder to learn. Despite its simplicity, the Bayes-optimal setting provides insights on how the depth, non-linearity and finite (proportional) width influence neural networks in the feature learning regime that are potentially relevant in much more general settings.

\end{abstract}

\maketitle


\section{Introduction}
Neural networks (NNs) are the powerhouse of modern machine learning, with applications in all fields of science and technology. Their use is now widespread in society much beyond the scientific realm. Understanding their expressive power and generalisation capabilities is therefore not only a stimulating intellectual activity, producing surprising results that seem to defy established common sense in statistics and optimisation~\citep{bartlett2021}, but is also of major practical and economic importance. 

One issue is that even the models dating back to the inception of deep learning \cite{lecun2015deep} are not theoretically well understood when operating in the ``feature learning regime'' (a task-dependent term that will be clear later). The simplest deep learning model is the multi-layer fully connected feed-forward neural network, also called \emph{multi-layer perceptron} (MLP). It corresponds to a function  $\mathcal{F}_\btheta(\bx)=\bv^\intercal \sigma(\bW^{(L)} \sigma(\bW^{(L-1)}\cdots \sigma(\bW^{(1)}\bx)\cdots)$ going from $\mathbb{R}^d$ to $\mathbb{R}$, parametrised by $L+1$ matrices $\btheta=(\bv\in \mathbb{R}^{k_L\times 1},(\bW^{(l)}\in \mathbb{R}^{k_l\times k_{l-1}})_{l\le L})$, with $k_l$ denoting the width of the $l$-th hidden layer (with $k_0=d$), and an activation function $\sigma(\,\cdot\,)$ applied entrywise to vectors. 

Until now, the quantitative theories for such NNs predicting which relevant features they can extract, how much data $n$ they need to do so and how well they generalise beyond their training data, relied on over-simplified architectural and/or data-abundance assumptions. This prevented to precisely capture the combined role of the depth and non-linearity of NNs when they are trained from sufficiently many data for them to fully express their representation power. This paper offers answers to these questions in a richer scenario than what current statistical approaches could tackle.

\subsection{A pit in the neural networks landscape}

\begin{figure*}[ttt]
    \centering    
    
    \vspace{-10pt}\includegraphics[width=\textwidth,trim={0 18.7cm 0.5cm 0cm},clip]{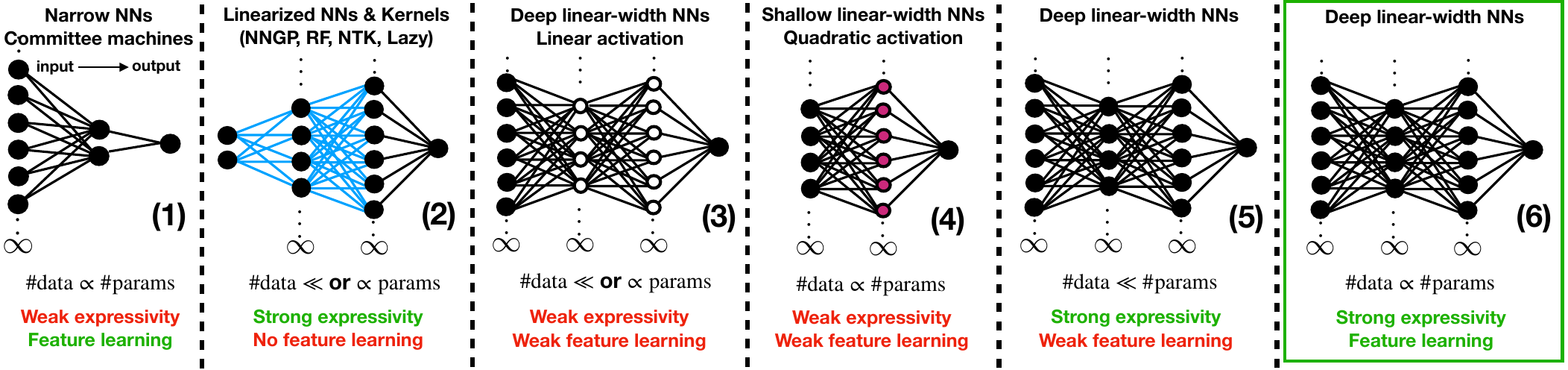}
    \caption{Classification of models of fully connected feed-forward neural networks analysed in the theoretical literature (see the main text for references). Class $\bf{(1)}$ models are very narrow, i.e., with a width independent of the large input dimension $d$. This includes the perceptron and committee machines studied in the statistical mechanics literature (the latter are linked to the so-called multi-index models). They are analysed near-interpolation, where the number of data and model's parameters are proportional. In this regime, feature learning emerges through phase transitions but these models suffer from their limited expressivity. $\bf{(2)}$ encompasses all ``kernel-like models'' whose inner weights are frozen to random values (which is represented by the blue colour) either by construction as in the random feature model, or as a consequence of their overwhelming width/overparametrisation (as in neural network Gaussian processes, or in gradient-based dynamics in the lazy regime, where the weights effectively remain at initialisation and the networks behave as neural tangent kernels). These models are expressive but do not learn task-relevant features due to their effectively frozen embedding layers. However, with readout weights (black last layer) to be $O(1/d)$ rather than the standard scaling $O(1/\sqrt{d})$, feature learning emerges despite the width being infinite. Another tractable simplification is $\bf{(3)}$ deep linear networks, where the weights are learnable but activation functions linear (white circled nodes), thus reducing expressivity to that of a linear model, which allows for limited feature learning. A recent simplification is the linear-width shallow network with quadratic activation (pink nodes), $\bf{(4)}$. Even if trained near interpolation, it can only learn a quadratic approximation of the target, which limits its expressivity. Moreover, we will see that in the teacher-student setting we consider, it cannot recover the target weights, thus preventing strong feature-learning in this sense. Models $\bf{(5)}$ are the same as studied here, but set in the strongly overparametrised ``proportional regime'': with a sample size only scaling as the width. Yet, weak forms of feature learning can emerge: in the teacher-student setting as used in this paper, by ``weak feature learning'' we mean regimes where feature learning does occur but the model's inner weights have zero overlap with the teacher's one. The present paper considers fully trainable proportional-width non-linear NNs trained \emph{near interpolation}, $\bf{(6)}$. This is the most challenging regime, where the model expressivity can fully manifest via strong task-adaptation (i.e., recovery of the target).}
    \label{fig:NNmodels}
    \vskip -0.1in
\end{figure*}
Given the difficulty of the theoretical analysis of NNs, a zoology of tractable simplifications reported here has emerged, each coming with pros and cons.

\vspace{5pt}\paragraph*{\textbf{(1)} Narrow networks.} Triggered by pioneering works employing spin glass techniques to study NNs \cite{amit1985spin,gardner1988,gardner1989teacher}, the interest of the statistical physics community for the equilibrium properties of the narrow \emph{committee machines} ($L=1$ with $k_1=\Theta(1)$ while $n=\Theta(d)\to\infty$, (1) in \figurename~\ref{fig:NNmodels}) rose quickly in the nineties \cite{seung1992query,engel1992storage,kang1993,okane1994,schwarze1993generalization,urbanczik1997,winther1997,schwarze1992generalization,schwarze1992gen2layer,mato1992generalization,monasson1995weight,schottky1995,engel1996,malzahn1997,malzahn1999}. This line of classical works is at the inception of the discovery of learning phase transitions (found concurrently also in single-layer architectures with constrained weights or peculiar activations \cite{sompolinsky1990learning,gyorgyi1990first,meir1992weightConstrained,barbato1993lesions,engel1994reliability,bex1995reversedWedge}). The main issue with narrow NNs is their restricted expressivity. Nevertheless, their analysis yielded important insights on NNs learning mechanisms, some of which are also occurring in more expressive models. One of particular importance is the so-called \emph{specialisation transition}~\cite{barkai1992broken,schwarze1993learning}, where hidden neurons start learning different features. However, as we will see, in more expressive models a richer phenomenology emerges. This field has since then remained very much alive, with the goal of treating more complex architectures as in the present paper, see~\cite{engel2001statistical,cui2025high} for reviews. Narrow NNs are \emph{multi-index functions}, i.e., functions projecting their argument on a low-dimensional subspace, see \cite{bruna2025survey} for a review. Their study allows to understand which properties of non-linearities make learning hard for gradient-based or message-passing algorithms \cite{arous2021online,damian2024computational,abbe2021staircase,abbe2023sgd,troiani2024fundamental}.

\begin{figure}[t!!]
    \centering
    \includegraphics[width=0.945\linewidth]{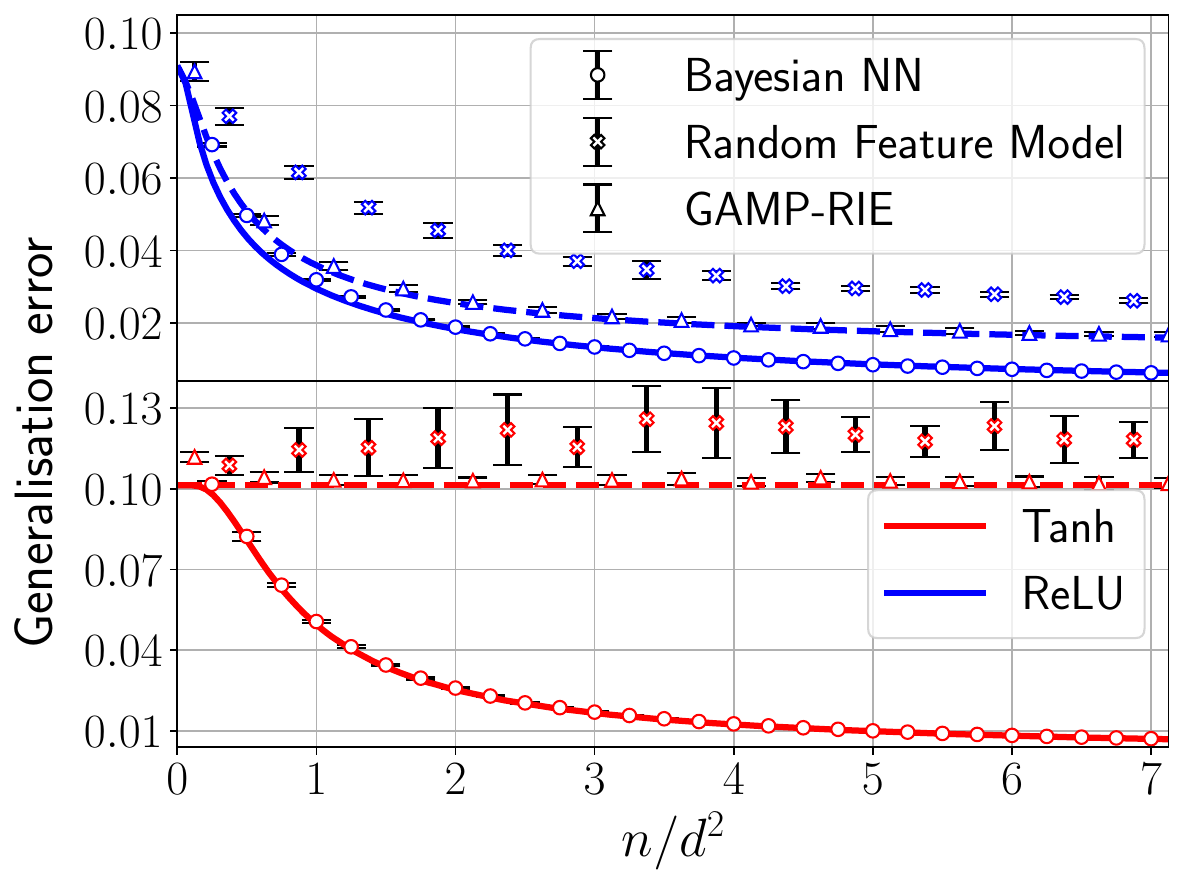}
    \vspace{-7pt}
    
    \caption{Bayes-optimal mean-square generalisation error achievable by a two-layer NN $\mathcal{F}_{\btheta}(\bx)=\bv^\intercal\sigma(\bW \bx)$ as a function of the amount of training data $n$ over the squared input dimension $d^2$ when $d,n$ and the NN width $k$ all diverge with $n/d^2$ and $k/d\to 0.5$ fixed (solid curves), with activation $\sigma(x)={\rm ReLU}(x)$ and $\tanh(2x)$ (same setting as right panel of \figurename~\ref{fig:gen_error_gauss}). These theoretical curves follow from the results in Sec.~\ref{sec:result}. The task is regression with standard Gaussian inputs $(\bx_\mu)_{\mu \le n}$ and noisy responses $(y_\mu)_{\mu \le n}$ generated by a target two-layer NN $\mathcal{F}_{\btheta^0}(\bx)$ with Gaussian random weights and same activation. In the experiments, $d=150$ and $k=75$. Empty circles are obtained by training the Bayesian NN using Hamiltonian Monte Carlo initialised close to the target (yielding the best achievable error), and then computing its generalisation error on $10^5$ test data (error bars are the standard deviation over $10$ instances of the training set and target). Crosses show the generalisation error empirically achievable by the random feature model $\mathcal{F}^{\rm RF}_{\ba^*}(\bx)=\ba^{*\intercal}\sigma(\bW^{\rm RF} \bx)$ trained by exact empirical risk minimisation $\ba^*={\rm argmin} (\sum_{\mu\le n} (y_\mu-
    \mathcal{F}^{\rm RF}_{\ba}(\bx_\mu))^2+t\|\ba\|^2)$ with optimised $\ba\in \mathbb{R}^{r}$ and $L_2$-regularisation strength $t$ picked by cross-validation. The fixed Gaussian features matrix $\bW^{\rm RF}\in\mathbb{R}^{r\times d}$ has width $r=3kd$ roughly three times larger than the total number $k(d+1)$ of parameters of the NN. Triangles are the error of GAMP-RIE \citep{maillard2024bayes} extended to generic activations (see App.~\ref{app:GAMP}), which reaches the performance of an optimally regularised kernel.}
    \label{fig:kernel_vs_NN}
\end{figure}

\vspace{5pt}\paragraph*{\textbf{(2)} Kernel limit: ultra wide and linearised NNs, and the mean-field regime.} On the other hand, in the ultra wide limit ($L$ fixed, $k_l\gg n$) fully connected Bayesian NNs behave as kernel machines (the so-called neural network Gaussian processes, NNGPs) \citep{neal1996,williams1996,lee2018gaussian,matthews2018gaussian,hanin2023infinite}, and hence suffer from these models' limitations. Indeed, kernel machines infer the decision rule by first embedding the data in a fixed a priori feature space, the renowned \emph{kernel trick}, then operating linear regression/classification over the features. In this respect, they do not learn task-relevant features and therefore need larger and larger feature spaces and training sets to fit their higher order statistics~\citep{yoon1998poly,dietrich1999svm,gerace2021,bordelon2021kernel,canatar2021spectral,xiao2022precise}. The same conclusions hold for NNs trained with gradient-based methods but linearised around initialisation \cite{10.1214/20-AOS1990}, i.e., with frozen weights represented in blue in \figurename~\ref{fig:NNmodels}. These include the random feature model (RF) with fixed random inner layer \cite{rahimi2007random} (which is a finite size approximation of kernel machines), or the closely related neural tangent kernel (NTK) \cite{jacot2018NTK} and lazy regimes \cite{chizat2019lazy}, see models (2) in \figurename~\ref{fig:NNmodels}. Such models are thus ``effectively linearised'' because only the readouts are learnt. \figurename~\ref{fig:kernel_vs_NN} illustrates the importance of feature learning: despite having a larger number of parameters, the best RF model or kernel are outperformed by an optimally trained NN, see also \cite{ghorbani2020neural,refinetti2021classifying}. 

One way to probe feature learning effects is through perturbative expansions around the ultra wide limit, where $k\gg n$ but $O(1/k)$ corrections are kept~\cite{dyerasymptotics,yaida2020finite,naveh2021predicting,zavatone2021asymptotics,grosvenor2022edge,fischer2024critical,banta2024structures,guillen2025finite}. This connects to expansions around free fields in quantum field theory, where diagrammatic rules are used to manage complex combinatorial sums, see \cite{Bahri_2024,ringel2025applications} for introductions. Another way to force feature-learning in infinitely wide models is the \emph{mean-field scaling} obtained by taking the readout weights $\bv$ vanishingly small ($O(1/k)$ rather than the standard $O(1/\sqrt k)$ scaling we consider). Originally proposed as a mean to escape the lazy regime of gradient-based dynamics~\citep{chizat2019lazy} via a specific weights initialisation ~\citep{mei2018meanfield,mei2019mean,yang2020feature,rotskoff2022,sirignano2020,bordelon2022,nguyen2023rigorous}, it was later extended to the Bayesian framework~\citep{seroussi2023,bassetti2024,rubin2024unified,vanmeegen2024,lauditi2025adaptive}. NNs in this scaling converge to kernel machines with \emph{data-dependent kernels} (rather than fixed a priori, as in the lazy regime). We also mention \cite{yang2023theory,rubin2024grokking} which alternatively rescale the Bayesian likelihood to induce feature learning at infinite width.

\vspace{5pt}\paragraph*{\textbf{(3)} Deep linear networks.} Another way to linearise networks, thus making them tractable, is by allowing fully trainable weights while placing linear activations in the inner layers. Linear networks are a major theoretical playground from the dynamical perspective \cite{Saxe2014Exact} but also at the equilibrium (Bayesian) level \cite{li2021,aitchison2020bigger,hanin2023bayesian,zavatone2022contrasting,bassetti2024}. In the same vein, theoreticians considered linear diagonal networks of the form $\mathcal{F}^{\rm diag}_{\mathbf{w},\bv}(\bx)=\bv^\intercal {\rm diag}(\mathbf{w})\bx$ \cite{neyshabur2015norm,pesme2023saddle}, which exhibit an implicit bias in gradient descent learning \cite{soudry2018implicit,pesme2021implicit,berthier2023incremental,labarriere2024optimization}. A main issue with linear networks, however, is their intrinsically small expressivity, so only weak notions of feature learning can manifest.

\vspace{5pt}\paragraph*{\textbf{(4)} Shallow quadratic networks.}
Various works have recently exploited the fact that a shallow NN with quadratic activation $\sigma(x)=x^2$ simplifies drastically \cite{du2018power,soltanolkotabi2018theoretical,venturi2019spurious,sarao2020optimization,gamarnik2024stationary,martin2024impact,arjevani2025geometry,maillard2024bayes,xu2025fundamental,erba2025nuclear,arous2025learning}. However, we will see that this prevents strong feature learning from emerging. The closest settings to ours are \cite{maillard2024bayes,xu2025fundamental,erba2025nuclear}. There, the analysis based on results for the GLM \cite{barbier2019glm} and matrix denoising \cite{barbier2022statistical,maillard2022perturbative,matrix_inference_Barbier,semerjian2024} follows thanks to a specific mapping to a \emph{linear matrix sensing problem}, where the goal is to infer a Wishart-like matrix given its projection along random rank-one matrices.

\vspace{5pt}\paragraph*{\textbf{(5)} Proportional data regime.} This overparametrised regime considers a sample size much smaller than the number of model parameters ($L$ fixed, $d$ large, $k_l,n=\Theta(d)$). Recent works show how a limited amount of feature learning makes the network equivalent to optimally regularised kernels \citep{li2021,pacelli2023,baglioni2024,rubin2024unified,ingrosso2025transfer} (see \cite{hanin2026gibbs} for a large depth analysis). MLPs thus reduce to linear networks (GLMs) in the sense conjectured in \cite{cui2023bayes} and proven in \cite{camilli2023fundamental,camilli2025inforeduction} and thus suffer from their limitations. This could be a consequence of the fully connected architecture, as, e.g., convolutional networks can learn more informative features in this regime~\citep{naveh2021,seroussi2023,aiudi2023,bassetti2024}. In a similar data regime, Yoshino and co-authors have developed a replica theory for overparametrised deep NNs with a non-standard architecture \cite{yoshino2020complex,yoshino2023spatially,huang2025liquid}.


\vspace{5pt}\paragraph*{\textbf{(6) A timely challenge:} Deep non-linear networks of linear width trained near interpolation.}
Despite the wealth of methods developed to study the aforementioned models, none is able to tackle NNs enjoying all the following realistic properties:

$(P_1)$ a width proportional to the input dimension;

$(P_2)$ with broad classes of non-linear activations;

$(P_3)$ with possibly multiple hidden layers;

$(P_4)$ learning in the \emph{interpolation regime}.

\noindent The property $(P_1)$ combined with $(P_2),(P_3)$ allows to capture finite-width effects in NNs that are highly expressive, while still allowing to take the large system limit needed to obtain sharp theoretical predictions. Even if it is not entirely clear whether a finite width improves the performance of Bayesian MLPs compared to their kernel limit \cite{yao2022empirical,lee2020finite}, it is certainly one of the most natural ways to allow for the emergence of representation learning, which really is the crux of deep \emph{learning} \cite{lecun2015deep}. 

The interpolation regime $(P_4)$ means $L$ fixed, $d$ large with $k_l=\Theta(d)$ (from $(P_1)$) and $n=\Theta(d^2)$, i.e., a sample size comparable to the number of trainable parameters. 
This regime is difficult to analyse for expressive models but also very interesting, because it forces them to adapt to the data in order to perform well. Hence, task-dependent feature learning emerges, thus escaping the reduction to linear models discussed above. Analysing MLPs in the interpolation regime has been an open problem for decades, and is widely recognised as one of the major theoretical challenges in the physics of learning \cite{cui2023bayes,maillard2024bayes}. That statistical mechanics is up to the task is an encouraging signal for physicists working on deep learning \cite{zdeborova2020understanding,bahri2020statistical}.

This setting is relevant and timely also from a practical perspective. Indeed, the latest NN architectures such as generative diffusion and large language models (LLMs) do operate near interpolation: the compute-optimal training scales parameters and tokens in equal proportion \cite{hoffmann2022empirical,ni2025training}, with typical sizes ranging in $10^{10}$--$10^{12}$.
These models of utmost interest are
highly expressive and also exhibit signs of feature learning \cite{lan2024sparse,li2025find}. From that perspective, it places them in a similar regime as considered in the present paper. LLMs are far more intricate than MLPs. Yet, they have things in common: in addition to the fact that one of the basic building blocks of LLMs is actually the MLP (together with the attention head), both correspond to deep non-linear architectures. We thus consider essential to tackle the interpolation regime of MLPs, with the hope that some insights brought forward by our theoretically tractable idealised setting remain qualitatively relevant for the NN architectures deployed in applications.

\subsection{Main contributions and setting}\label{sec:setting}

We address questions pertaining to the foundations of learning theory for NN models possessing all four properties $(P_1)$--$(P_4)$. The first one is \emph{information-theoretic}: 

\vspace{3pt}
\noindent \hypertarget{Q:Q1}{{\bf Q1}}:\  \emph{Assuming the training data are generated by a target MLP, how much is needed to achieve a certain generalisation performance using an MLP with the same architecture and Bayes-optimally trained in a supervised manner?} 
\vspace{3pt}

\noindent The answer, provided analytically by Result~\ref{res:gen_error}
in 
Sec.~\ref{sec:result}, yields the Bayes-optimal limits of learning an MLP target function, thus bounding the performance of \emph{any} model trained on the same data. The setting where the data-generating process is itself an MLP may look artificial at first, but given the high expressivity and universal approximation property of neural networks, studying their learning provides insights applicable to very general classes of functions. For this reason, and the analytical tractability of the \emph{teacher-student scenario} explained below, this question has always been a starting point in the statistical physics literature on NNs~\citep{gardner1989teacher}.

Secondly, using statistical physics we will answer another important question concerning \emph{interpretability}: 

\vspace{3pt}
\noindent \hypertarget{Q:Q2}{{\bf Q2}}:\  \emph{Given a certain data budget, which target features can the MLP learn?}
\vspace{3pt}

\noindent This is key in order to understand the evolution of the best learning strategy for an MLP as a function of the amount of available data. Consequently, we will precisely explain what a perfectly trained MLP does to beat the random feature model or an optimally regularised kernel (see \figurename~\ref{fig:kernel_vs_NN}). In few words, the reason is that given enough data, strong feature learning emerges within the NN, in the sense of recovery of the target weights. This happens through a \emph{specialisation phase transition}; in the deep case there will be one transition per layer. This mechanism is not possible with the random feature model, which explains the gap in performance. These insights will follow from the detailed analysis of the sufficient statistics (i.e., the order parameters, OPs) of the model as the data increases, obtained 
from the large deviation perspective
provided in Results~\ref{res:free_entropy_shallow}, \ref{res:free_entropy_L=2} and \ref{res:free_entropy_deep}. The OPs carry more information than merely computing the achievable generalisation performance (see \hyperlink{Q:Q1}{{\bf Q1}}). 

Finally, based on experiments, we provide in Sec.~\ref{sec:numerics} \emph{algorithmic insights} for MLPs with $L\le 2$ layers:

\vspace{3pt}
\noindent \hypertarget{Q:Q3}{{\bf Q3}}:\  \emph{Given a reasonable compute and data budget, can practical training algorithms reach optimal performance or are they blocked by statistical-computational gaps?}
\vspace{3pt}

\noindent The short answer is that it depends on the target, in particular on whether its readout weights are discrete.

Answering these questions will provide a phase diagram depicting the optimal performance, the features that are learnt to attain it, and the limitations faced by algorithms, as a function of the data budget, see Sec.~\ref{sec:numerics}. Before presenting the setting needed to do so, let us emphasise once more that the theoretical component of this paper will be only concerned with \emph{static} aspects, i.e., generalisation capabilities of \emph{trained} networks (after a manageable or unconstrained compute time). We do not provide any theoretical claims on how learning occurs \emph{during} training.

\vspace{5pt}\paragraph*{Teacher-student set-up.} 

We consider the supervised learning of an MLP with $L$ hidden layers when the data-generating model, i.e., the target function (or ``teacher''), is also an MLP of the same form with unknown weights. These are the readouts $\bv^0\in\R^k$ and inner weights $\bW^{(l)0}\in\R^{k_l\times {k_{l-1}}}$ for $\ell\le L$ (with $k_0=d$), drawn entrywise i.i.d. from $P_v^0$ and $P^0_W$, respectively (the latter being the same law for all $\ell\le L$). We assume $P^0_W$ to be centred while $P^0_v$ has mean $\bar{v}$, and both priors have unit second moment. We denote the set of unknown parameters of the target as $\btheta^0=(\bv^0,(\bW^{(l)0})_{l\le L})$.

For a given input vector $\bx_\mu\in\R^{d}$, for $\mu\leq n$, the response/label $y_\mu$ is drawn from a kernel $P^0_{\rm out}$:
\begin{talign} \label{eq:teacher}
    y_\mu\sim P^0_{\rm out}(\,\cdot\mid \lambda^0_\mu) \quad \text{with} \quad \lambda^0_\mu:=\mathcal{F}^{(L)}_{\btheta^0}(\bx_\mu),
   \end{talign} 
where the MLP target function is defined as
\begin{talign*}
        \mathcal{F}^{(L)}_{\btheta^0}(\bx):=\frac{\bv^{0\intercal}}{\sqrt{k^L}} \sigma\big(\frac{\bW^{(L)0}}{\sqrt{k_{L-1}}} \sigma\big(\frac{\bW^{(L-1)0}}{\sqrt{k_{L-2}}}\cdots \sigma\big(\frac{\bW^{(1)0}}{\sqrt{k_0}}\bx\big)\cdots\big).
\end{talign*}
We will analyse the case of an arbitrary number of layers $L$ (that remains $d$-independent).
However, we will pay special attention to the shallow, one hidden layer MLP (we drop layer indices in this case)
\begin{talign}
\mathcal{F}^{(1)}_{\btheta^0}(\bx)= \frac{1}{\sqrt{k}} \bv^{0\intercal} \sigma\big(\frac{1}{\sqrt{d}}\bW^{0} \bx\big)
\end{talign}
as well as the MLP with two hidden layers:
\begin{talign}
\mathcal{F}^{(2)}_{\btheta^0}(\bx)= \frac{1}{\sqrt{k_2}} \bv^{0\intercal} \sigma\big(\frac{1}{\sqrt{k_1}}\bW^{(2)0}  \sigma(\frac{1}{\sqrt{d}}\bW^{(1)0} \bx\big)\big).   
\end{talign}

The kernel can be stochastic or model a deterministic rule if $P^0_{\rm out}(y\mid\lambda)=\delta(y-\mathsf{f}^0(\lambda))$ for some function $\mathsf{f}^0$. Our main example $P^0_{\rm out}(y\mid\lambda) =\exp(-\frac1{2\Delta}(y-\lambda)^2)/\sqrt{2\pi\Delta}$ is the \emph{linear readout} with Gaussian label noise. 

We will first consider i.i.d. standard Gaussian vectors as inputs $\bx_\mu$. In that case the whole data structure is dictated by the input-output relation only, allowing us to focus solely on the influence of the target function on the learning. In Sec.~\ref{sec:numerics} we generalise the results to include structured data: Gaussian with a covariance and real data (MNIST). The input/output pairs $\mathcal{D}=\{(\bx_\mu,y_\mu)\}_{\mu \leq n}$ form the training set for a student network with matching architecture. 

\begin{figure}
    \centering
    \includegraphics[width=1\linewidth,trim={0 19.5cm 0cm 0cm},clip]{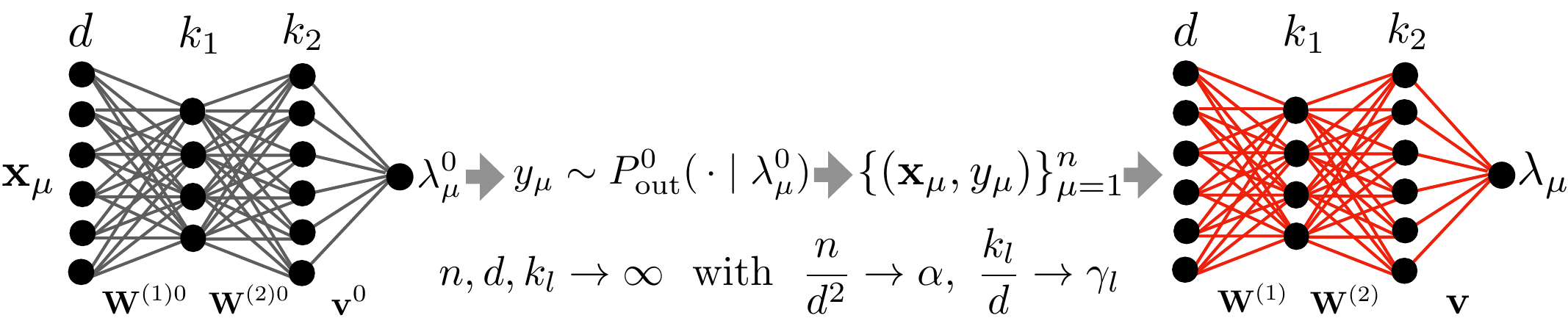}
    \caption{The teacher-student scenario for the case of two hidden layers. The teacher NN is used to produce the responses given the inputs. A student NN with matched architecture (but who is not aware of the parameters of the teacher) is then trained in a Bayesian manner given the training data. We also display the scaling limit considered given by \eqref{thermolim}.}
    \label{fig:TS}
    \vskip -0.1in
\end{figure}

The Bayesian student learns via the posterior distribution of the weights matrices $\btheta=(\bv,(\bW^{(l)})_{l\le L})$ (of the same respective sizes as the teacher's) given the training data:
\begin{talign*}
&dP(\btheta\mid\mathcal{D}):=\mathcal{Z}(\mathcal{D})^{-1}dP_\theta(\btheta)\prod_{\mu\le n}P_{\rm out}\big(y_\mu\mid \lambda_\mu(\btheta)\big) 
\end{talign*}
where $dP_\theta(\btheta):=dP_v(\bv) \prod_{l\le L}dP_W(\bW^{(l)})$ (with the notation $dP(\bM):=\prod_{i,j}dP(M_{ij})$),  with post-activations
\begin{talign*}
\lambda_\mu(\btheta):= \mathcal{F}^{(L)}_{\btheta}(\bx_\mu), \quad \mu\le n.
\end{talign*}
The posterior normalisation $\mathcal{Z}(\mathcal{D})=\mathcal{Z}^{(L)}(\mathcal{D})$ for the model with $L$ hidden layers is the partition function, and $P_W,P_v$ are the priors assumed by the student. We focus on the \emph{Bayes-optimal setting} $P_W=P_W^0$, $P_v=P_v^0$ and $P_{\rm out}=P_{\rm out}^0$, but the approach can be extended to account for a mismatch.

As stated above, we study the \emph{linear-width regime with quadratically many samples}, which places the model in the interpolation regime, i.e., a large size limit 
\begin{talign}
 d,k_l,n\to+\infty \ \ \text{with} \ \  \frac{k_l}{d}\to\gamma_l \ \ \text{for} \ \ l\le L , \ \ \frac{n}{d^2}\to\alpha  .\label{thermolim}
\end{talign}
 
Given the cost of training deep Bayesian MLPs and specific difficulties discussed below associated with an increasing number of layers, we distinguish the cases of one, two and more than two hidden layers for what concerns the hypotheses we impose on the activation $\sigma$.

\hypertarget{H:H1}{$(H_1)$} For shallow NNs with $L=1$ hidden layer our results are valid for an arbitrary activation function as long as it admits an expansion in Hermite polynomials with coefficients $(\mu_\ell)_{\ell\geq0}$, see App.~\ref{app:hermite}: 
\begin{talign}
\sigma(x) = \sum_{\ell \ge 0 } \frac{\mu_\ell}{\ell !} \,\He_\ell (x) .    \label{eq:HermiteSig}
\end{talign}
We also assume it has vanishing 0th Hermite coefficient $\mu_0=0$, i.e., that it is centred $\EE_{z\sim\mathcal{N}(0,1)}\sigma(z)=0$; in App.~\ref{app:non-centered} we relax this assumption. We will mainly consider $\tanh$, ${\rm ReLU}$ and Hermite polynomial activations. 

Through Hermite expansion, the MLP function can be decomposed as
\begin{talign*}
\mathcal{F}^{(1)}_{\btheta}(\bx) \!=\! \frac{\mu_1}{\sqrt{d}}\frac{\bv^\intercal\bW}{\sqrt{k}}\bx  + \frac{\mu_2}{2d}{\rm Tr}(\frac{\bW^\intercal{\rm diag}(\bv)\bW}{\sqrt{k}}(\bx^{\otimes 2} - I_d )) + \cdots
\end{talign*}
where 
$\cdots$ contains 
terms made of tensors of all orders constructed from $\btheta$, contracted with input rank-one tensors $(\bx^{\otimes\ell})_\ell$. In each such term, \emph{at least one tensor is of order $\ell\ge 3$}. Therefore an equivalent interpretation of the learning problem of an MLP target is that of a ``tensor sensing problem'' where the tensors entering the observed responses, $\bv^{0\intercal}\bW^0\in \mathbb{R}^d, \bW^{0\intercal}{\rm diag}(\bv^0)\bW^0\in \mathbb{R}^{d\times d},\ldots$ are all constructed from the same fundamental parameters $\btheta^0$ (see, e.g.,  \cite{mondelli2019connection}). The first term in the above expansion is called the ``linear term/component''. The one of the target is perfectly learnable in the quadratic data regime we consider. The second term is the ``quadratic term/component''. Both will play a special role because, as we will see, the terms $\cdots$ effectively behave as (Gaussian) noise when $n=\Theta(d^2)$, \emph{unless} $\btheta^0$ is partially recovered. Learning through recovery of $\btheta^0$ is called \emph{specialisation}. 
In contrast, the linear and quadratic terms are learnable without specialisation. This is reassuring given that we will argue that for many targets, it takes a time growing as $\exp(c\, d)$ for the network to specialise, for some positive $\sigma$-dependent constant $c<1$. This separation in algorithmic learnability of first and second components versus all the others is at the root of the emergence of different learning strategies employed by the network, and the crux of the generalisation of a learning algorithm in App.~\ref{app:GAMP} coined GAMP-RIE (generalised approximate message-passing with rotational invariant estimator) \cite{maillard2024bayes}.

\hypertarget{H:H2}{$(H_2)$} For $L= 2$ we require $\mu_0=\mu_2=0$, which is e.g. the case for odd activations. Our main example is $\tanh$. In the tensor inference problem appearing when expanding all activations, $\mu_2=0$ means that no quadratic term is present. However, a ``product term'' $\bW^{(2)}\bW^{(1)}$ appears (in addition to $\bv^\intercal\bW^{(2)}\bW^{(1)}$). We will see in Sec.~\ref{sec:numerics} that skipping the quadratic term implies that learning terms beyond the linear ones will be possible only through specialisation. However, the presence of the product term will have interesting consequences on the learning curves. Importantly, $\bW^{(2)}\bW^{(1)}$ is a matrix learnable partly independently of its factors, and consequently requires its own OP in the analysis.

\hypertarget{H:H3}{$(H_{3})$} For $L\ge 3$ we require $\mu_0=\mu_1=\mu_2=0$. This does not include standard activations and we consider the hyperbolic tangent after setting $\mu_1=0$ in its Hermite decomposition. $\mu_2=0$ again entails that learning beyond-linear terms requires the network to specialise, and $\mu_1=0$ prevents the multiplication of OPs by avoiding the presence of many product terms.

Related to this last comment, we wish to emphasise that these hypotheses are not due to restrictions of the techniques we develop. The issue is purely practical: relaxing them while increasing the number of layers yields a combinatorial explosion (in $L$) of the number of OPs to track in the theory as well as cumbersome formulas. We have therefore decided to leave for future work the analysis of the most general case, and focus here on these special ones which already yield an extremely rich picture while remaining interpretable.

\subsection{Replica method and HCIZ combined}\label{sec:relatedLit}

A key component of our approach is the way we blend tools from spin glasses (the replica method \cite{mezard1987spin}) and matrix models, in particular, the so-called Harish Chandra--Itzykson--Zuber (HCIZ) ``spherical'' integral~\cite{itzykson1980planar,Matytsin_1994,guionnet2002large}. Here, we review the growing corpus of works utilising it jointly with the replica method. Let us first define this matrix integral: 
\begin{talign}\label{eq:HCIZ-def}
    \mathcal{Z}^{(\beta)}_{\rm HCIZ}(\bA,\bB):=\int d\mu^{(\beta)}(\bO)\exp\big(\frac{\beta N}{2}\Tr[\bO\bA\bO^\intercal\bB]\big)
\end{talign}where $\beta=2$ if $\bA,\bB$ are $N\times N$ Hermitian matrices, $\beta=1$ if real symmetric. Respectively the integral is over the unitary $\mathbb{U}(N)$ or orthogonal group $\mathbb{O}(N)$, w.r.t. the corresponding uniform Haar measure $\mu^{(\beta)}$. For $\beta=2$ it admits a closed form for any $N$ \cite{itzykson1980planar} and a known large $N$ limit for $\beta =1,2$ \cite{Matytsin_1994,guionnet2002large,guionnet2004first,zuber2008large}. It is crucial to analyse matrix models in physics and random geometry \cite{kazakov2000solvable,brezin2016random,anninos2020notes}.
In spite of having an ``explicit'' limit, it can be tackled only in few cases \cite{bun2014instanton,maillard2022perturbative}. However, if one matrix, say $\bA$, has small rank compared to $N\gg 1$, the corresponding \emph{low-rank spherical integral} is simple \cite{potters2020first,husson2025spherical}.

Spherical integrals were used in the replica method for spin glasses with correlated disorder in the seminal paper \cite{parisi1995mean}. It triggered a long series of works in spin glasses \cite{opper2001adaptive,opper2016theory,fan2022tap,barbier2022marginals,fan2024replica,barbier2022marginals}, in analysing simple NNs \cite{kabashima2008inference,gabrie2018entropy} or in inference and message-passing algorithms \cite{takeda2006analysis,tulino2013support,hou2022sparse,barbier2025information,rangan2019vector,ma2017orthogonal,maillard2019high,liu2022memory,takeuchi2022convergence,takahashi2022macroscopic,fan2022approximate,barbier2018mutual,gerbelot2020asymptotic,gerbelot2022asymptotic,dudeja2023universality,barbier2023fundamental,dudeja2024optimality}. In these papers the degrees of freedom are few vectors (e.g., the replicas of the system forming a low-rank matrix $\bA$) interacting with a quenched \emph{rotationally invariant matrix} $\bB$ of rank $\Theta(N)$. Rotational invariance, a crucial property for employing spherical integrals, means distributional invariance under orthogonal transformations, i.e., $\mathbb{P}(\bB) = \mathbb{P}(\bO\bB\bO^\intercal) \ \forall \ \bO\in \mathbb{O}(N)$ if $\bB$ is symmetric. 
Consequently, only the \emph{low-rank} spherical integral intervenes when integrating $\bB$'s eigenvectors. 

An active research line tries to include models where the degrees of freedom themselves are linear-rank matrices in addition to the quenched disorder. This presents a whole new challenge. Seminal papers in the context of matrix denoising are \cite{ledoit2011eigenvectors,bun2016rotational} which provided a spectral denoising algorithm (on which the GAMP-RIE \cite{maillard2024bayes} relies), also analysed in \cite{matrix_inference_Barbier,semerjian2024}. Extensions to non-symmetric matrix denoising exist \cite{pourkamali2023rectangular,troiani2022optimal}. An early attempt at
combining linear-rank spherical integration (where both $\bA,\bB$ in \eqref{eq:HCIZ-def} have rank $\Theta(N)$) with the replica method is \cite{thesis_schmidt}, which tried to improve on the replica approach for matrix denoising in \cite{sakata2013,kabashima2016phase} that was missing important correlations among variables. It was followed by two concurrent papers yielding intractable \cite{barbier2022statistical} or perturbative \cite{maillard2022perturbative} results for non-Gaussian signals.   

\begin{remark}
No method in the aforementioned papers is satisfactory beyond the realm of denoising problems involving strictly rotationally invariant signal matrices (Gaussian, Wishart,...). E.g., the HCIZ/replica combination in the latest works \cite{maillard2024bayes,erba2025bilinear} requires it, because after using the replica trick to integrate the quenched disorder, the HCIZ is used directly to integrate the annealed matrix degrees of freedom (representing the replicas of the signal matrix), which is possible by rotational invariance.
\end{remark}

Recently, matrix denoising \emph{without} rotational invariance was analysed in \cite{barbier2024phase} by assuming that the model behaves as a pure matrix model (due to an ``effective rotational invariance'') in a first phase, and then as a ``standard'' planted mean-field spin glass in a second. The phases were thus treated \emph{separately} via different formalisms --HCIZ in one phase, a cavity method under mean-field decoupling assumptions in the other-- and then joined using a criterion to locate the transition. This approach yielded a good match with numerics. However, we now understand that this treatment can be improved, because the ``matrix nature'' of the model and associated correlations discarded by mean-field methods do play a role also in the second phase. Thus, a major conceptual (and technical) issue remained: whether there exists a theory based on a \emph{unified formalism} able to describe the whole phase diagram of inference/learning problems involving linear-rank matrices which lack rotational invariance. Ideally, it should be able to handle the correlations induced by the matrix nature of the problem while still capturing the phase transitions and symmetry breaking effects connected to its mean-field component. The present paper provides this theory in the context of NNs, through a replica/HCIZ combination of a different nature than previous works, see Sec.~\ref{sec:theory}. 
 
Related to this last point, we emphasise that previous works on extensive-widths shallow NNs ($k=\Theta(d^c)$ for $0<c\le 1$) considered either purely quadratic activation \cite{gamarnik2024stationary,maillard2024bayes,xu2025fundamental,erba2025nuclear,martin2024impact,arous2025learning} or, on the contrary, with $\mu_{\le 2}=0$ \cite{ren2025scalinglaws}. Both settings enjoy intrinsic simplifications. On one hand, the quadratic NN reduces to a matrix sensing problem \cite{maillard2024bayes,erba2025nuclear}. It is therefore a ``pure matrix model'' with rotational invariance when considering Gaussian weights: the target (and model) only depends on them via $\bW^{0\intercal}\bW^0$. Therefore, by rotational invariance (from the left and right) of the Gaussian matrix $\bW^0$, it cannot be recovered, so no specialisation transitions can occur. The advantage is that a large toolbox from random matrix theory is then available: the HCIZ integral to study static aspects \cite{maillard2024bayes,erba2025nuclear,xu2025fundamental}, or Oja’s flow and matrix Riccati differential equations for the dynamics \cite{bodin2023gradient,martin2024impact,arous2025learning}. On the other hand, \cite{ren2025scalinglaws} considers $\mu_\ell\neq 0$ for $\ell\ge 3$ only. In this case, the model is ``purely mean-field'': strong decoupling phenomena take place which allow a treatment in term of an effective one-body equivalent system as in mean-field spin systems. 

In contrast, the techniques we develop in the present paper can deal with truly hybrid models where the two types of characteristics manifest and are taken into account using a single formalism: the correlations among the entries of the matrix degrees of freedom entering the problem, and specialisation phase transitions induced by mean-field terms. The emerging phase diagram will consequently be extremely rich. In particular, we are able to treat the shallow MLP with \emph{generic} activation function (Result~\ref{res:free_entropy_shallow}), or the two-layer MLP with $\mu_1\neq 0$ (Result~\ref{res:free_entropy_L=2}). The case of $L\ge 3$ requires hypothesis (\hyperlink{H:H3}{$H_3$}) on $\sigma$ which, in turn, makes the model ``purely mean-field''.


\subsection{Organisation of the paper}

$\bullet$ Section~\ref{sec:result} first discusses the main hypothesis underlying the theory and the meaning of functions entering it. We then present the theoretical results: replica symmetric formulas for the free entropy and OPs for shallow NNs (Result~\ref{res:free_entropy_shallow}), with two hidden layers (Result~\ref{res:free_entropy_L=2}), or arbitrary $L$ ( Result~\ref{res:free_entropy_deep}). These provide an answer to \hyperlink{Q:Q2}{{\bf Q2}}. The Bayes generalisation error is deduced automatically from Result~\ref{res:gen_error} in all cases, thus answering \hyperlink{Q:Q1}{{\bf Q1}}.

$\bullet$ Section~\ref{sec:numerics} is the core experimental part. It validates the theory through the numerical exploration of the rich learning phase diagram. Our main message concerning \hyperlink{Q:Q2}{{\bf Q2}} is as follows. As $\alpha$ increases two phases appear: 

\paragraph*{$(i)$ Universal phase.} Before a critical sample rate $\alpha_{\rm sp}$, the NN makes predictions by exploiting specific non-linear combinations of the teacher’s features without disentangling them; effectively, the student learns the best ``quadratic network approximation'' of the target. In this phase, performance is (asymptotically) independent of the detailed law of the target hidden weights (hence the term ``universal''). Yet, the (effectively quadratic) NN outperforms kernel ridge regression (and thus the random feature model too, \figurename~\ref{fig:kernel_vs_NN}), see \cite{maillard2024bayes,erba2025nuclear}.

\paragraph*{$(ii)$ Specialisation phase.}  Increasing the data beyond $\alpha_{\rm sp}$ triggers specialisation transitions: individual hidden units start aligning with target units. Which features specialise first is governed by the readout strengths of the target: stronger features (larger readout amplitudes) emerge earlier. For heterogeneous readouts, this yields a sequence of specialisation events; for homogeneous readouts, a collective transition occurs. If $L\ge 2$, possible heterogeneity both in the rows and columns of individual weight matrices induces non-trivial specialisation profiles in each layer. In turn, different layers can experience different phases and do not necessarily specialise concurrently. We will also show that learning propagates from inner to outer hidden layers, because deeper layers require more data to be recovered through specialisation. Consequently, deeper target functions appear harder to learn than shallow ones.

In summary, despite the model's ``matrix nature'' at the source of the universal phase, additional mean-field-like terms in the free entropy (in information theory parlance, Gaussian scalar channels) imply the existence of specialisation events. These terms depend explicitly on the weights prior and interact with the matrix degrees of freedom, and ultimately break the numerous effective symmetries holding before the transition.

The theoretical phase diagram will be extensively tested against various training algorithms: two Monte Carlo-based Bayesian samplers, a first-order optimisation procedure (ADAM), and a mixed spectral/approximate message-passing algorithm generalising the GAMP-RIE of \cite{maillard2024bayes} to accommodate general activation functions $\sigma$ when $L=1$. The performance of these algorithms belonging to different classes, even when sub-optimal, can be exactly (or, for ADAM, at least accurately) predicted by non-equilibrium solutions of the theoretical equations.

Focusing on $L\le 2$ for what pertains algorithmic hardness of learning, \hyperlink{Q:Q3}{{\bf Q3}}, we will show empirically that specialisation is potentially hard to reach for some target functions, in particular when the readouts are discrete. The tested algorithms fail to find it and instead get trapped by sub-optimal non-specialised solutions, probably due to statistical-computational gaps. 

We will also generalise the theory to structured data, i.e., Gaussian with a covariance. It will capture the model's performance when trained from non-Gaussian inputs too. Tests with real (MNIST images) and synthetic data generated by one layer of a NN will confirm it.

$\bullet$ Section~\ref{sec:theory} contains the main steps of our replica theory, with an emphasis on its novel ingredients. Along the derivation, the mixed matrix model/mean-field planted spin glass nature of the problem will become apparent.

$\bullet$ Finally, Section~\ref{sec:conclusion} summarises our contributions and discusses the numerous perspectives this work opens.

The appendices are found after the references.

$\bullet$ Appendix~\ref{app:auxiliaryresults} gathers some important pre-requisites: App.~\ref{app:notations} summarises all  notations used in the paper (we advise the reader to give it a look before reading the main results); the definition of the Hermite polynomials and Mehler's formula are found in App.~\ref{app:hermite}; the Nishimori identities in Bayes-optimal inference in App.~\ref{app:nishiID}; the link between free entropy and mutual information in App.~\ref{app:mutual_info}; and a simplification of the expression for the optimal mean-square generalisation error in App.~\ref{app:gen_err}.

$\bullet$ Appendix~\ref{app:shallowMLP} groups all sub-appendices related to the shallow MLP: App.~\ref{app:replicas} details all the steps of the replica calculation; App.~\ref{app:alternatives} proposes alternative routes to take care of the entropy of the order parameters associated with the matrix degrees of freedom in the model; App.~\ref{app:large_alpha} analyses the large sampling rate limit of the theoretical free entropy; App.~\ref{app:GAMP} provides the generalisation of the GAMP-RIE algorithm needed to deal with general $\sigma$; App.~\ref{app:hardness} is an empirical analysis of the hardness of learning shallow targets; App.~\ref{sec:partial_proof} is a partial proof for a special case of activation function; finally, App.~\ref{subsec:learnable_readouts} provides additional experimental validations of the fact that the readout weights of the model being learnable or fixed has no effect on its optimal performance.

$\bullet$ Appendix~\ref{app:deep} concerns only the deep MLP: App.~\ref{app:replicas_deep} is the replica calculation; App.~\ref{app:structured_data} shows the consistency of the formulas provided for structured inputs with $L=1$ in the main, and the ones for a special case of non-Gaussian data obtainable from the theory for two hidden layers when freezing the first one (which induces a structure for the inputs of the second, learnable layer).

$\bullet$ Appendix~\ref{app:numericsdetails} provides all information needed to reproduce the simulations with the provided codes \cite{github}.

\section{Main results: theory of the MLP}\label{sec:result}

We aim at evaluating the expected optimal generalisation error in the teacher-student setting of \figurename~\ref{fig:TS}. Let $(\bx_{\rm test}, y_{\rm test}\sim P_{\rm out}(\,\cdot\mid \lambda^0_{\rm test}))$ be a test sample independent of $\mathcal{D}$ drawn using the teacher, where $\lambda_{\rm test}^0$ is defined as in \eqref{eq:teacher} with $\bx_\mu$ replaced by $\bx_{\rm test}$ (and similarly for $\lambda_{\rm test}(\btheta)$). Given a prediction function $\mathsf{f}$, the Bayes estimator for the test response is $\hat{y}^{\mathsf{f}}(\bx_{\rm test},\calD)
    :=\langle \mathsf{f}(\lambda_{\rm test}(\btheta)) \rangle$, where $\langle \,\cdot\, \rangle :=\EE[ \,\cdot\, |\, \mathcal{D}]$. Then, for a
performance measure $\mathcal{C}:\mathbb{R}\times \mathbb{R}\mapsto\mathbb{R}_{\ge 0}$ the Bayes generalisation error is
\begin{align}
\varepsilon^{\mathcal{C},\mathsf{f}}:=\EE_{\btheta^0,\calD,\bx_{\rm test},y_{\rm test}}\mathcal{C}\big(y_{\rm test}, \big\langle \mathsf{f}(\lambda_{\rm test}(\btheta)) \big\rangle\big).\label{eq:Bayes_error_def}
\end{align}
The case of square loss $\mathcal{C}(y,\hat y)=(y-\hat y)^2$ with the choice $\mathsf{f}(\lambda)=\int dy\, y\, P_{\rm out}( y\mid \lambda)=:\EE[y\mid \lambda]$ yields the Bayes-optimal mean-square generalisation error:
\begin{align}\label{eq:gen_error_def}
    \varepsilon^{\rm opt} &:= \EE_{\btheta^0,\calD,\bx_{\rm test},y_{\rm test}}\big(y_{\rm test} - \big\langle\EE[y\mid \lambda_{\rm test}(\btheta)]\big\rangle\big)^2.
\end{align}

In order to access $\varepsilon^{\mathcal{C},\mathsf{f}},\varepsilon^{\rm opt}$ and other relevant observables, one can tackle the computation of the average log-partition function, or \emph{free entropy} in statistical physics: 
\begin{talign}
f_n:=\frac1n\EE_{\btheta^0,\mathcal{D}}\ln\mathcal{Z}(\mathcal{D}).
\label{eq:free_entropy_def}
\end{talign}
The mutual information $I(\btheta^0;\mathcal{D})$ between the target and data is related to the free entropy $f_n$, see App.~\ref{app:mutual_info}.

Before presenting the results we will first detail the main hypothesis for their derivation and explain the physical meaning of the quantities entering them. This will ease their interpretation. We postpone the core of the theoretical derivations to Sec.~\ref{sec:theory}.

\vspace{5pt}\paragraph*{Main hypothesis.}  Let $s\in\mathbb{N}_{>0}$ independent of $d$ and define a Gaussian vector $(\lambda^a)_{a=0}^s := (\lambda^0,\lambda^1,\cdots,\lambda^s)^\intercal \sim \mathcal{N}(\mathbf{0},\mathbf{K}^*)$ with covariance (for $a,b=0,\ldots,s$)
\begin{talign}
 (\mathbf{K}^*)_{ab}:=\EE \lambda^a \lambda^{b} = K^*+(K_{\rm d}-K^*)\delta_{ab}.\label{eq:covLam}
\end{talign}
Let $(\btheta^a)_{a=1}^s$ be i.i.d. from the posterior $dP(\,\cdot \mid\mathcal{D})$ and $\btheta^0$ are the random target weights. Our main assumption is that there exists a non-random $\mathbf{K}^*$ s.t., under the randomness of a common test input $\bx_{\rm test}  \notin \mathcal{D}$ and $(\btheta^a)_{a=0}^s$, the post-activations $(\lambda_{\rm test}(\btheta^a))_{a=0}^s$  (called ``replicas''), converge in law towards $(\lambda^a)_{a=0}^s$ in the limit~\eqref{thermolim}:
\begin{talign}\label{eq:hypothesis_gaussian_test}
   \textbf{Hypothesis:}\ \exists \ \mathbf{K}^* \ | \ (\lambda_{\rm test}(\btheta^a))_{a=0}^s  \xrightarrow{{\rm Law}} (\lambda^a)_{a=0}^s.
\end{talign}
The goal of the replica method will be to derive $\mathbf{K}^*$ in terms of fundamental low-dimensional OPs capturing the statistical dependences among $(\btheta^a)_{a=0}^s$. The above convergence can be equivalently assumed conditionally on $(\btheta^a)_{a=0}^s$, if well sampled, by concentration of the OPs. $\EE_{(\lambda^a)}[\,\cdot\,]$ must therefore be interpreted as the asymptotic equivalent of the expectation w.r.t. the ``quenched Gibbs measure'' (i.e., the whole randomness): given a function $f:\mathbb{R}^{s+1}\mapsto \mathbb{R}^{\bar s}$ of $s+1$ replicas of the post-activation (with $s,\bar s$ independent of $d$), 
\begin{talign*}
    \EE_{\btheta^0,\calD,\bx_{\rm test},y_{\rm test}}\langle f((\lambda_{\rm test}(\btheta^a))_a) \rangle\to \EE_{(\lambda^a)}\,f((\lambda^a)_a).
\end{talign*} Notice that the covariance for $a\neq b$ does not depend on whether one of the indices is the teacher's index $0$. That the teacher is statistically indistinguishable from the other replicas $a\ge 1$ is a consequence of the Bayes-optimal setting and the Nishimori identities, see App.~\ref{app:nishiID}. In the non-Bayes-optimal setting, which is treatable with a similar approach, the covariance would be more complicated, with the teacher playing a special role.

\begin{figure}
    \centering
    \includegraphics[width=0.45\linewidth]{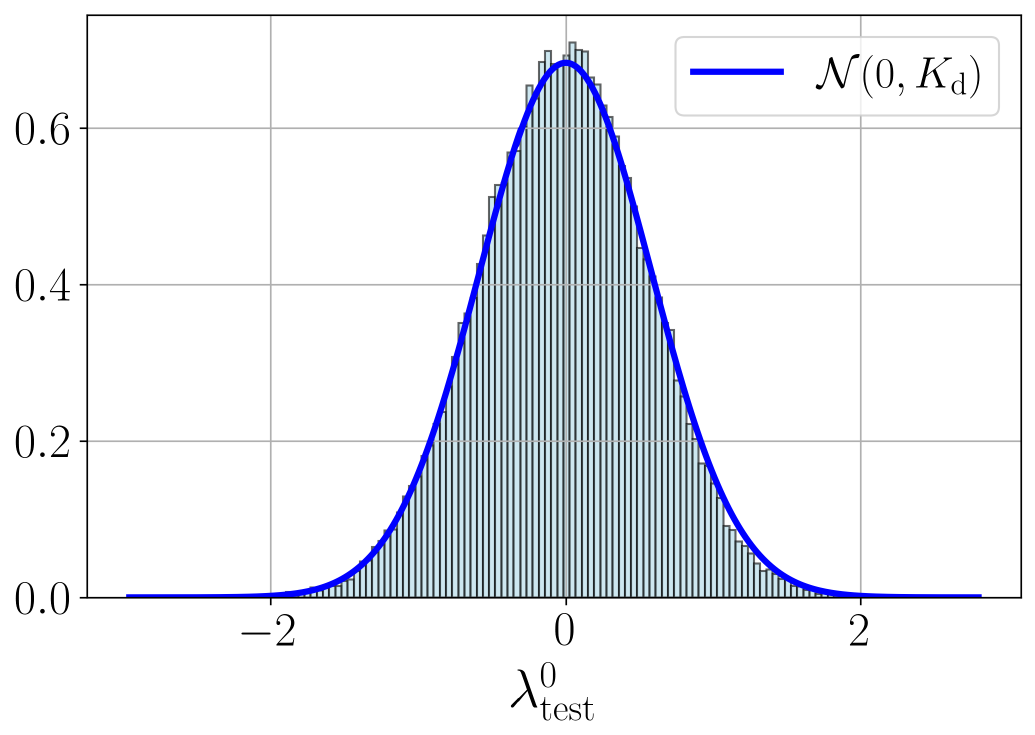}
    \includegraphics[width=0.45\linewidth]{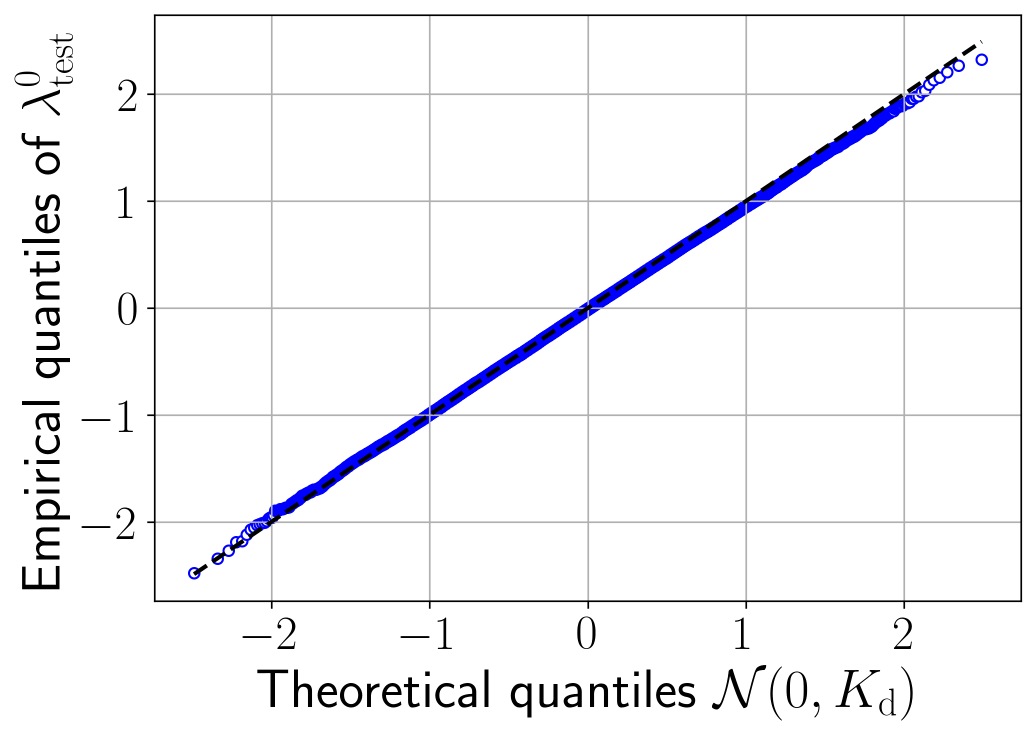}
    \includegraphics[width=0.45\linewidth]{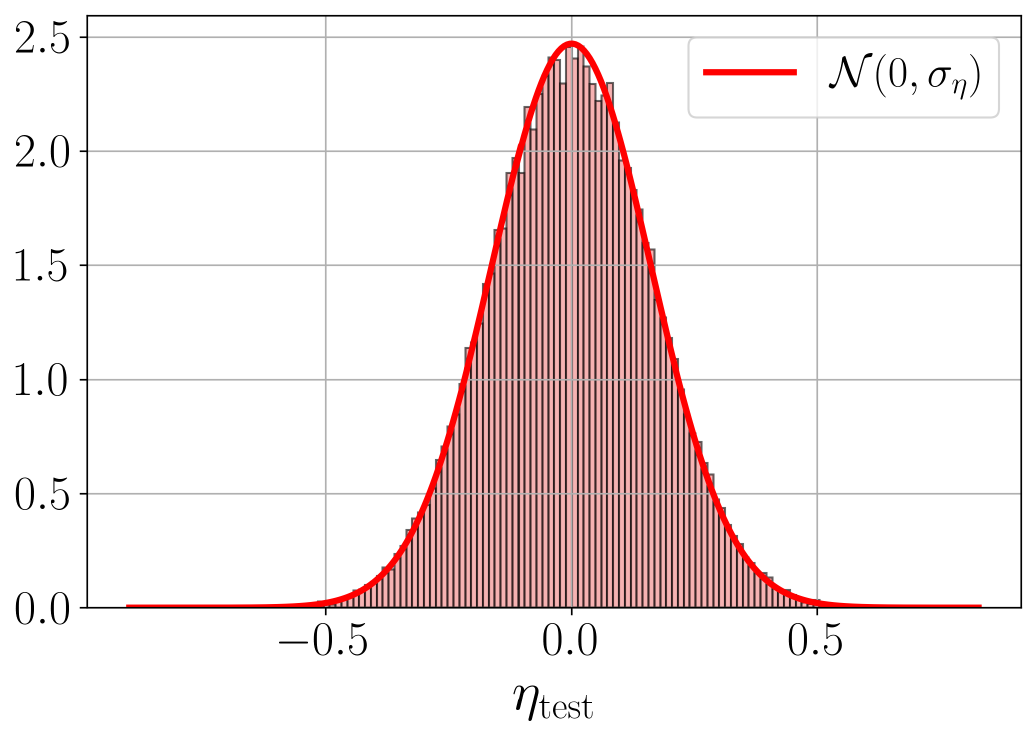}
    \includegraphics[width=0.45\linewidth]{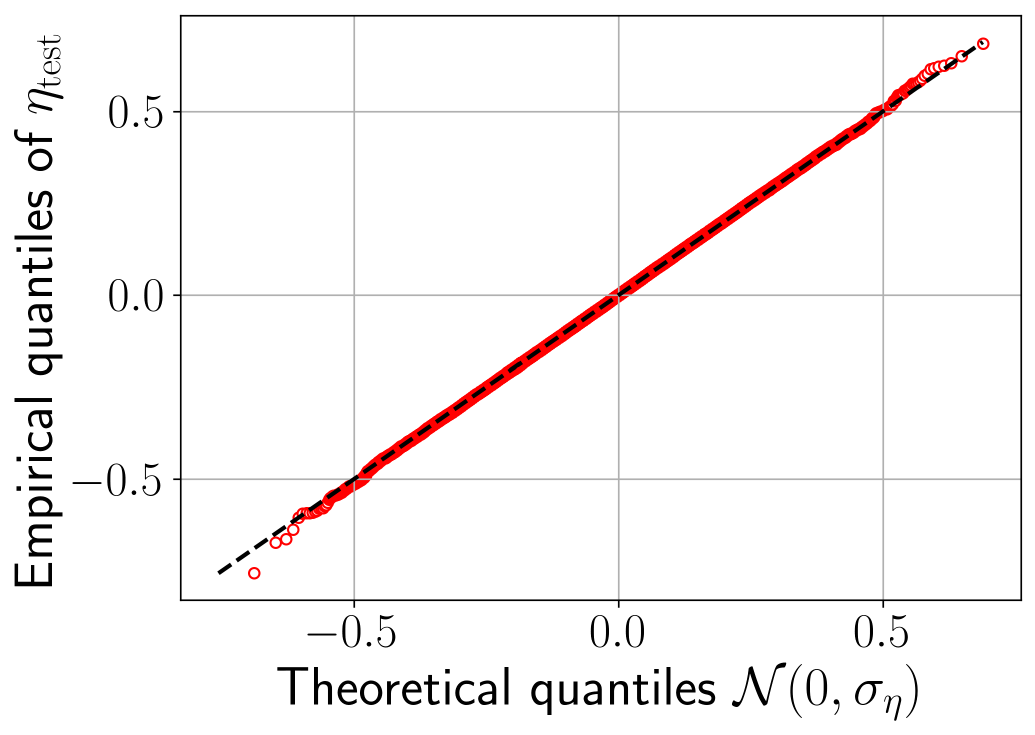}
    \caption{Experimental evidence for the Gaussian hypothesis. In all experiments, $d=300, \gamma=0.5, \alpha=3.0, \Delta=0.1,\sigma(x) = \relu(x) - 1/\sqrt{2 \pi}$, both readout and inner weights have standard Gaussian prior. Empirical evaluations are based on a test set of size $5\times 10^4$. The results have been averaged over 10 instances of the training set and teacher. \textbf{Top left}: Histogram of the teacher post-activations $\lambda^0_{\rm test}$ evaluated on $\bx_{\rm test}$ compared with the theoretically predicted Gaussian density $\mathcal{N}(0,K_{\rm d})$ (see \eqref{eq:covariance_vars_main} for the definition of $K_{\rm d}$). 
\textbf{Top right}: Quantile-quantile plot comparing the theoretical quantiles of $\mathcal{N}(0,K_{\rm d})$ with the empirical ones of $\lambda^0_{\rm test}$.  
\textbf{Bottom left}: Histogram of the student's projection along the orthogonal direction to the teacher: $\eta_{\rm test} = \lambda_{\rm test} - [\EE_{\bx_{\rm test}}\lambda^0_{\rm test}\lambda_{\rm test}/\EE_{\bx_{\rm test}}(\lambda^0_{\rm test})^2] \lambda^0_{\rm test} \approx \lambda_{\rm test} - (K^*/K_{\rm d})\lambda^0_{\rm test}$, where $\lambda_{\rm test}(\bv,\bW)$ is the student post-activation with both $\bv$ and $\bW$ sampled from the posterior via Hamiltonian Monte Carlo, evaluated on the same test set, and compared with the theoretical density $\mathcal{N}(0,\sigma_\eta)$, where $\sigma_\eta = K_{\rm d} - K^{*2}/K_{\rm d}$ (see \eqref{eq:covLam_2} for $K^*$). 
\textbf{Bottom right}: Quantile-quantile plot comparing the theoretical quantiles of $\mathcal{N}(0,\sigma_\eta)$ with the empirical quantiles of $\eta_{\text{test}}$.  } 
    \label{fig:gaussianPostactiv}
    \vskip -0.1in
\end{figure}

The Gaussian hypothesis \eqref{eq:hypothesis_gaussian_test} will be justified a posteriori in Sec.~\ref{sec:numerics}, by the excellent match between our predictions for the learning curves and OPs and the experimental ones. It can also be tested directly: \figurename~\ref{fig:gaussianPostactiv} displays the histogram of the teacher post-activation for multiple test inputs (blue) and of the projection of the student post-activation along the orthogonal direction to the teacher, when trained by Hamiltonian Monte Carlo and evaluated on the same test data (red). Our hypothesis implies that they should both be Gaussian distributed and indeed they are, with laws correctly predicted by the theory (see next two sections). 
We also compare the empirical moment generating function  of $(\lambda_{\rm test}^0, \lambda_{\rm test})$ and its theoretical prediction based on the Gaussian hypothesis, respectively given by $M_{\rm emp}(t_0, t_1) = \EE_{\bx_{\rm test}} \exp (t_0\lambda_{\rm test}^0 + t_1 \lambda_{\rm test} )$, where  $\E_{\bx_{\rm test}}$ is an average over a test set of size $10^5$ and $M_{\rm th}(t_0, t_1) = \exp (K_{\rm d}(t_0^2 + t_1^2)/2 + K^* t_0 t_1)$. Their relative error $|1 - M_{\rm emp}(t_0, t_1)/M_{\rm th}(t_0, t_1)|$, computed over a $21\times21$ regular grid in $[0,1]^2$, has mean 0.015 and standard deviation 0.016. This confirms that the theoretical Gaussian laws provide a remarkably accurate fit of the observed ones.

\begin{remark}
\label{rmk:gep}
    Gaussian assumptions on the post-activations are at the core of a fruitful series of works mapping the generalisation capabilities of random feature models~\cite{mei2019precise,goldt2020modeling,hastie2022surprises,goldt2022gaussian,hu2022universality,montanari2022universality} and overparametrised NNs~\cite{cui2023bayes,camilli2023fundamental,camilli2025inforeduction} to the ones of equivalent Gaussian covariate models. In these settings, formal proofs support the hypothesis, but the covariance of the post-activations matches the one of a statistically equivalent model which is \emph{linear} in the input data.
    For this reason, many results in these settings go under the name of ``Gaussian equivalence principle'' (GEP) or ``theorem'' (GET). The failure of the approaches based on the GEP to capture non-linear effects in learning around interpolation has been attributed to non-Gaussian corrections becoming relevant in this regime~\cite{cui2023bayes}. Instead, we show here that the Gaussian hypothesis~\eqref{eq:hypothesis_gaussian_test}, once non-linear effects are taken into account in the form of $\bK^*$ as in our Results below, works remarkably well to make predictions in the interpolation asymptotics, to the point we conjecture it to be exact in some cases, see Remark~\ref{rmk:quadratic}. A recent rigorous work provides examples where GEPs break down and describes how they can be redeemed \cite{wen2025does}.
\end{remark}


\vspace{5pt}\paragraph*{Auxiliary potentials and their interpretation.}
As usual with the replica formalism used in the context of inference \cite{nishimori2001statistical,zdeborova2016statistical}, the derived formulas are expressed in terms of auxiliary potential functions that are related to the log-normalisation constants of the posterior distributions of auxiliary inference problems. These potentials shall be denoted by $\psi_{P_W}$, $\phi_{P_{\rm out}}$, $\iota$ and $\tilde\iota$. We describe their meaning hereby.

$\bullet$ Let $w^0,w \sim P_W$ and $\xi\sim \calN(0,1)$ all independent. We define the potential
\begin{talign*}
        &\psi_{P_W}(x) := \mathbb{E}_{w^0,\xi} \ln\mathbb{E}_w \exp(-\frac{1}{2}xw^2 + x w^0 w + \sqrt{x}\, \xi w).
\end{talign*}
This is the free entropy of a scalar Gaussian observation channel $y_G=\sqrt{x} \,w^0+\xi$ with prior $P_W$ on the signal $w^0$. Parameter $x$ plays the role of signal-to-noise ratio (SNR).

$\bullet$ Let $\xi,u,u^0\sim \calN(0,1)$ be all independent. Define
\begin{talign*}
    &\phi_{P_{\rm out}}(x;r) := \int dy\, \mathbb{E}_{\xi,u^0}P_{\rm out}(y\mid \sqrt{x}\,\xi+\sqrt{r-x}\,u^0)\\
        &\qquad\qquad\qquad \qquad\times\ln \mathbb{E}_{u}P_{\rm out}(y\mid \sqrt{x}\,\xi+\sqrt{r-x}\,u).
\end{talign*}
This is the free entropy associated with the scalar observation channel $y_{\rm out}\sim P_{\rm out}(\,\cdot\mid\sqrt{x}\,\xi+\sqrt{r-x}\,u^0 )$ with Gaussian signal $u^0$, given a quenched variable $\xi$. 

$\bullet$ In contrast with the two previous free entropies, which are associated with scalar inference problems, $\iota(x)$ is the mutual information between signal and data of a \emph{high-dimensional}, yet tractable, problem: matrix denoising. In this inference problem the goal is to recover a \emph{generalised Wishart matrix} $\tilde\bS^0:=\tilde \bW^{0\intercal}\diag(\bv^0)\tilde\bW^0/\sqrt{kd}\in\mathbb{R}^{d\times d}$ given the matrix observation $\bY(x)=\sqrt{x}\,\tilde \bS^0+\bZ$. Here, $\tilde \bW^{0}\in\mathbb{R}^{k\times d}$ has i.i.d. standard Gaussian entries, the noise $\bZ$ is a GOE matrix (symmetric with upper triangular part made of entries i.i.d. from $\mathcal{N}(0,(1+\delta_{ij})/d)$) and $x$ is the SNR. The potential is then defined as
$\iota(x):=\lim_{d\to\infty}\frac1{d^2} I(\bY(x),\tilde \bS^0)$. It was conjectured \cite{barbier2022statistical,maillard2022perturbative} and proven \cite{matrix_inference_Barbier} that this mutual information is linked to the HCIZ integral:
\begin{talign*}
    \iota(x)&= \frac x2 \int s^2 \rho_{\tilde \bS^0}(s)ds -\lim_d \frac{1}{d^2}\ln\mathcal{Z}^{(1)}_{\rm HCIZ}(\sqrt{x}\,\tilde \bS^0,\bY(x)) ,
\end{talign*}
where $\rho_{\tilde \bS^0}$ is the limiting spectral density of $\tilde \bS^0$ as $d\to\infty$. The limit of the log-HCIZ integral is generally intractable in practice despite admitting a dimension-independent variational expression \cite{Matytsin_1994,guionnet2002large}. Luckily, the one needed in the present setting is explicit \cite{maillard2022perturbative,matrix_inference_Barbier}. The most convenient expression for numerical evaluation is based on the I-MMSE relation \cite{guo2005mutual,matrix_inference_Barbier} which requires an expression for the minimum mean-square error (MMSE).

Using the results of \cite{matrix_inference_Barbier}, the limiting MMSE for matrix denoising verifies
\begin{talign*}
    {\rm mmse}_S(x)&:=\lim_{d\to \infty}\frac{1}{d}\EE\|\tilde\bS^0-\EE[\tilde\bS^0\mid\bY(x)]\|^2\nonumber\\
    &\hspace{3pt}=\frac{1}{x}\big(1-\frac{4\pi^2}{3}\int \rho_{\bY(x)}(y)^3 dy\big).
\end{talign*}
Using this, $\iota(x)$ admits a compact expression:
\begin{talign*}
    \iota(x)= \frac{1}{4}\int_0^x {\rm mmse}_S(t)dt.
\end{talign*}

$\bullet$ Consider now a \emph{rectangular} matrix denoising problem with observations $\tilde\bY(x)=\sqrt{x/(pk)}\bU^0\bV^0+\bN/\sqrt{p}\in\mathbb{R}^{p\times d}$, where $\bU^0\in\mathbb{R}^{p\times k}$, $\bV^0\in\mathbb{R}^{k\times d}$ and $\bN\in\mathbb{R}^{p\times d}$ are all made of i.i.d. standard Gaussian entries. $\tilde\iota(x;\eta,\gamma)$ is then defined as the limit (when $d,k,p\to\infty$) of the mutual information $\frac1{pd}I(\tilde \bY(x);\bU^0\bV^0)$ while fixing $p/d\to\eta$, and $k/d\to\gamma$. Similarly to its symmetric version, the mutual information is computed by means of a ``rectangular spherical integral'' \cite{GUIONNET2023_rect}, see \eqref{eq:iota_v_deep}. The noise being Gaussian, we can again exploit the I-MMSE relation \cite{guo2005mutual}. The MMSE function for this problem \cite{troiani2022optimal,pourkamali2023rectangular} is
\begin{talign*}
    &{\rm mmse}(x;\eta,\gamma)\!:=\!\lim_{d}\frac{1}{pkd} \EE\|\bU^0\bV^0-\EE[\bU^0\bV^0\mid\tilde\bY(x)]\|^2\nonumber\\
    &\ \ =\frac{1}{x}\big[
    1-\int\big(\eta(\frac{1}{\eta}-1)^2y^{-2}\tilde\rho_{\tilde\bY(x)}(y) -\frac{\pi^2\eta}{3} \tilde\rho_{\tilde\bY(x)}(y)^3\big)dy
    \big].
\end{talign*}Here $\tilde\rho_{\tilde\bY(x)}$ is the limiting singular value density of $\tilde\bY(x)$, which is the so-called rectangular free convolution \cite{GUIONNET2023_rect,pourkamali2023rectangular} 
between the asymptotic singular value density of $\sqrt{x/(pk)}\bU^0\bV^0$ and a Marchenko-Pastur distribution of parameter $\eta$. The potential $\tilde\iota(x;\eta,\gamma)$ is then given by
\begin{talign*}
   \tilde\iota(x;\eta,\gamma)& =\frac{1}{2}\int_0^x {\rm mmse}(t;\eta,\gamma) dt.
\end{talign*}

\subsection{Shallow MLP}

Starting with $L=1$, our first result is a formula for the free entropy based on the Gaussianity assumption \eqref{eq:hypothesis_gaussian_test}. The strategy to evaluate it, based on the replica method, relies on identifying the sufficient statistics (order parameters), the free entropy being related to their large deviations rate function. This formula will therefore also give us access to their equilibrium values. We note that before the present work, no existing method could tackle linear-width NNs in the interpolation regime with a generic activation  \emph{even for this shallow case}. 

\vspace{5pt}\paragraph*{Order parameters.} 
In the definitions below, a superscript $*$ emphasises that the student is sampled at equilibrium, $\btheta\sim dP(\,\cdot\mid\mathcal{D})$, and that the thermodynamic limit \eqref{thermolim} is taken (even if not explicit).
All the OPs involve the target $\btheta^0$ and student $\btheta$ weights. By Bayes-optimality and the Nishimori identities (App.~\ref{app:nishiID}), the target weights can be equivalently replaced by $\btheta'\sim dP(\,\cdot\mid\mathcal{D})$ coming from another independent student. 

$\bullet$ $R_2^*\propto {\rm Tr}(\bW^{0\intercal}{\rm diag}(\bv^0)\bW^0\bW^{\intercal}{\rm diag}(\bv)\bW)$ measures the alignment between the teacher's and student's quadratic terms, which is non trivial with $n=\Theta(d^2)$ data even when the student is not able to reconstruct $\bW^0$ itself (i.e., to specialise). 

$\bullet$ $\mathcal{Q}^*(\mathsf{v})\propto \sum_{\{i \mid v_i^0 = \mathsf{v}\}} (\bW^{0}\bW^{\intercal})_{ii}$ measures the overlap between the teacher and student's inner weights that are connected to readouts with the same amplitude $\mathsf{v}$: $\mathcal{Q}^*(\mathsf{v})\neq 0$ signals that the student learns part of $\bW^0$. Thus, the \emph{specialisation transition} for the neurons connected to readouts with amplitude $\mathsf{v}$ is defined as 
\begin{talign}
\alpha_{\rm sp, \mathsf{v}}(\gamma):=\sup
\,\{\alpha \mid \mathcal{Q}^*(\mathsf{v}) =0\}  .  \label{eq:v-depSpecTrans}
\end{talign}
For non-homogeneous readouts, \emph{the} specialisation transition is defined as $$\alpha_{\rm sp}(\gamma):=\min_{\mathsf{v}}\alpha_{\rm sp, \mathsf{v}}(\gamma)=\min_{\mathsf{v}}\sup
\,\{\alpha \mid \mathcal{Q}^*(\mathsf{v}) =0\}.$$ 

Associated with these OPs, the ``hat variables'' $\hat{R}_2^*,\hat{\mathcal{Q}}^*(\mathsf{v})$ in Result~\ref{res:free_entropy_shallow} are conjugate OPs. Their meaning is that of effective fields (called ``cavity fields'' in spin glasses), which self-consistently determine the OPs through the replica symmetric saddle point equations given in the result.

To state our first result we need additional definitions. Let $\mathcal{Q}(\mathsf{v}),\hat{\mathcal{Q}}(\mathsf{v})\in\mathbb{R}$ for $\mathsf{v}\in {\rm Supp}(P_v)$, $\mathcal{Q} :=\{\mathcal{Q}(\mathsf{v}) \mid  \mathsf{v}\in {\rm Supp}(P_v)\}$ and similarly for $\hat{\mathcal{Q}}$. Let also (see \eqref{eq:g_func_first} for a more explicit expression of $g$)
\begin{talign}
        &g(x) := \sum_{\ell \ge  3}^\infty x^\ell{\mu_{\ell}^2}/{\ell !},\nonumber \\
        &K(x,\mathcal{Q}) := \mu_1^2 + {\mu_2^2} \,x/2 + \EE_{v\sim P_v}\,v^2 g(\mathcal{Q}(v)),\label{eq:covariance_vars_main}\\
        &K_{\rm d} := \mu_1^2+ {\mu_2^2}(1 + \gamma \bar{v}^2)/2 + g(1).\nonumber
\end{talign} 
    The physical meaning of $K(\,\cdot,\cdot\,)$, when evaluated at the equilibrium $R_2^*,\calQ^*$, is that of the covariance $K^*$ appearing in \eqref{eq:covLam} (i.e., the large  $d$ limiting covariance between two post-activations $\lambda_{\rm test}(\btheta^a),\lambda_{\rm test}(\btheta^b)$ evaluated from the same test input $\bx_{\rm test}$ but with weights $\btheta^a,\btheta^b$ i.i.d. from the posterior); $K_{\rm d}$ is instead their variance, which matches that of the target by Bayes-optimality.

 \vspace{5pt}\paragraph*{Replica symmetric formulas.}
We are ready to state the replica symmetric (RS) formula giving access to the equilibrium order parameters. From now on, we denote the joint $d,k,n\to\infty$ limit with rates \eqref{thermolim} simply by ``${\lim}$''.

\begin{result}[Replica symmetric free entropy for the MLP with $L=1$] \label{res:free_entropy_shallow}
Assume that $\mu_0=0$ in the Hermite decomposition \eqref{eq:HermiteSig}. Let the functional 
\begin{talign}
\tau(\mathcal{Q}) := {\rm mmse}_S^{-1}(1 - \EE_{v\sim P_v}v^2 \mathcal{Q}(v)^2). \label{eq:deftau}   
\end{talign}
The replica symmetric formula for the limiting free entropy ${\lim}\,f_n$ is $f_{\rm RS}^{(1)}( R_2^*,\hat{R}_2^*,\mathcal{Q}^*,\hat{\mathcal{Q}}^*)$
with RS potential $f^{(1)}_{\rm RS}=f^{(1)}_{\rm RS}( R_2,\hat{R}_2,\mathcal{Q},\hat{\mathcal{Q}})$ which, given $(\alpha,\gamma)$, reads
\begin{talign}
f^{(1)}_{\rm RS}&:= \phi_{P_{\rm out}}(K(R_2,\mathcal{Q});K_{\rm d}) + \frac{1}{4\alpha}(1 + \gamma \bar{v}^2-R_2) \hat{R}_2\nonumber
    \\
    &\qquad + \frac{\gamma}{\alpha}\EE_{ v\sim P_v}\big[
     \psi_{P_W}(\hat{\mathcal{Q}}(v))-\frac{1}{2}\mathcal{Q}(v) \hat{\mathcal{Q}}(v) \big]\nonumber \\
     &\qquad +\frac{1}\alpha\big[\iota(\tau(\mathcal{Q})) - \iota(\hat R_2 + \tau(\mathcal{Q}))\big].\label{eq:fRS}
\end{talign}
The order parameters equilibrium values $( R_2^*,\hat{R}_2^*,\mathcal{Q}^*,\hat{\mathcal{Q}}^*)$ are obtained from the RS saddle point equations \eqref{eq:NSB_equations_generic_ch} derived from the extremisation condition $\nabla f^{(1)}_{\rm RS}=\mathbf{0}$, as a solution (there may be more than one) maximising $f^{(1)}_{\rm RS}$. 
\end{result}


Result~\ref{res:free_entropy_shallow} provides $R_2^*,\calQ^*$ through the solution of a tractable variational problem. Consequently, under our joint-Gaussianity hypothesis~\eqref{eq:hypothesis_gaussian_test} on $(\lambda_{\rm test}(\btheta^a))_{a=0}^s$ with i.i.d. $\btheta^a\sim dP(\,\cdot\mid \mathcal{D})$ for $a=1,\ldots,s$, we can also access their asymptotic covariance (and thus their law) given by 
\begin{talign}
     (\mathbf{K}^*)_{ab}= K^*+(K_{\rm d}-K^*)\delta_{ab}, \ K^*=K(R_2^*,\calQ^*).\label{eq:covLam_2}
\end{talign}
The Bayes error can then be computed as in App.~\ref{app:gen_err}.

\begin{result}[Bayes generalisation error]
\label{res:gen_error} Let $(\lambda^a)_{a\ge 0}\sim \mathcal{N}(\mathbf{0},\mathbf{K}^*)$ with covariance \eqref{eq:covLam_2}, $y_{\rm test}|\lambda^0\sim  P_{\rm out}(\,\cdot\mid \lambda^0)$. Assume $\mathcal{C}$ has series expansion $\mathcal{C}(y,\hat y)=\sum_{i\ge 0} c_i(y)\hat y^i$. The RS formula for the $\smash{\lim\varepsilon^{\mathcal{C},\mathsf{f}}}$ of the Bayes error \eqref{eq:Bayes_error_def} is
\begin{talign}\label{eq:Bayes_error_result}
   \EE_{(\lambda^a)}\EE_{y_{\rm test}\mid\lambda^0} \sum_{i\ge 0} c_i(y_{\rm test})\prod_{a=1}^i\mathsf{f}(\lambda^a).
\end{talign}
Letting $\EE[\,\cdot\mid \lambda]=\int dy \,(\,\cdot\,)\, P_{\rm out}(y\mid \lambda)$, the RS formula for the $\smash{\lim \varepsilon^{\rm opt}}$ of the Bayes-optimal mean-square generalisation error \eqref{eq:gen_error_def} is 
  \begin{talign}
\EE_{(\lambda^0,\lambda^1)}\big(\EE[y^2\mid \lambda^0] - \EE[y\mid \lambda^0]\EE[y\mid \lambda^1] \big).\label{eq:gen_err_result}
  \end{talign}
\end{result}
\noindent This result assumed $\mu_0=0$; see App.~\ref{app:non-centered} if $\mu_0\neq 0$. We remind the reader that Sec.~\ref{sec:numerics} provides a generalisation of the theory able to tackle structured input data (valid also for the deep case $L\ge 2$). 

The presence of the HCIZ matrix integral in our replica formulas suggests that the usual asymptotic decoupling of the finite marginals of the posterior in terms of products of the single-variable marginals does not occur here, in contrast with standard Bayes-optimal inference problems \cite{barbier2022strong}. In the related context of matrix denoising, this may explain why the approximate message-passing algorithms proposed in \cite{parker2014bilinear,krzakala2013phase,kabashima2016phase} are, as stated by the authors, not properly converging nor matching their corresponding theoretical predictions based on the cavity method, as it relies on such decouplings.

\begin{remark}\label{rmk:readouts}
No OP related to the readout weights appears in our results. The reason is the following. 
The $kd=\Theta(d^2)$ inner weights $\bW^0$ and $n=\Theta(d^2)$ data are
overwhelmingly many compared to the $k$ unknowns $\bv^0$, which thus contribute trivially to the leading order of the  thermodynamic equilibrium quantities we aim for. Let us prove that the mutual information stays the same at leading order if the readouts are fixed to $\bv^0$ rather than learnable/unknown. By the chain rule for mutual information $I((\bW^0,\bv^0);\mathcal{D})=I(\bW^0;\mathcal{D}\mid \bv^0)+I(\bv^0;\mathcal{D})$. Moreover $I(\bv^0;\mathcal{D})= H(\bv^0)- H(\bv^0\mid \mathcal{D})$. For a discrete-valued $\bv^0$ both these Shannon entropies are non-negative. Additionally $H(\bv^0)=O(k)$. Because $H(\bv^0)\ge H(\bv^0\mid \mathcal{D})$, then $H(\bv^0\mid \mathcal{D})=O(k)$ too. Therefore, in the limit~\eqref{thermolim},
\begin{talign*}
\frac1{n}I((\bW^0,\bv^0);\mathcal{D})=\frac1{n}I(\bW^0;\mathcal{D}\mid \bv^0)+O(1/d)  
\end{talign*}
and similarly for the free entropy. The same also holds for the generalisation error given its link with the mutual information, see \cite{barbier2019glm}. The argument can be extended to continuous-valued readouts and $L\ge 2$.

Another way to understand this is through the symmetry of the NNs under permutation of their $k$ hidden neurons. It implies that only the law of $\bv^0$ matters. Consequently, if one draws $\bv'$ from the correct $P_v$ and fixes it in the student (thus only learning $\bW$), it will have the same law as $\bv^0$ (up to small fluctuations) and is therefore equally good as $\bv^0$ when $d\gg 1$. This implies that in the Bayes-optimal setting, knowing $P_v$ is equivalent to $\bv^0$ for large $d$. 
For additional illustration, in the paper we test our theory with numerical experiments both with fixed and learnable readouts, as explicit from the caption of each figure. Moreover, \figurename~\ref{fig:gen_error_learnable_readouts} of App.~\ref{app:hardness} (obtained with learnable readouts), to be compared directly with \figurename~\ref{fig:gen_error_gauss} (right) and \ref{fig:v_non-constant} (bottom), obtained with fixed readouts, is showing that equilibrated Bayesian NNs achieve the same generalisation performance independently of whether the readouts are trainable or fixed to the truth. The same holds for $L=2$ (see \figurename~\ref{fig:gen_error_learnable_readouts_L=2}). \figurename~\ref{fig:gaussianPostactiv} is another confirmation because the theory there, which describes well the empirical distribution of post-activations over test samples, is done fixing directly the readouts, while they are trained in the experiments for this figure. However, the readouts being fixed or learnable does influence the learning dynamics, see e.g. the difference for ADAM in \figurename~\ref{fig:hardness_main} and \ref{fig:hardness_adam_runs}, but its theoretical analysis is out of the scope of the paper.
\end{remark}

 \vspace{5pt}\paragraph*{Main steps of the replica approach.}
 Before moving to the results for deep networks, let us provide a summary of the main steps of the derivation of Result~\ref{res:free_entropy_shallow}, highlighting the conceptual novelties (the main ideas are similar for the deeper architectures we consider later on). More details are given in Sec.~\ref{sec:theory} and App.~\ref{app:shallowMLP}. We adopt the replica method, but encounter and resolve
a major technical obstacle that is absent in standard applications.

\textbf{1) Standard replica setup.} Having quenched the readouts, which is possible by the last remark, the replica trick
$\lim_{n\to \infty} \frac{1}{n}\mathbb{E}\ln Z = \lim_{s\to 0^+}\lim_{n\to \infty} \frac{1}{ns}\ln \mathbb{E}Z^s$ \cite{mezard1987spin} introduces $s$ independent posterior samples of the inner weights (replicas) $\{\bW^a\}_{a=1}^s$
alongside the teacher's $\bW^0$. Under our main Gaussian hypothesis \eqref{eq:hypothesis_gaussian_test}, the key quantity governing learning is the covariance
of the replicated post-activations
$\lambda^a = \frac{1}{\sqrt{k}}\bv^{0\intercal}\sigma(\frac1{\sqrt{d}}\bW^a \bx)$
under the shared input $\bx$: $K^{ab}=\EE_\bx[\lambda^a \lambda^b]$. Using the Hermite expansion of $\sigma$ and Mehler's formula~\eqref{eq:mehler}, this covariance decomposes as an infinite series of overlap order parameters
$R_\ell^{ab} := \frac{1}{k}\sum_{i,j}v^0_i v^0_j (\Omega^{ab}_{ij})^\ell$,
where $\Omega^{ab}_{ij} := \bW^{a\intercal}_i \bW^b_j/d$.

\textbf{2) Reduction to fundamental order parameters.}
A first non-trivial step is reducing this infinite family of overlaps $(R_\ell^{ab})$ to a small
manageable set. The first ingredient is to assume that specialization of neurons can only happen in a one-to-one manner by Bayes-optimality (i.e., a neuron of the student cannot align with more than one teacher neuron). Consequently, all but one entry per row or column of the overlap matrix $(\Omega^{ab}_{ij})_{i,j\le k}$ are $O(1/\sqrt{d})$. The second ingredient is that the permutation symmetry among neurons can be broken in a way that the $k$ entries of the overlap matrix that are potentially $O(1)$ are put on the diagonal $(\Omega^{ab}_{ii})_{i\le k}$, which amounts to select the ``natural'' permutation of student neurons w.r.t. the teacher (and similarly in any given layer in the $L\ge 2$-hidden layers case). The permutation symmetry can then be discarded, which is justified by the fact that the number of such permutations grows much slower than $O(\exp(n))$, and therefore plays no role at leading exponential order in the large system limit. The third ingredient is that the ``Wishart-like'' overlap matrix -- with $O(1)$ diagonal and $O(1/\sqrt d)$ off-diagonal -- taken at the $\ell$-th Hadamard power $(\Omega^{ab}_{ij})^\ell$ can be considered diagonal in
the thermodynamic limit~\eqref{thermolim} for $\ell \geq 3$: the off-diagonal becomes too weak to be relevant. This procedure leaves only three types of OPs: the
linear overlap $R_1^{ab}\to 1$ (i.e., the linear component of the target is perfectly
learnable at $n=\Theta(d^2)$ and requires no further tracking), the quadratic
overlap $R_2^{ab}$ associated with the second Hermite component, and the
overlaps $\Omega^{ab}_{ii}=:\calQ^{ab}(\mathsf{v})$ for any hidden unit $i \in [k]$ connected to a readout with amplitude $\mathsf{v}$ (by symmetry among these neurons). This yields the reduction of OPs via $R_\ell^{ab} \approx \frac{1}{k}\sum_{i}(v^0_i)^2 (\Omega^{ab}_{ii})^\ell \approx \EE_{v\sim P_v}v^2 \calQ^{ab}(v)^\ell$ for $\ell \ge 3$.
The covariance then takes the compact form $K^{ab} = K(R_2^{ab}, \calQ^{ab}( \cdot))$ given in~\eqref{eq:covariance_vars_main}.

\textbf{3) The entropic obstacle for matrix degrees of freedom.}
In a standard narrow-network replica calculation, the weights
are few vectors. Consequently the overlap order parameters are few neuron-neuron overlaps and their entropic
degeneracy factorises over components, yielding a scalar free entropy term of the type
$\psi_{P_W}$. Here the situation is fundamentally different: the second-order
overlap $R_2^{ab} = \frac{1}{d^2}\mathrm{Tr}\,\bS^a_2 \bS^b_2$
depends on the \emph{matrix}
$\bS^a_2 := \bW^{a\intercal}\mathrm{diag}(\bv^0)\bW^a/\sqrt{k}\in\mathbb{R}^{d\times d}$.
Fixing $R_2^{ab}$ therefore imposes a \emph{quartic} constraint on the weight
entries $(W^a_{i\alpha})$, rather than the quadratic constraints arising in
standard problems. A direct computation of the entropy of $(R_2^{ab})$ \emph{conditional} on $\bm{\mathcal{Q}}:=\{\mathcal{Q}^{ab}(\mathsf{v})\mid \mathsf{v}\in\mathsf{V}, a\le b\}$, which is of the form
\begin{talign*}
     \ln\int dP((\bS_2^a)_a\mid\bm{\mathcal{Q}}) \prod_{a\leq b}^{0,s}\delta(d^2R_2^{ab}-{\Tr \,\bS_{2}^{a} \bS_{2}^b})
\end{talign*}
for some complicated measure $dP((\bS_2^a)_a\mid\bm{\mathcal{Q}})$, is therefore out of reach via standard methods. In contrast, the entropy of $\bm{\mathcal{Q}}$ is easy to compute using Gaussian tricks.

\textbf{4) Measure relaxation by moment matching.}
The resolution is to shift perspective: rather than integrating over the weight
entries $W^a_{i\alpha}$ subject to the constraints on $R_2^{ab}$, we work
directly with the matrices $\bS^a_2$ as the fundamental degrees of freedom. Their
true conditional law $dP((\bS_2^a)_a\mid\bm{\mathcal{Q}})$ is too complex -- it is a
coupled, non-Gaussian measure over generalized Wishart-like matrices. We replace it
by a simpler effective measure $d\tilde P((\bS_2^a)_a\mid\bm{\mathcal{Q}})=d\tilde P((\bS_2^a)_a\mid\tau(\bm{\mathcal{Q}}))$. The effective measure enforces what we claim to be the only relevant characteristics in the large system limit of the original measure $dP((\bS_2^a)_a\mid\bm{\mathcal{Q}})$, but it has the advantage of being tractable. It depends on a tilt parameter $\tau(\bm{\mathcal{Q}})$ determined by enforcing exact \emph{moment matching}: the second moment
$\mathbb{E}_{\tilde{P}}\mathrm{Tr}\,\bS^a_2 \bS^b_2$ must reproduce
the value computed under the true measure, which is accessible. The measure simplification we introduce differs crucially from the earlier approach~\cite{sakata2013} (developed for matrix factorization), which considers a \emph{factorised
Gaussian effective measure}, thereby discarding the Wishart-type correlations among
entries of $\bS^a_2$. Our Wishart-based relaxation retains the essential matrix correlations while
remaining tractable, and leads to improved agreement with experiments.

\textbf{5) Spherical integration.}
With the effective measure in place, the entropic contribution of $R_2^{ab}=R_2$ reduces -- after
standard replica algebra, Hubbard--Stratonovich linearisation of inter-replica
couplings, and the $s\to 0^+$ limit under the replica symmetric
ansatz~\cite{mezard1987spin} -- to the free entropy of a
\emph{rotationally invariant matrix denoising problem}. This free entropy is computable via the
HCIZ spherical integral, yielding the two $\iota(\cdot)$ terms
in $f^{(1)}_{\mathrm{RS}}$~\eqref{eq:fRS}. The function $\mathrm{mmse}_S(\cdot)$ is the MMSE of this auxiliary denoising problem, and the
moment matching equation relating true to effective measure is \eqref{eq:deftau}. Via this equation, the matrix denoising problem interacts with the specialisation order parameter $\calQ^{ab}(\,\cdot\,)=\calQ(\,\cdot\,)$. The entropy of $\calQ$ yields instead the free entropy of a standard scalar Gaussian denoising problem (one per $\mathsf{v}$), i.e. the terms $\psi_{P_W}$.

\textbf{Summary of the structure of the replica symmetric free entropy.}
The resulting free entropy $f^{(1)}_{\mathrm{RS}}$ admits a transparent
decomposition. The term $\phi_{P_{\mathrm{out}}}$ is the standard output
channel free entropy, identical in form to what appears in the GLM calculation \cite{barbier2019glm}. The terms $\psi_{P_W}$
are the scalar Gaussian channel free entropies encoding the prior on the
weights and governing the specialisation transition(s). The two $\iota$
terms encode the matrix nature of the problem through the auxiliary matrix denoising
problem, and are the only place where the HCIZ integral enters. The interplay
between the mean-field specialisation mechanism (encoded in $\psi_{P_W}$)
and the matrix model (encoded in $\iota$) is what generates the rich phase
diagram of the model, and constitutes the principal
conceptual novelty of the approach with respect to prior works that treated
these two aspects separately or only partially.

\subsection{Two hidden layers MLP} \label{res:2layers}

For $L=2$ we consider activations without 0th and 2nd Hermite components, see (\hyperlink{H:H2}{$H_2$}). The results are obtained by an expansion of the nested activations in the Hermite basis. This produces different terms that can be interpreted as equivalent sub-networks with ``effective'' readouts and inner weights built as combinations of the original ones, as detailed in Sec.~\ref{sec:replicasL=2}. When the linear component of the last activation is involved, the readouts $\bv$ combine with the second layer inner weights and give rise to ``effective readouts'' $\bv^{(2)}:=\bW^{(2)\intercal}\bv/\sqrt{k_2}$ that act on the non-linear first layer. By binning through finite discretisation the distribution of the components of this vector, we denote the admitted amplitudes as $\mathsf{v}^{(2)}$. Similarly, when the linear component of the first layer is considered, the two sets of inner weights combine in an effective layer with weights $\bW^{(2:1)}:=\bW^{(2)}\bW^{(1)}/\sqrt{k_1}$, which can be reconstructed partly independently of its factors $\bW^{(2)}$ and $\bW^{(1)}$, and thus comes with an OP.

\vspace{5pt}\paragraph*{Order parameters.} Already with two hidden layers, the OPs detailed below describe a much richer phase diagram than in the shallow case. Until now it was unclear what OPs should be tracked.

$\bullet$ ${\mathcal{Q}}_{1}^*(\mathsf{v^{(2)}})
\propto \sum_{\{i \mid v^{(2)0}_i = \mathsf{v^{(2)}}\}} (\bW^{(1)0}\bW^{(1)\intercal})_{ii}
$ is the overlap between teacher and student's first layer weights connected to the effective readouts $\bv^{(2)}$ with amplitude $\mathsf{v^{(2)}}$. As in Remark~\ref{rmk:readouts}, the vector $\bv^{(2)}$ can be treated as quenched on the teacher's. In virtue of this, from its definition, $\mathsf{v}^{(2)}$ has Gaussian distributed entries by the central limit theorem.

$\bullet$ $\mathcal{Q}_{2}^*(\mathsf{v},\mathsf{v}^{(2)})\propto\sum_{\{i,j \mid v_i^0 = \mathsf{v}, v^{(2)0}_j = \mathsf{v^{(2)}}\}}W^{(2)0}_{ij} W^{(2)}_{ij} $ is the overlap for the second layer. It is labelled by two values. The first, $\mathsf{v}$, as for the shallow case, is the value of a readout. It takes into account the learning inhomogeneity along the output dimension ($i\le k_2$) of the second layer weight matrix induced by the readouts $\bv$. $\mathsf{v}^{(2)}$ is instead the \emph{same} variable labelling ${\mathcal{Q}}^*_{1}(\mathsf{v^{(2)}})$. It captures the inhomogeneity along the input dimension ($j\le k_1$) of the second layer induced by the inhomogeneity of the first layer output, itself induced by (and therefore labelled according to) the effective readouts $\bv^{(2)}$. Notice that this implies a non-trivial feedback loop of interactions: inhomogeneities of $\bW^{(2)}$ influence $\bW^{(1)}$ via $\bv^{(2)}$, and at the same time the inhomogeneities in $\bW^{(1)}$'s rows influence the columns of $\bW^{(2)}$ directly. 

We wish to emphasise a conceptually important point. $\mathcal{Q}_{2}^*(\mathsf{v},\mathsf{v}^{(2)})$ being a matrix may lead to believe that it does not help in reducing the dimensionality of the problem, because the ``microscopic degrees of freedom'' are weight \emph{matrices}, too. However,  $\mathsf{v}$ and $\mathsf{v}^{(2)}$ are indexing intensive, $d$-independent dimensions. Indeed, the binning of $\mathcal{K}_{12}:=\{1,\ldots,k_1\}\times \{1,\ldots,k_2\}$ in terms of the non-overlapping sets $\{i,j \mid v_i^0 = \mathsf{v}, v^{(2)0}_j = \mathsf{v^{(2)}}\}$ entering $\mathcal{Q}_{2}^*$'s definition (i.e., the mapping from $\mathcal{K}_{12}$ to $\{\mathsf{v}\}\times \{\mathsf{v}^{(2)}\}$) is done as follows. Firstly, $\mathcal{K}_{12}$ is  partitioned into finitely-many ``macroscopic'' sets; secondly the thermodynamic limit $d\to+\infty$ is taken, see \eqref{thermolim}; finally, only after this limit is the number of bins allowed to diverge. This implies that each set always includes a number of terms growing to infinity as $d^2$ times a small constant. Consequently, $\mathcal{Q}_{2}^*(\mathsf{v},\mathsf{v}^{(2)})$ (or any OP function with continuous argument) is a proper ``macroscopic (or intensive)'' OP summarising the behaviour of a large assembly of degrees of freedom, for each pair of arguments. Dimensionality reduction therefore takes place and justifies the use of saddle point integration w.r.t. to the OPs when evaluating the (replicated) log-partition function in Sec.~\ref{sec:theory}. 


$\bullet$ Lastly, ${\mathcal{Q}}^*_{2:1}(\mathsf{v})\propto \sum_{\{i \mid v_i^0 = \mathsf{v}\}} (\bW^{(2:1)0}\bW^{(2:1)\intercal})_{ii}$ is the teacher-student overlap for specific rows of $\bW^{(2:1)}$. This OP arises from the linear term in the Hermite expansion of the inner activation. It is needed because the product $\bW^{(2:1)}$ between first and second layer weights can in principle be learned partly independently from $\bW^{(1)}$ and $\bW^{(2)}$. Observe that $\bW^{(2:1)}$ ``connects'' the input directly to the output, which is why it is labelled only by the readout values $\mathsf{v}$.

The reader eager to already gain intuition on the behaviour of these functional ``vector'' and ``matrix order parameters'' can look at \figurename~\ref{fig:3LNN_NSB_FSB} and \ref{fig:matrixOPs}. We remark that extrapolating to the linear-width setting the replica techniques successful for narrow NNs also yields overlap matrix order parameters, but with a prohibitive dimension. However, if in the shallow case one takes $k\to\infty$ \emph{after} $d\to\infty$, simple parametrisations of the $k\times k$ overlaps allow to solve it~\cite{schwarze1993learning}. This double limit is however \emph{not} an extensive-width limit and, indeed, the resulting formulas are similar to those for GLMs \cite{aubin2018committee,baldassi2019properties,barbier2025generalization}.

For $L=2$ specialisation transitions can happen layer-wise: we defined the specialisation transitions as
\begin{talign*}
 \alpha_{{\rm sp},l}:=\sup\,\{\alpha \mid \calQ_l^* \equiv 0 \} \ \ \text{for} \ \ l=1,2  
\end{talign*}
where $\mathcal{Q}_l^* \equiv 0$ means the constant null function. Keep in mind that a non-vanishing overlap ${\mathcal{Q}}^*_{2:1}$ entails another kind of learning mechanism than specialisation.

Our result for the two hidden layers MLP requires the following function: letting $v^{(2)}\sim\calN(0,1), v\sim P_v$,
\begin{talign*}
    &K^{(2)}(\bar \calQ):=\mu_1^4 
    + \mu_1^2 \EE_{v^{(2)}} (v^{(2)})^2 g \big(\mathcal{Q}_{1}(v^{(2)}) \big) \nonumber\\
    &
    + \EE_{v } v^2 g \Big(\mu_1^2 \mathcal{Q}_{2:1}(v) +\EE_{v^{(2)} }\mathcal{Q}_{2}(v,v^{(2)}) g\big(\mathcal{Q}_{1}(v^{(2)})\big)\Big),
\end{talign*}with $\bar \calQ:=\{\mathcal{Q}_{1},\calQ_2,\calQ_{2:1}\}$, which are functions of $\mathsf{v},\mathsf{v}^{(2)}$. Analogous notations hold for the conjugate OPs $\hat{\mathcal{Q}}_{1},\hat{\calQ}_2, \hat{\calQ}_{2:1}$. The meaning of $K^{(2)}({\bar \calQ}^*)$ evaluated at equilibrium is, as in the shallow case, that of asymptotic covariance between different replicas of the post-activation with same test input entering \eqref{eq:covLam}:
\begin{talign}
     (\mathbf{K}^*)_{ab}= K^*+(1-K^*)\delta_{ab}, \ K^*=K^{(2)}({\bar \calQ}^*).\label{eq:covLam_3}
\end{talign}
That the variance is $1$ is a consequence of our convention $\EE_{z\sim \mathcal{N}(0,1)}\sigma(z)^2=1$ which greatly simplifies notations in the deep case, see App.~\ref{app:replicas_deep} for an explanation.


\vspace{5pt}\paragraph*{Replica symmetric formula.}
Recall the definitions of ${\rm mmse}(x;\eta,\gamma),\tilde\iota(x;\eta,\gamma)$ in Sec.~\ref{sec:result}. The equilibrium OPs are determined by the following RS formula:
\begin{result}[Replica symmetric free entropy for the MLP with $L=2$]\label{res:free_entropy_L=2}
Consider an activation with $\mu_0=\mu_2=0$ in \eqref{eq:HermiteSig} and $\EE_{z \sim \mathcal{N}(0,1)} \sigma(z)^2 = 1$. Let $v^{(2)}\sim\calN(0,1), v\sim P_v$. Define ${\rm mmse}_\mathsf{v}:={\rm mmse}(\cdot\,;\,\gamma_2P_v(\mathsf{v}),\gamma_1)$, $\tilde\iota_\mathsf{v}:=\tilde\iota(\cdot\,;\,\gamma_2P_v(\mathsf{v}),\gamma_1)$ and $\tau_\mathsf{v}=\tau_\mathsf{v}(\calQ_1,\calQ_2)$ solves
\begin{talign*}
    {\rm mmse}_\mathsf{v}(\tau_\mathsf{v}) := 1- \EE_{v^{(2)}}\mathcal{Q}_{2}(\mathsf{v},v^{(2)}) \mathcal{Q}_{1}(v^{(2)}).
\end{talign*}

The RS formula for the limiting free entropy ${\lim}\,f_n$ for the MLP with $L = 2$ hidden layers is given by $f_{\rm RS}^{(2)}(\mathcal{Q}_{1}^*,\hat{\mathcal{Q}}_{1}^*,\mathcal{Q}_{2}^*,\hat{\mathcal{Q}}_{2}^*,\mathcal{Q}_{2:1}^*,\hat{\mathcal{Q}}_{2:1}^*)$
with RS potential 
\begin{talign*}
    &f^{(2)}_{\rm RS}:=  \phi_{P_{\rm out}}(K^{(2)}(\bar{\calQ});1)\nonumber\\
    &+\frac{\gamma_1}{\alpha} \EE\big[\psi_{P_{W_1}}(\hat{\mathcal{Q}}_{1}(v^{(2)}))-\frac{1}{2} \mathcal{Q}_{1}(v^{(2)}) \hat{\mathcal{Q}}_{1}(v^{(2)}) \big]\nonumber\\
    & +\frac{\gamma_1 \gamma_2}{\alpha} \EE\big[\psi_{P_{W_2}}(\hat{\mathcal{Q}}_{2}(v, v^{(2)})) -\frac{1}{2} \mathcal{Q}_{2}(v,v^{(2)}) \hat{\mathcal{Q}}_{2}(v,v^{(2)})) \big] \nonumber\\
    &+\frac{\gamma_2}{\alpha}\EE \big[\frac{\hat{\calQ}_{2:1}(v)}{2}(1-{\calQ}_{2:1}(v))-\tilde\iota_v(\tau_v+\hat\calQ_{2:1}(v))+\tilde\iota_v(\tau_v)\big].
\end{talign*}
The order parameters equilibrium values $(\mathcal{Q}_{1}^*,\hat{\mathcal{Q}}_{1}^*,\mathcal{Q}_{2}^*,\ldots)$ are obtained from the RS saddle point equations \eqref{eq:deep_Saddlepoint_eq} derived from the extremisation condition $\nabla f^{(2)}_{\rm RS}=\mathbf{0}$, as a solution (there may be more than one) maximising $f^{(2)}_{\rm RS}$.
\end{result}

Deducing the Bayes error is done as in the shallow case: from Result~\ref{res:free_entropy_L=2} we get ${\bar \calQ}^*$ and thus the covariance $\mathbf{K}^*$ given by \eqref{eq:covLam_3}, which simply replaces \eqref{eq:covLam_2} in Result~\ref{res:gen_error}.

\subsection{Three or more hidden layers MLP} 

\vspace{5pt}\paragraph*{Order parameters.} For $L \ge 3$ we consider activations verifying (\hyperlink{H:H3}{$H_{3}$}). In this setting, our theory predicts specialisation of all layers as the only non-trivial learning mechanism. Accordingly, the OPs are:

$\bullet$ $Q^*_l \propto \Tr( \bW^{(l)0}\bW^{(l)\intercal})$ for $l\le L-1$ are the teacher-student layer-wise overlaps. They are simple scalars rather than functions: indeed, the neurons in all layers but the last enter the theory in a symmetric way, such that we can freely sum over their indices.

$\bullet$ ${\mathcal{Q}}_{L}^*(\mathsf{v})
\propto \sum_{\{i \mid v^{0}_i = \mathsf{v}\}} (\bW^{(L)0}\bW^{(L)\intercal})_{ii}
$ is the overlap between teacher and student's $L$th layer weights connected to the readouts $\bv^{0}$ with amplitude $\mathsf{v}$. As before, weights connected to larger readouts are learned from less data.

For the considered class of activation functions, a single specialisation transition occurs jointly for all layers at $$\alpha_{{\rm sp}}:=\sup
\,\{\alpha \mid Q_1^* =\cdots =Q_{L-1}^*=0 \ {\rm and} \ \calQ_L^*\equiv 0\}.$$

Redefining $\bar{\calQ}:=\{(Q_l)_{l\le L-1}, \calQ_L\}$, and letting $v\sim P_v$, we introduce 
\begin{talign*}
    &K^{(L)}(\bar{\calQ})  := \EE_{v} v^2 g\Big(\calQ_L(v)g \big(Q_{L-1} g(\cdots Q_{2} g(Q_1) \cdots ) \big) \Big).
\end{talign*}
The asymptotic covariance $\bK^*$ between replicas of the post-activation with same test input is of the form \eqref{eq:covLam_3} with $K^*=K^{(L)}({\bar \calQ}^*)$. 

\vspace{5pt}\paragraph*{Replica symmetric formula.} The equilibrium order parameters and Bayes error are derived from the replica symmetric formula below, which applies to MLPs with any number of layers as long as $\sigma$ verifies (\hyperlink{H:H3}{$H_{3}$}) and has normalised variance.

\begin{result}[Replica symmetric free entropy for the MLP with arbitrary $L$]\label{res:free_entropy_deep}
Consider an activation $\sigma$ with $\mu_0=\mu_1=\mu_2=0$ in \eqref{eq:HermiteSig} and such that $\EE_{z \sim \mathcal{N}(0,1)} \sigma(z)^2 \!= \!1$. The replica symmetric formula for the limiting free entropy ${\lim}\,f_n$ for the MLP with $L$ hidden layers is given by $f_{\rm RS}^{(L)}(Q_1^*,\hat{Q}_1^*, \ldots,Q_{L-1}^*,\hat{Q}_{L-1}^*,\calQ_{L}^*,\hat{\calQ}_{L}^*)$
with RS potential 
\begin{talign*}
    &f^{(L)}_{\rm RS}:=  \phi_{P_{\rm out}}( K^{(L)}(\bar{\calQ});1)\nonumber\\
    &+\frac{\gamma_{L-1} \gamma_L}{\alpha} \EE_{v\sim P_v}\big[\psi_{P_{W_L}}(\hat{\calQ}_{L}(v)) -\frac{1}{2} \calQ_{L}(v) \hat{\calQ}_{L}(v) \big]\nonumber\\
    & +\sum_{l=1}^{L-1}\frac{\gamma_{l-1} \gamma_l}{\alpha} \big[\psi_{P_{W_l}}\!(\hat{Q}_{l}) -\frac{1}{2} Q_{l} \hat{Q}_{l} \big] ,
\end{talign*}
where $\gamma_0 := 1$. The order parameters equilibrium values $\ldots^*$ are obtained from the RS saddle point equations 
derived from the extremisation condition $\nabla f^{(L)}_{\rm RS}=\mathbf{0}$, as a solution (there may be more than one) maximising $f^{(L)}_{\rm RS}$.
\end{result}

The Bayes error follows by plugging $\bK^*$ in Result~\ref{res:gen_error}. Our results provide a precise quantitative theory for the sufficient statistics and generalisation capabilities of shallow and deep Bayesian MLPs with data generated by a random MLP target with matched architecture, for broad classes of activations and weight distributions.

\begin{remark}\label{rmk:quadratic}
For $L=1$ we conjecture that our theory is \emph{exact} for activations $\sigma$ with $\mu_2=0$. This is strengthened by a partial proof in App.~\ref{sec:partial_proof}. The case $\mu_2\neq 0$ is special as it involves the HCIZ integral with possibly approximative steps, see the discussion in App.~\ref{app:structured_S2}. When the theory does not rely on matrix integrals, the assumptions we make and which we believe are exact are mostly (and in order) $(i)$ the Gaussian hypothesis \eqref{eq:hypothesis_gaussian_test} on pre-activations which can be accurately tested; $(ii)$ that entries of Wishart-like overlap matrices can be considered small w.r.t. to their diagonal when taken at a large enough power; $(iii)$ the identification and indexing of the OPs as well as their concentration in the thermodynamic limit \eqref{thermolim} (i.e., replica symmetry), which is justified in the Bayes-optimal setting we consider \cite{barbier2021overlap,barbier2022strong}.

    For NNs with arbitrary $L$ we are confident that  Result~\ref{res:free_entropy_deep} is exact as, again, matrix integrals do not appear. See \figurename~\ref{fig:deep-tanh} that confirms its high accuracy. Another nice property is its simplicity. Even more so if $P_v=\delta_1$ when all OPs are scalars. Yet, it takes into account all key aspects of the model: its depth, the linear-width of the layers and the interpolation regime. Consequently, despite not capturing all intricacies emerging when considering a more general $\sigma$, it has a high pedagogical value.
    
    Beyond these cases, when matrix integrals appear in the formulas, due to the unconventional nature of their derivation we cannot confidently assess nor discard their exactness despite their excellent match with numerics. One reason is that it is numerically difficult to test our theory against the rigorous result \cite{xu2025fundamental} for the special case $L=1,\sigma(x)= x^2,P_W=\mathcal{N}(0,1)$ that they cover. When numerically solving the extremisation of \eqref{eq:fRS}, the saddle point equations seem to predict a maximiser at $\calQ(\mathsf{v})>0$ when $\gamma\lesssim 1$. The equations of \cite{xu2025fundamental} instead match the universal branch of the theory, i.e., $\mathcal{Q}(\mathsf{v})=0\ \forall\ \mathsf{v}$, for any $(\alpha,\gamma)$. Yet, we cannot confidently discard the exactness of the theory because the difference between the correct free entropy and the predicted one never exceeds $\approx 1\%$: our RS potential is \emph{very} flat in $\mathcal{Q}$. It could be that the true maximiser is at $\mathcal{Q}(\mathsf{v})=0$ even when $\gamma\lesssim 1$, and that we observe otherwise due to numerical errors. Indeed, evaluating the spherical integrals $\iota(\,\cdot\,)$ in $f^{(1)}_{\rm RS}$ is challenging, in particular when $\gamma$ is small. Actually, for $\gamma \gtrsim 1$ we correctly get that $\mathcal{Q}(\mathsf{v})=0$ is the maximiser.
\end{remark}

\section{Testing the theory, and algorithmic insights}\label{sec:numerics}

\paragraph*{Experimental setting.}

In this section we compare our theory with simulations. For all experiments but the ones in the dedicated paragraph on structured data, we use standard Gaussian input vectors $\bx_\mu\sim\mathcal{N}(\mathbf{0},I_d)$. We tested both the case of frozen and learnable readouts. For the equilibrium values obtained through sampling algorithms it makes no difference, as explained in Remark~\ref{rmk:readouts} and App.~\ref{subsec:learnable_readouts}, and further tested in \figurename~\ref{fig:gaussianPostactiv} and \ref{fig:gen_error_learnable_readouts}. For the ADAM optimiser we tested, this can change its dynamics but the overall conclusions remain the same. 

\begin{figure}[t!!]
\begin{center}
\centerline{\includegraphics[width=1\linewidth,trim={0cm 0cm 0cm 0.4cm},clip]{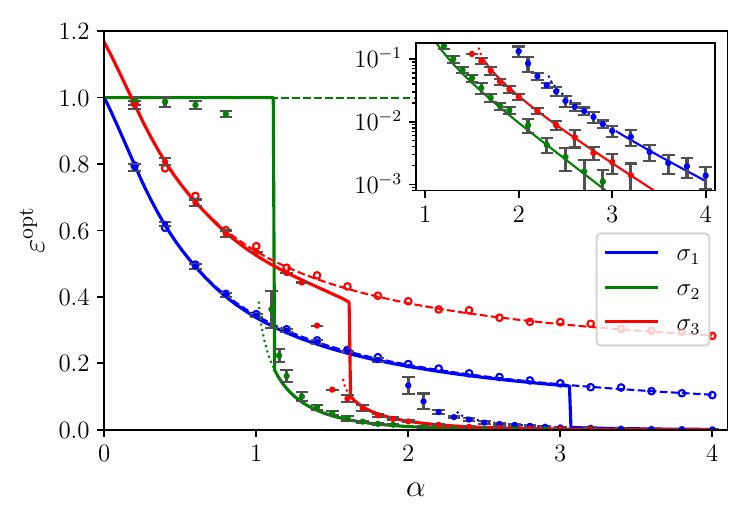}}
\vspace{-10pt}
    \caption{Theoretical prediction (solid curves) of the Bayes-optimal mean-square generalisation error for $L=1$ with binary inner weights and polynomial activations:
    $\sigma_1 = \He_2/\sqrt 2$, $\sigma_2 = \He_3/\sqrt 6$, $\sigma_3 = \He_2/\sqrt 2 + \He_3/6$,
    with $\gamma = 0.5, d=150,\Delta=1.25$, and quenched homogeneous readouts $\bv= \mathbf{1}$. 
    Dots are the half Gibbs error computed using the 
    Metropolis-Hastings algorithm initialised informatively.
    Circles are the error of GAMP-RIE
    \citep{maillard2024bayes} extended to generic activation.
    Points are averaged over 16 data and teacher instances. Error bars for MCMC are the standard deviation over instances (omitted for GAMP-RIE, but of the same order). Dashed and dotted lines denote, respectively, the universal and specialisation branches where they are metastable.}
    \label{fig:gen_errors_univ_spec}
\end{center}
\vskip -0.1in
\end{figure}

We consider three different priors for the readouts: the standard Gaussian prior $P_v=\mathcal{N}(0,1)$, homogeneous readouts $P_v=\delta_1$, and the 4-point prior $P_v = \frac{1}{4}(\delta_{-3/\sqrt{5}} + \delta_{-1/\sqrt{5}} + \delta_{1/\sqrt{5}} +\delta_{3/\sqrt{5}})$ (which is centred and has unit variance). To reduce finite-size sampling fluctuations we \emph{fix} the empirical frequencies of the entries in each readout vector, rather than sampling them. For the 4-point prior this means enforcing $1/4$ frequencies to each symbol. For the Gaussian prior we use an almost-deterministic Gaussian readout: the $k$ entries are set to the population quantiles of a standard normal. The case of random Gaussian readouts is presented in App.~\ref{subsec:learnable_readouts}. For the activation functions, we remind the reader about the hypotheses (\hyperlink{H:H1}{$H_{1}$}), (\hyperlink{H:H2}{$H_{2}$}), (\hyperlink{H:H3}{$H_{3}$}) depending on the NN depth $L$. We consider polynomial activations made of sums of Hermite polynomials used in conjunction with Rademacher inner weights $P_W=\frac12(\delta_{-1}+\delta_{1})$ in \figurename~\ref{fig:gen_errors_univ_spec}. For the other figures on standard Gaussian inner weights for all hidden layers we take $\sigma(x)={\rm ReLU}(x)$ as an example of activation with both $\mu_1\neq 0$ and $\mu_2\neq 0$, and $\sigma(x)=\tanh(2x)$ with $\mu_1\neq 0$ but $\mu_2=0$. We also consider its normalised version which is analytically convenient (but not necessary) when there are more than one hidden layer: $\sigma(x)=\tanh(2x)/\sigma_{\tanh}$ with $\sigma_{\tanh}$ enforcing $\EE_{z \sim \mathcal{N}(0,1)}\sigma(z)^2 =1$. 

In all experiments we consider the regression task with linear readout and Gaussian label noise of variance $\Delta$. We thus focus on the mean-square generalisation error. We always remove from it the irreducible error present in definition \eqref{eq:gen_err_result} for the linear readout, $\varepsilon^{\rm opt}\to\varepsilon^{\rm opt}-\Delta$, still denoting it $\varepsilon^{\rm opt}$ by a slight abuse of notation.

\vspace{5pt}\paragraph*{Probing the solutions of the RS saddle point equations.}
The various theoretical errors (with $\Delta$ removed) we will analyse are all obtained from the same formula:
\begin{talign}
\varepsilon^{\square}=\EE_{(\lambda^0,\lambda)}[(\lambda^0)^2-\lambda^0\lambda]=K_{\rm d}-K^\square, \label{eq:varEpsBox}
\end{talign} 
where $(\lambda^0,\lambda)\sim \mathcal{N}(\mathbf{0},\mathbf{K}^{\square})$, with $\square\in\{*,{\rm uni},{\rm sp}\}$ and $K^\square,K_{\rm d}$ are respectively the covariance off-diagonal and diagonal (the latter being $1$ for $L\ge 2$). For $L=1$, $\mathbf{K}^\square$ has the form \eqref{eq:covLam_2}, but where the equilibrium solution $*$ of the RS saddle point equations (simply called ``RS equations'' from now on) can also be replaced by the \emph{universal solution} (or branch), yielding $\varepsilon^{\rm uni}$, or by the \emph{specialisation solution}, yielding $\varepsilon^{\rm sp}$. The latter probes the performance of a Bayesian student initialised in the vicinity of the target rather than completely randomly. The equilibrium solution corresponds to the Bayes-optimal error: ${\varepsilon}^*=\varepsilon^{\rm opt}$. In the same way, for $L\ge 2$, $\mathbf{K}^\square$ generalises \eqref{eq:covLam_3} using $K^{(2)},K^{(L)}$ and we get the errors similarly. 

We now explain how to concretely find the solutions of the RS equations. For any $L$ the \emph{universal solution} is obtained using the fully uninformative initialisation, i.e., setting all physical order parameters to $0$ in the RS equations and then solving them by fixed point iterations (the conjugate OPs never require an initialisation). In contrast, the \emph{specialisation solution} is obtained from the fully informative initialisation where all physical OPs start from a strictly positive value (generally close to $1$ to speed up convergence). When the universal solution is the equilibrium one (i.e., maximises the RS potential among all fixed points) it defines the universal phase.  Similarly, when the specialisation solution is the equilibrium one it defines the specialisation phase. 


For $L=1$, inhomogeneous readouts imply multiple specialisation transitions associated with different solutions of the RS equations, in addition to the one discussed above. Each one is associated with a different ``state'' where only some of the (macroscopic) sub-populations of neurons connected to the same readout value have specialised, see \eqref{eq:v-depSpecTrans}. These ``partially specialised solutions'' are accessed by initialising the RS equations with $\calQ(\mathsf{v})= c\,\mathbbm{1}(\mathsf{v}\ge \bar{\mathsf{v}})$ for some $\bar{\mathsf{v}}$ and constant $c$ close to $1$.

For $L= 2$ each layer can live in a different phase (i.e., specialise at different sampling rates), which are defined similarly as for $L=1$ but layer-wise using the layer-indexed overlaps. Additionally, in a given layer, partial specialisation as described above is also possible, making the overall picture extremely rich: \emph{for deep NNs, specialisation transitions can happen inhomogeneously across layers, but also across neurons in a given layer}. The theory predicts these two types of learning inhomogeneities observed in simulations.

\begin{figure*}[t]
  \centering
  \setlength{\tabcolsep}{4pt} 
  \centerline{
    \includegraphics[width=0.3469\textwidth,trim={0 0 0.2cm 0.2cm},clip]{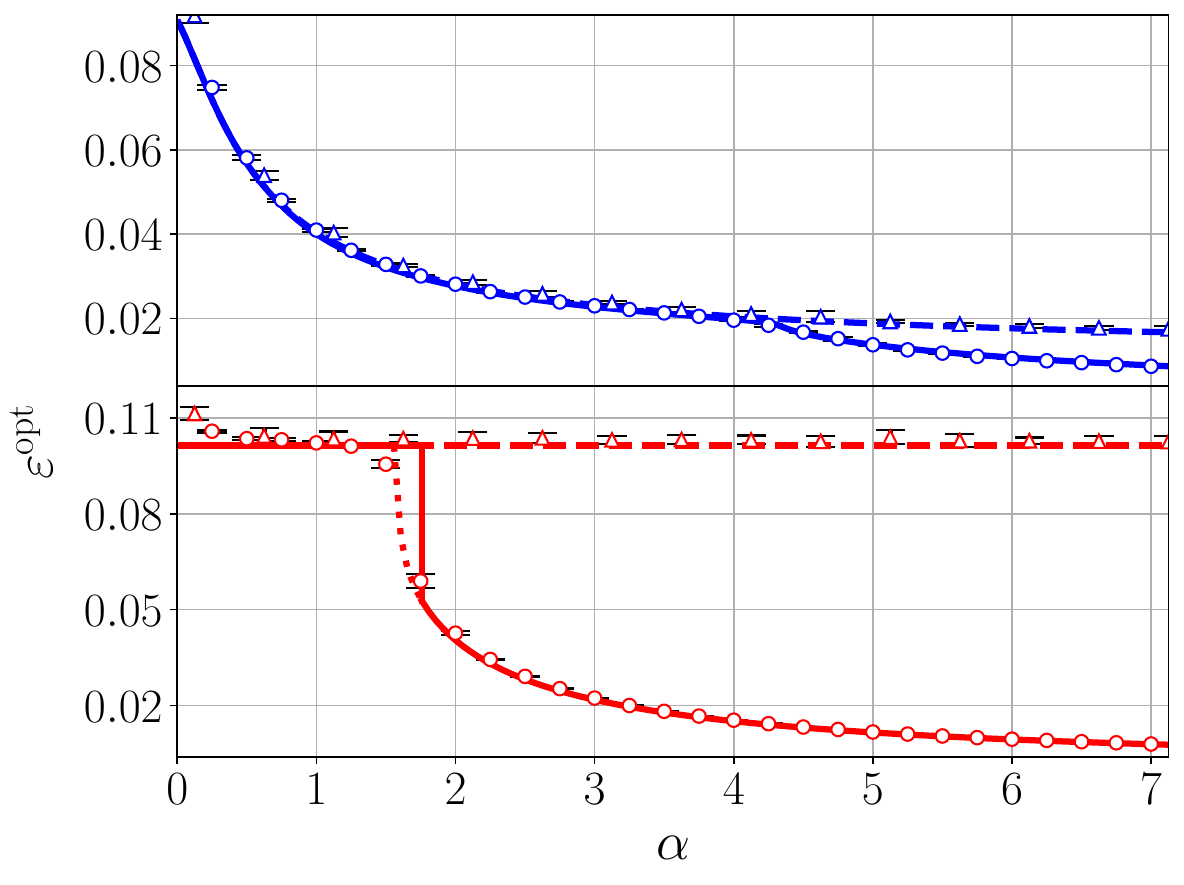} 
    \includegraphics[width=0.2975\textwidth,trim={1.2cm 0 0.2cm 0.2cm},clip]{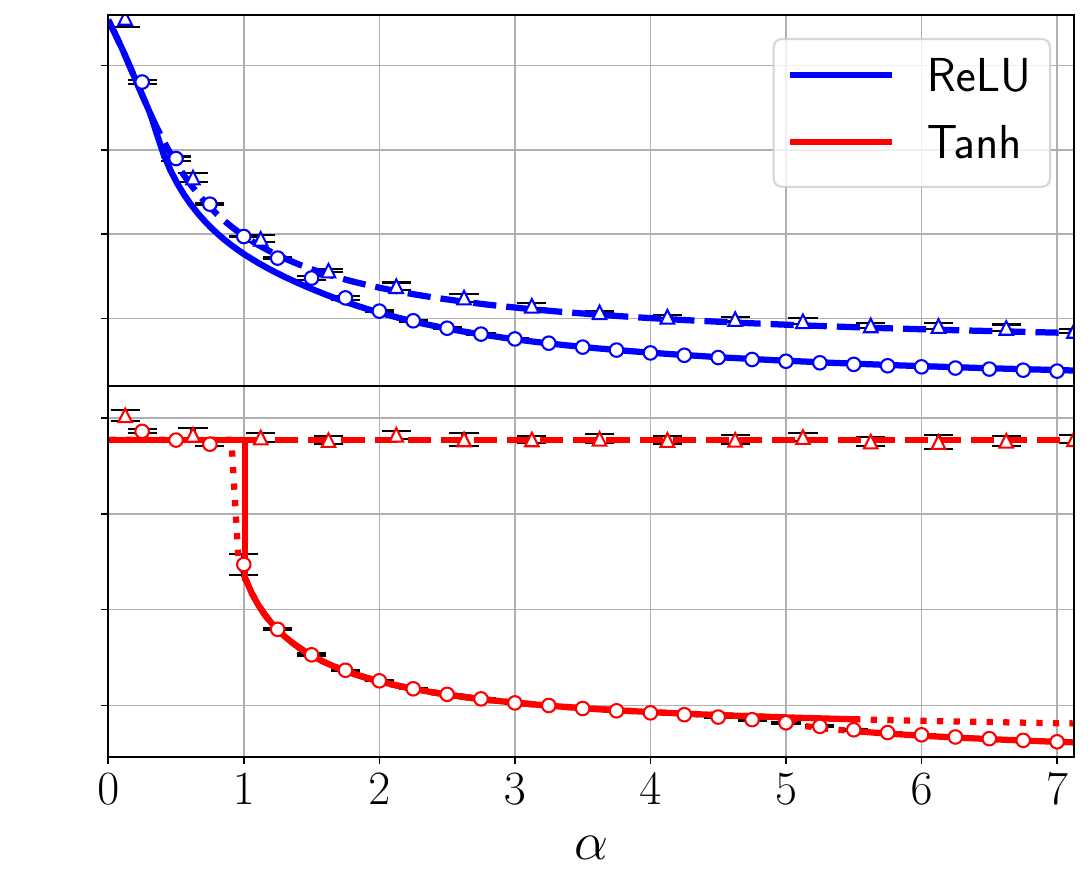} 
    \includegraphics[width=0.32\textwidth,trim={1.1cm 0 0.2cm 0,2cm},clip]{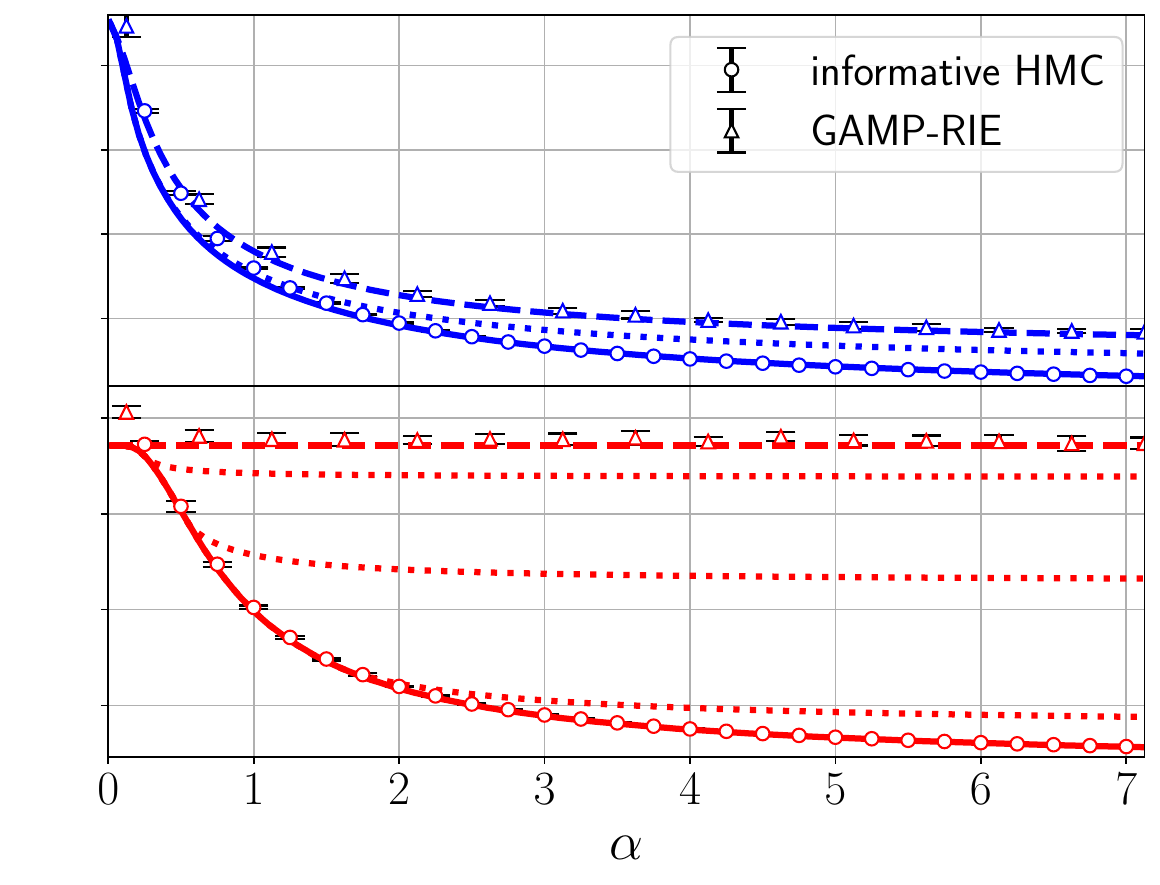} 
    }
  \vspace{-10pt}

  \caption{Theoretical prediction (solid curves) of the Bayes-optimal mean-square generalisation error (with the irreducible error $\Delta$ removed) for $L=1$ with Gaussian inner weights, ${\rm ReLU}(x)$ (blue curves) and $\tanh(2x)$ activation (red curves), $d=200, \gamma=0.5,\Delta=0.1$ and different $P_v$ laws. Dashed and dotted lines denote, respectively, the universal and multiple specialisation branches where they are metastable (i.e., a solution of the RS equations not corresponding to the equilibrium). The readouts are fixed to the teacher's during sampling. \textbf{Left}: Homogeneous readouts. \textbf{Centre}: 4-point readouts. \textbf{Right}: Gaussian readouts. For these two latter cases, the specialisation transitions correspond to partial specialisation (of just some neurons). The numerical points correspond to the half Gibbs error obtained with HMC with informative initialisation on the target. Triangles are the error of GAMP-RIE \citep{maillard2024bayes} extended to generic activation, see App.~\ref{app:GAMP}. Each point has been averaged over 12 instances of the training set (including the teacher). Error bars are the standard deviation over instances. 
  The empirical test error is computed empirically from $10^5$ i.i.d. test samples.}
  \label{fig:gen_error_gauss}
  \vskip -0.1in
\end{figure*}

In order to get the corresponding solutions of the RS equations, we proceed as before by playing with the initialisation of the OPs. We focus on three representative specialisation scenarios across layers, the equilibrium solution always corresponding to one of them: $(i)$ $\mathcal{Q}_{2:1}> 0$ (i.e., positive for any argument value) and $\mathcal{Q}_{1},\mathcal{Q}_{2}\equiv 0$, because the product matrix $\bW^{(2)0}\bW^{(1)0}$ can be learned without specialisation; $(ii)$ $\mathcal{Q}_{1}>0$ and $\mathcal{Q}_{2:1},\mathcal{Q}_{2}\equiv 0$ to probe partial specialisation of the first layer only; and recall $(iii)$ the fully informative initialisation $\mathcal{Q}_{1},\mathcal{Q}_{2},\mathcal{Q}_{2:1}> 0$ for complete specialisation. Other initialisations converge either to the universal solution or match the solution reached from $(iii)$. Having access to this rich family of solutions, the corresponding errors are again obtained by plugging them in $\bK^{\square}$ in \eqref{eq:varEpsBox}.

For $L\ge 3$, under the (\hyperlink{H:H3}{$H_{3}$}) hypothesis, specialisation of all layers occurs concurrently so we only consider the universal and specialisation solutions.

Notice that \eqref{eq:varEpsBox} relies on the simplification of the optimal mean-square generalisation error in App.~\ref{app:gen_err}, which is a direct consequence of the Nishimori identity at equilibrium (App.~\ref{app:nishiID}). By construction of the theory, \emph{any} solution of the RS equations verifies the Nishimori identities, which justifies using \eqref{eq:varEpsBox} beyond the equilibrium solution $*$. This reflects the property that metastable states ``behave as the equilibrium'' for what concerns the validity of the Nishimori identities and concentration properties, see Remark~\ref{rmk:sampling} for a discussion and \figurename~\ref{fig:HMC_hardness}, \ref{fig:ADAM_HMC_hardness_L=2} and \ref{fig:NishiBreak} for numerical confirmations.

\vspace{5pt}\paragraph*{Tested algorithms.}

The theory is tested against four algorithms: the first two are based on Monte Carlo, the third is a spectral algorithm combined with approximate message-passing, and the last is a popular first-order optimiser: ADAM. We thus cover different classes of algorithms, and we will see that our theory is linked to all.

$({\rm Algo}_1)$ Hamiltonian Monte Carlo (HMC) initialised uninformatively, i.e., from a random initialisation, will be used to sample the posterior when the inner weights have a Gaussian prior. We will also use HMC to sample it but starting from an informative (i.e., on the teacher) initialisation. These may lead to different results and are used to probe the two solutions of the theory: universal and specialisation.

$({\rm Algo}_2)$ Another algorithm used for sampling the posterior but with binary valued weight matrices is the standard Metropolis-Hastings algorithm. It will also be tested from the two kinds of initialisations.

\begin{remark}\label{rmk:sampling}
The optimal way to construct a predictor for a test sample using these Monte Carlo sampling algorithms is Bayesian, i.e., through an empirical average of the network output over sampled configurations: $\langle \lambda_{\rm test}(\btheta)\rangle_{\rm Monte\, Carlo}$. This is costly, as we would need to do that for many instances of the problem and hyperparameters. A computationally more efficient alternative, but in general sub-optimal, is a one-shot estimator $\lambda_{\rm test}(\btheta)$: a student constructed from one sample $\btheta$ of the parameters. The average mean-square generalisation error of the latter is called \emph{Gibbs error}: $\varepsilon^{\rm Gibbs}:=\EE_{\btheta^0, \mathcal{D},\bx_{\rm test}}\langle (\lambda_{\rm test}(\btheta)-\lambda_{\rm test}^0)^2\rangle$. At equilibrium, $\varepsilon^{\rm Gibbs}/2 = \varepsilon^{\rm opt}$, see Remark~\ref{rem:Gibbs_error} or \cite{barbier2019glm} for a justification based on the Nishimori identities. 

For the experiments, we use this formula also \emph{during} sampling. In practice we compute the half Gibbs error as $\EE_{\bx_{\rm test}} (\lambda_{\rm test}(\btheta_t)-\lambda_{\rm test}^0)^2/2$ based on a single sample $\btheta_t$ at time $t$ (per $\alpha$ value and dataset), where $\EE_{\bx_{\rm test}}$ is an empirical average over many test inputs ($10^4-10^5$). When the chains have mixed and the samples are correctly drawn according to the posterior, which is guaranteed for long enough times, this replacement is justified if also assuming the concentration of the square-error w.r.t. $\btheta_t$ onto the Gibbs error, i.e., $\EE_{\bx_{\rm test}} (\lambda_{\rm test}(\btheta_t)-\lambda_{\rm test}^0)^2=\EE_{\bx_{\rm test}} \langle (\lambda_{\rm test}(\btheta)-\lambda_{\rm test}^0)^2\rangle +o_d(1)$. This concentration is numerically verified to hold for $t$ sufficiently large. Consequently, our way to evaluate the Bayes error holds at equilibrium for large $d$.

However, out of equilibrium these guarantees are lost. This is a priori an issue given that we will need to evaluate errors in ``metastable states'' which are empirically the only reachable ones in polynomial time. Yet, we claim that when probing the error \emph{at} a metastable state, the relation $\EE_{\btheta^0, \mathcal{D},\bx_{\rm test}}\langle (\lambda_{\rm test}(\btheta)-\lambda_{\rm test}^0)^2\rangle_{\rm meta}/2 =  \EE_{\btheta^0, \mathcal{D},\bx_{\rm test}} (\langle\lambda_{\rm test}(\btheta)\rangle_{\rm meta}-\lambda_{\rm test}^0)^2 $, where $\langle\, \cdot \,\rangle_{\rm meta}$ means ``sampling at the metastable state'',
remains valid, but not \emph{while} dynamically reaching it (where half the error of the one-shot estimator is merely a proxy for the Bayesian one). In other words, a hypothesis we make for the rest of the discussion is that metastable states (if present) ``behave'' as the equilibrium for what concerns Nishimori identities and concentration properties, which are the only ones we need to use this relation. This is verified in other Bayes-optimal inference problems \cite{barbier2019glm}, and it will be justified a posteriori by the match of the theoretical predictions (which always agree with the Nishimori identities) associated with a metastable state and the numerics. 

One can also build $s$-shot estimators from HMC samples of (meta)stable states, namely $\lambda_{\rm test}((\btheta^{(p)})_{p\leq s}):=\frac{1}{s}\sum_{p=1}^s\lambda_{\rm test}(\btheta^{(p)})$ where $\btheta^{(p)}$ are HMC samples for long enough times. Assuming, as above, that Nishimori identities hold also at metastable states, and that the generalization error of such an estimator concentrates in $(\btheta^{(p)})_{p\leq s}$, i.e.\ $\EE_{\bx_{\rm test}}(\lambda_{\rm test}((\btheta^{(p)})_{p\leq s})-\lambda^0_{\rm test})^2=\EE_{\btheta^0,\calD,\bx_{\rm test}} \langle(\lambda_{\rm test}((\btheta^{(p)})_{p\leq s})-\lambda^0_{\rm test})^2\rangle_{\rm meta}+o_d(1)$, then we can predict the generalization error of an $s$-shot estimator. It suffices to expand the square in the $\langle\,\cdot\,\rangle_{\rm meta}$ in the last expression to find $\EE_{\btheta^0,\calD,\bx_{\rm test}} \langle(\lambda_{\rm test}((\btheta^{(p)})_{p\leq s})-\lambda^0_{\rm test})^2\rangle_{\rm meta}=\frac{s+1}{s} \EE_{\btheta^0,\calD,\bx_{\rm test}} (\langle\lambda_{\rm test}(\btheta)\rangle_{\rm meta}-\lambda^0_{\rm test})^2$. The latter expression, for $s=1$, indeed yields the correct relation for one-shot estimators. It is also clear that when $s$ becomes larger, and HMC is sampling the equilibrium state, the above is approaching $\varepsilon^{\rm opt}$.

As a numerical test of these claims, \figurename~\ref{fig:HMC_hardness}, \ref{fig:overparam_s-shot} and \ref{fig:ADAM_HMC_hardness_L=2} show that the theory (which is ``Nishimori-compliant'') captures the HMC error, both for one and $s$-shot estimators, not only at long times, but also during the earlier plateau it experiences when sampling a metastable state. On top of this evidence, in Appendix~\ref{app:nishiID} and~\figurename~\ref{fig:NishiBreak} we also plot the evolution of a deviation from the Nishimori identities during posterior sampling by HMC. We see that on the states that HMC finds for long enough times, the Nishimori identities are verified, regardless of them being stable, or only metastable.

\end{remark}

$({\rm Algo}_3)$ 
We extended the GAMP-RIE of~\cite{maillard2024bayes}, publicly available at~\cite{maillard2024github}, to obtain a polynomial-time predictor for test data in the shallow networks case. Extending this algorithm, initially proposed for quadratic activation, to a generic one is possible thanks to the identification of an \emph{effective} GLM onto which the learning problem can be mapped, see App.~\ref{app:GAMP} (the mapping being exact only when $\sigma(x)=x^2$, \cite{maillard2024bayes}). The key observation is that our effective GLM representation holds not only from a theoretical perspective when describing the universal phase, but also algorithmically. The GAMP-RIE is $P_W$-independent as it exploits only the asymptotic spectral law of $\bW^{0\intercal}{\rm diag}(\bv^0)\bW^{0}$, which is the same for Gaussian or binary weight matrices by spectral universality \cite{tao2010random}. It is therefore in general sub-optimal. In order to evaluate the generalisation error of GAMP-RIE in the experiments, we plug the estimator \eqref{eq:output_GAMP_RIE} in \eqref{eq:gen_error_def}.

$({\rm Algo}_4)$ We also test the standard Python implementation of the ADAM optimiser~\cite{ADAM} initialised uninformatively for Gaussian teacher weights. 
The generalisation error for ADAM for a given training set is evaluated as $\EE_{\bx_{\rm test}} (\lambda_{\rm test}(\btheta_t)-\lambda_{\rm test}^0)^2$ using parameters $\btheta_t$ obtained by training a student network through empirical risk minimisation with non-regularised cost function $C(\btheta)=\frac1n\sum_{\mu\le n}(\lambda_{\mu}(\btheta)-y_{\mu})^2$. Adding weight decay does not change the global picture. Notice that in contrast with the Monte Carlo algorithms where the Gibbs error (divided by $2$) is a computationally simpler way to access their mean-square generalisation error, the error of ADAM is not divided by two because it provides a one-shot estimator and is used as such for predictions.

The codes needed to reproduce our experiments are accessible online \cite{github}.

\subsection{Shallow MLP}

\paragraph*{Generalisation error and specialisation transition.} Starting with the shallow case, in \figurename~\ref{fig:gen_error_gauss} and \ref{fig:gen_errors_univ_spec} we report the theoretical generalisation errors from Result~\ref{res:gen_error} for both the universal and specialisation solutions.

\figurename~\ref{fig:gen_error_gauss} considers networks with Gaussian inner weights sampled with informatively initialised HMC in order to focus on the specialisation solution. Tests with uninformative initialisation are discussed later on. Experiments and theory show that HMC initialised close to the target precisely follows the theoretical specialisation solution $\varepsilon^{\rm sp}$ (which is not always the equilibrium). In contrast, GAMP-RIE's generalisation error follows the universal branch of the theory $\varepsilon^{\rm uni}$. It can actually be shown analytically that it is the case when $d\to\infty$. 
An interesting observation is that non-homogeneous readouts trigger the appearance of the specialisation transition earlier and shrink the region where the equilibrium solution coexists with a metastable one (dotted line). In the case of 4-point prior (middle panel), we see two partial specialisation transitions as defined in \eqref{eq:v-depSpecTrans}. The first corresponds to the specialisation of the neurons connected to readouts with the largest amplitude only, and thus yields a greater improvement in the error than the second. 

With continuous readouts, there are infinitely many specialisation transitions; \figurename~\ref{fig:gen_error_gauss} right shows different representative branches. The equilibrium is obtained iteratively by initialising $\calQ(\mathsf{v})=c\mathbbm{1}(\mathsf{v}\ge\bar{\mathsf{v}})$ with $c \lesssim 1$ for several values of $\bar{\mathsf{v}}$, selecting the fixed point with the largest free entropy. This selected equilibrium may coincide with the specialised solution (reached from a fully informative initialisation), as in the case of $\relu$, or it may slightly differ from it, as for $\tanh$. This distinction leads to very small differences in the generalisation curves. Furthermore, experimental findings in App.~\ref{app:hardness} suggest that, while for discrete readouts (left and centre panel) the specialised solution is hard to reach, continuous readouts ease the learning task, with no evidence of hardness for standard algorithms (see paragraph below). 
 

\figurename~\ref{fig:gen_errors_univ_spec} concerns networks with Rademacher inner weights. The numerical points are of two kinds: the dots, obtained from Metropolis-Hastings sampling, and the circles for the GAMP-RIE. We report analogous simulations for $\relu$ and $\elu$ activations in \figurename~\ref{fig:GAMP-RIE_eLU-ReLU}, App.~\ref{app:GAMP}. Notice an important thing here: although our theoretical framework presented in Sec.~\ref{sec:theory} uses the HCIZ integral, which relies on the strict rotational invariance of the matrices involved, it is able to accommodate any prior on the weights. It is thus able to deal with non-rotationally invariant matrices, as in the case of Rademacher weights.

In the two considered set-ups (Gaussian $P_W$ of \figurename~\ref{fig:gen_error_gauss} and Rademacher of \figurename~\ref{fig:gen_errors_univ_spec}), when data are scarce, $\alpha<\alpha_{\rm sp}$, the student cannot break the numerous symmetries of the problem, resulting in an ``effective rotational invariance'' at the source of the prior universality of the free entropy and OPs, with posterior samples having a vanishing overlap with $\bW^0$. In this \emph{universal phase}, feature learning occurs because the student tunes its weights to match a quadratic approximation of the teacher, rather than aligning to those weights themselves. This phase is universal in the law of the i.i.d. teacher inner weights (centred, with unit variance): our numerics obtained both with binary and Gaussian inner weights match well the theory. This phase is superseded at $\alpha_{\rm sp}$ by a \emph{specialisation phase} where the prior $P_W$ matters. There, a finite fraction of the student weights aligns with the teacher's, which lowers the generalisation error.

\begin{figure}[t!!]
  \centering
  \includegraphics[width=0.955\linewidth,trim={0 0 0.2cm 0},clip]{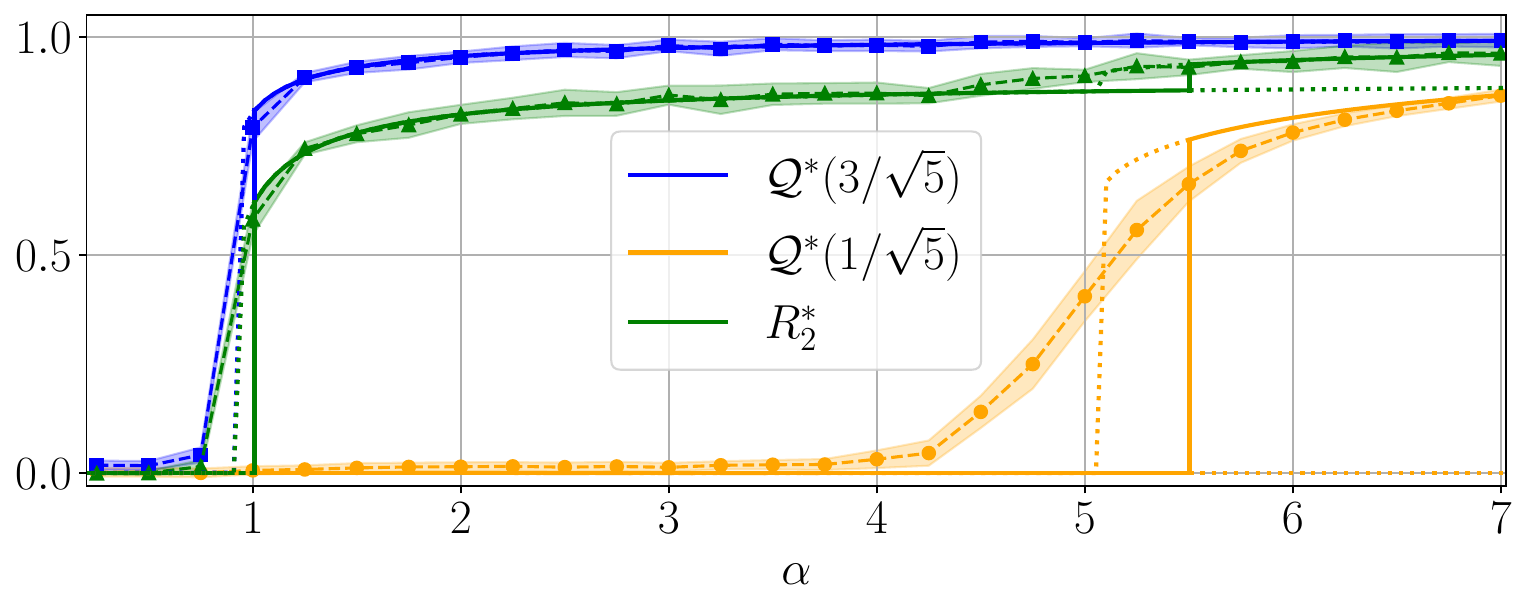} \\
  \vspace{-3pt}
  \includegraphics[width=1\linewidth,trim={0.3cm 0 0 0},clip]{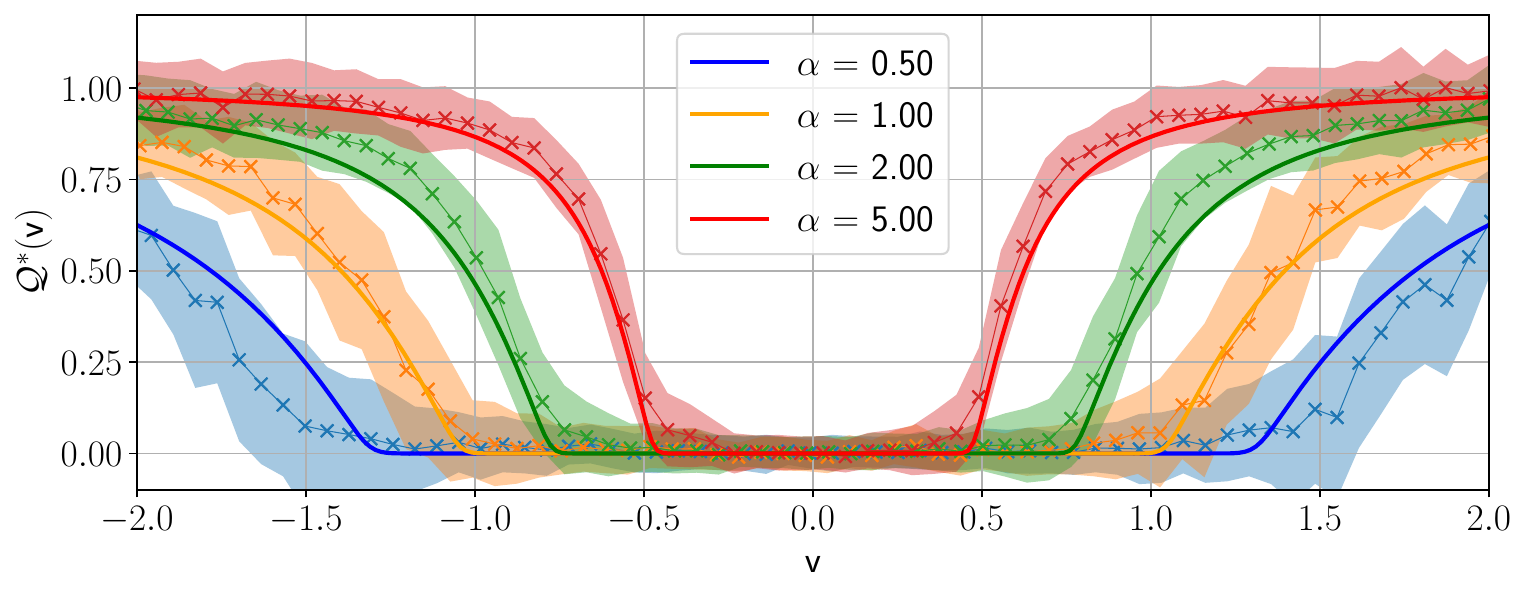}\hspace*{-0.1cm}
  \vspace{-5pt}
  \caption{Theoretical prediction (solid curves) for the equilibrium overlaps as function of the sampling ratio $\alpha$ for $L=1$ with Gaussian inner weights, $d=200, \gamma=0.5,\Delta=0.1$. The empirical crossed curves were obtained from informed HMC using a single posterior sample $\bW$ (per $\alpha$ and data instance), and shaded regions around them correspond to 1 standard deviation w.r.t. to data instances. \textbf{Top}: $\sigma(x)=\tanh(2x)$ and 4-point readouts, and average over 12 instances of the data. \textbf{Bottom}: $\sigma(x)={\rm ReLU}(x)$ and Gaussian readouts. $\mathcal{Q}(\mathsf{v})$ is evaluated numerically by dividing the interval $[-2,2]$ into bins and then computing the value of the overlap associated with the readout value in that bin. We averaged over 100 data instances. Readouts are fixed to $\bv^0$.}
  \label{fig:v_non-constant}
\vskip -0.1in
\end{figure}

The phenomenology depends on the activation function due to the following reason. Recall the interpretation in terms of a tensor inference problem discussed in Sec.~\ref{sec:setting}, in particular that before the specialisation, components in the Hermite expansion of the target beyond the first two play the role of effective noise when learning. Only with $\alpha>\alpha_{\rm sp}$ the student can realise that they are informative and exploit them. Consequently, for odd activation ($\tanh$ in \figurename~\ref{fig:gen_error_gauss}, $\sigma_2$ in \figurename~\ref{fig:gen_errors_univ_spec}), where $\mu_2=0$, we observe that the generalisation error is constant for $\alpha<\alpha_{\rm sp}$, whereas at the phase transition it suddenly drops. This is because the learning of the second component is skipped entirely, and the only way to perform better is to learn all terms jointly through specialisation.

We emphasise that our theory is consistent with \cite{cui2023bayes}, which considers the simpler regime of strong over-parametrisation $n=\Theta(d)$ rather than the interpolation one $n=\Theta(d^2)$: our generalisation curves at $\alpha\to0$ match theirs at $\alpha_1:=n/d\to+\infty$, which is when the student learns perfectly the linear component $\bv^{0\intercal}\bW^0$ of the target but nothing more. This is also the best a network can do in the quadratic data regime when $\mu_2=0$ \emph{if} it does not specialise.

\vspace{5pt}\paragraph*{Order parameters and learning mechanisms.}

\figurename~\ref{fig:v_non-constant} reveals sequences of phase transitions in $\alpha$. The top panel shows the evolution of the two relevant overlaps $\mathcal{Q}^*(\mathsf{v})$ in the case of readouts with discrete values: as $\alpha$ increases, the student weights start aligning with the target weights with highest readout amplitude, marking the first phase transition. At the same time $R^*_2$ jumps, indicating that learning of the quadratic term of the target occurs concurrently. As these alignments strengthen, the last transition occurs when the weights corresponding to the next largest readout amplitude are learnt. We see the (relatively small) effect of this latter transition also at the level of the generalisation error, see red middle curve in \figurename~\ref{fig:gen_error_gauss} at $\alpha\approx 5$. Through the same mechanism, continuous readouts produce an infinite sequence of learning transitions in the limit \eqref{thermolim}, as supported by the lower part of \figurename~\ref{fig:v_non-constant} for Gaussian readouts. From these observations, we conclude that the readout amplitudes $|v_j|$, controlling the strength with which the responses $(y_\mu)$ depend on feature (neuron) $\bW_j^0$, play the role of an SNR.

\vspace{5pt}\paragraph*{Algorithmic hardness of specialising, and ${\rm ADAM}$ as an approximate Bayesian sampler.}

\begin{figure}[t!!]
  \centering
  \includegraphics[width=0.975\linewidth,trim={0 0 0.0cm 0},clip]{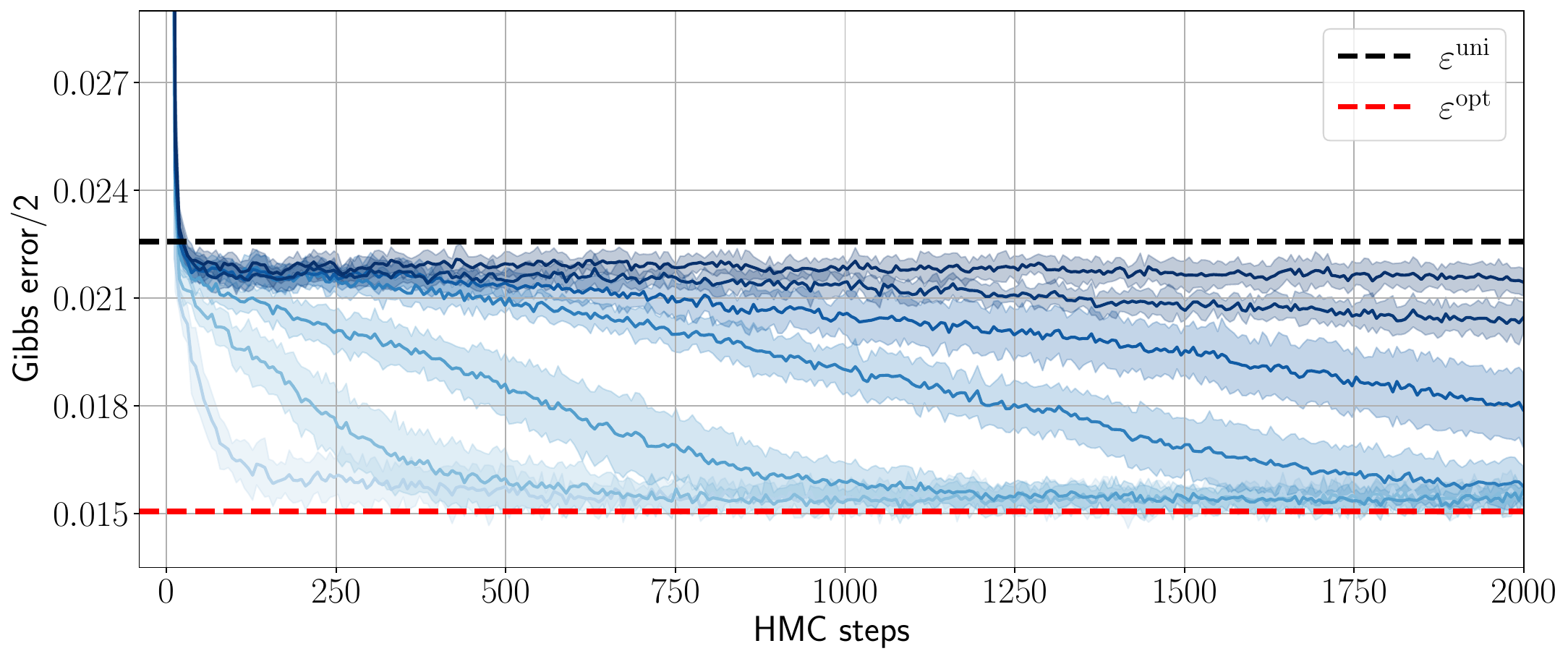} \\
  \vspace{-14pt}
  \hspace*{0.11cm}\includegraphics[width=0.975\linewidth,trim={0.0cm 0 0 0},clip]{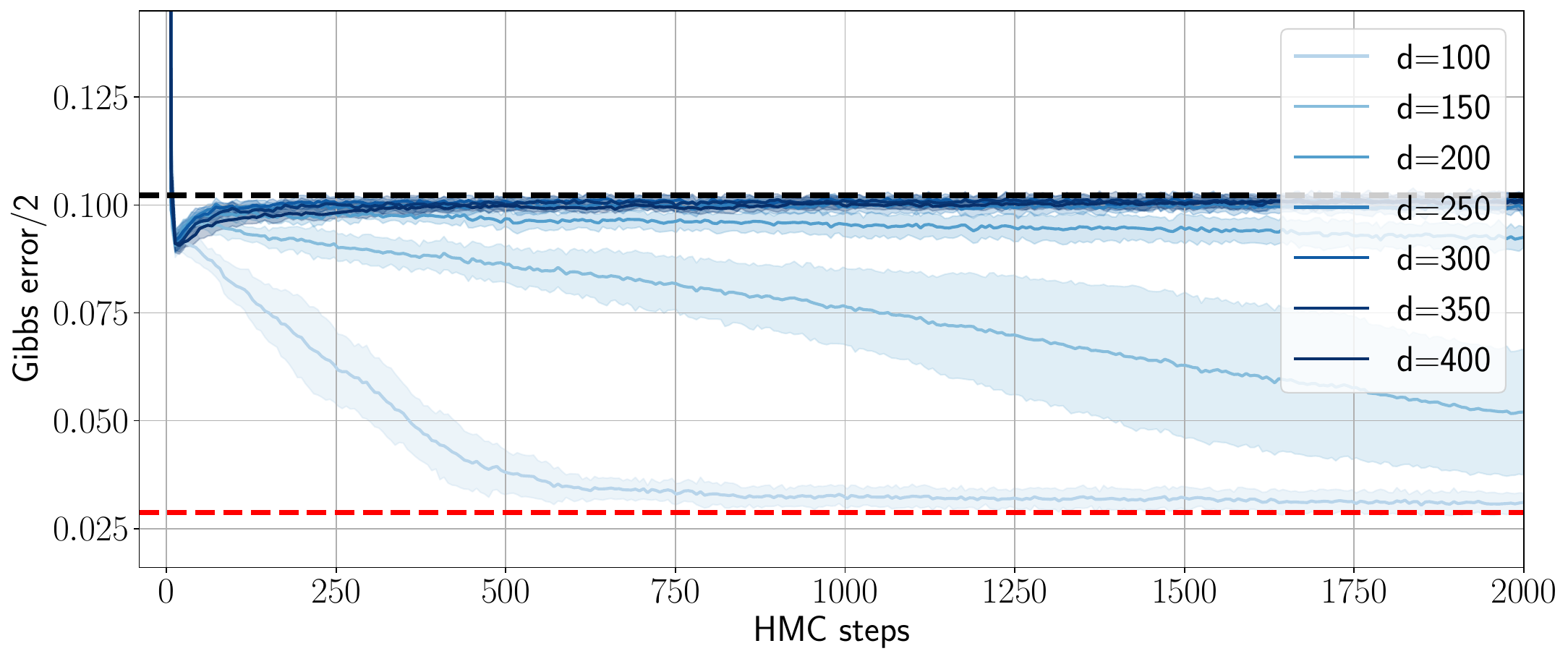}
  \vspace{-5pt}
  \caption{Half Gibbs error of HMC from random initialisation as a function of the number of updates for various $d$ with $L=1$ and Gaussian inner weights. 
  The errors are averaged over 10 data instances and shaded regions represent 1 standard deviation. The black dashed line corresponds to the error associated with the universal solution while the red corresponds to the specialised solution. \textbf{Top:} $\sigma(x)=\mathrm{ReLU}(x), \alpha = 3.0, \gamma = 0.5,\Delta = 0.1$ and 4-point quenched readouts. \textbf{Bottom:} $\sigma(x)=\tanh(2x),\alpha = 2.5,\gamma = 0.5,\Delta = 0.1$ and homogeneous quenched readouts. For both activations, larger $d$ lead to an important slowing down of the convergence towards the specialised solution happening precisely when crossing the error predicted by the universal solution $\varepsilon^{\rm uni}$ of the theory.}
  \label{fig:HMC_hardness}
\vskip -0.1in
\end{figure}

\begin{figure}
    \centering
    \includegraphics[width=1\linewidth]{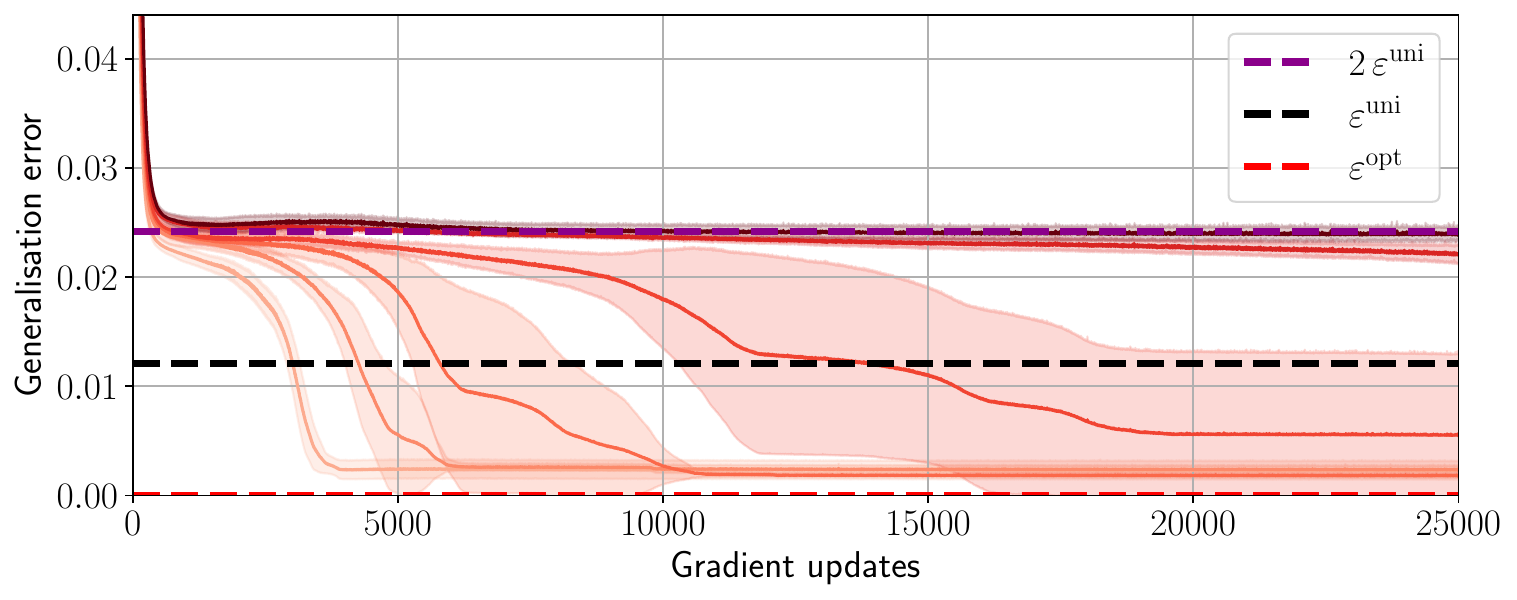}\vspace{-9pt} \\
    
    \hspace*{-0.2cm}\includegraphics[width=0.973\linewidth]{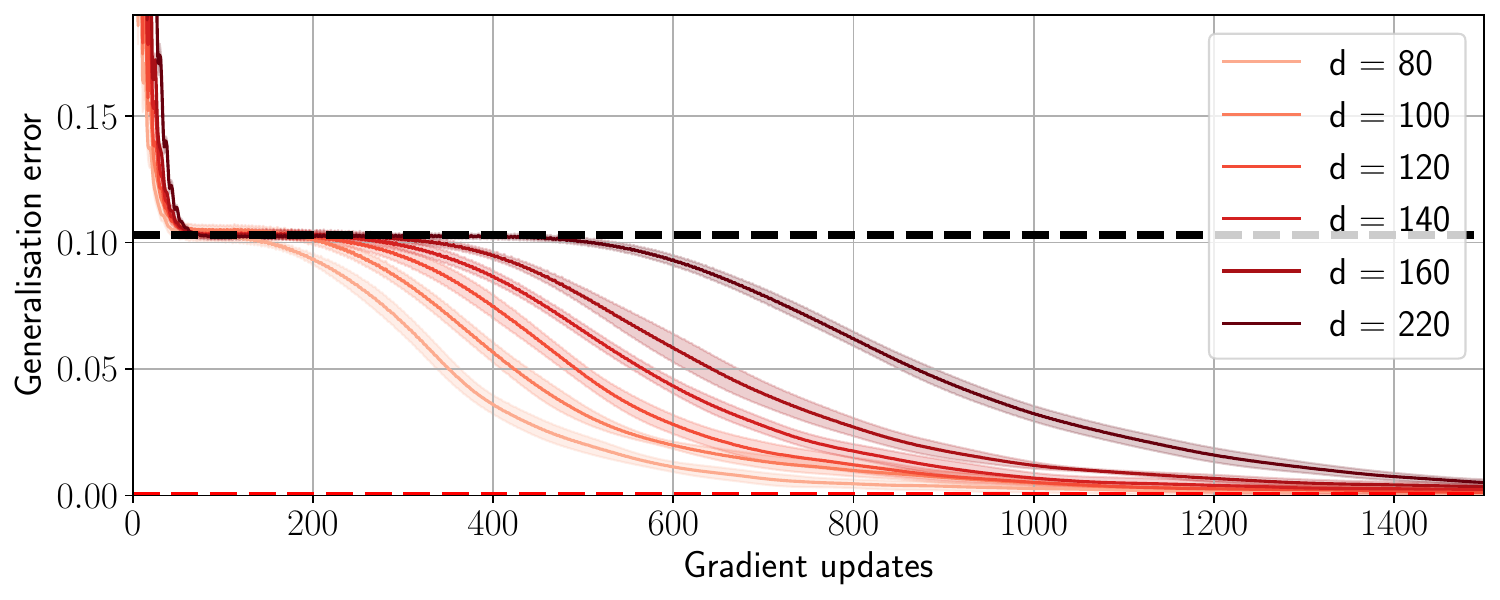}
    
    \caption{Generalisation error of ADAM from random initialisation as a function of the gradient updates for various $d$ with $L=1$ and Gaussian inner weights.
    The initial learning rate is $0.01$ and batch size $\lfloor n/4 \rfloor$. The error is averaged over 10 data instances and shaded regions represent 1 standard deviation and computed empirically from $10^4$ i.i.d. test samples. In both plots $\alpha = 5.0, \gamma = 0.5, \Delta = 10^{-4}$, the target readouts are homogeneous while the student has learnable readouts. \textbf{Top:} $\sigma(x)=\mathrm{ReLU}(x)$. The error plateaus at the purple dashed line corresponding to twice the error associated with the universal solution of the theory $\varepsilon^{\rm uni}$. \textbf{Bottom:} $\sigma(x)=\tanh(2x)$. The error plateaus at the black dashed line corresponding to the universal solution. The number of gradient updates necessary to improve upon the universal solution (or twice its value) grows exponentially with $d$, see App.~\ref{app:hardness}.    
    }
    \label{fig:hardness_main}
    \vskip -0.1in
\end{figure}

Even when dominating the posterior measure, the specialisation solution can be algorithmically hard to reach. With discrete readouts, simulations for binary inner weights exhibit specialisation only when sampling with informative initialisation. Moreover, even in cases where algorithms (such as ADAM or HMC for Gaussian inner weights) are able to find the specialisation solution, they manage to do so only after a training time increasing exponentially with $d$.
For the continuous distribution $P_v=\calN(0,1)$, our tests are inconclusive on hardness and deserve numerical investigation at a larger scale. We refer to App.~\ref{app:hardness} for a detailed discussion and systematic tests. As an illustration of the conclusions reached in this appendix, \figurename~\ref{fig:HMC_hardness} and \figurename~\ref{fig:hardness_main} display the evolution of the generalisation error reached with HMC and ADAM, respectively (recall that for HMC we plot the half Gibbs error proxy).

\figurename~\ref{fig:HMC_hardness} shows that HMC, for a discrete readout prior, converges fast to the universal solution where it is abruptly stopped (for $d$ large), before very slowly approaching the specialisation solution. The time it takes to escape the plateau scales exponentially with the dimension, meaning that improving upon $\varepsilon^{\rm uni}$ is hard given quadratically many data. The behaviour is the same for both ${\rm ReLU}(x)$ and $\tanh(2x)$ activations. We observe the same phenomenology when the teacher’s inner weights are drawn from a binary distribution while HMC sampling wrongly assumes a Gaussian prior, indicating prior universality of the metastable state.

Concerning ADAM, \figurename~\ref{fig:hardness_main}, the picture remains globally the same: an initial fast convergence followed by a long plateau before a slow descent toward a close-to-specialisation solution (the precise analysis of where ADAM lands in this case and its associated generalisation error is out of the scope of the paper). In the case of $\sigma={\rm ReLU}$ (top panel) and homogeneous readouts the plateau is at $2\varepsilon^{\rm uni}$. It takes again exponentially many updates to escape it. ADAM thus reaches precisely the same error as the Gibbs error of HMC, i.e., when also using HMC as a one-shot estimator. This suggests that ADAM is essentially sampling the posterior the best it can given only a polynomial-in-$d$ number of updates, and ends up in a similar metastable state as HMC. The same observation was made for pure gradient-descent in the special case of one hidden layer with quadratic activation \cite{maillard2024bayes}. Our observations contribute, in a quantitatively precise manner within a rather general NN model, to the recent line of works on ``stochastic gradient descent behaves as a Bayes sampler'' \cite{hennick2025almost,mingard2021sgd,smith2018stochastic,mandt2017stochastic} but for ADAM.

There is however one major difference compared to HMC in the case of $\sigma(x)={\rm tanh}(2x)$ (bottom panel): ADAM plateaus close to $\varepsilon^{\rm uni}$, not twice this value. By calling $\btheta_{\rm algorithm}$ a sample from that algorithm, this means that the ADAM one-shot estimator $\lambda_{\rm test}(\btheta_{\rm ADAM})$ performs almost as a Bayesian, ensemble-averaged estimator $\langle \lambda_{\rm test}(\btheta_{\rm HMC}) \rangle_{{\rm meta }}$ sampling the metastable state. The performance of the latter is what we conjecture to be the best achievable in polynomial time. Notice instead that HMC, when also used as a one-shot estimator, does not perform as well (see again \figurename~\ref{fig:HMC_hardness}, bottom panel). This is at odds with the ${\rm ReLU}$ case, where ADAM and one-shot HMC were comparable, and both worse than the Bayesian estimator.

We believe that these different behaviours are a consequence of the fact that $\mu_2\neq 0$ for ${\rm ReLU}$, while it is vanishing for $\tanh $, a crucial element also in our replica analysis of Sec.~\ref{sec:theory}. Indeed, with no 2nd component in the activation the theory predicts the linear term as the only one learnable without specialising. Thus, the non-specialised NN effectively behaves as a linear model trying to fit a noisy linear target. In this respect, the picture is similar to what happens in the proportional regime $n=\Theta(d)$, where the mapping to a GLM is known to hold in the whole phase diagram~\cite{cui2023bayes,camilli2023fundamental,camilli2025inforeduction}: in~\cite{cui2023bayes}, it is shown that optimisers, such as optimally regularised ridge regression, achieve the performance of the Bayes estimator. Here we observe that the picture changes with $n=\Theta(d^2)$ when also the quadratic term is present, as in the top panel of \figurename~\ref{fig:hardness_main}. A gap between the performance of optimisers and Bayesian estimators has been recently shown in~\cite{erba2025nuclear} limited to purely quadratic activation, and requires future investigation in our more general setting.

\begin{figure}
    \centering
    \includegraphics[width=0.985\linewidth]{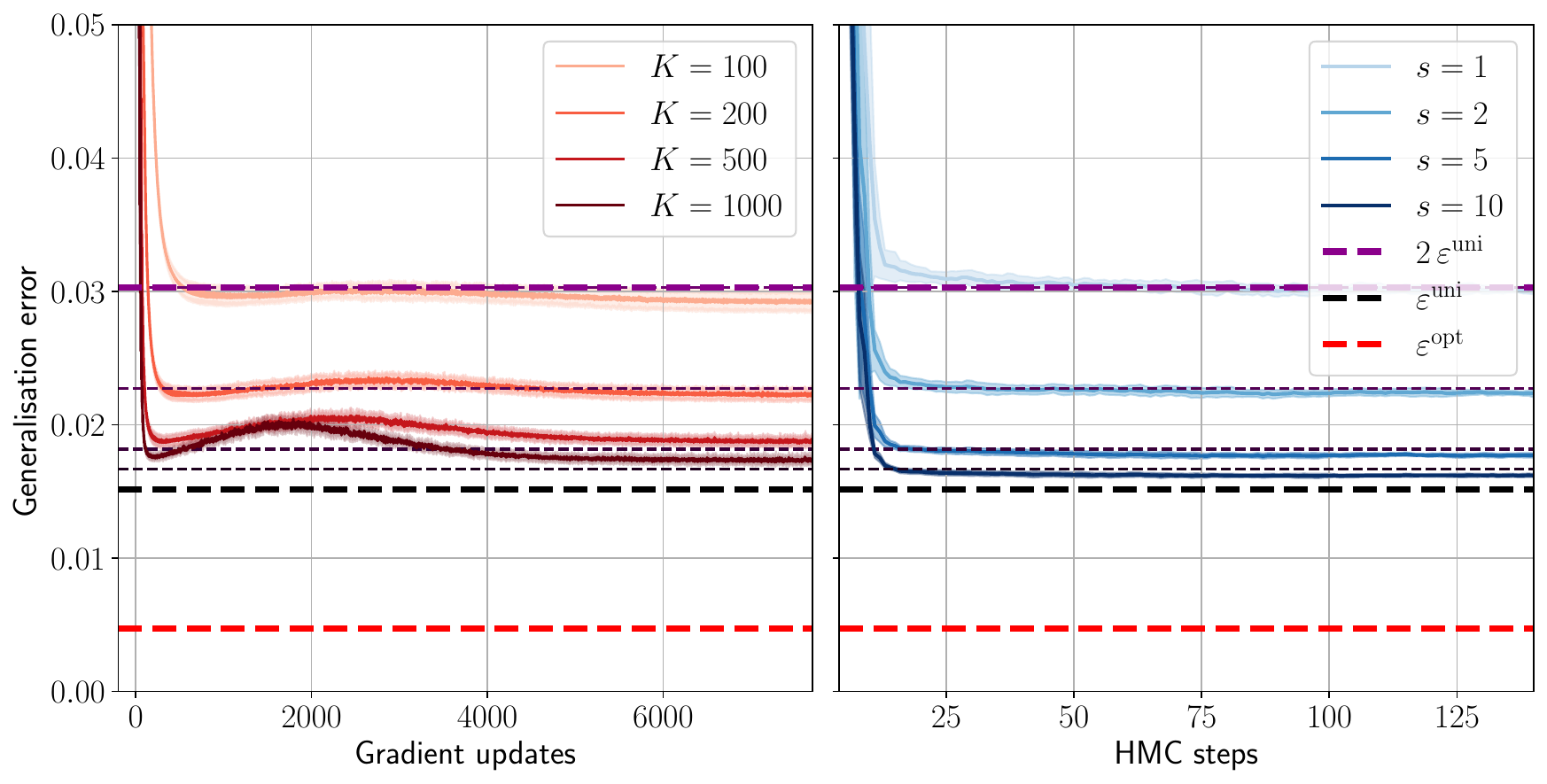}
    
    \caption{Generalisation error of different estimators, initialised randomly, as a function of the number of gradient updates or HMC steps. The errors are averaged over 10 data instances and shaded regions represent 1 standard deviation. The black and purple dashed lines correspond, respectively, to the error associated with the universal solution and twice the universal solution, while the red dashed line corresponds to the specialised solution. The dashed lines between the black and purple lines indicate the universal performance of $s$-shot estimators, which is simply $[(s+1)/s] \varepsilon^{\rm uni}$, see Remark~\ref{rmk:sampling}. In both panels $\sigma(x)=\mathrm{ReLU}(x), \alpha = 4.0, \gamma = 0.5,\Delta = 0.03$, $d=200$ and readouts are homogeneous. The number of hidden units of the teacher is $k=100$. \textbf{Left}: Generalisation error of an overparametrised student trained with ADAM as a function of gradient updates; the readouts are learnable during training. $K$ represents the width of the possibly mismatched student. \textbf{Right}: Generalisation error of an $s$-shot estimator, obtained averaging the output of $s$ posterior HMC samples, as a function of HMC steps; the readouts are fixed during sampling.}
    \label{fig:overparam_s-shot}
    \vskip -0.1in
\end{figure}

In \figurename~\ref{fig:overparam_s-shot}, we show the impact of overparameterisation with respect to the target function in ADAM. This figure refers to a setting where $L=1$ and the number of hidden units for the target is fixed to $100$. $K$ in the left panel is instead the number of hidden neurons of the (possibly mismatched) student. In the right panel we show the performance that an HMC sampler would attain with $s$-shot estimator. Both plots display generalisation errors for the ReLU activation. The dashed lines correspond to the theoretical predictions for $s$-shot estimators as discussed in Remark~\ref{rmk:sampling}. As we can see, overparameterised students are able to reach, with ADAM, the same performance predicted with an $s$-shot estimator produced via HMC. There is an intuitive reason for this: when the student has, say, $K=500$ hidden units, which means 5 times more than the target, then ADAM is somehow picking 5 sets of weights which, when combined via the readouts, yield the performance of a 5-shot estimator. This suggests that ADAM is effectively \enquote{sampling} $s=5$ i.i.d.\ configurations from the metastable state; this aligns with the results of \cite{hennick2025almost,mingard2021sgd,smith2018stochastic,mandt2017stochastic}.

This experiment shows that introducing overparameterisation can help to reduce the generalisation error to $\varepsilon^{\rm uni}$ in polynomial time with ADAM, even when matched students would get stuck at $2\varepsilon^{\rm uni}$. However, the specialisation value remains out of reach in polynomial time according to this picture.

In summary, these experiments show a clear timescale separation between the universal solution (or twice the universal solution) reachable in polynomial time and the specialisation solution requiring exponential time (when it corresponds to the equilibrium). The sub-optimal solution is due to the presence of an attractive metastable state experienced by both algorithms. Moreover, ADAM behaves similarly to a Bayesian sampler like HMC.

\begin{figure}[t!!]
    \centering
    \includegraphics[width=\linewidth]{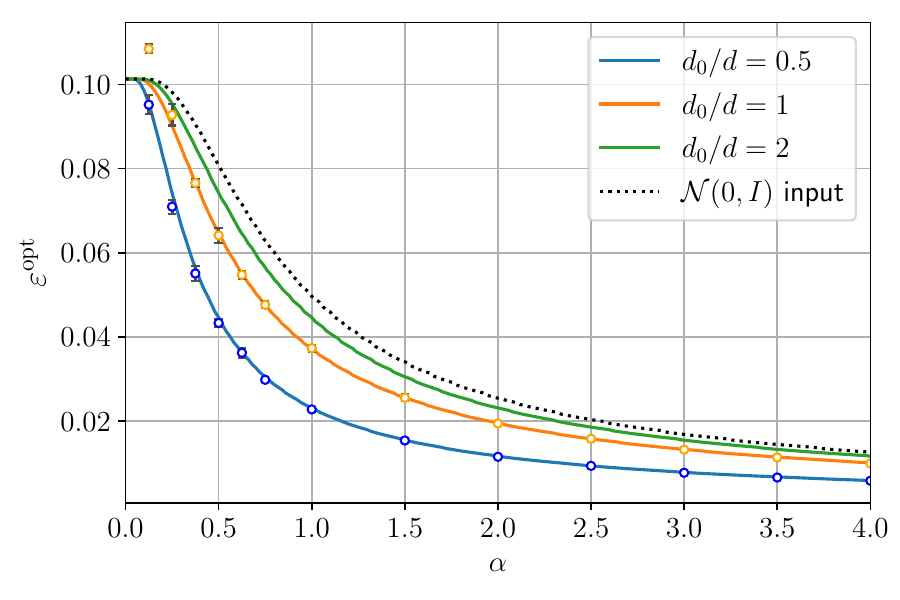}
    
    \vspace{-7pt}
    
    \caption{Theoretical prediction for the Bayes-optimal mean-square generalisation error for $L=1$ with fixed Gaussian readouts, Gaussian inner weights, structured inputs drawn from $\mathcal N(\mathbf{0}, \bC)$, where $\bC = \bW_0 \bW_0^\intercal/d_0 $, where $\bW_0 \in \R^{d \times d_0}$ is a Gaussian matrix, $\tanh(2x)$ activation, $d=150, \gamma=0.5, \Delta=0.1$. The dotted line shows the theoretical result for standard Gaussian inputs, with the other settings unchanged. Experimental points are obtained with informative HMC, by averaging over 9 instances of data, with error bars representing the standard deviation. As the ratio $d_0/d$ grows, the data become less structured and the theoretical curve rapidly approaches that of standard Gaussian input.}
    \label{fig:gaussian-vector-input}
    \vskip -0.1in
\end{figure}

\begin{figure}[t!!]
    \centering
    \includegraphics[width=\linewidth]{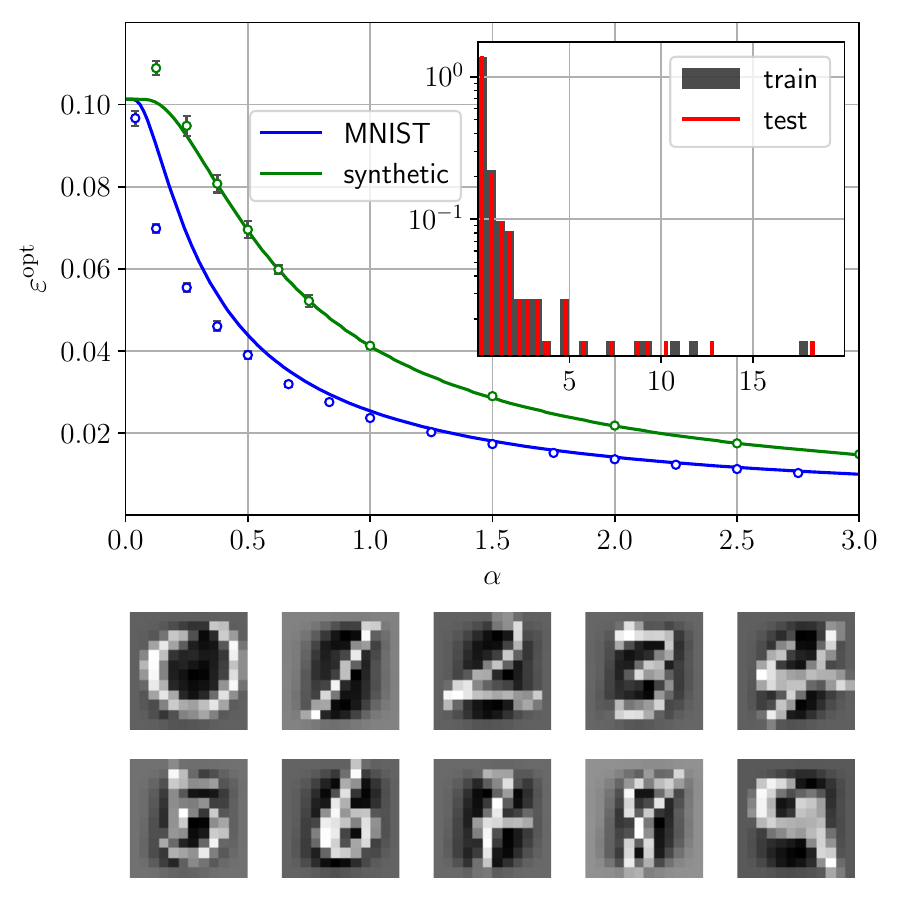}
    
    \vspace{-7pt}
    
    \caption{\textbf{Top:} Theoretical prediction of the optimal mean-square generalisation error for non-Gaussian data. The inputs are taken from the MNIST image dataset or as the outputs of one layer of another NN fed with standard Gaussian vectors (synthetic data). More precisely, the synthetic data are generated as $\bx_\mu=\sigma_0(\bW^{(0)} \bx_\mu^{(0)} / \sqrt{d_0})$ where $\bW^{(0)} \in \R^{d \times d_0}$ is a Gaussian matrix with $d_0/d=0.5, \bx_\mu^{(0)}\sim \mathcal{N}(\mathbf{0},I_{d_0}), \sigma_0(x)=(\text{ReLU}(x)-\mu_0)/c_0$ where $\mu_0$ is the 0-th Hermite coefficient of $\text{ReLU}(x)$ and $c_0$ enforces $\E_{z \sim \mathcal N(0,1)} \sigma_0(z)^2 = 1$. $(\bx_\mu)$ are then passed through the random MLP target with $L=1,\sigma(x)=\tanh(2x), \gamma=0.5$ and Gaussian weights (inner and readouts) to generate the noisy responses $(y_\mu)$ with $\Delta=0.1$. The trainable NN has fixed $\bv=\bv^0$. The MNIST dataset consists of $60000$ training samples and $10000$ test samples. Each one is a $28 \times 28$ pixel image representing a digit from $0$ to $9$. To make the dataset manageable for HMC, each image is downsampled to a $12 \times 12$ resolution. This is achieved by partitioning each side of the original image into blocks of sizes $4, 2, 2 \dots, 2, 4$, resulting in $12 \times 12$ regions, over which pixel values are averaged. The maximum value for the sampling rate is thus $60000/144^2 \approx 2.9$. Subsequently, the images are centred and normalised to have zero mean and a covariance matrix $\mathbf{C}$ satisfying $\Tr(\bC) = d = 144$. Importantly, the responses are still generated by a random target NN with the same architecture as the trained one: the purpose of this experiment is to test input data with realistic correlations. \textbf{Inset:} Histogram of the eigenvalues of covariance matrices computed from the training and test datasets. Both exhibit a few eigenvalues that are significantly larger than the rest which may explain the discrepancy between theoretical and experimental results at low $\alpha$. \textbf{Bottom:} Examples of MNIST images after being downsampled, centred and normalised, showing that their integrity is preserved after the process. 
    }
    \label{fig:mnist-input}
    \vskip -0.1in
\end{figure}

\vspace{5pt}\paragraph*{Structured data.}
Let us consider structured data where the input distribution is different from the standard Gaussian. The most basic example of this is Gaussian inputs with covariance $\bC$. In this case, the model can be transformed into one with standard Gaussian inputs at the cost of losing independence among entries within the same row $\bW$. Indeed, by writing $\bx_\mu = \bC^{1/2} \tilde \bx_\mu$, where $\tilde \bx_\mu \sim \mathcal N(\mathbf{0}, I_d)$, the model can be viewed as having $\tilde \bx_\mu$ as input and $\tilde{\bW}= \bW \bC^{1/2}$ as first layer weights. The weight matrix $\tilde \bW$ has independent rows but dependent entries within the same row.

With this relaxed condition, where $\tilde{\bW}$'s rows are independent and follow a law $P_{\mathbf{w}}$ in $\R^d$, for activations that have zero second Hermite coefficient, Result \ref{res:free_entropy_shallow} still holds, with the only modification being the replacement of the scalar free entropy function $\psi_{P_W}$ by its vector version:
\begin{talign}\label{eq:vector-channel}
    \lim_{d \rightarrow \infty} \frac{1}{d}\mathbb{E}_{\mathbf w^0,\boldsymbol{\xi}} \ln\mathbb{E}_{\mathbf w} e^{-\frac{1}{2}x\|\mathbf w\|^2 + x  \mathbf w^\intercal \mathbf w^0 + \sqrt{x}\,  \boldsymbol{\xi}^\intercal \mathbf w }
\end{talign}
where $\mathbf w, \mathbf w^0$ are i.i.d. from $P_{\mathbf{w}}$ in $\mathbb R^d$ and $\boldsymbol{\xi} \sim \mathcal N(\mathbf{0}, I_d)$. This is evident from the replica computation, where the i.i.d. assumption on the weights is required in equation (\ref{eq:VW_app}) to factorise the integral over the weights, yielding the log-integral term that later becomes the scalar free entropy. If dependences within the same row are allowed, the factorisation can only be performed over the rows, yielding  the free entropy (\ref{eq:vector-channel}). For activations with a non-zero second Hermite coefficient, analysing the model with structured data is an open problem. Even for $\mathcal N(0,\mathbf C)$ input, the task requires solving a denoising problem for the matrix $\mathbf C^{1/2}\mathbf W^\intercal \diag(\bv)\mathbf W \mathbf C^{1/2}$, which is not analytically tractable due to the lack of rotational invariance.

In general there is no analytical simplification for \eqref{eq:vector-channel} except when $\mathbf w$ has i.i.d. entries or when $\mathbf w$ is a Gaussian vector. In the latter case, suppose $P_{\mathbf{w}} = \mathcal N(\mathbf{0}, \bC)$, then $\psi_{P_{\mathbf{w}}}(x) = \lim_{d \rightarrow \infty} \frac{1}{2 d} (x \Tr(\bC) - \ln \det (I_d + x \bC))$. If $\bC$ admits a limiting spectral density $\rho_{\mathbf{C}}$,
\begin{talign}\label{eq:psi_P_W_structureddata}
\psi_{P_{\mathbf{w}}}(x) = \frac{1}{2} \int (xs-\ln(1+xs)) \rho_{\mathbf{C}}(s)ds.
\end{talign}
In App.~\ref{app:structured_data}, we show that by quenching the first hidden layer's weights and taking the first activation to be linear, via a procedure similar to that followed to derive Result~\ref{res:free_entropy_L=2}, it is possible to describe such structured data within our replica formalism.

\figurename~\ref{fig:gaussian-vector-input} shows that structure in the input helps reduce the optimal generalisation error. Here, the covariance matrices are of the Wishart form $\mathbf{W}_0 \mathbf{W}_0^\intercal / d_0$, where $\bW_0 \in \R^{d \times d_0}$ is a Gaussian matrix, with varying ratios $d_0 /d$. It is known that as $d_0 / d$ increases, the spectrum of the Wishart matrix approaches that of the identity matrix, meaning the inputs become less structured. The figure demonstrates that as $d_0/d$ grows, the curves quickly approach that of the standard Gaussian inputs.

\begin{figure*}[t!!!]
  \centering
  \setlength{\tabcolsep}{4pt} 
  \begin{tabular}{ccc}
    \includegraphics[width=0.452\textwidth,trim={0 0 0.2cm 0.2cm},clip]{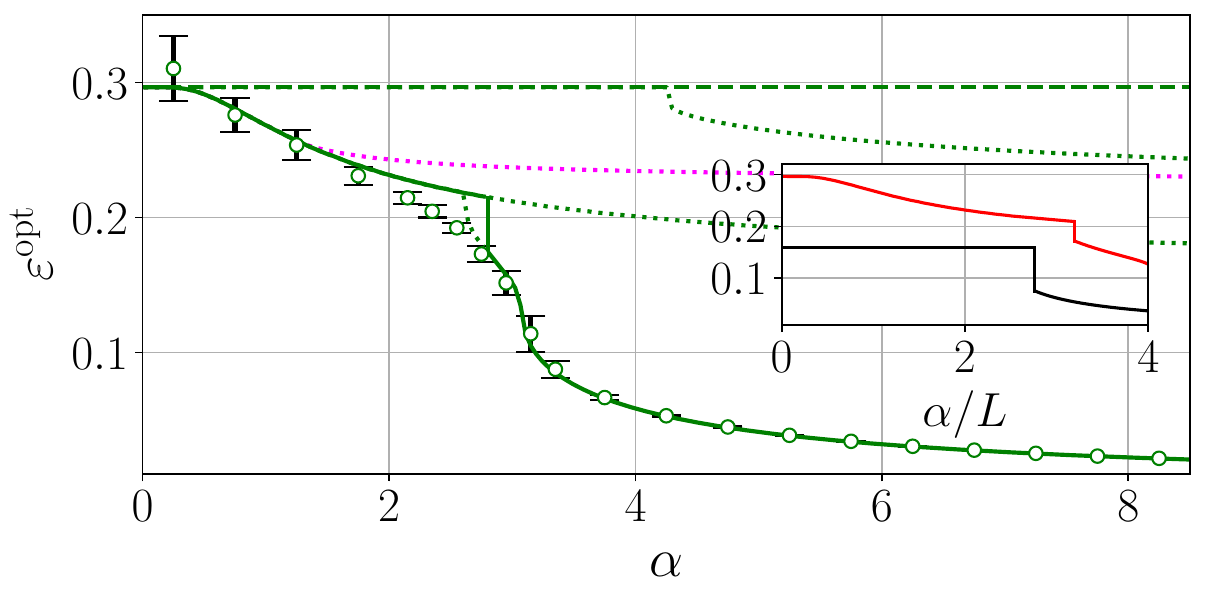} &
    \includegraphics[width=0.42\textwidth,trim={1.2cm 0 0.2cm 0.2cm},clip]{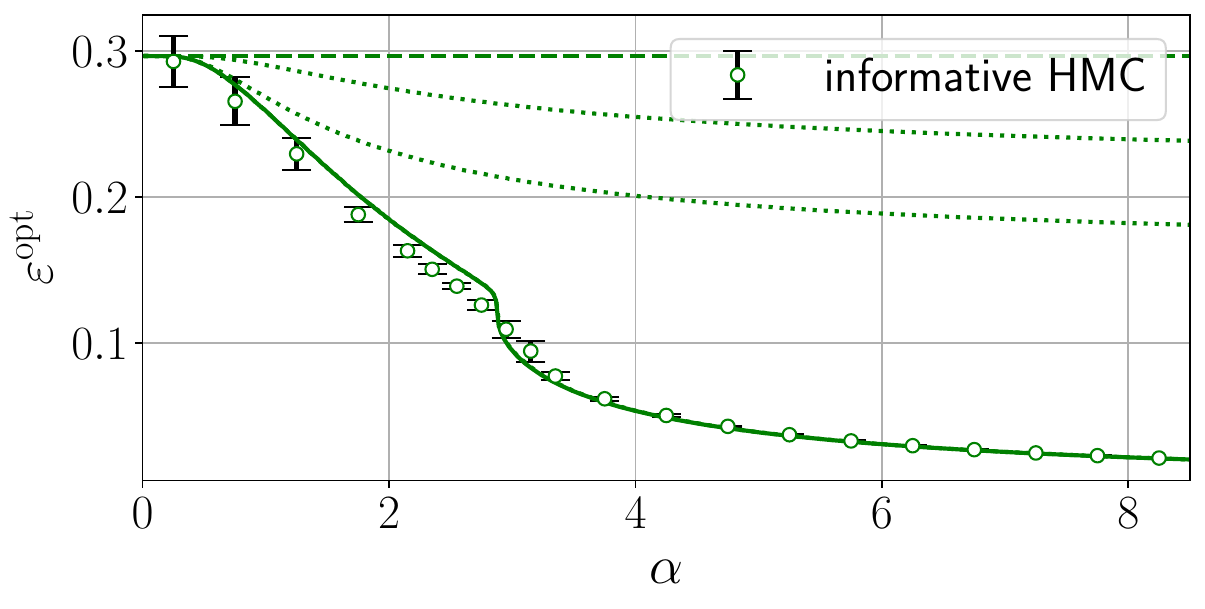} 
  \end{tabular}
  \vspace{-10pt}

  \caption{Theoretical prediction (green solid curve) of the Bayes-optimal mean-square generalisation error for $L=2$ with Gaussian inner weights, $\sigma(x)=\tanh(2x)/\sigma_{\tanh}, d=200, \gamma_1=\gamma_2=0.5,\Delta=0.2$ and different $P_v$ laws. The dashed line represents the universal branch. Dotted lines denote metastable specialisation branches of the RS saddle-point equations reached from different initialisations for the overlaps. From the top, the first green dotted line represents the solution reached by initialising $\mathcal{Q}_{2:1}> 0$ (i.e., positive for any argument value) and $\mathcal{Q}_{1},\mathcal{Q}_{2}\equiv 0$ ($0$ for all arguments), the second to $\mathcal{Q}_{1},\mathcal{Q}_{2:1},\mathcal{Q}_{2} >0$ (it yields in the left panel the small metastable solution just before the transition around $\alpha=2.8$; in the right panel, this solution collapses on the equilibrium curve), the third is $\mathcal{Q}_{1}> 0,\mathcal{Q}_{2:1},\mathcal{Q}_{2} \equiv 0$. The magenta dotted curve corresponds to initialisation $\mathcal{Q}_1(\mathsf{v}^{(2)})>0$ only for sufficiently large $\mathsf{v}^{(2)}$, while $\calQ_{2:1},\calQ_2 \equiv 0$: this solution is an inhomogeneous specialisation across layers (only the first specialises), and across the neurons in that layer (only some neurons specialise). Points are obtained with Hamiltonian Monte Carlo with informative initialisation. Each point has been averaged over 20 instances of the data, with error bars representing 1 standard deviation. The generalisation error is computed empirically from $10^4$ i.i.d. test samples. The readouts are fixed to the teacher’s during sampling. \textbf{Left}: Homogeneous readouts.  \textbf{Inset}: Optimal generalisation error of a shallow MLP (black line: $L=1$) and deep MLP (red line: $L=2$). The activation for both curves is $\sigma(x)=\tanh(2x)/\sigma_{\tanh}$, $\gamma,\gamma_1,\gamma_2 = 1$ and $\Delta = 0.2$, while $\alpha$ is divided by $L$ for comparison. \textbf{Right}: Gaussian readouts.}
  \label{fig:3LNNgen_error_gauss}
  \vskip -0.1in
\end{figure*}

\begin{figure}[t!!]
  \centering
  \includegraphics[width=0.955\linewidth,trim={0 0 0.0cm 0},clip]{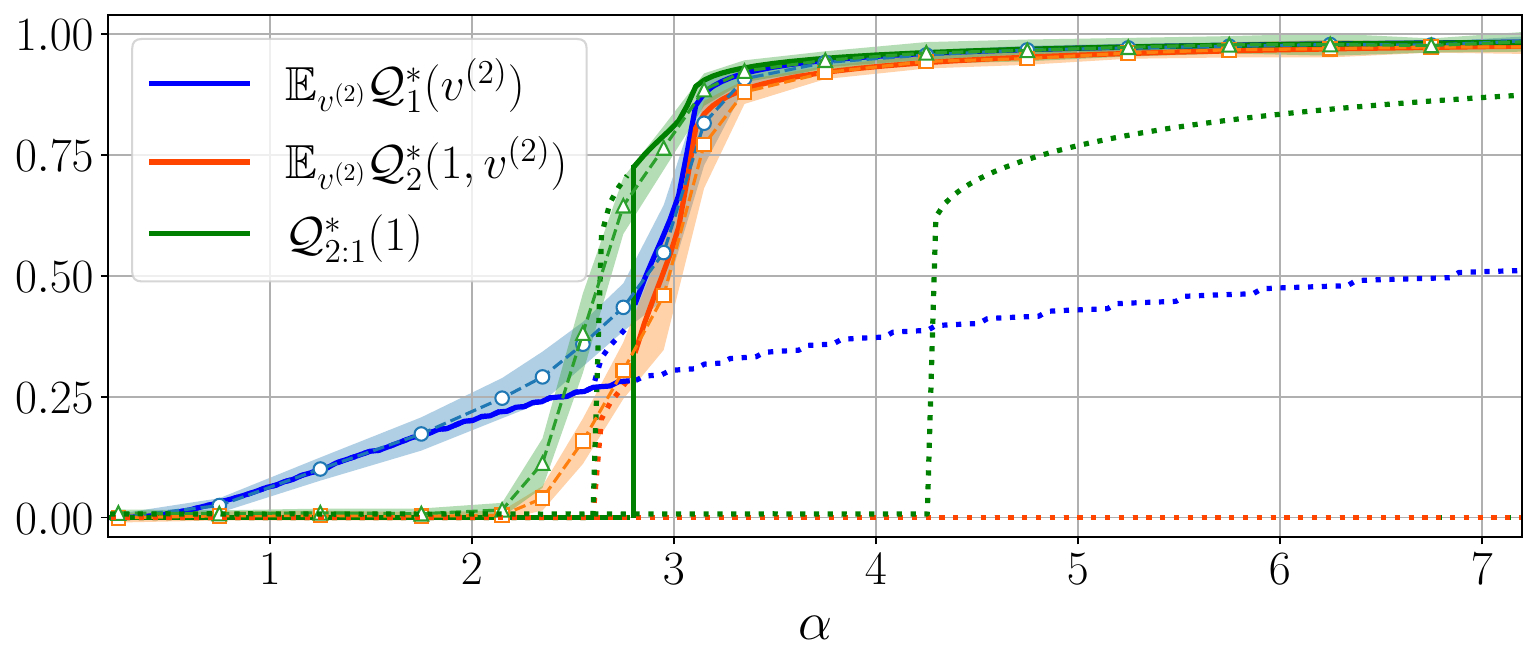} \\
  \vspace{-20pt}
  \includegraphics[width=0.955\linewidth,trim={0.0cm 0 0 0},clip]{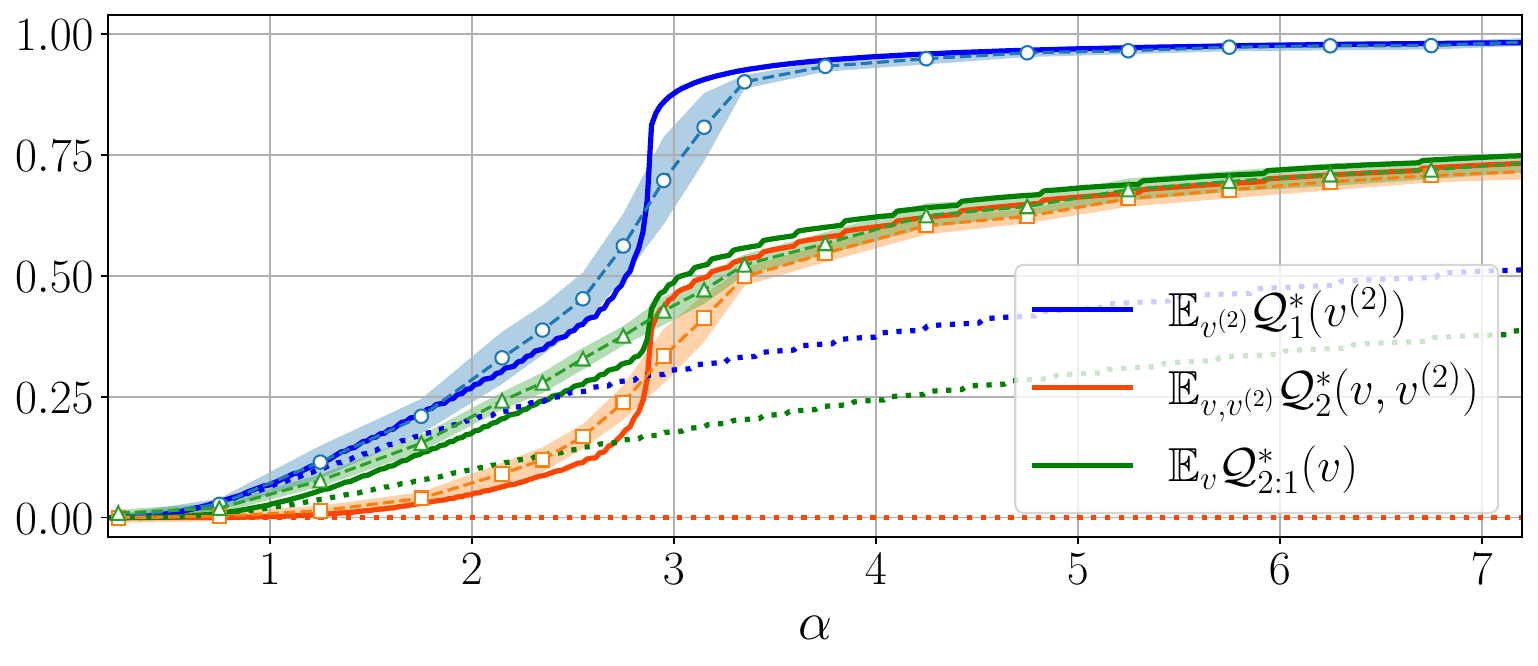}\hspace*{-0.1cm}
  \vspace{-5pt}
  \caption{Solid and dotted curves represent, respectively, the mean of different overlaps at equilibrium and in metastable specialised states, as function of the sampling ratio $\alpha$ for $L=2$ with Gaussian inner weights, $\sigma(x)=\tanh(2x)/\sigma_{\tanh},d=200, \gamma_1=\gamma_2=0.5,\Delta=0.2$. The shaded curves were obtained from informed HMC. Each point has been averaged over 20 instances of the training set, with 1 standard deviation depicted. The readouts are fixed to the teacher’s during sampling. \textbf{Top}: Homogeneous  readouts. \textbf{Bottom}: Gaussian  readouts.}
  \label{fig:v_non-constant_2}
\vskip -0.1in
\end{figure}

For a broad class of non-Gaussian inputs, our experiments show that the Bayes-optimal error is unchanged if the inputs entering the random MLP target are replaced by Gaussian vectors with the same covariance. This behaviour, implied in our theory by the Gaussian hypothesis \eqref{eq:hypothesis_gaussian_test}, was verified using inputs generated by feeding standard Gaussian vectors into one NN layer with Gaussian weights as well as real data from MNIST, see \figurename~\ref{fig:mnist-input}. The discrepancy between theoretical and experimental results at low $\alpha$ for MNIST data can be attributed to the few large outlier eigenvalues, while our theory is only dependent on the spectral density of the data covariance, which the outliers do not influence.

\subsection{Two hidden layers MLP}

In this section we present experiments for $L=2$ hidden layers, similar to the ones conducted for the shallow case. Adding one layer already makes the picture richer, in particular in terms of the learning phase transitions taking place and the information flow across layers.

\vspace{5pt}\paragraph*{Learning propagates from inner towards outer layers.} \figurename~\ref{fig:3LNNgen_error_gauss} displays the mean-square generalisation error for a network with hyperbolic tangent activation. The general picture is similar to the shallow case: at small $\alpha$ a single prior-independent solution exists; with more data, the specialisation solution branches out continuously. A clear transition occurs in the specialisation solution. The mechanism behind it is explained by \figurename~\ref{fig:v_non-constant_2} showing the evolution of the average over neurons of the overlap profile in each layer. 

We first discuss the case of Gaussian $\bv^0$, bottom of \figurename~\ref{fig:v_non-constant_2} and right of \figurename~\ref{fig:3LNNgen_error_gauss}. The ordering of the specialisations along layers is clear: $(i)$ The network starts specialising from the inner layer $\bW^{(1)}$. The information then propagates outwards. $(ii)$ The deep layer $\bW^{(2)}$ is then getting learned. This experimentally observed shallow-to-deep layer ordering of the learning is encoded in our formulas (including for $L\ge 3$): the RS equations imply that the overlap for the second layer can become non-zero only if the one for the first layer is itself non-vanishing.

Concerning the recovery of $\bW^{(2)}\bW^{(1)}$, it should be understood that products of weight matrices from different layers are learnable partly independently of their factors. This is similar to the shallow case, where the quadratic term $\bW^{0\intercal}{\rm diag}(\bv^0)\bW^{0}$ in the target can be partially recovered without learning $\bW^0$, and thus comes with its own OP which can be non-zero even if the one for $\bW^{0}$ vanishes (in the universal phase). The equations consistently encompass this possibility. 

The learning transitions occur more abruptly with homogeneous readouts, top of \figurename~\ref{fig:v_non-constant_2} and left of \figurename~\ref{fig:3LNNgen_error_gauss}: the learning of the product matrix and the deep one occur jointly and sharply. The homogeneity of the readouts is the source of the discontinuity of the learning transition which makes learning harder. Interestingly, continuous rather than discrete readouts induce smoother transitions. Nevertheless, notice the richer behaviour of the generalisation error and overlaps right after the first transition for homogeneous readouts, and also that in that region of $\alpha$, the ordering of overlap values is different between the two readout distributions.

We mention that \cite{yoshino2020complex,yoshino2023spatially,huang2025liquid} also predicted learning inhomogeneities across layers in a teacher-student setting but in a strongly overparametrised data~regime.

\vspace{5pt}\paragraph*{Inhomogeneous learning profile across neurons and matrix order parameters.} When there are two or more hidden layers, non-trivial overlap profiles emerge in each. This effect is a joint consequence of the depth and linear width of the network but also of the complex interactions among the layers (recall the discussion when we introduced $\mathcal{Q}_{2}^*(\mathsf{v},\mathsf{v}^{(2)})$ in Sec.~\ref{res:2layers}).

\begin{figure}[t!!]
  \centering
  \includegraphics[width=1\linewidth,trim={0.0cm 0 0 0},clip]{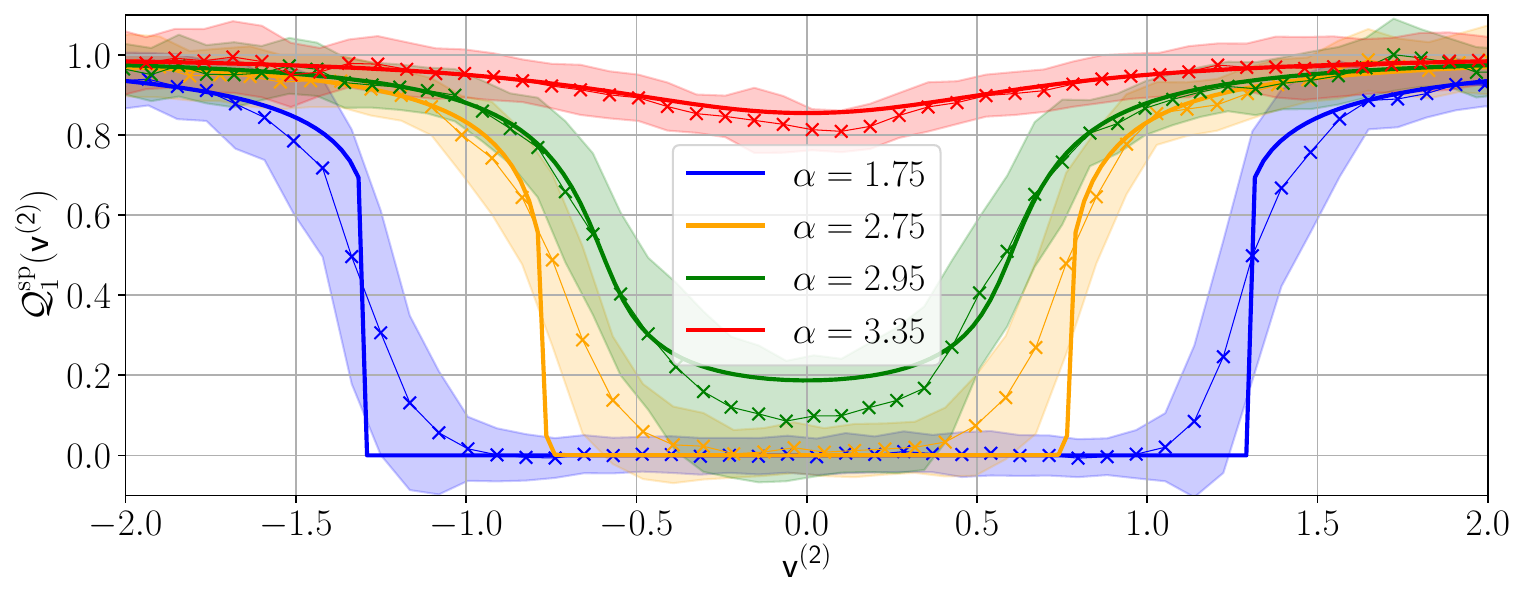} \\
  \vspace{-16.8pt}

 \includegraphics[width=1\linewidth,trim={0 0 -0.0cm 0},clip]{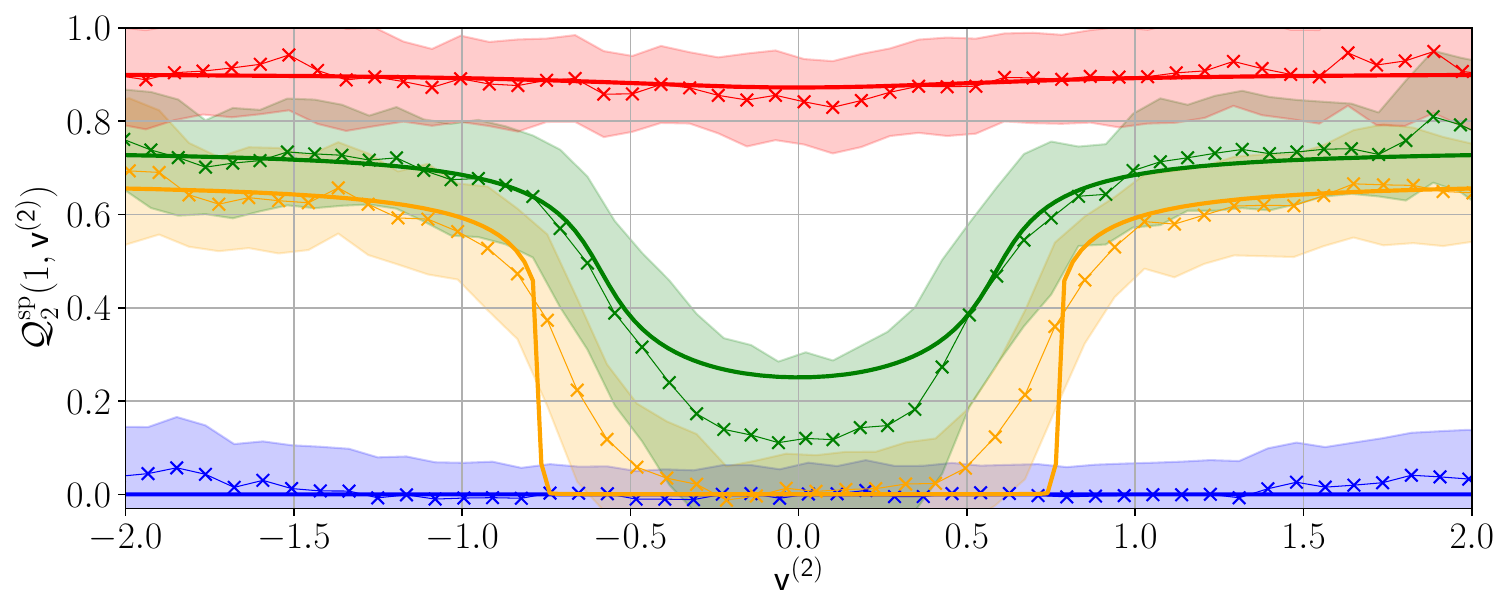}
  \vspace{+0pt}

  \includegraphics[width=1\linewidth,trim={0 0 -0.5cm 0},clip]{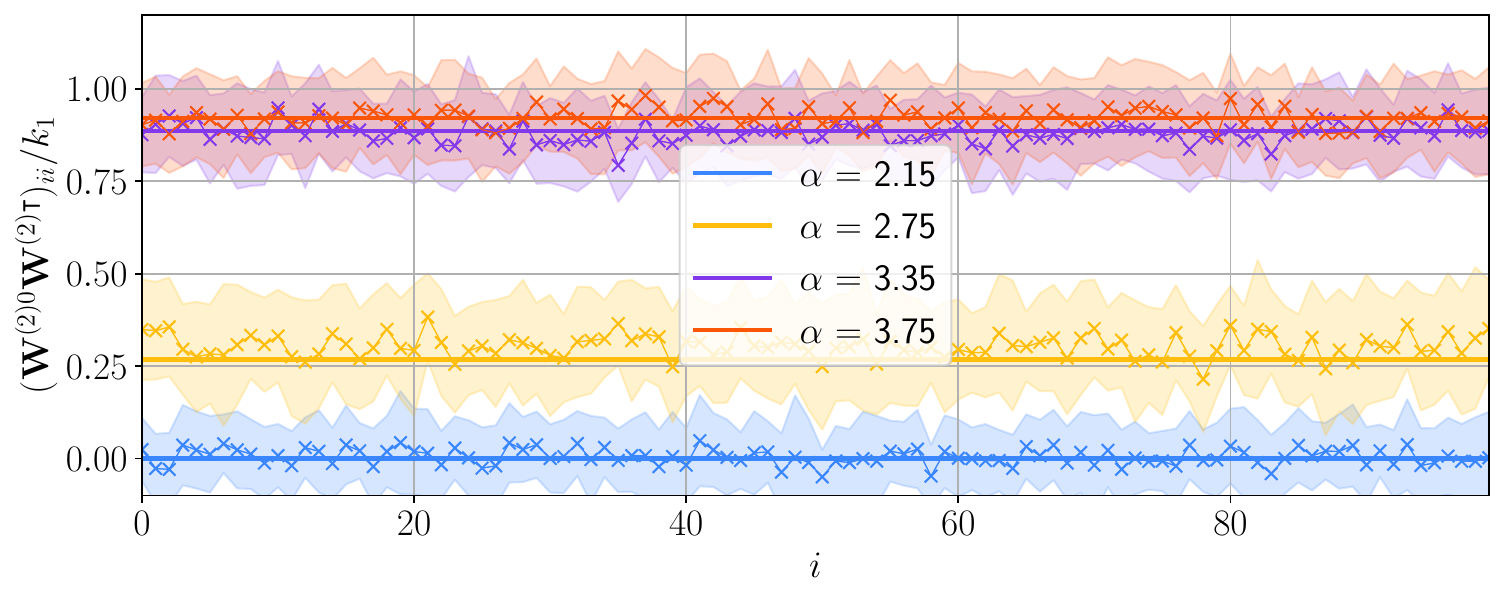}
  \vspace{-25.9pt}

  \includegraphics[width=1\linewidth,trim={0 0 -0.5cm 0},clip]{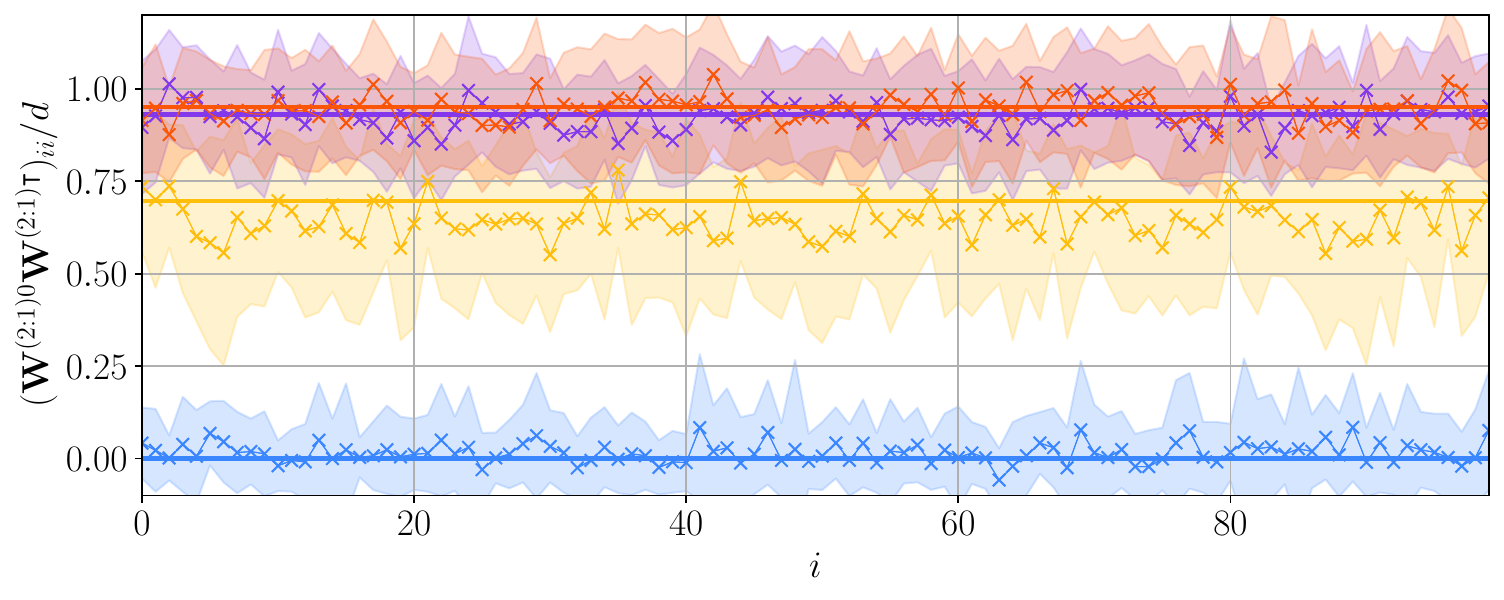}

  \caption{Theoretical predictions (solid curves) for the overlaps obtained from informative initialisation as functions of $\mathsf{v}^{(2)0}$ or $i=1,\dots,k_2$ for $L=2$ with activation $\tanh(2x)/\sigma_{\tanh},d=200, \gamma_1=\gamma_2=0.5,\Delta=0.2$, Gaussian inner weights, homogeneous quenched readouts and different $\alpha$ values. The shaded curves were obtained from informed HMC. Using singles posterior samples, the overlaps have been evaluated numerically by dividing the interval $[-2,2]$ into bins and by computing their value in each bin. Each point has been averaged over 20 instances of the data, and shaded regions around them correspond to 1 standard deviation.
  \textbf{First (top)}: First layer overlap $\mathcal{Q}^*_{1}(\mathsf{v^{(2)}})$ profile ordered according to the amplitude of the effective readouts $\bv^{(2)0}$. \textbf{Second}: The input (or column)-indexed overlap for the second layer $\mathcal{Q}^*_{2}(1,\mathsf{v^{(2)}})$ also ordered according to the effective readouts. \textbf{Third}: The neuron (i.e., output or row)-indexed overlap profile for the second layer. \textbf{Last}: The output-indexed overlap profile for the product matrix $\bW^{(2:1)}$.}
  \label{fig:3LNN_NSB_FSB}
\vskip -0.1in
\end{figure}

For the first hidden layer, top panel of \figurename~\ref{fig:3LNN_NSB_FSB}, the overlap inhomogeneity is related to the fluctuations in the effective readouts of the target $\bv^{(2)0}:=\bW^{(2)0\intercal}\bv^0/\sqrt{k_2}$: its components are Gaussian random variables. It implies that neurons are not all ``measured equally well'' (in particular through the linear term in the Hermite expansion of $\sigma$ in the first layer). The profile therefore manifests itself along the output (row) dimension.

For the second layer, with homogeneous readouts, the only inhomogeneity is along its input (column) dimension and is induced by the output profile of the first layer weights, second panel. For completeness we checked that, consistently with the theory, the profile of the second layer overlap along its output dimension is indeed constant, third panel. Due to the homogeneity of readouts, a similar constant overlap profile along the output dimension of the product matrix $\bW^{(2:1)}$ appears (last panel).

These experiments probing all possible overlap inhomogeneities in a three layers MLP vindicate our definitions and indexing of OPs in the theory.

\begin{figure}[ht!]
  \centering
  \includegraphics[width=0.73\linewidth]{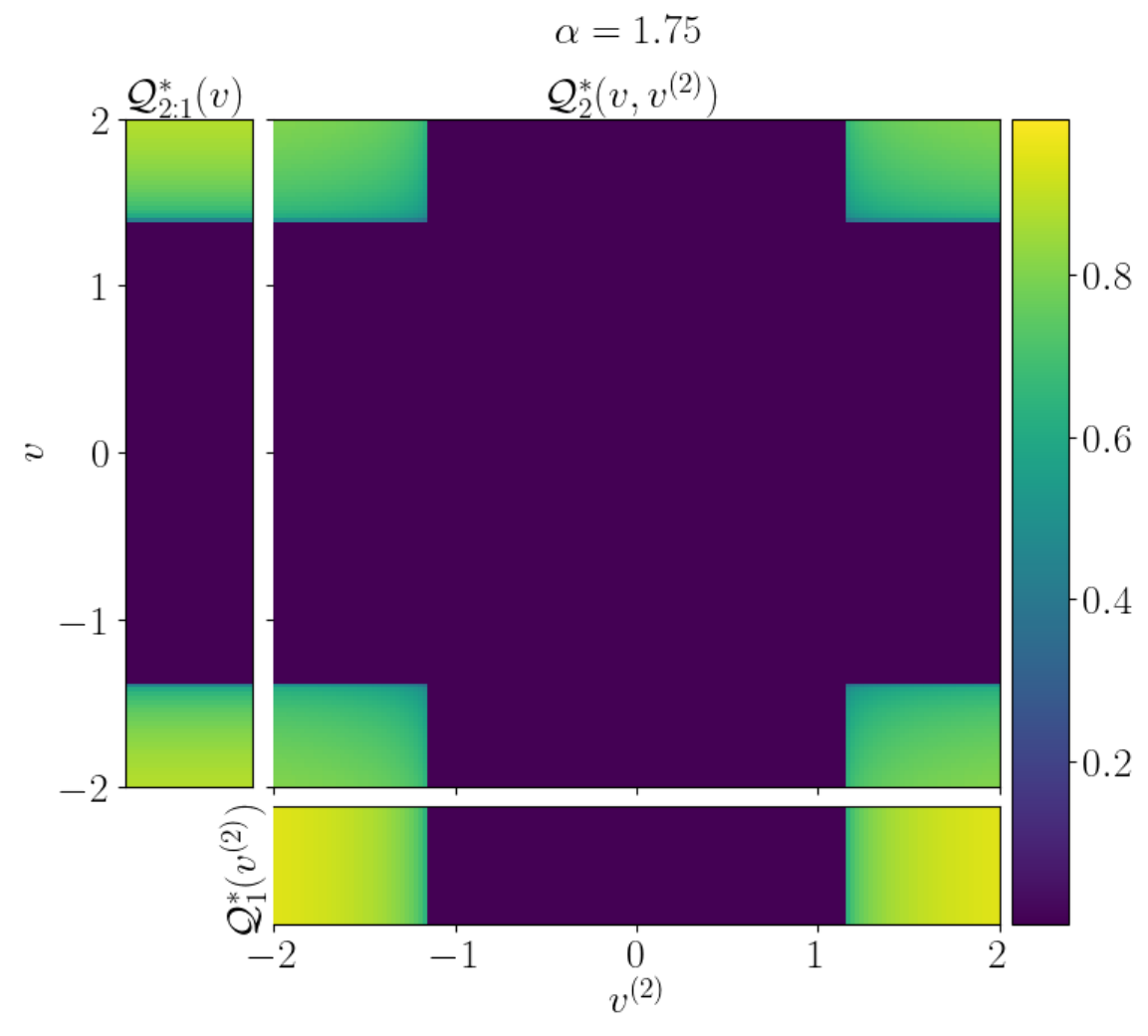} \
  \includegraphics[width=0.73\linewidth]{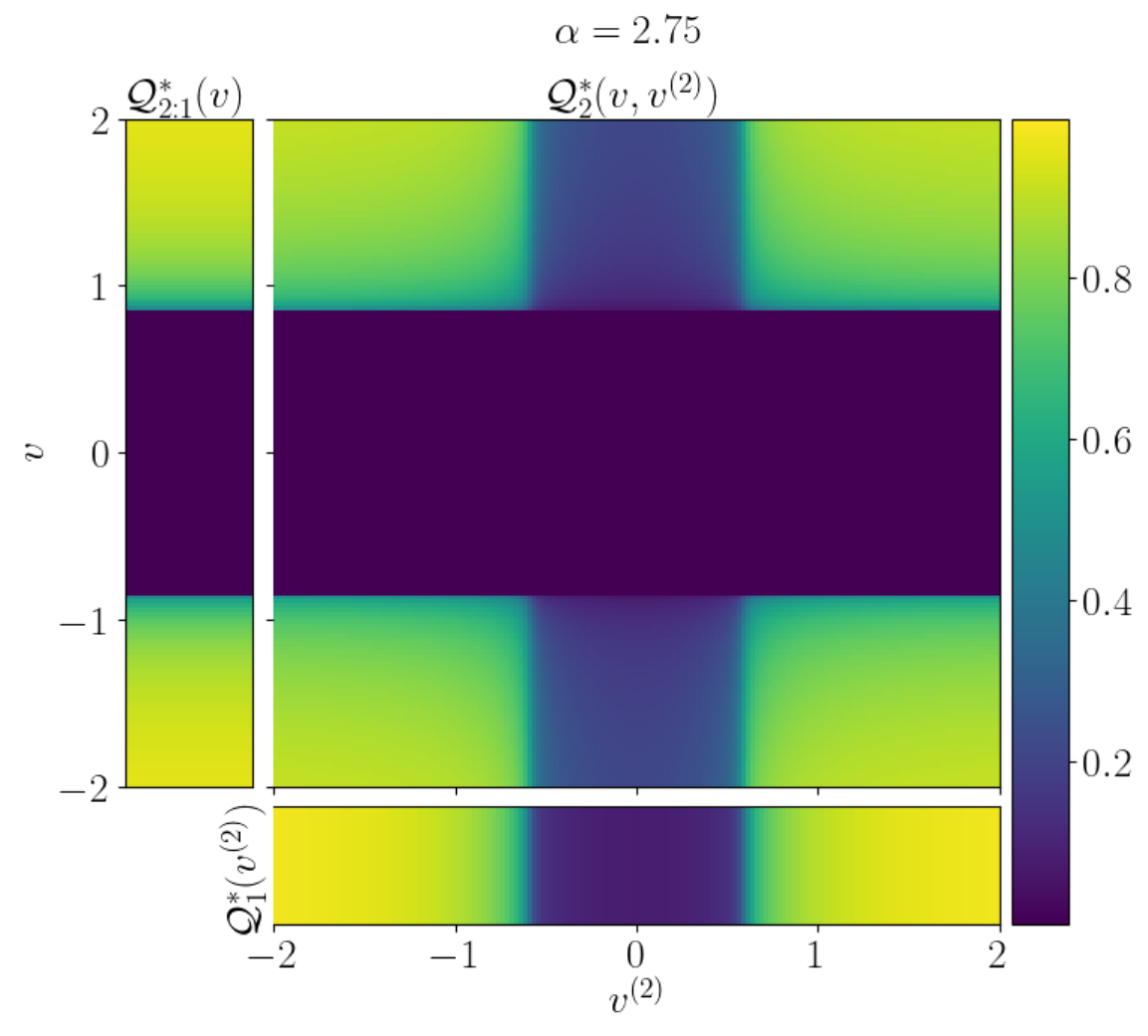} \\
  \includegraphics[width=0.73\linewidth]{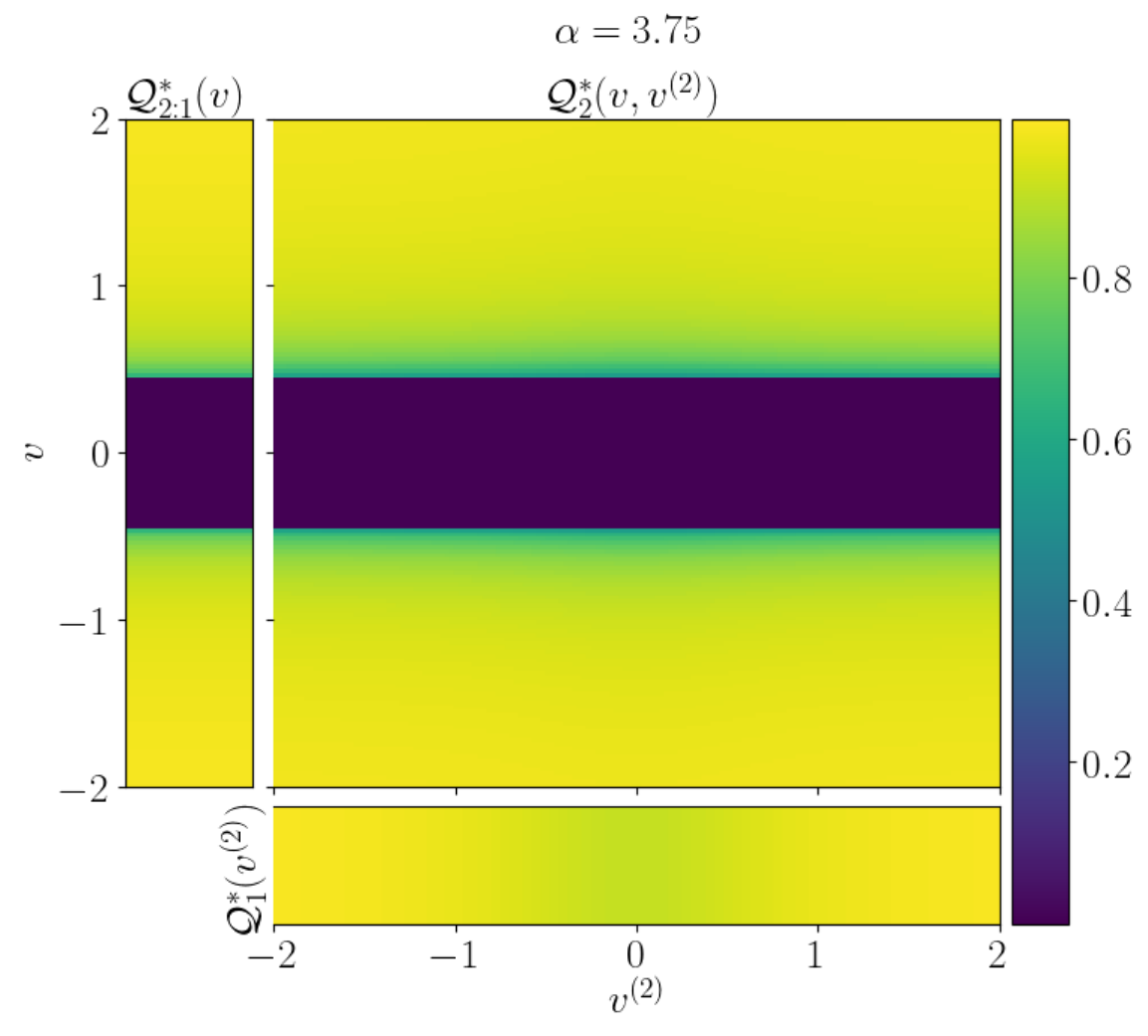}\hspace*{-0.1cm}
  \vspace{-5pt}
  \caption{Heat-maps of all the theoretical equilibrium overlaps as function of the sampling rate $\alpha\in\{1.75, 2.75, 3.75\}$ (increasing from top to bottom) for $L=2$ with Gaussian inner and readout weights, $\sigma(x)=\tanh(2x)/\sigma_{\tanh}, \gamma_1=\gamma_2=0.5,\Delta=0.2$, which is the same setting as the right panel of  \figurename~\ref{fig:3LNNgen_error_gauss} and bottom panel of \figurename~\ref{fig:v_non-constant_2}. \textbf{Left column:} Product matrix overlap $\mathcal{Q}^*_{2:1}(v)$. \textbf{Bottom row:} First layer overlap $\mathcal{Q}^*_1(v^{(2)})$. \textbf{Central square:} Second layer overlap $\mathcal{Q}^*_2(v,v^{(2)})$. The overlap arguments in the theory are amplitudes $\mathsf{v},\mathsf{v}^{(2)}>0$. However, we plot them here as a function of the actual signed readouts and effective readouts values $v,v^{(2)}$ for better visualisation of what is going on in the network. These figures fully capture the features learnt along the layers, and across the rows and columns of each layer weight matrix, in a three layers neural network.}
  \label{fig:matrixOPs}
\vskip -0.1in
\end{figure}

As an illustration of what is learnt in the deep NN when increasing the data, we plot in \figurename~\ref{fig:matrixOPs} the three functional overlap OPs using heat-maps. Two carry a single argument and the one for the second inner layer possesses two, as it captures at a macroscopic level the learning inhomogeneities along both the rows and columns of $\bW^{(2)}$ (we restricted the domains of the OPs for visualisation despite the true ones are in principle infinite). As $\alpha$ exceeds the smooth transition happening around $\alpha=2.8$ in the same setting of bottom panel of \figurename~\ref{fig:v_non-constant_2}, we see that specialisation nucleates in $\bW^{(2)}$ starting from its neurons (i.e., rows) indexed by the largest readout amplitudes, $|v|$, and concurrently from its ``dual neurons'' (i.e., its columns) connected to the largest effective readout amplitudes $|v^{(2)}|$. Specialisation then propagates towards lower values as $\alpha$ increases. The figures emphasise how the learning of the other matrices (the first layer weights and the product matrix, which both display learning inhomogeneities along one dimension only) interact with the deep one and yield such an intricate behaviour. 

\begin{figure}[ht!]
    \centering
    \includegraphics[width=0.73\linewidth]{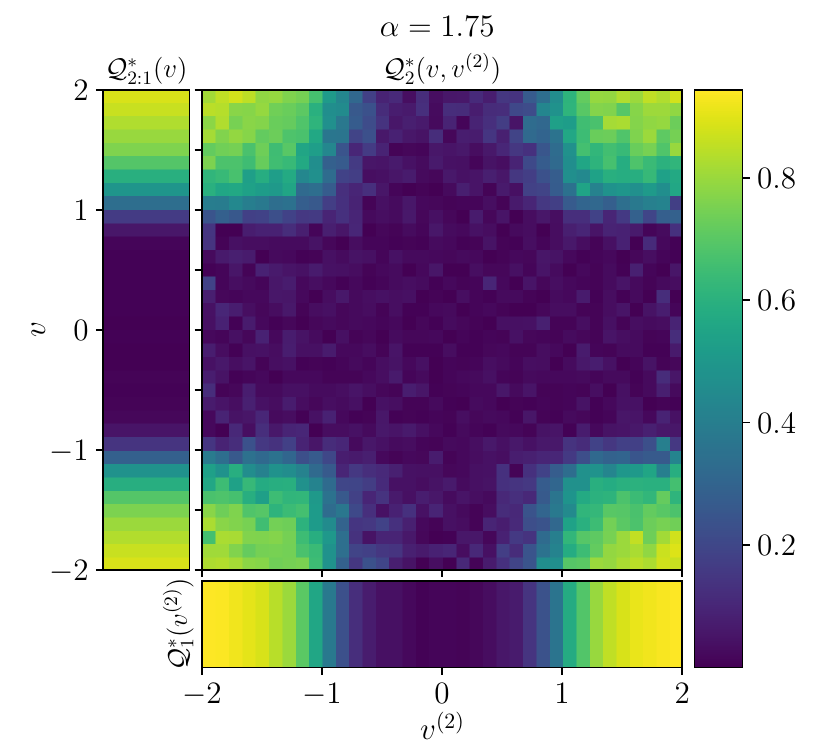}\\
    \includegraphics[width=0.73\linewidth]{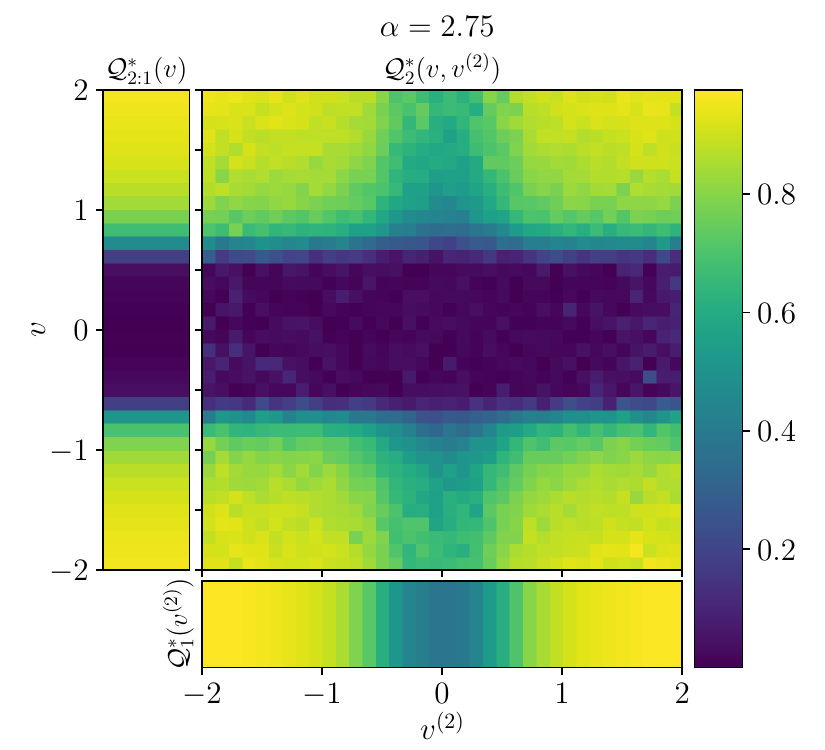}\\
    \includegraphics[width=0.73\linewidth]{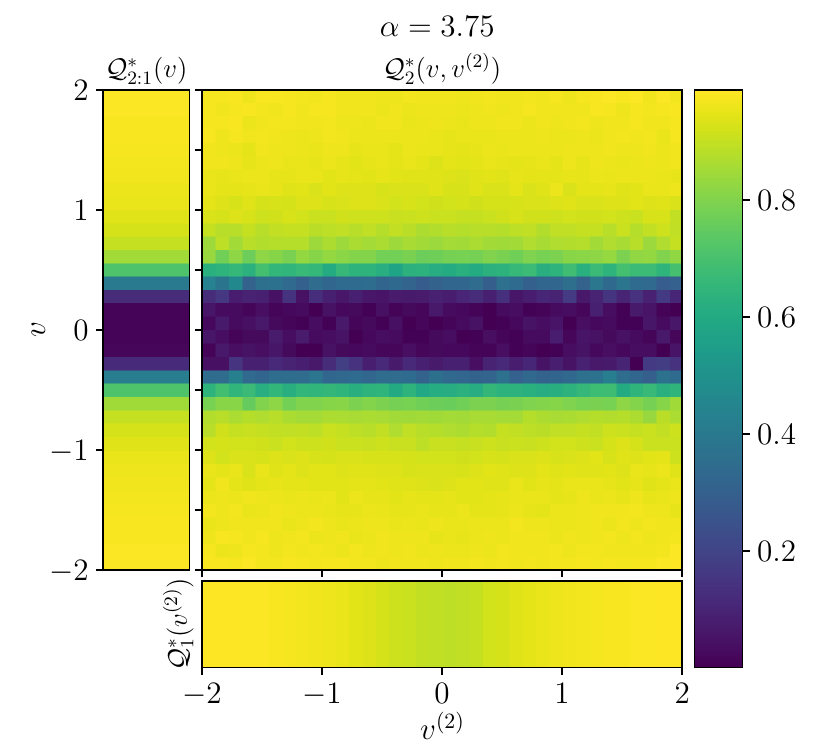}
    \vspace{-5pt}
    \caption{Heat-map of all the empirical overlaps on grids of (effective) readouts $(\mathsf{v},\mathsf{v}^{(2)})$. Besides the local average performed by binning the distribution of the weights according to the grid of readouts, we average over $100$ instances of data-student pairs. 
    Here $d=200$, $\alpha=1.75$ (\textbf{top}), $\alpha=2.75$ (\textbf{middle}) and $\alpha=3.75$ (\textbf{bottom}), while the rest is as in \figurename~\ref{fig:matrixOPs}. For the second layer overlap, the values are rearranged putting those associated with highest readout values on the corners of the image, as in \figurename~\ref{fig:matrixOPs}. Up to finite size fluctuations, the same patterns as the theoretically predicted ones, top and bottom panels of \figurename~\ref{fig:matrixOPs}, clearly appear.}
    \vspace{32pt}
    \label{fig:local_averages_overlaps}
\end{figure}

Finally, in \figurename~\ref{fig:local_averages_overlaps}, we display a result of a numerical experiment for $\calQ_2(\mathsf{v},\mathsf{v}^{(2)})$. This figure was realised by averaging over different instances of a data-student pair, with the overlap values ordered as those of \figurename~\ref{fig:matrixOPs} for each pair, and then by performing a ``local average'' of neighbouring indices on the grid $(\mathsf{v},\mathsf{v}^{(2)})$ in order to suppress the ``microscopic fluctuations''. The latter should be interpreted in the thermodynamic limit as being over a relatively small patch, which still contains $\Theta(d^2)$ weights. Remarkably, this figure could have also been generated from a single instance of the student, as the local average alone is sufficient to reproduce the patterns of \figurename~\ref{fig:matrixOPs}.

Notice that in contrast with the shallow case where $\calQ(\mathsf{v})$ vanishes for $\mathsf{v}$ small even at large sampling ratios (see \figurename~\ref{fig:v_non-constant}), the top panel of \figurename~\ref{fig:3LNN_NSB_FSB} and \figurename~\ref{fig:matrixOPs}, \ref{fig:local_averages_overlaps} show that, for sufficiently large $\alpha$, overlaps indexed by $\mathsf{v}^{(2)}$ become non-zero for any value of the index. This occurs because the effective readouts enter the covariance $K^{(2)}(\bar \calQ)$ from Result~\ref{res:free_entropy_L=2} in a different way compared to the actual readouts $\mathsf{v}$.

\begin{figure}[t!!]
  \centering
  \includegraphics[width=0.945\linewidth,trim={0 0 0.0cm 0},clip]{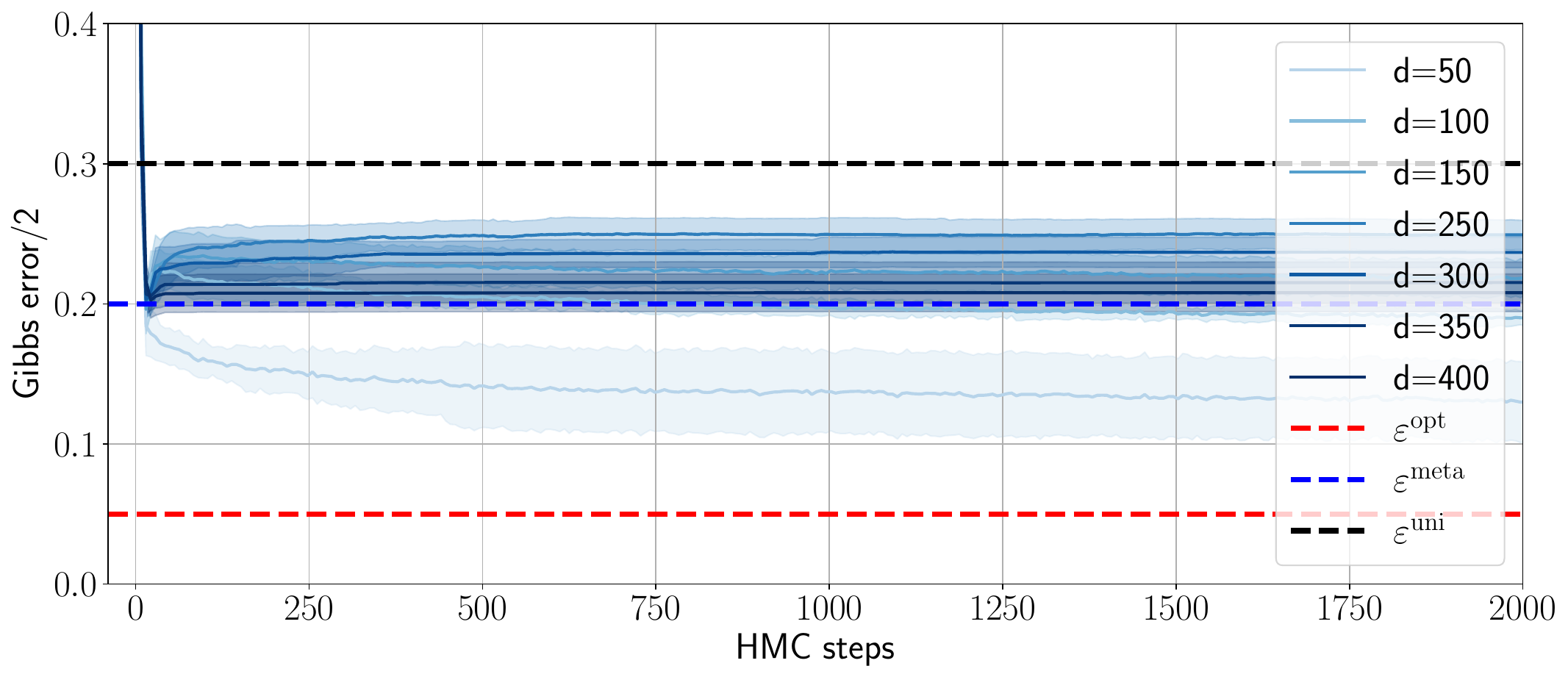} \\
  \vspace{0pt}
  \includegraphics[width=0.97\linewidth,trim={0.0cm 0 0 0},clip]{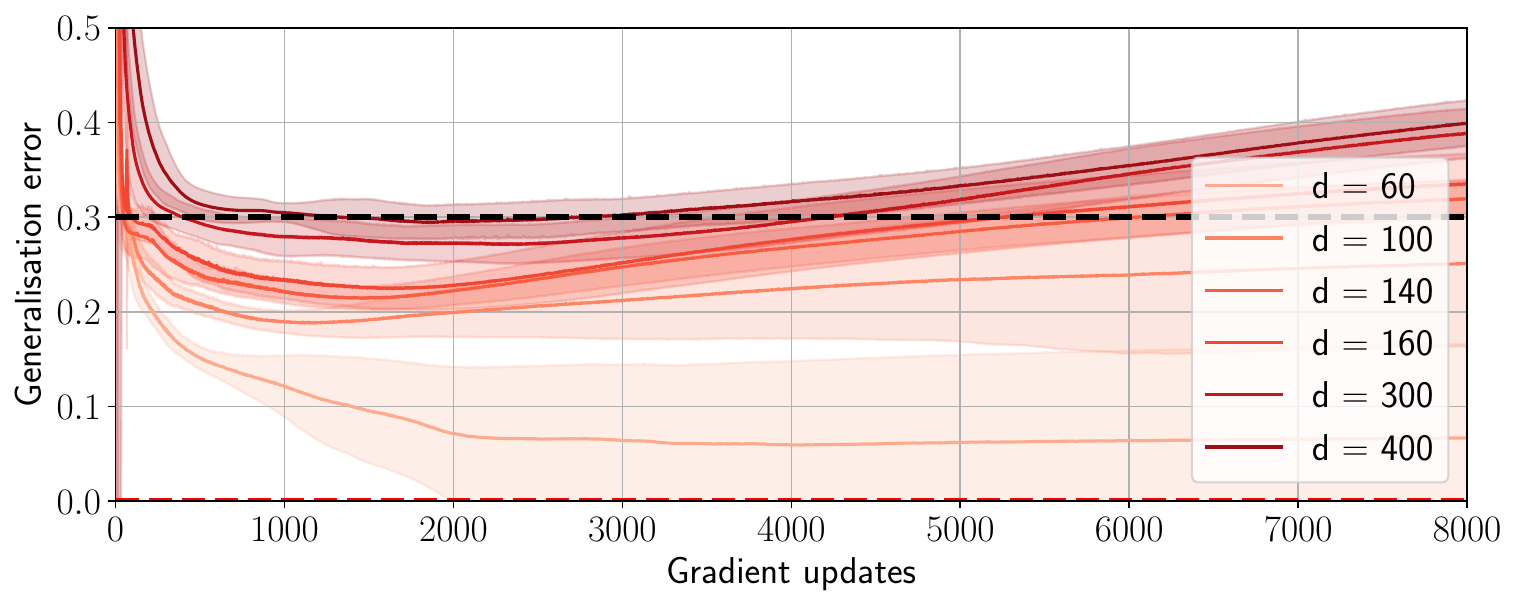}\hspace*{-0.1cm}
  \vspace{-5pt}
  \caption{Generalisation errors, computed empirically from $10^4$ i.i.d. test samples, of HMC (ADAM) as function of the number of steps (gradient updates). Errors are averaged over 10 instances; shaded areas indicate 1 standard deviation. Here $L=2,\sigma(x)=\tanh(2x)/\sigma_{\tanh}, \gamma_1=\gamma_2=0.5,\alpha=4.0$ with Gaussian inner weights and homogeneous target readouts for both plots. Dashed lines represent the theoretical errors associated with equilibrium and metastable solutions.   \textbf{Top}: Half Gibbs error of HMC from random initialisation as a function of the number of updates for various $d$ with $\Delta=0.2$. The readouts are quenched during sampling. \textbf{Bottom}: Generalisation error of ADAM from random initialisation as a function of the gradient updates for various $d$ with $\Delta=10^{-4}$. The initial learning rate is $0.01$ and batch size $\lfloor n/4 \rfloor$. The student has learnable readout layer.}
  \label{fig:ADAM_HMC_hardness_L=2}
\vskip -0.1in
\end{figure}

\vspace{5pt}\paragraph*{Algorithmic hardness and partial specialisation.} We repeated experiments probing the behaviour of HMC and ADAM similar to the shallow case, see \figurename~\ref{fig:ADAM_HMC_hardness_L=2}. Starting with HMC, two noticeable differences appear compared to $L=1$. Firstly, reaching the specialised equilibrium state from uninformative initialisation seems \emph{much} harder/costly than in the shallow case; we do observe the descent towards it only for a rather small size ($d=50$).

The most striking difference, however, is the nature of the state that HMC experiences. For $L=1$, HMC was getting attracted by the metastable state associated with the universal solution (top panel of \figurename~\ref{fig:HMC_hardness}), with the single inner layer not specialising. With one more hidden layer a richer picture emerges. The top panel of \figurename~\ref{fig:ADAM_HMC_hardness_L=2} shows that in polynomial-in-$d$ time, before complete specialisation ultimately occurs when the chain equilibrates, HMC is now stuck in a \emph{partially specialised metastable state}. There, the \emph{first} layer has specialised (i.e., has been partly recovered) while the second has not. We tracked the experimental overlaps, not depicted here, and they confirm this picture. The theory correctly predicts this state: it is the third green dotted curve from the top in \figurename~\ref{fig:3LNNgen_error_gauss} left panel, which at $\alpha=4$ corresponds to the dashed blue curve $\varepsilon^{\rm meta}= 0.2$ in \figurename~\ref{fig:ADAM_HMC_hardness_L=2} (the two plots are done in the same setting). This is very interesting: somehow, depth helps from that perspective, since this mechanism cannot be observed in the shallow case. The reason is the presence of the effective readouts in the target $\bv^{(2)0}=\bW^{(2)0\intercal}\bv^0/\sqrt{k_2}$ when $L=2$, that inhomogeneously ``measure'' the inner layer.

Apart from this stable metastable state, there are other solutions of the RS equations. The magenta curve in \figurename~\ref{fig:3LNNgen_error_gauss} depicts a case where only the first layer specialises, and only for a subset of its neurons. The same effect is observed in shallow MLPs for their single hidden layer (see \figurename~\ref{fig:gen_error_gauss}, right panel). Although these solutions exist, their stability is not systematically analysed here, as this poses a significant challenge both for the theory and even more so in the numerical experiments due to strong finite-size effects and the difficulty in training large deep Bayesian NNs. We leave an investigation of their stability for future work. 

When learning with ADAM (bottom panel of \figurename~\ref{fig:ADAM_HMC_hardness_L=2}), we instead observe a similar scenario as with $L=1$ with $\tanh$ activation (lower panel of \figurename~\ref{fig:hardness_main}): as $d$ increases ADAM gets stopped by the metastable state associated with the universal solution, and does not reach the better partially specialised one. This suggests that HMC is able to ``see'' metastable states that outperform $\varepsilon^{\rm uni}$, while ADAM is not in the tested cases. It would be interesting to understand why.

\subsection{Three or more hidden layers}

\begin{figure*}[t!!]
  \centering
  \setlength{\tabcolsep}{4pt} 
  \begin{tabular}{ccc}   
    \includegraphics[width=0.4\textwidth]{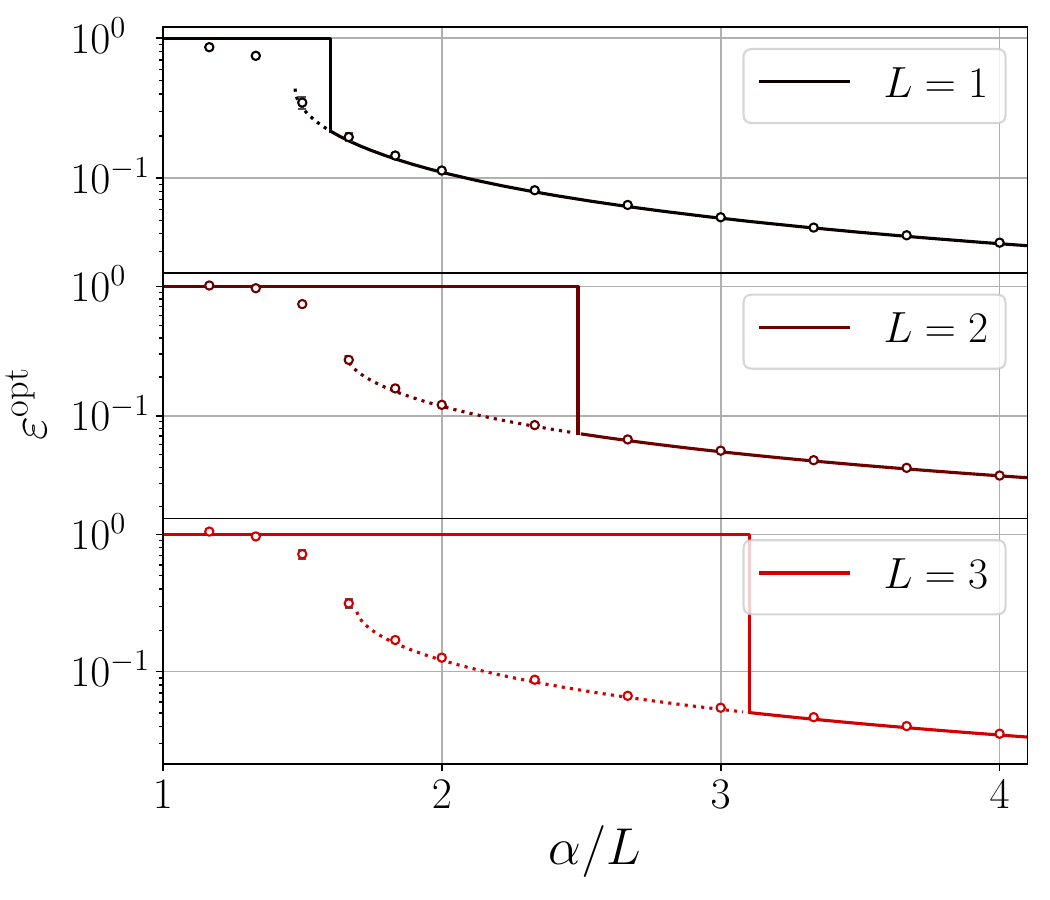}  &
    \includegraphics[width=0.4\textwidth]{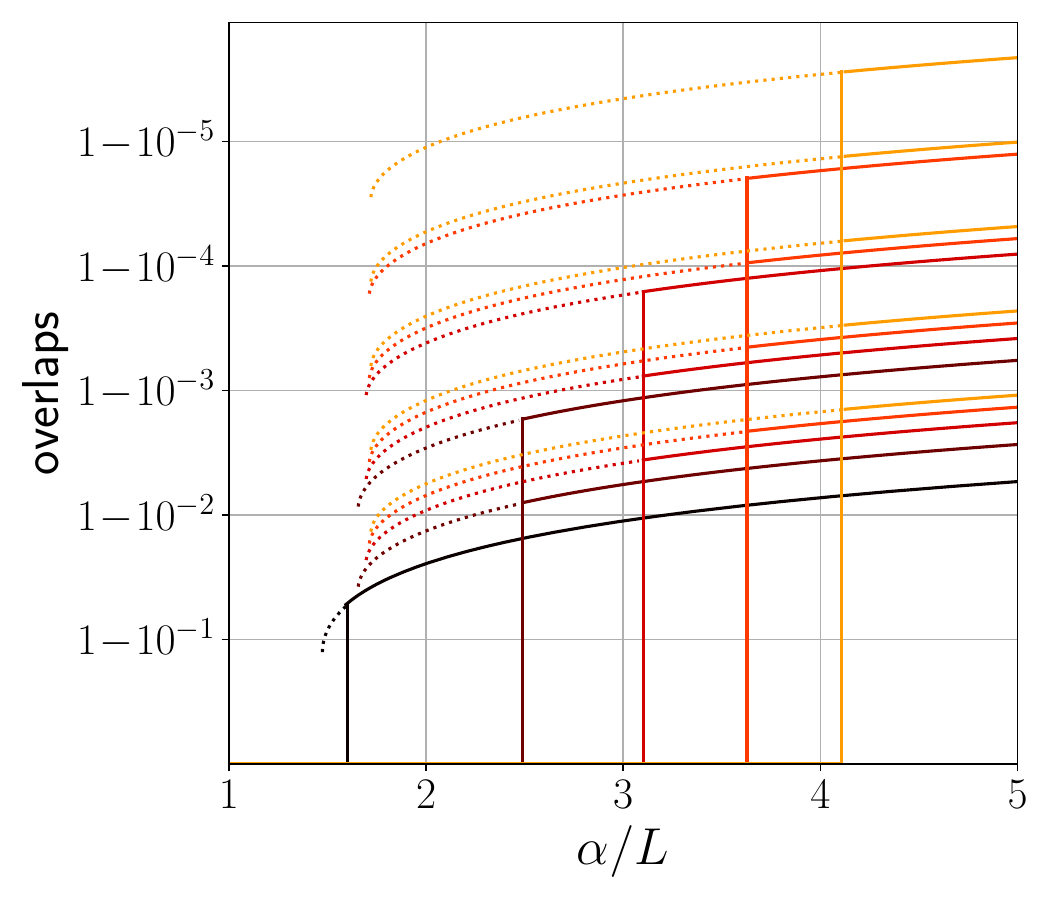} 
  \end{tabular}

\caption{Theoretical predictions for deep NNs with homogeneous readout $v_i=1$ and the activation on every layer given by $\sigma(x) =  (\tanh(2x) - \mu_1 x)/c$, where $\mu_1$ is the first Hermite coefficient of $\tanh(2x)$ and the constant $c$ is s.t. $\E_{z \sim \mathcal N(0,1)} \sigma(z)^2 = 1$; $\gamma_L=\dots=\gamma_1=1$, and the number of hidden layers $L\in\{1,\ldots,5\}$. \textbf{Left:} Bayes-optimal error (solid curves) as function of $\alpha/L$ for $L\le 3$, while dotted lines are metastable solutions of the RS equations. The points and error bars represent the mean and standard deviation of half the Gibbs errors, evaluated on 9 data instances at $d = 50$ for all points except the second and third points (from left to right) for $L = 3$, which are computed at $d = 30$ since HMC remains stuck at initialisation for higher $d$. \textbf{Right:} Overlap $Q^{(L)*}_{l}$ of the $l$-th layer weights in the $L$-hidden layers NN for each pair $(l\le L,L)$. For each $L$, all phase transitions for different $Q^{(L)*}_l$ occur concurrently, with the overlaps decreasing with the layer index: $Q^{(L)*}_1 > \dots > Q^{(L)*}_L$ after the transition.} 
  \label{fig:deep-tanh}
  \vskip -0.1in
\end{figure*}

\vspace{5pt}\paragraph*{Deeper is harder.}  \figurename~\ref{fig:deep-tanh} displays the theoretical and experimental Bayes-error and the theoretical overlaps corresponding to informative initialisation. A number of observations can be made. Let us start with the left panel. Note that the sampling rate $\alpha$ in the abscissa is rescaled by the number of layers to make a fair comparison between targets with different depths. We conclude based on the location of the specialisation transition (common to all layers under (\hyperlink{H:H3}{$H_{3}$})) and the ordering of the generalisation error curves with $L$ that \emph{the deeper the target, the more data per layer it requires to be recovered, and the higher is the $\varepsilon^{\rm opt}$ at given} $\alpha/L$. This vindicates information-theoretically the intuitive picture that the more ``non-linear'' is a task/target, in the present case through more layers, the harder it should be to learn it. Another confirmation of this fact is provided by the inset of \figurename~\ref{fig:3LNNgen_error_gauss}: also in the case of normalised $\tanh$ activation the amount of data per layer remains greater for the deep case, in spite of the fact that $\mu_1\neq 0$ implies the presence of effective Gaussian readouts $\bv^{(2)0}:=\bW^{(2)0\intercal}\bv^0/\sqrt{k_2}$ ``measuring'' the inner layer $\bW^{(1)0}$, which allow it to specialise for small sampling rate.

Another way to see from the results that depth is linked to hardness is through the right panel depicting the overlaps in each layer: for a given $L\ge 2$ and at fixed sampling rate (non rescaled by $L$ this time), overlaps $Q_l^*$ are monotonically decreasing with the layer index $l$. In other words, deeper features are harder to learn than the shallow ones, which confirms for generic $L$ the shallow-to-deep ordering of the learning observed for $L=2$ in the previous section. This is again rather intuitive and matches what is observed when deploying neural networks on real tasks \cite{raghu2017svcca}; see also \cite{cagnetta2024deep} for the role of depth in NNs when learning in a hierarchical task. 

In addition to the ordering of overlaps across layers within the same NN, another ordering is also evident from \figurename~\ref{fig:deep-tanh}. Letting $Q^{(L)*}_{l}$ be the $l$-th layer equilibrium overlap for a NN with $L$ hidden layers we have
\begin{talign*}
Q^{(L+1)*}_{l+1} \leq Q^{(L)*}_{l} \quad \text{for}  \quad  1\leq l \leq L.
\end{talign*}
It follows from $Q^{(L+1)*}_{l+1} \leq Q^{(L+1)*}_{l+1} |_{\bW^{(1)}=\bW^{(1)0}} = Q^{(L)*}_{l}$, where the equality is a consequence of the fact that, since $\sigma$ has $\mu_0=\mu_1=\mu_2=0$ and $\E_{z \sim \mathcal N(0,1)} \sigma(z)^2 = 1$, the data after the first layer have a covariance indistinguishable from $I_d$ from the perspective of the NN. Thus, the $(L+1)$-hidden layer NN with quenched first layer is equivalent to one with $L$ hidden layers and standard Gaussian inputs.

\section{Replicas plus HCIZ, revamped}\label{sec:theory}
The goal is to compute the asymptotic free entropy by the replica method \cite{mezard1987spin}, a powerful approach from spin glass theory also used in machine learning \cite{engel2001statistical}, combined with the HCIZ integral. We focus first on the derivation of the results for the shallow case $L=1$ which comes with its own set of difficulties due to the presence of $\mu_2\neq 0$ in $\sigma$. We will later move on to the deep case where, even when considering $\mu_2=0$, a different kind of difficulty will appear due to the multi-layer structure.

Our derivation is based on three key ingredients. 

$(i)$ The first ingredient is a Gaussian ansatz on the replicated post-activations, which generalises Conjecture~3.1 of~\cite{cui2023bayes}, now proved in~\cite{camilli2025inforeduction}, where it is specialised to the case of linearly many data ($n=\Theta(d)$). To obtain this generalisation, we will write the kernel arising from the covariance of the post-activations as an infinite series of scalar OPs derived from the expansion of the activation function in the Hermite basis, following an approach recently devised in~\cite{aguirre2024RF} in the context of the random feature model (see also \cite{hu2024asymptotics} and \cite{10.1214/20-AOS1990}).

$(ii)$ The second ingredient, exposed in ``Simplifying the order parameters'', amounts to a drastic reduction of the number of OPs entering the covariance of the post-activations through the realisation that infinitely many of them are expressible in terms of a few, more fundamental (functional) OPs.

$(iii)$ The last ingredient is a generalisation of an ansatz used in the replica method by \cite{sakata2013} for dictionary learning, which will allow us to capture important correlations discarded by these earlier approaches \cite{sakata2013,kabashima2016phase}. Our ansatz, explained in subsection ``Tackling the entropy'', is the crux for capturing the lack of rotational invariance and matrix nature of the problem when $\sigma$ possesses $\mu_2\neq 0$ when $L=1$ or $\mu_1\neq 0$ when $L=2$. We will see that, surprisingly, the HCIZ integral remains central despite the absence of rotational symmetry. App.~\ref{app:alternatives} provides a comparison with the approach of \cite{sakata2013}. 

For the sake of presentation, we discuss in the main only the non-standard steps corresponding to these ingredients. The complete derivations are presented in App.~\ref{app:replicas} for the shallow case and App.~\ref{app:replicas_deep} for the deep.

\vspace{5pt}\vspace{5pt}\paragraph*{Fixing the readouts.} 

We use the fact that, having as goal the computation of the leading order of the free entropy, the readouts $\bv$ of the learner can be fixed from the beginning to those of the target $\bv^0$. The proof has been given in Remark~\ref{rmk:readouts} for the mutual information (and further discussed and tested in App.~\ref{subsec:learnable_readouts}), and implies directly at the level of free entropy that
\begin{talign*}
\frac1n\mathbb{E}\ln \mathcal{Z}_{\bv=\bv^0}=\frac1n\mathbb{E}\ln \mathcal{Z}_{\bv\, \rm{learnable}}+O(1/d).    
\end{talign*}
Consequently, for the rest of the derivations we set without loss of generality $\bv=\bv^0$, thus leaving as learnable parameters the (many more) inner weights. If keeping learnable readouts, the (equivalent) replica calculation would be more cumbersome and ultimately yield that the overlap between $\bv$ and $\bv^0$ is irrelevant for what matters all the other OPs in the universal phase; and once the student specialises, the readouts corresponding to specialised neurons are concurrently exactly recovered.

\subsection{Shallow MLP}

We start with the shallow case $L=1$. Having directly fixed the readouts, the partition function is re-defined as
\begin{talign*}
\calZ(\mathcal{D}):=\calZ_{\bv=\bv^0}=\int dP_W(\bW)\prod_{\mu\le n}P_{\rm out}\big(y_\mu\mid \lambda_\mu(\btheta)\big)   
\end{talign*}
with  $\lambda_\mu(\btheta):=\mathcal{F}_{\btheta}^{(1)}(\bx_\mu)$ and $\btheta=(\bW,\bv^0)$. The quenched variables, averaged by the symbol $\EE=\EE_{\mathcal{D}}$, are the data $\mathcal{D}$ which depend on inputs and teacher. Equivalently, $\EE[\,\cdot\,]=\EE_{(\bx_\mu)} \EE_{\btheta^0}\EE_{(y_\mu)\mid (\bx_\mu),\btheta^0}[\,\cdot\,]$.

\vspace{5pt}\vspace{5pt}\paragraph*{Replicated system and order parameters.}

The starting point to tackle the data average is the usual replica trick:
\begin{talign*}
 {\lim}\frac1n\EE \ln \calZ(\mathcal{D}) &={\lim} \lim_{s\to 0^+}\frac{1}{ns}\ln\EE\mathcal{Z}^s \\
 &= \lim_{s\to 0^+}{\lim}\frac{1}{ns} \ln\EE\mathcal{Z}^s  
\end{talign*}
assuming the limits commute. Consider first $s\in \mathbb{N}^+$. Let $\btheta^a=(\bW^a,\bv^0)$, $(\bx,y)=(\bx_1,y_1)$ and the ``replicas'' of the post-activation, including the teacher's $a=0$:
\begin{talign*}
\big\{\lambda^a(\btheta^a):=\frac1{\sqrt k}{\bv^{0\intercal}}\sigma\big(\frac1{\sqrt d}{\bW^a\bx}\big)\big\}_{a=0}^s. 
\end{talign*}
We directly get
\begin{talign}
\EE\mathcal{Z}^s
&=\EE_{\bv^0}\int \prod_{a}^{0,s}dP_W(\bW^a)\nonumber\\
&\quad \times\big[\EE_{\bx}\int dy \prod_{a}^{0,s} P_{\rm out}(y\mid \lambda^a( \btheta^a ) )\big]^n.
\label{eq:EZ_replicas}
\end{talign}
The key is to identify the law of the replicas $\{\lambda^a\}_{a=0}^s$,
which are dependent random variables due to the common random Gaussian input $\bx$, conditionally on $(\btheta^a)$. As explained and checked numerically in Sec.~\ref{sec:result}, our main \emph{hypothesis} is that $\{\lambda^a\}$ are jointly Gaussian for the $(\btheta^a)$ that dominate the partition function (i.e., posterior samples), an ansatz we cannot prove but that we validate a posteriori thanks to the excellent match between the theory and the empirical generalisation curves, see~\eqref{eq:hypothesis_gaussian_test}, Remark~\ref{rmk:gep} and Sec.~\ref{sec:numerics}.

Given two replica indices $a,b\in\{0,\ldots,s\}$ we define the neuron-neuron overlap matrix 
\begin{talign*}
\Omega^{ab}_{ij}:=\frac{1}{d}{\bW_{i}^{a\intercal}\bW^b_{j}}    , \quad i,j\in[k].
\end{talign*}
Recalling $\sigma$'s Hermite expansion, Mehler's formula (see App.~\ref{app:hermite}) implies that the post-activations covariance is
\begin{talign}
K^{ab}:=\EE[\lambda^a\lambda^b\mid \btheta^a,\btheta^b]=\sum_{\ell\ge 1}^\infty\frac{\mu^2_\ell}{\ell!} R_\ell^{ab}, 
\label{eq:K}
\end{talign}
with the infinitely many overlap OPs
\begin{talign}
R_\ell^{ab}:=\frac{1}k\sum_{i,j\le k} v_i^0v_j^0(\Omega^{ab}_{ij})^\ell, \quad \ell \ge 1.
\label{eq:Rell_def}
\end{talign}
This covariance $\bK$ is complicated but, as we argue hereby, simplifications occur as $d\to\infty$. In particular, the first two overlaps $R_1^{ab},R_2^{ab}$ are special.
We claim that higher order overlaps $(R_\ell^{ab})_{\ell\ge 3}$ can be simplified as functions of simpler OPs.

\vspace{5pt}\vspace{5pt}\paragraph*{Simplifying the order parameters.}
In this section we show how to drastically reduce the number of OPs~\eqref{eq:Rell_def} to track. {To build some intuition, it is convenient to define the symmetric tensors $\bS_\ell^a$ with
\begin{talign}   
S^a_{\ell;\alpha_1\ldots\alpha_\ell} := \frac{1}{\sqrt{k}}\sum_{i\le k} v_i^0 W^a_{i\alpha_1} \cdots W^a_{i\alpha_\ell} .   \label{eq:def_S}
\end{talign}
Indeed, the generic $\ell$-th overlap~\eqref{eq:Rell_def} can be written as $R^{ab}_\ell= (\bS^a_\ell \cdot \bS_\ell^b)/d^\ell$ (where ``$\cdot$'' is the inner product among tensors obtained by contracting all the indices), e.g., $R_2^{ab} =  \Tr \,\bS_{2}^{a}\bS_{2}^{b}/d^2$. The following assumptions amount in considering how these tensors behave for $\ell=1, 2$ and $\ell\ge 3$. Let us start from the latter case.

First, we assume that for Hadamard powers $\ell \ge 3$, the off-diagonal of the overlap $(\bOmega^{ab})^{\circ \ell}$, obtained from i.i.d. weight matrices sampled from the posterior, is small enough to be discarded:
\begin{talign}\label{eq:Omega_ansatz_i}    
(\Omega_{ij}^{ab})^\ell \approx\delta_{ij}(\Omega_{ii}^{ab})^\ell \quad \text{if}  \quad  \ell\geq 3 .
\end{talign}
Approximate equality is up to a matrix with $o_d(1)$ operator norm. In other words, the vector $\bW^a_i$ of a student (replica) is assumed to possibly align, for each $i$, only with a single $\bW^b_j$ of the teacher (or, by Bayes-optimality, of another replica) indexed by $j=\pi_i$, with $\pi$ a permutation. The model is symmetric under permutations of hidden neurons with the same readout value, we thus take $\pi$ to be the identity without loss of generality. The same ``concentration on the diagonal'' happens, e.g., for a standard Wishart matrix, which is the extreme case for $\bOmega^{ab}$ if $\bW^a = \bW^b$ and $P_W=\calN(0,1)$: its eigenvectors and those of its Hadamard square are delocalised, while higher Hadamard powers $\ell\ge 3$ have strongly localised eigenvectors \cite{potters2020first} (consequently, $R_2^{ab}$ will require a separate treatment). 

Moreover, assume that the readout prior has discrete support ${\rm Supp}(P_v)=:\mathsf{V}=\{ \mathsf{v} \}$; this can be relaxed by binning to a continuous support, as mentioned in Sec.~\ref{sec:result}. By exchangeability among neurons with the same readout value, we further assume that all diagonal elements $\{\Omega_{ii}^{ab}\mid i \in \mathcal{I}_\mathsf{v}\}$ concentrate onto the constant $\mathcal{Q}^{ab}(\mathsf{v})$, where $\mathcal{I}_\mathsf{v}:=\{i\le k\mid v_i^0=\mathsf{v}\}$: 
\begin{talign}\label{eq:Omega_ansatz_v}    
(\Omega_{ij}^{ab})^\ell \approx\delta_{ij}\mathcal{Q}^{ab}(\mathsf{v})^\ell \ \ \text{if}  \  \ell\geq 3 \  \text{and}  \ i \  \text{or} \  j\in \mathcal{I}_\mathsf{v}.
\end{talign}
Equivalently, under the neuron exchangeability assumption, by summing over the indices $i \in \mathcal{I}_\mathsf{v}$ and dividing by their number the constant $\mathcal{Q}^{ab}(\mathsf{v})$ can be written as
\begin{talign*}
    \mathcal{Q}^{ab}(\mathsf{v}) := \frac{1}{|\mathcal{I}_\mathsf{v}| d} \sum_{i \in \mathcal{I}_\mathsf{v}} (\bW^{a} \bW^{b\intercal})_{ii}.
\end{talign*}
This definition is directly related to the way we measure overlaps in numerical experiments, as empirical averages are less affected by finite size effects than specific choices of $i \in \mathcal{I}_\mathsf{v}$; thus, we adopt this definition also in our theoretical analysis. The advantage in switching from~\eqref{eq:Omega_ansatz_i} to~\eqref{eq:Omega_ansatz_v}, i.e. in labelling the neurons with their readout value, is an expression suitable for the asymptotic regime we are considering, where the neurons are infinitely many. Indeed, with these simplifications we can write 
\begin{talign}
\label{eq:Qell_NSB}
   R_\ell^{ab}=\EE_{v\sim P_v}v^2 \calQ^{ab}(v)^\ell +o_d(1) \ \ \text{for}\ \  \ell\geq 3.
\end{talign}
This assumption has also a natural interpretation in terms of the tensors $\bS_\ell^a$: in absence of specialisation, $\mathcal{Q}^{ab}(\mathsf{v})=0$ for all $\mathsf{v}$, so $R_\ell^{ab} = (\bS^a_\ell \cdot \bS^b_\ell)/d^\ell = 0$ according to~\eqref{eq:Qell_NSB}. Indeed, a non-specialised model with $kd=\Theta(d^2)$ parameters and $n=\Theta(d^2)$ data cannot learn these tensors, as this would require the knowledge of $\Theta(d^\ell/\ell!)$ entries and a comparable amount of tunable parameters; on the contrary, once specialisation occurs the model is able to factorise them using the r.h.s. of \eqref{eq:def_S}.
}

\begin{figure}[t!!]
    \centering
    \includegraphics[width=0.48\linewidth]{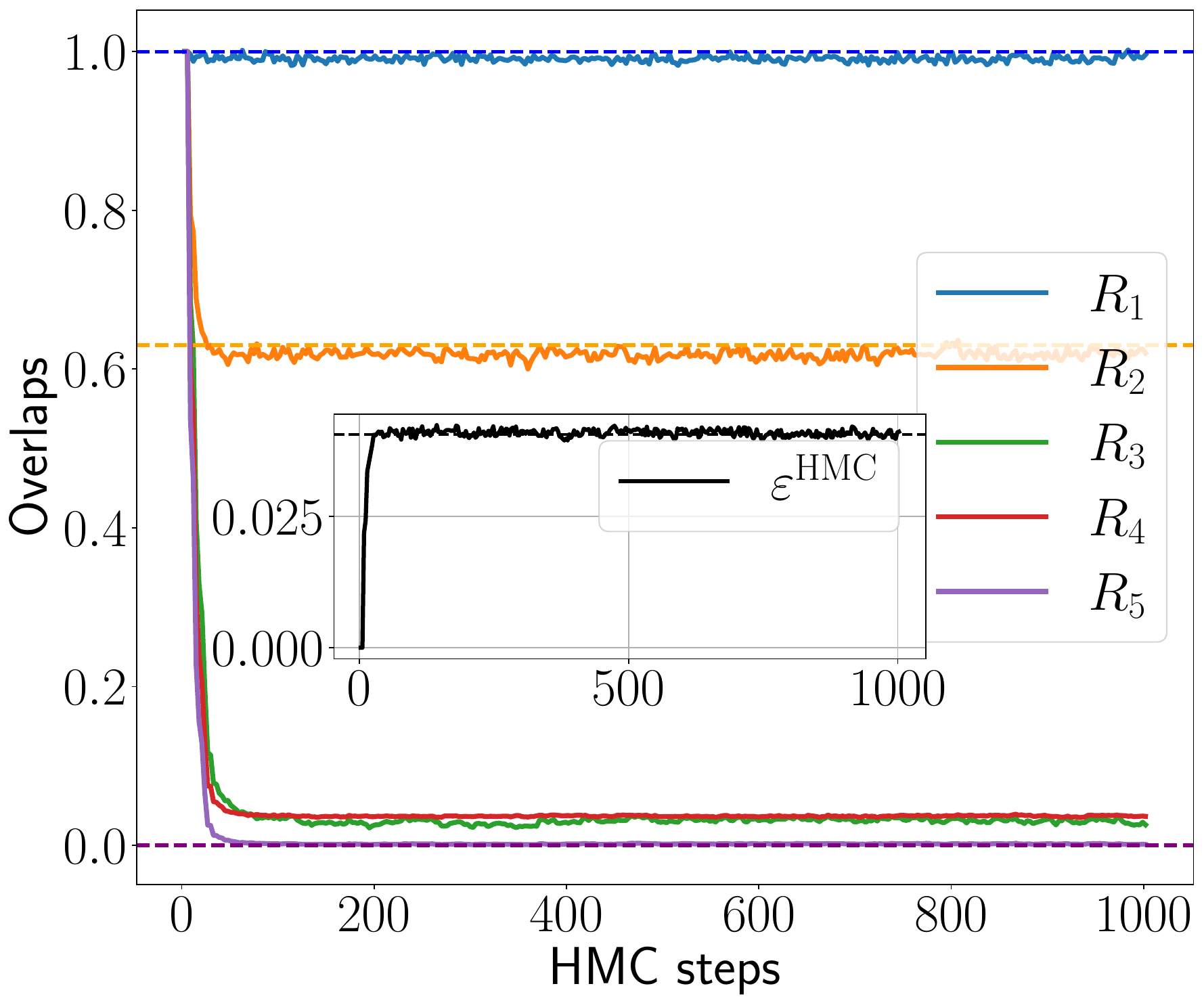}
    \hfill
    \includegraphics[width=0.48\linewidth]{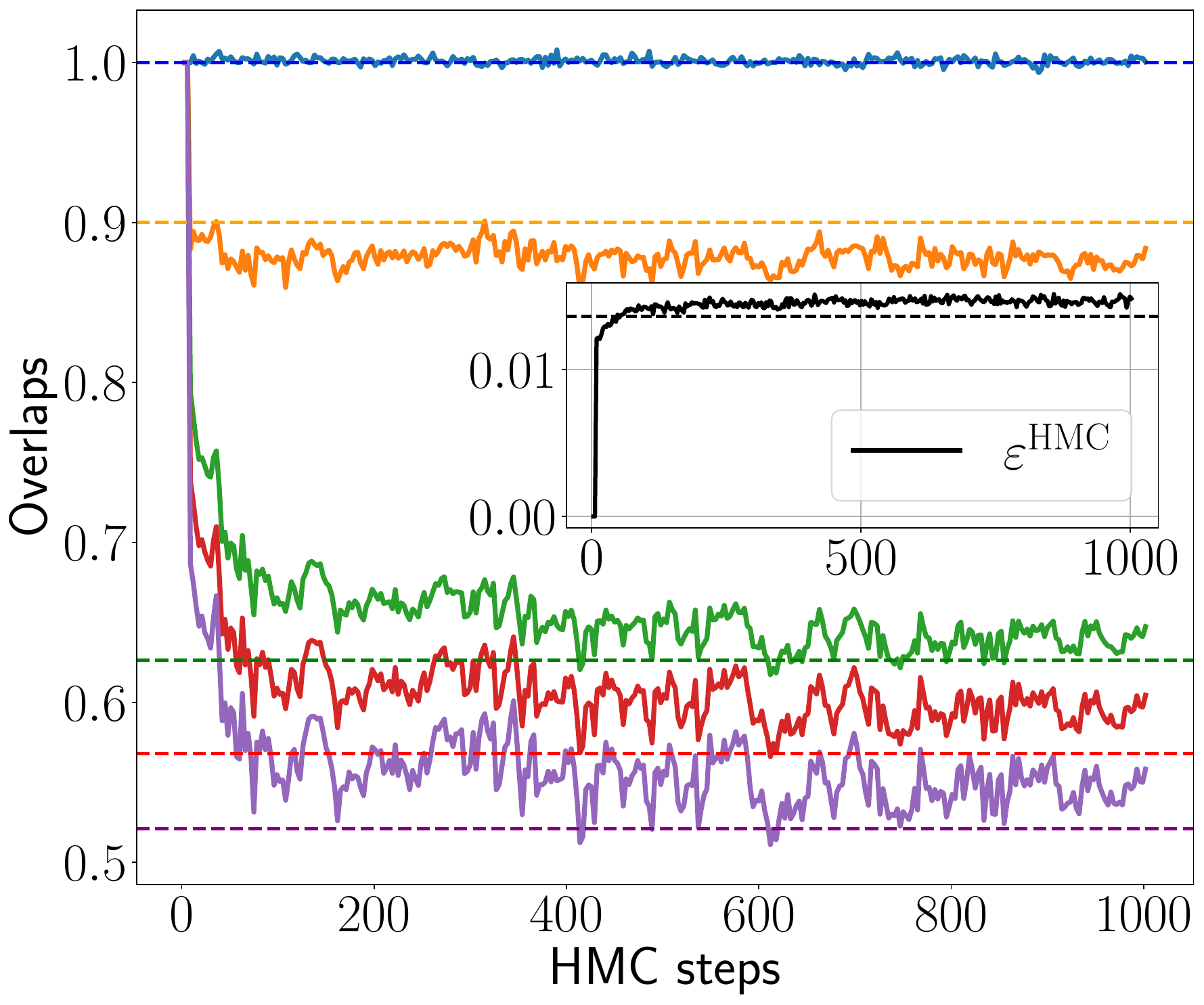}
    \vspace{-5pt}
    \caption{Hamiltonian Monte Carlo dynamics of the overlaps $R_\ell=R_\ell^{01}$ between student and teacher weights for $\ell \in [5]$, with activation function for $L=1$ with ${\rm ReLU}(x)$ activation, $d=200, \gamma=0.5$, linear readout with $\Delta=0.1$ and two choices of sample rates and readouts: $\alpha=1.0$ with $P_v = \delta_{1}$ (\textbf{Left}) and $\alpha=3.0$ with $P_v = \mathcal{N}(0,1)$ (\textbf{Right}). The teacher weights $\bW^0$ are Gaussian and the readouts are fixed during sampling to the teacher ones. The dynamics is initialised informatively, i.e., on $\bW^0$. The overlap $R_1$ always fluctuates around 1. \textbf{Left}: The overlaps $R_\ell$ for $\ell  \ge 3$ at equilibrium converge to 0, while $R_2$ is well estimated by the theory (orange dashed line). \textbf{Right}: At higher sample rate $\alpha$, also the $R_\ell$ for $\ell \ge 3$ are non zero and agree with their theoretical prediction (dashed lines). Insets show the mean-square generalisation error attained by HMC (solid) and the theoretical prediction $\varepsilon^{\rm opt}$ (dashed).}
    \label{fig:overlaps}
    \vskip -0.1in
\end{figure}

Our assumption is verified numerically a posteriori as follows. Identity \eqref{eq:Qell_NSB} is true (without $o_d(1)$) for the predicted theoretical values of the OPs by construction of our theory. \figurename~\ref{fig:v_non-constant} verified the good agreement between  theoretical and experimental overlap profiles $\mathcal{Q}^{01}(\mathsf{v})$ for all $\mathsf{v}\in\mathsf{V}$ (which is statistically the same as  $\smash{\mathcal{Q}^{ab}(\mathsf{v})}$ for any $a\neq b$ by the so-called Nishimori identity following from Bayes-optimality, see App.~\ref{app:nishiID}), while \figurename~\ref{fig:overlaps} checks
the agreement at the level of $(R_\ell^{ab})$. Consequently, \eqref{eq:Qell_NSB} also holds for the experimental overlaps.

Having simplified the $\ell\ge 3 $ terms in the series~\eqref{eq:K}, let us pass to the $\ell=1$ case. Given that the number of data $n=\Theta(d^2)$ and that the corresponding $(\bS_1^a)$ are only $d$-dimensional, they are reconstructed perfectly (the same argument was used in Remark~\ref{rmk:readouts} to argue that the readouts $\bv$ can be quenched). We thus assume right away that at equilibrium the overlaps $R_1^{ab}=1$ (or saturate to their maximum value; if tracked, the corresponding saddle point equations end up being trivial and do fix this). In other words, in the quadratic data regime, the $\mu_1$ contribution in the Hermite decomposition of $\sigma$ for the target is perfectly learnable, while higher order ones play a non-trivial role. In contrast, \cite{cui2023bayes} studies the regime $n=\Theta(d)$ where $\mu_1$ is the only learnable term.

Then, the average replicated partition function reads 
\begin{talign*}
  \EE\mathcal{Z}^s = \int d\bR_2 d\bm{\mathcal{Q}} \exp(F_S\!+\! nF_E)  
\end{talign*}
where $F_E, F_S$ depend on $\bR_2=(R_2^{ab})$ and $\bm{\mathcal{Q}}:=\{\mathcal{Q}^{ab} \mid a\le b\}$, where $\mathcal{Q}^{ab}:=\{\mathcal{Q}^{ab}(\mathsf{v})\mid \mathsf{v}\in\mathsf{V}\}$. 

The ``energetic potential'' is defined as
\begin{talign*}
    &e^{nF_E}
    :=\big[\int dyd\blambda\frac{\exp(-\frac{1}{2}\blambda^\intercal\bK^{-1}\blambda)}{((2\pi)^{s+1}\det \bK)^{1/2}}\prod_{a}^{0,s} P_{\rm out}(y\mid \lambda^a)\big]^n.
    \end{talign*}
   It takes this form due to our Gaussian assumption on the replicated post-activations and is thus easily computed, see App.~\ref{app:energetic_potential}. 
   
   The ``entropic potential'' $F_S$ taking into account the degeneracy of the OPs entering the covariance of the post-activations is obtained by averaging delta functions fixing their definitions w.r.t. the ``microscopic degrees of freedom'' $(\bW^a)$. It can be written compactly using the following conditional law over the tensors $(\bS_2^a)_{a=0}^s$:
\begin{talign}    &P((\bS_2^a)\mid \bm{\mathcal{Q}}):=
    V_W^{kd}(\bm{\mathcal{Q}})^{-1}\int \prod_a^{0,s}dP_W(\bW^a)\nonumber\\
    &\qquad\times\prod_{a\leq b}^{0,s}\prod_{\mathsf{v}\in \mathsf{V}}\delta(|\mathcal{I}_\mathsf{v}|d\,\mathcal{Q}^{ab}(\mathsf{v})-\sum_{i\in \mathcal{I}_\mathsf{v}}{\bW^{a\intercal}_i\bW_i^{b}})\nonumber\\
    &\qquad \times\prod_{a}^{0,s}\delta(\bS^a_2-
    \bW^{a\intercal}{\rm diag}(\bv^0)\bW^a/\sqrt{k}),\label{eq:truePS2}
\end{talign}
with normalisation $V_W^{kd}=V_W^{kd}(\bm{\mathcal{Q}})$ given by
\begin{talign*}
   V_W^{kd} = &\int \prod_{a}^{0,s}dP_W(\bW^a)\\
   &\times\prod_{a\le b}^{0,s}\prod_{\mathsf{v}\in\mathsf{V}}\!\delta(|\mathcal{I}_\mathsf{v}|d\,\mathcal{Q}^{ab}(\mathsf{v})-\sum_{i\in \mathcal{I}_\mathsf{v}}{\bW^{a\intercal}_i\bW_i^{b}}).
\end{talign*}
The entropy of $(\bR_2,\bm{\mathcal{Q}})$, which is the challenging term to compute, then reads   
\begin{talign*}
    e^{F_S}:= 
    V_W^{kd} \int dP((\bS_2^a)\mid\bm{\mathcal{Q}}) \prod_{a\leq b}^{0,s}\delta(d^2R_2^{ab}-{\Tr \,\bS_{2}^{a} \bS_{2}^b}).
\end{talign*}

\vspace{5pt}\paragraph*{Tackling the entropy: measure simplification by moment matching.}

The delta functions above fixing $R_2^{ab}$ in the entropy of $\bR_2$ conditional on $\bm{\mathcal{Q}}$ induce \emph{quartic} constraints between the degrees of freedom $(W_{i\alpha}^a)$ instead of quadratic as usual. A direct computation thus seems out of reach. However, we will exploit the fact that the constraints are quadratic in the matrices $(\bS_2^a)$. Consequently, shifting our focus towards $(\bS_2^a)$ as the basic degrees of freedom to integrate rather than $(W_{i\alpha}^a)$ will allow us to move forward by simplifying their measure \eqref{eq:truePS2}. Note that while  $(W_{i\alpha}^a)_{i,\alpha}$ are i.i.d. under the prior $P_W$, $\bS_2^a$ has dependent entries under its corresponding prior. This important fact is taken into account as follows.

Define $P_{S}$ as the probability density of a generalised Wishart random matrix $\tilde \bW^{\intercal}{\rm diag}(\bv^0)\tilde\bW/\sqrt{k}$, where $\tilde \bW \in \mathbb{R}^{k\times d}$ is made of i.i.d. standard \emph{Gaussian} entries. The simplification we consider consists in replacing \eqref{eq:truePS2} by the effective measure
\begin{talign}
\label{eq:effectivePS2}
 &\tilde P((\bS_2^a)\mid \bm{\mathcal{Q}}) :=\nonumber\\ &\quad\tilde V_W^{kd}(\bm{\mathcal{Q}})^{-1} \prod_a^{0,s} P_{S}(\bS_2^a) \prod_{a<b}^{0,s} e^{\frac 12 \tau(\mathcal{Q}^{ab}) \Tr\,\bS^a_2\bS^b_2}
\end{talign}
where $\tilde V_W^{kd}=\tilde V_W^{kd}(\bm{\mathcal{Q}})$ is a normalisation constant, and 
\begin{talign}
\tau(\mathcal{Q}^{ab}):= \text{mmse}_S^{-1}(1 - \EE_{v\sim P_v}[v^2 \mathcal{Q}^{ab}(v)^2]) . \label{eq:tauDef}  
\end{talign}

The rationale behind this choice goes as follows. The matrices $(\bS_2^a)$ are, under the measure \eqref{eq:truePS2}, $(i)$ similar to generalised Wishart matrices, but instead constructed from $(ii)$ non-Gaussian factors $(\bW^a)$, which $(iii)$ are coupled between different replicas, thus inducing a coupling among replicas $(\bS^a)$. The proposed simplified measure captures all three aspects while remaining tractable, as we explain now.

The first assumption is that in the measure \eqref{eq:truePS2} the details of the (centred, unit variance) prior $P_W$ enter only through $\bm{\mathcal{Q}}$ at leading order. Due to the conditioning, we can thus relax it to Gaussian (with the same two first moments) by universality, as is often the case in random matrix theory. $P_W$ will instead explicitly enter in the entropy of $\bm{\mathcal{Q}}$ related to $V_W^{kd}$. Point $(ii)$ is thus taken care of by the conditioning. Then, the generalised Wishart prior $P_{S}$ encodes $(i)$ and, finally, the exponential tilt in $\tilde P$ induces the replica couplings of point $(iii)$. 

It now remains to capture the correct dependence of measure \eqref{eq:truePS2} on $\bm{\mathcal{Q}}$. This is done by realising that 
\begin{talign*}
\int dP((\bS_2^a)\mid \bm{\mathcal{Q}}) \frac 1{d^2}\Tr\bS_2^a \bS_2^b=    \EE_{v\sim P_v}[v^2 \mathcal{Q}^{ab}(v)^2]+\gamma \bar{v}^2.
\end{talign*}
This is shown in App.~\ref{app:second_moment_P(S2|QW)}. The Lagrange multiplier $\tau(\mathcal{Q}^{ab})$ to plug into $\tilde P$ enforcing this moment matching condition between true and simplified measures as $s\to0^+$ is \eqref{eq:tauDef}, see App.~\ref{app:entropic_contribution}. For completeness, we provide in App.~\ref{app:alternatives} alternatives to the simplification \eqref{eq:effectivePS2}, whose analysis is left for future~work.

\vspace{5pt}\vspace{5pt}\paragraph*{Final steps and spherical integration.}
Combining all our findings, the average replicated partition function is simplified as
\begin{talign*}
&\EE\mathcal{Z}^s =  \int d\bR_2 d\bm{\mathcal{Q}} e^{nF_E+kd \ln V_W(\bm{\mathcal{Q}})-kd\ln \tilde V_W(\bm{\mathcal{Q}})}\\
&\qquad\times \prod_a^{0,s}\! P_{S}(\bS_2^a) \prod_{a<b}^{0,s} e^{\frac 12 \tau(\mathcal{Q}^{ab}) \Tr\,\bS^a_2\bS^b_2}\\
&\qquad\times\prod_{a\le b}^{0,s}\delta(d^2R_2^{ab}\!-\!{\Tr \,\bS_{2}^{a} \bS_{2}^b}).
\end{talign*}
The equality should be interpreted as holding at leading exponential order $\exp (\Theta(n))$, assuming the validity of our previous measure simplification. All remaining steps but the last are standard:

$(i)$ Express the delta functions fixing $\bm{\mathcal{Q}},\bR_2$ in exponential form using their Fourier representation; this introduces additional Fourier conjugate OPs $\hat{\bm{\mathcal{Q}}},\hat \bR_2$ of same respective dimensions. 

$(ii)$ Once this is done, the terms coupling different replicas of $(\bW^a)$ or of $(\bS^a)$ are all quadratic. Using the Hubbard-Stratonovich transformation (i.e.,  $\E_{\bZ} \exp(\frac d 2 \Tr\,\bM \bZ) = \exp(\frac d 4 \Tr\,\bM^2)$
for a $d\times d$ symmetric matrix $\bM$ with $\bZ$ a standard GOE matrix) therefore allows us to linearise all replica-replica coupling terms, at the price of introducing new Gaussian fields interacting with all replicas. 

$(iii)$ After these manipulations, we identify at leading exponential order an effective action $\mathcal{S}$ depending on the OPs only, which allows a saddle point integration w.r.t. them as $n\to \infty$: 
\begin{talign*}
    &\lim \frac{1}{ns}\ln\E \mathcal{Z}^s  \! =\! \lim  \frac{1}{ns}\ln \int   d\bR_2 d\hat\bR_2 d\bm{\mathcal{Q}} d\hat{\bm{\mathcal{Q}}}  e^{n \mathcal{S}} \!=\! \frac 1s {\rm extr} \,\mathcal{S}.
\end{talign*}

$(iv)$ Next, the replica limit $s\to 0^+$ of the previously obtained expression has to be considered. To do so, we make a replica symmetric assumption, i.e., we consider that at the saddle point, all OPs entering the action $\mathcal{S}$, and thus $K^{ab}$ too, take a simple form of the type $R^{ab} = R_{\rm d} \delta_{ab} + R (1-\delta_{ab})$. Replica symmetry is rigorously known to be correct in Bayes-optimal learning and is thus justified here, see \cite{barbier2022strong,barbier2019adaptive}. 

$(v)$ The resulting expression still includes two high-dimensional integrals related to the $\bS_2$ matrices. They correspond to the free entropies associated with the Bayes-optimal denoising of a generalised Wishart matrix, described above Result~\ref{res:free_entropy_shallow}, for two signal-to-noise ratios. The last step deals with these matrix integrals over rotationally invariant matrices using the HCIZ integral, whose form is tractable in this case \cite{maillard2022perturbative,matrix_inference_Barbier}. These free entropies yield the two last terms $\iota(\,\cdot\,)$ in $f_{\rm RS}^{(1)}$, \eqref{eq:fRS}.

The complete derivation in App.~\ref{app:replicas} gives Result~\ref{res:free_entropy_shallow}. From the meaning of the OPs, this analysis also yields the post-activations covariance $\bK$ and thus Result~\ref{res:gen_error}.

As a final remark, we emphasise again a key difference between our approach and earlier works on extensive-rank systems. If, instead of taking the generalised Wishart prior $P_S$ as the base measure over the matrices $(\bS_2^a)$ in the simplified $\tilde P$ with moment matching, one takes a factorised Gaussian measure, thus entirely forgetting the dependences among $\bS_2^a$ entries, this mimics the Sakata-Kabashima replica method \cite{sakata2013}. Our ansatz thus captures important correlations neglected in \cite{sakata2013,krzakala2013phase,kabashima2016phase,barbier2024phase} in the context of linear-rank matrix inference. For completeness, we show in App.~\ref{app:alternatives} that our ansatz indeed improves the prediction compared to these earlier approaches.

\subsection{Two hidden layers MLP}\label{sec:replicasL=2}

We now move to the deep MLP, by first considering the $L=2$ case. We highlight here only the crucial steps that make the derivation different with respect to the one sketched above, referring the reader to App.~\ref{app:replicas_deep} for more details. We assume this time that $\mu_2=0$ (in addition to $\mu_0=0$): in this way, our approach simplifies considerably, as the matrix degrees of freedom involved in the 2nd Hermite components of the activation functions do not appear in the theory. We will see however that, due to the deep structure of the network, matrix degrees of freedom coming from combinations of the model's hidden weights are still entering the theory, and will require the use of a rectangular spherical integral (see App.~\ref{app:replicas_deep}). Popular activation functions (e.g., all the odd ones) comply with the requirement $\mu_2=0$.

\vspace{5pt}\vspace{5pt}\paragraph*{Replicated system, order parameters and their simplification.} Replicas of the post-activations can now be written recursively as
\begin{talign*}
    &\big\{\lambda^a(\btheta^a):=\frac1{\sqrt{k_2}}{\bv^{0\intercal}}\sigma^{(2)}( \bh^{(2)a} )\big\}_{a=0}^s,\\
    &\big\{\bh^{(l)a} := \frac{1}{\sqrt{k_{l-1}}}\bW^{(l)a}\sigma^{(l-1)}(\bh^{(l-1)a} ) \big\}_{a=0,\ldots,s;\,l=1,2},
\end{talign*} 
where we allowed different activations at each layer, we used the notation $\sigma^{(0)}(x) := x$, $k_0 = d$ and $\bh^{(0)a} = \bx$ for all $a$. For the sake of presentation, we further require the normalisation $\EE_{z\sim\calN(0,1)}\sigma^{(l)}(z)^2=1$ for all $l$, to avoid tracking this variance in the following (the case of generic variance can be derived from App.~\ref{app:hermite}). The expectation over the input $\bx$ at given weights can be done as in \eqref{eq:EZ_replicas}, by assuming the same joint-Gaussianity of the post-activations $(\lambda^a)$ as in the shallow case. Moreover, to use recursively Mehler's formula we also assume that the pair of pre-activations $(h_i^{(2)a},h_j^{(2)b})$ is jointly Gaussian for any choice of $a,b=0,\ldots,s$ and $i,j\le k_2$. With these assumptions the covariances $K^{ab} := \EE_{\bx} \lambda^a \lambda^b$ and $\bOmega^{(l)ab} := \EE_{\bx} \bh^{(l)a} \bh^{(l)b\intercal} $ can be written as
\begin{talign*}
    &K^{ab} = \frac{(\mu_1^{(2)})^2}{k_{2}} \bv^{0\intercal} \bOmega^{(2)ab} \bv^0 + \frac{1}{k_{2}} \bv^{0 \intercal} g^{(2)} (\bOmega^{(2)ab})\bv^{0} \, , \nonumber\\
    & \bOmega^{(2)ab} = \frac{(\mu_1^{(1)})^2}{k_{1}} \bW^{(2)a} \bOmega^{(1)ab} \bW^{(2)b \intercal} \nonumber
    \\
    &\hspace{8em}+ \frac{1}{k_{1}} \bW^{(2)a} g^{(1)} (\bOmega^{(1)ab} )\bW^{(2)b \intercal}, \nonumber\\
    &\bOmega^{(1)ab} = \frac{1}{k_{0}} \bW^{(1)a} \bW^{(1)b \intercal},
\end{talign*}
where the functions 
\begin{talign*}
g^{(l)}(x) := \sum_{\ell=3}^{\infty} \frac{(\mu_\ell^{(l)})^2}{\ell!} \,x^{\ell}  
\end{talign*}
are applied entry-wise to matrices and $\mu_\ell^{(l)}$ is the $\ell$-th Hermite coefficient of $\sigma^{(l)}$.

Unfolding the above recursion, the covariance $K^{ab}$ can be written in terms of overlaps of ``effective'' hidden weights and readout vectors
\begin{talign*}
    &\bW^{(2:1)a} := \frac{\bW^{(2)a} \bW^{(1)a}}{\sqrt{k_{1}}},\\
    &\bv^{(1)a} := \frac{\bW^{(1) a \intercal}\bW^{(2) a \intercal}\bv^0}{\sqrt{k_2 k_1 }} , \quad \bv^{(2)a} := \frac{\bW^{(2) a \intercal}\bv^0}{\sqrt{k_2}},
\end{talign*}
each of them arising from combinations of the activation's linear components. We also set $\bv^{(3)a} :=\bv^{0}$. Moreover, simplifications can be taken along the three following lines: 

$(i)$ $\bv^{(1)a  \intercal } \bv^{(1)b }/k_0$ is an overlap of $d$-dimensional vectors: as explained above (see the discussion on $R_1^{ab}$ in the shallow case), it can be directly taken to be 1 in the quadratic data regime.

$(ii)$ Wherever a function $g^{(l)}$, involving only Hadamard powers greater than 2, is applied to a matrix overlap, we assume the resulting matrix to be diagonal in the limit, $g^{(l)}(\bOmega^{(l)ab})_{ij} \approx \delta_{ij} g^{(l)}(\bOmega^{(l)ab})_{ii}$, as we did in~\eqref{eq:Omega_ansatz_i}. 

$(iii)$ The components $v^{(2)a}_i$, $v^{(2)b}_i$ of the effective readouts enter the above expressions only if $\Omega^{(1)ab}_{ii}\neq 0$, that is if some specialisation has occurred in the previous layer. As these components are $\Theta(k_1)$ in number, they can be reconstructed exactly with $\Theta(d^2)$ data. We can thus take these vectors as given, $\bv^{(2)a  \intercal } = \bv^{(2)0  \intercal }$. From central limit theorem, the components of $\bv^{(2)0  \intercal }$ are standard Gaussian variables, $v^{(2)0}_i\sim \calN(0,1)$.

From point $(ii)$, and extending the approach we followed for the shallow case, we are naturally led to consider as OPs the diagonal profiles of the overlap matrices, $(\Omega^{(l)ab}_{ii})_i$. Moreover, from point $(iii)$ we can label internal neurons (say, the ones in layer $l$) with the value of the effective readout to which they are connected ($v^{(l+1)0}_i$) rather than with their index ($i\le k_l$). By binning the distribution of the elements of $\bv^{(l)0}$, we define the sets of indices $\calI_{\mathsf{v}^{(l)}}:= \{i \le k_{l-1} \mid v^{(l)0}_i = \mathsf{v}^{(l)}\}$, while keeping $\calI_{\mathsf{v}}$ (with no layer label) as in~\eqref{eq:Omega_ansatz_v}. As before, in order to define the OPs we further assume exchangeability among neurons with the same effective readout value (e.g., $(\bW^{(1)a}\bW^{(1)b\intercal})_{ii}/d =: {\mathcal{Q}}_{1}^{ab}(\mathsf{v^{(2)}})$ for all $i \in \calI_{\mathsf{v}^{(2)}}$). Equivalently, by summing over these indices and normalising by their number, we obtain:
\begin{talign*}
    &{\mathcal{Q}}_{1}^{ab}(\mathsf{v^{(2)}})
:= \frac{1}{|\calI_{\mathsf{v}^{(2)}}| d} \sum_{i \in \calI_{\mathsf{v}^{(2)}} } (\bW^{(1)a}\bW^{(1)b\intercal})_{ii} ,\\
&\mathcal{Q}_{2}^{ab}(\mathsf{v},\mathsf{v^{(2)}})
:= \frac{1}{|\calI_{\mathsf{v}^{(2)}}| |\calI_{\mathsf{v}}|} \sum_{i \in \calI_{\mathsf{v}^{(2)}}, j \in \calI_{\mathsf{v}}} W^{(2)a}_{ji} W^{(2)b}_{ji},\\
&{\mathcal{Q}}_{2:1}^{ab}(\mathsf{v}) := \frac{1}{|\calI_{\mathsf{v}} | d}\sum_{i \in \calI_{\mathsf{v}} } (\bW^{(2:1)a}\bW^{(2:1)b\intercal})_{ii}.
\end{talign*}
The bold notations $\bm{\mathcal{Q}}_{1}$, $\bm{\mathcal{Q}}_{2}$ and $\bm{\mathcal{Q}}_{2:1} $ are defined analogously to the shallow case. In terms of these, the covariance of the post-activations reads
\begin{talign*}
    K^{ab} &= (\mu_1^{(2)} \mu_1^{(1)})^2 \\
    &+ ( \mu_1^{(2)})^2 \EE_{v^{(2)} \sim \calN(0,1)} (v^{(2)})^2 g^{(1)} \big(\mathcal{Q}_{1}^{ab}(v^{(2)}) \big)  \\
    & + \EE_{v \sim P_{v}} v^2 g^{(2)} \big[(\mu_1^{(1)})^2 \mathcal{Q}^{ab}_{2:1}(v) \\
    &\hspace{3em}+ \EE_{v^{(2)} \sim \calN(0,1)}\mathcal{Q}^{ab}_{2}(v,v^{(2)}) g^{(1)}\big(\mathcal{Q}^{ab}_{1}(v^{(2)})\big)\big] \, .
\end{talign*}
The structure of this covariance informs us on how the learning can or cannot take place. E.g., if the first layer does not specialise, $\mathcal{Q}^{ab}_{1}(v^{(2)})=0$, then the second layer cannot either because its associated overlap disappears.

From this point on, the energetic potential $F_E$ follows: it is the very same as for $L=1$ but using the above covariance.

\vspace{5pt}\vspace{5pt}\paragraph*{Tackling the entropy using the rectangular spherical integral.}
The entropic potential, accounting for the degeneracy of the OPs, requires some care. We first define the conditional law over the matrices $(\bW^{(2:1)a})$:
\begin{talign*}
    &P((\bW^{(2:1)a})\mid \bm{\calQ}_{1},\bm{\calQ}_2) \propto \int\prod_{a=0}^s\prod_{l=1}^2 dP_{W_l}(\bW^{(l)a})\\
    &\times \delta(\bW^{ (2:1)a}-{
    \bW^{ (2)a}\bW^{(1)a}}/\sqrt{k_1}) \nonumber\\
    &\times\prod_{a\leq b}^{0,s} \prod_{\mathsf{v^{(2)}}\in \mathsf{V^{(2)}}} \delta(|\mathcal{I}_\mathsf{v^{(2)}}| d\,\mathcal{Q}_{1}^{ab}(\mathsf{v^{(2)}})\nonumber\\
    &\qquad\qquad\qquad\qquad-\sum_{i\in\mathcal{I}_\mathsf{v^{(2)}}}  ({\bW^{(1)a}\bW^{(1)b\intercal }})_{ii}) \nonumber\\
    &\times \prod_{\mathsf{v}\in \mathsf{V}}\delta(|\mathcal{I}_\mathsf{v^{(2)}}||\mathcal{I}_\mathsf{v}| \mathcal{Q}_{2}^{ab}(\mathsf{v}, \mathsf{v^{(2)}})\nonumber\\
    &\qquad\qquad\qquad\qquad-\sum_{i\in\mathcal{I}_\mathsf{v^{(2)}},j\in\mathcal{I}_\mathsf{v}}{W^{(2)a}_{j i}W_{j i}^{(2)b}}),
\end{talign*}
where $\mathsf{V^{(2)}}$ is the binned support of $\calN(0,1)$. In this way, we can write the entropic contribution as
\begin{talign*}
    &e^{F_s} = V_{2:1}^{k_2d} \int dP((\bW^{(2:1)a})\mid \bm{\calQ}_{1},\bm{\calQ}_2) \prod_{a\le b}^{0,s} \prod_{\mathsf{v} \in \mathsf{V}} \\
    &\hspace{3em}\times \delta(|\calI_{\mathsf{v}} |d {\mathcal{Q}}_{2:1}^{ab}(\mathsf{v}) - \sum_{i \in \calI_{\mathsf{v}} } (\bW^{(2:1)a}\bW^{(2:1)b\intercal})_{ii}),
\end{talign*}
where $V_{2:1}^{k_2d}$, depending implicitly on $ \bm{\calQ}_{1}$, $\bm{\calQ}_2$, is the normalisation factor of $P((\bW^{(2:1)a})\mid \bm{\calQ}_{1},\bm{\calQ}_2)$.

The evaluation of the last integral involves coupled replicas of matrices with correlated entries. We deal with it by relaxing the measure $P((\bW^{(2:1)a})\mid \bm{\calQ}_{1},\bm{\calQ}_2)$ to a tractable one, still able to capture the correlations between these degrees of freedom. To this aim, we first observe that, asymptotically, under the true conditional measure
\begin{talign*}
    &\frac{1}{|\mathcal{I}_{\mathsf{v}}|d} \sum_{i\in\mathcal{I}_{\mathsf{v}}} \EE[  (\bW^{(2:1)a} \bW^{(2:1)b\intercal})_{ii} \mid \bm{\calQ}_{1},\bm{\calQ}_2]\\
    &\hspace{8em}
    \approx 
    \EE_{v^{(2)}\sim \calN(0,1)}\calQ^{ab}_2(\mathsf{v},v^{(2)})\calQ^{ab}_{1}(v^{(2)}).
\end{talign*}
In order to match this moment in our relaxation, we take a tractable base measure with exponential tilts for each value $\mathsf{v}\in\mathsf{V}$:
\begin{talign*}
    &d\bar{P}((\bW^{(2:1)a})\mid\bm{\calQ}_{1},\bm{\calQ}_2)= 
    \prod_{\mathsf{v}\in\mathsf{V}} V(\bm{\tau}_\mathsf{v})^{-1}\\
    &\qquad\times\prod_{a=0}^s dP_{UV}(\bW^{(2:1)a}_\mathsf{v})
    e^{
    \sum_{a<b,0}^s \tau^{ab}_\mathsf{v}\Tr\bW^{(2:1)a}_{\mathsf{v}} \bW^{(2:1)b\intercal}_{\mathsf{v}}
    }
\end{talign*}
where $\bW^{(2:1)a}_\mathsf{v}=(\bW^{(2:1)a}_{i})_{i\in\mathcal{I}_\mathsf{v}}$
and $dP_{UV}$ is the law of the product of two matrices with i.i.d. Gaussian entries ($\bU\in\mathbb{R}^{|\mathcal{I}_\mathsf{v}|\times k_1}$, $\bV\in\R^{k_1\times d}$), $\bm{\tau}_\mathsf{v}=(\tau_\mathsf{v}^{ab})_{a,b}$ is a function of $ \bm{\calQ}_{1}$, $\bm{\calQ}_2$ fixed by the previous moment matching, and $V(\bm{\tau}_\mathsf{v})$ is a normalisation factor. With this relaxation, the entropic contribution can be evaluated explicitly using the rectangular spherical integral, leading eventually to Result~\ref{res:free_entropy_L=2} (see App.~\ref{app:replicas_deep} for more details).

\subsection{Three or more hidden layers MLP}
To tackle the $L\ge 3$ case, one could push forward the approach we presented in the previous section. However, even with the simplification of considering only activations with no second Hermite component, the analysis remains very challenging, as the post-activation covariance involves the ``effective weights''
\begin{talign*}
    \bW^{(l':l)a} := \frac{1}{\sqrt{k_{l'-1}k_{l'-2}\dots k_l}}\bW^{(l')a} \bW^{(l'-1)a}\dots \bW^{(l)a},
\end{talign*}
as shown in App.~\ref{app:replicas_deep}. These terms appear through combinations of the linear components of the activation functions $\sigma^{(l'-1)}, \ldots, \sigma^{(l)}$. To evaluate the entropic contributions of the OPs $\bm{\calQ}_{l':l}$, defined as overlaps between replicas of the above effective weights, one has to consider all the possible learning mechanisms the network could adopt: a totally unspecialised strategy, where no $\bW^{(l'')a}$ entering $\bW^{(l':l)a}$ is learned by itself, but still $\bW^{(l':l)a}$ is learned as a whole; a totally specialised strategy, where the model is able to learn separately all the $\bW^{(l'')a}$; mixed strategies where some subsets of layers are specialised while others are not. All these mechanisms give explicit contributions to the correlation between teacher and student, and could correspond to different phases of the system. We leave the study of this rich phase diagram for future work.

To simplify the picture, we require activations such that $\mu_0^{(l)} = \mu_1^{(l)} = \mu_2^{(l)} = 0$ and set $\EE_{z\sim\calN(0,1)}\sigma^{(l)}(z)^2=1$ for all $l\le L$, which this time implies $g^{(l)}(1) = 1$. In this case, the effective weights $\bW^{(l':l)a}$ do not enter the post-activations covariance, simplifying considerably our analysis because the only possible learning strategy is specialisation at all layers. Indeed, 
\begin{talign*} 
\{\lambda^a(\btheta^a):=\frac1{\sqrt{k_L}}{\bv^{0\intercal}}\sigma^{(L)}( \bh^{(L)a} )\}_{a=0}^s, \ \text{and} \ K^{ab}:=\EE_{\bx} \lambda^a \lambda^b    
\end{talign*}
can now be written recursively as
\begin{talign}
    &K^{ab} = \frac{1}{k_{L}} \bv^{0 \intercal} g^{(L)} (\bOmega^{(L)ab})\bv^{0} \, , \nonumber\\
    & \bOmega^{(l)ab} =\frac{1}{k_{l-1}} \bW^{(l)a} g^{(l-1)} (\bOmega^{(l-1)ab} )\bW^{(l)b \intercal},   
    \label{eq:K_deep_recursion}
\end{talign}
with $\bOmega^{(0)ab} = I_d$ for all $a, b$ and $g^{(0)}={\rm id}$ (the identity map). We make the same concentration assumption as before, namely $g^{(l)}(\bOmega^{(l)ab})_{ij} \approx \delta_{ij} g^{(l)}(\bOmega^{(l)ab})_{ii}$. Moreover, we notice that only the neurons in the $L$-th layer can contribute with different importances to the output of the network, being connected to the (potentially) non-homogeneous readout vector $\bv^0$; nothing, in the post-activation covariance, distinguishes neurons in a layer $l< L$. For this reason, the OPs in this case are
\begin{talign*}
    Q_l^{ab} &:= \frac{1}{k_l k_{l-1}} \Tr \bW^{(l)a} \bW^{(l)b\intercal} \ \text{for}\ l=1,\ldots,L-1,\\
    \calQ^{ab}_L(\mathsf{v}) &:= \frac{1}{| \calI_{\mathsf{v}}|k_{L-1}} \sum_{i\in \calI_{\mathsf{v}}} (\bW^{(L)a} \bW^{(L)b\intercal})_{ii} .
\end{talign*}
In terms of these, the post-activations covariance reads
\begin{talign}
    &K^{ab} = \EE_{v\sim P_v} v^2 g^{(L)} \Big(\calQ^{ab}_L(v)g^{(L-1)}\big(Q_{L-1}^{ab}\nonumber\\
    &\hspace{7em}\times g^{(L-2)}(\cdots Q_{2}^{ab} g^{(1)}(Q_1^{ab}) \cdots ) \big) \Big).
    \label{eq:K_deep_recursion_Q}
\end{talign}

The entropic contribution of the OPs is easily computable, as they are all independent of each other. The calculation, reported in App.~\ref{app:replicas_deep}, ultimately yields the free entropy reported in Result~\ref{res:free_entropy_deep}.

\section{Conclusion and perspectives}\label{sec:conclusion}

In this paper we have derived a quantitatively accurate statistical physics description of the optimal generalisation capability of fully-trained linear-width MLPs with an arbitrary number of layers, for broad classes of activation functions, in the challenging scaling regime where the number of parameters is comparable to that of the training data. Even for shallow MLPs, this feature learning regime has resisted for a long time to mean-field approaches used, e.g., in the study of the narrow committee machines \cite{barkai1992broken,engel1992storage,schwarze1992generalization,mato1992generalization}\footnote{The work \cite{afanah2025unified} concurrent with ours proposes an annealed approximation of the free entropy (rather than the more accurate quenched computation proposed in the present paper) to study shallow MLPs of extensive width with ReLU activation.}. 

Our theory has been validated through extensive numerical experiments using Monte Carlo samplers, a generalisation we proposed of the GAMP-RIE algorithm, and a popular algorithm used to train NNs, ADAM, showing in all cases a phenomenological picture consistent with our predictions.

Phase transitions in supervised learning are known in the statistical physics literature at least since~\cite{gyorgyi1990first}, when the analysis was limited to linear models. In this sense, our theory contributes to enrich this landscape, unveiling numerous phase transitions in the learning of different layers' weights for MLPs. Such transitions can occur heterogeneously across layers, and inside each layer. This rich behaviour is captured by functional order parameters with up to two arguments, which nevertheless allow a tractable dimensionality reduction of the problem.


Concerning limitations, for $L\geq 2$ we made some simplifying assumptions on the activation functions in order to reduce the number of order parameters and possible ``learning strategies'' accessible to the trained network. A direction we intend to pursue is to relax them. Specifically, we aim at a theory encompassing a broader class of activation functions for an arbitrary finite number of layers. An additional difficulty is the systematic stability analysis of the potentially many fixed points of the RS equations, each corresponding to a different learning strategy of the network. Our theory opens an avenue for these extensions, as all the order parameters needed are now identifiable, but the analytical treatment and numerical exploration of the problem require further effort. Moreover, as hinted at in App.~\ref{sec:partial_proof}, for $L=1$ and activations without the second Hermite coefficient, we foresee a path for a possible rigorous proof of our results.

A key novelty of our approach is the way we blend matrix models and spin glass techniques in a unified formalism, which is able to handle matrix degrees of freedom which are not necessarily rotationally invariant. It applies to NNs, but also to matrix sensing problems. We foresee that it will be useful beyond the realm of inference/learning problems. Another limitation of the approach is linked to the restricted class of solvable matrix models \cite{kazakov2000solvable,anninos2020notes}. Indeed, as explained in App.~\ref{app:alternatives}, possible improvements would require additional order parameters. Taking them into account yields matrix models when computing their entropy which, to the best of our knowledge, are not currently solvable. 
This is an exciting program at the crossroad of matrix models, inference and learning of extensive-rank~matrices.

Accounting for structured inputs is another challenging perspective. Here we took into consideration a rather simple data model, i.e., Gaussian data with a covariance in the vein of~\citep{monasson1992,loureiro2021real}. It would be desirable to study richer data models like mixture models~\citep{delgiudice1989,loureiro2021gmm}, hidden manifolds~\citep{goldt2020modeling}, object manifolds and simplexes~\citep{lopez1995storage,chung2018manifold,rotondo2020beyond,pastore2020structure}, hierarchical data~\citep{cagnetta2024deep,sclocchi2025hierarchical}.

We have considered the idealised matched teacher-student setting as a first step, with the goal of tackling the methodological bottlenecks associated with the depth and linear-width of the NN. Even in this simpler Bayes-optimal scenario, which in particular prevents replica symmetry breaking \cite{mezard1987spin}, the solution required the development of a non-standard approach. A natural next step is to consider targets belonging to a different function class than the trainable MLP, and focus on training by empirical risk minimisation (zero temperature) rather than Bayesian (finite temperature). We believe this generalisation can be carried out without major modifications of the theory, at least in the case where the target remains an MLP but with a different architecture. A complementary natural continuation is to push our formalism further in order to tackle deep architectures beyond the MLP, such as convolutional networks, restricted Boltzmann machines or transformers.

The identification of the relevant order parameters for the characterisation of the equilibrium state carried out in our contribution paves the way for the study of the learning dynamics of first-order methods in similar settings. Indeed, there exist classical methods rooted in physics to study the learning dynamics of NNs \cite{solla1993online,NIPS1995_a1519de5,Goldt_2020,cugliandolo2023recentapplicationsdynamicalmeanfield}. Recently, \cite{montanari2025dynamical} exploits these techniques to study the learning dynamics of a large NN trained on a GLM target, observing a separation of timescales between generalisation and over-fitting: it could be interesting to use the insights from our equilibrium analysis to extend their approach to more expressive targets. 
In the context of the dynamics of learning, it is also relevant to consider power-law distributed readouts in the target, as many groups are currently doing in order to capture neural scaling laws \cite{ren2025scalinglaws,bordelon2024dynamicalmodelneuralscaling,paquette202543phasescomputeoptimalneural,lin2025scalinglawslinearregression,oko2024learningsumdiversefeatures} in extensive-width shallow NNs.

On a more speculative footing, we would like to point out that the specialisation transition we uncovered in the interpolation asymptotics is only one of the known mechanisms for feature learning: kernel adaptation~\cite{ringel2025applications,aiudi2025local}, grokking~\cite{power2022grokking,rubin2024grokking}, etc. One of the ambitions of an asymptotic theory is to provide a good approximation for large but realistic networks, where however these different mechanisms can be at play together. We thus believe that understanding the interplay between them, as proposed in~\cite{rubin2024unified,rubin2026mitigating}, and eventually unifying their taxonomy, is of practical relevance for the future.


\section*{Acknowledgements}
F.C. was affiliated with the Abdus Salam International Centre for Theoretical Physics during the conduction of this work. J.B., F.C., M.-T.N. and M.P. were funded by the European Union (ERC, CHORAL, project number 101039794). Views and opinions expressed are however those of the authors only and do not necessarily reflect those of the European Union or the European Research Council. Neither the European Union nor the granting authority can be held responsible for them. M.P. thanks Vittorio Erba and Pietro Rotondo for interesting discussions and suggestions.

\bibliography{refs_PRX}

\appendix

\clearpage

\onecolumngrid

\begin{center}
\textbf{APPENDICES}
\end{center}

\tableofcontents

\section{Notations, pre-requisites and auxiliary results}\label{app:auxiliaryresults}

\subsection{Notations} \label{app:notations}

$\bullet$ \textbf{Fonts:} \ Bold symbols are reserved for vectors and matrices. For the order parameters, a calligraphic symbol such as $\calQ$ will emphasise that it is a function, while $Q$ is a scalar. 

$\bullet$ \textbf{Thermodynamics limit:} \ The limit $\lim$ without further specification will always correspond to the joint limit of large input dimension, NN layer widths and number of data all diverging, $d,k_l,n\to+\infty$, with scaling \eqref{thermolim}; it will be called the \emph{thermodynamic limit}. 

$\bullet$ \textbf{Hermite decomposition of the activation:}  $(\mu_\ell)$ are the Hermite coefficients of the activation $\sigma$ when expressed in the orthogonal basis of Hermite polynomials $(\He_\ell(x))$, see \eqref{eq:HermiteSig}. When different activations at each layer are considered, $\sigma^{(l)}$ denotes the activation at layer $l$ and $\mu^{(l)}_\ell$ its $\ell$-th Hermite coefficient.

$\bullet$ \textbf{Vectors and matrices:} \ A vector $\bx$ is always considered to be in column form, its transpose $\bx^\intercal$ is a row and the inner product is thus $\bu^\intercal \bv=\sum_{i} u_iv_i$. The norm $\|\bA\|=(\sum_{ij}A_{ij}^2)^{1/2}$ is the Frobenius norm and the Euclidean norm for a vector. The trace operator for matrices is $\Tr$. The $\ell$-th Hadamard (entry-wise) power of a matrix is denoted with a superscript $\circ \ell$. 

$\bullet$ \textbf{Replicas:} \ The superscript $0$ will always indicate a quantity associated with the target function, while superscript $a,b=1,\ldots,s$ will be used for ``replicas'' in the replica method, or for i.i.d. samples from the posterior distribution.

$\bullet$ \textbf{Probability and expectations:} \ Symbol $\sim$ expresses that a random variable is drawn from a certain law. $\mathbb{P}(\,\cdot\,)$ is a probability, $P(\,\cdot\,)$ is a density function w.r.t. Lebesgue measure, $dP(\,\cdot\,)$ the associated probability, $P(\,\cdot\mid Y)$ is the conditional density given $Y$. $\mathcal{N}(m,\Sigma)$ is the density of a Gaussian with mean $m$ and variance $\Sigma$; $\mathcal{N}(\mathbf{m},\bSig)$ is the multivariate version.  The expectation operator w.r.t. to a generic random variable $X$ is denoted $\EE_X$, the conditional expectation of $X$ given $Y$ is $\EE[\,\cdot\mid Y]$, and the expectation w.r.t. to all ensuing random variables entering an expression is simply $\EE$. The bracket notation is reserved for an expectation w.r.t. an arbitrary (but $d$-independent) number of samples $(\btheta^a)_a$ from the posterior given the training data $dP(\,\cdot\, \mid\mathcal{D})^{\otimes \infty}$: $\langle f((\btheta^a)_a) \rangle :=\EE[ f((\btheta^a)_a) \mid \mathcal{D}]$. 

$\bullet$ \textbf{Integrals and densities:} \ When unspecified, the integration domain of an integral $\int f(X) dX$ is $\mathbb{R}^{{\rm dim}(X)}$. For a sequence of symmetric real matrices $\bX=\bX_d$ indexed by the dimension $d$, the density $\rho_\bX(s)$ w.r.t. to Lebesgue measure is the weak limit of the empirical law of its (real) eigenvalues $\lim_{d\to\infty}\frac1d\sum_{i\le d} \delta(s-\lambda_i(\bX_d))$, where $\delta(\,\cdot\,)$ is the delta Dirac function; the Kronecker delta is denoted $\delta_{ij}$. 

$\bullet$ \textbf{Scalings and proportionalities:} \ Symbol $\propto$ means equality up to a multiplicative constant (which may be $d$-dependent). We use the standard BigO and smallO notations $O(\,\cdot\,), o(\,\cdot\,)$. In particular, $o_d(1)$ means a sequence vanishing as $d\to\infty$. The notation $f_d=\Theta(g_d)$ means that the two sequences verify $f_d/g_d\to C$ for some constant $C\in (0,+\infty)$ as $d\to\infty$. $\approx$ means equality up to a correction $o_d(1)$. 

$\bullet$ \textbf{Information-theoretic notions:} The mutual information between two random variables $X,Y$ with joint law $P_{X,Y}$ and marginals $P_X,P_Y$ is the Kullback-Leibler divergence $I(X;Y)=I(Y;X):=D_{\rm KL}(P_{X,Y}\|P_X\otimes P_Y)=H(Y)-H(Y\mid X)=H(X)-H(X\mid Y)$, where $H(X)$ is the Shannon entropy if $X$ is discrete, and it is instead the differential entropy if continuous-valued. Similarly, $H(Y\mid X)$ is the conditional Shannon or differential entropy. We refer to \cite{cover1999elements} if these information-theoretic notions are not familiar.

\subsection{Hermite basis and Mehler's formula\label{app:hermite}}
Recall the Hermite expansion of the activation:
\begin{equation}
    \sigma(x) = \sum_{\ell = 0}^{\infty} \frac{\mu_\ell}{\ell !}\He_\ell (x).
    \label{eq:sigma_hermite}
\end{equation}
We are expressing it on the basis of the probabilist's Hermite polynomials, generated through
\begin{equation}
    \He_\ell(z) = \frac{\diff^\ell}{{\diff t}^\ell} \exp\big(t z - t^2/2 \big)\Big|_{t=0}.
    \label{eq:hermite_G}
\end{equation}
The Hermite basis has the property of being orthogonal with respect to the standard Gaussian measure, which is the distribution of the input data. Specifically, if $z\sim\mathcal{N}(0,1)$
\begin{equation}
    \EE \,\He_k(z) \He_\ell(z) = \ell!\,\delta_{k\ell}.
\end{equation}
By orthogonality, the coefficients of the expansions can be obtained as
\begin{equation}
    \mu_\ell = \EE\,\He_\ell(z)\sigma(z).
\end{equation}
Moreover,
\begin{equation}
    \label{eq:avg_sigma}
    \E[\sigma(z)^2]=\sum_{\ell=0}^{\infty} \frac{\mu_\ell^2}{\ell!}.
\end{equation}
These coefficients for some popular choices of $\sigma$ are reported in Table~\ref{tab:Hermite} for reference.
\begin{table}[b]
    \caption{First Hermite coefficients of some activation functions reported in the figures. $\theta$ is the Heaviside step function.}
    \label{tab:Hermite}
    \vskip 0.15in
    \begin{center}
    \begin{small}
    \begin{tabular}{l|cccccccc}
    \toprule
        $\sigma(z)$ & $\mu_0$ & $\mu_1$ & $\mu_2$ & $\mu_3$ & $\mu_4$ & $\mu_5 $ &$\cdots$ & $\E_{z\sim\mathcal{N}(0,1)}[\sigma(z)^2]$  \\
        \midrule
        $\relu(z) = z \theta(z)$ & $1/\sqrt{2 \pi }$ & $1/2$ & $1/\sqrt{2 \pi }$  & 0 & $-1/\sqrt{2 \pi }$& 0 & $\cdots$ & 1/2 \\
        $\tanh(2z)$ & 0 & 0.72948 & 0  & -0.61398 & 0 & 1.5632 & $\cdots$ & 0.63526 \\
        $\tanh(2z)/\sigma_{\tanh} $ & 0 & 0.91524 & 0 & -0.77033 & 0& 1.9613 & $\cdots$ & 1 \\
        \bottomrule
    \end{tabular}
    \end{small}
    \end{center}
    \vskip -0.1in
\end{table}
The Hermite basis can be generalised to an orthogonal basis with respect to the Gaussian measure with generic variance. Let $z\sim\mathcal{N}(0,r)$, then
\begin{equation}
    \He_\ell^{[r]}(z) = \frac{\diff^\ell}{\diff t^\ell}  \exp\big(tz -  t^2 r/2\big)\Big|_{t=0}.
\end{equation}
For this basis one has
\begin{equation}
    \EE\, \He_k^{[r]}(z) \He_\ell^{[r]}(z) =  \ell! \,r^\ell\delta_{k\ell}.
\end{equation}

Consider now a couple of jointly Gaussian random variables $\bx=(u,v)\sim\mathcal{N}(0,\bC)$ with
\begin{align}
    \bC=
    \begin{pmatrix}
        r&q\\
        q&r
    \end{pmatrix}\,.
\end{align}
Then, by Mehler's formula
\begin{equation}
    \frac{1}{2\pi\sqrt{r^2-q^2}} \exp\!\Big[-\frac{1}{2} \bx^\intercal\, \mathbf{C}^{-1}\bx \Big] = \frac{e^{-\frac{u^2}{2r}}}{\sqrt{2\pi r}} \frac{e^{-\frac{v^2}{2r}}}{\sqrt{2\pi r}} \sum_{\ell = 0}^{+\infty} \frac{q^{\ell}}{\ell! r^{2\ell}} \He_{\ell}^{[r]}(u) \He_{\ell}^{[r]}(v),
    \label{eq:mehler}
\end{equation}
and by orthogonality of the Hermite basis, \eqref{eq:K} readily follows by noticing that the variables 
$(h_i^a = (\bW^a \bx)_i/\sqrt{d})_{i,a}
$
at given $(\bW^a)$ are Gaussian with covariances $\Omega^{ab}_{ij}=\bW_i^{a\intercal}\bW^b_j/d$:
\begin{equation}
    \EE \,\sigma(h_{i}^a)\sigma(h_{j}^b) = \sum_{\ell=0}^{\infty} \frac{(\mu_\ell^{[r]})^2}{\ell!r^{2\ell}} (\Omega_{ij}^{ab})^\ell,\qquad \mu_\ell^{[r]} = \EE_{z\sim\calN(0,r)} \He^{[r]}_\ell(z)\sigma(z).
\end{equation}
Moreover, as $\Omega^{aa}_{ii}$, to be identified with $r$ above, converges to the variance of the prior of $\bW^0$ for large $d$ by Bayes-optimality, whenever $\Omega^{aa}_{ii} \to 1$ we can specialise this formula to the simpler case $r=1$ we reported in the main text.

\subsection{Nishimori identities}\label{app:nishiID}
The Nishimori identities are a set of symmetries arising in inference in the Bayes-optimal setting as a consequence of Bayes' rule. To introduce them, consider a test function $f$ of the teacher weights, collectively denoted by $\btheta^0$, of $s-1$ replicas of the student's weights $(\btheta^a)_{2\leq a\leq s}$ drawn conditionally i.i.d. from the posterior, and possibly also of the training set $\mathcal{D}$: $f(\btheta^0,\btheta^2,\dots,\btheta^s ;\mathcal{D})$. Then
\begin{align}
    \mathbb{E}_{{\boldsymbol{\theta}}^0,\mathcal{D}}\langle f(\btheta^0,\btheta^2,\dots,\btheta^s ;\mathcal{D})\rangle=
    \mathbb{E}_{{\boldsymbol{\theta}}^0,\mathcal{D}}\langle f(\btheta^1,\btheta^2,\dots,\btheta^s ;\mathcal{D})\rangle.
\end{align}The Nishimori identities thus allow us to replace the teacher's weights with another replica from the posterior measure. The proof follows from Bayes' theorem, see e.g. \cite{barbier2019glm}.

The Nishimori identities have some consequences also on our replica symmetric ansatz for the free entropy. In particular, they constrain the values of the asymptotic mean of some OPs. For instance, consider the shallow case
\begin{align}
    R_2^{a0}= \lim
    \frac{1}{d^2}\EE_{\mathcal{D},\btheta^0}\langle\Tr[\bS_2^a\bS_2^0]\rangle=
    \lim\frac{1}{d^2}\EE_{\mathcal{D}}\langle\Tr[\bS_2^a\bS_2^b]\rangle= R_2^{ab},\quad \text{for }a\neq b.
\end{align}In addition, within the replica symmetry assumption, all the above overlaps are equal to one another.

Combined with the concentration of OPs, which can be proven in great generality in Bayes-optimal inference \cite{barbier2021overlap,barbier2022strong}, the Nishimori identities fix the values of some of them. For instance, we have that with high probability
\begin{align}
    \frac{1}{d^2}\Tr[(\bS_2^a)^2]\to R_{\rm d}=\lim\frac{1}{d^2}\EE_{\mathcal{D}}\langle\Tr[(\bS_2^a)^2]\rangle=\lim\frac{1}{d^2}\EE_{\btheta^0}\Tr[(\bS_2^0)^2]=1+\gamma\bar v^2,
\end{align} with $\EE v=\bar v$. When this happens, as for $R_{\rm d}$, then the respective Fourier conjugates $\hat R_{\rm d}$ vanish, since the desired constraints were already asymptotically enforced without the need of additional delta functions. This is because the configurations in which the OPs take those values exponentially (in $n$) dominate the posterior measure, so these constraints are automatically imposed by the measure. Another OP for which we have a similar consequence is
\begin{align}
    \calQ^{aa}(\mathsf{v})=\lim \frac{1}{d|\mathcal{I}_\mathsf{v}|} \sum_{i\in \mathcal{I}_\mathsf{v}} \EE_\calD\langle \bW^a_i\cdot\bW^a_i \rangle=\lim \frac{1}{d|\mathcal{I}_\mathsf{v}|} \sum_{i\in \mathcal{I}_\mathsf{v}} \EE_{\btheta^0}\|\bW^0_i\|^2=1
\end{align}and consequently $\hat \calQ^{aa}(\mathsf{v})=0 $.

\begin{figure}[b!]
    \centering
    \includegraphics[width=.6\linewidth,trim={0 0 -0.5cm 0},clip]{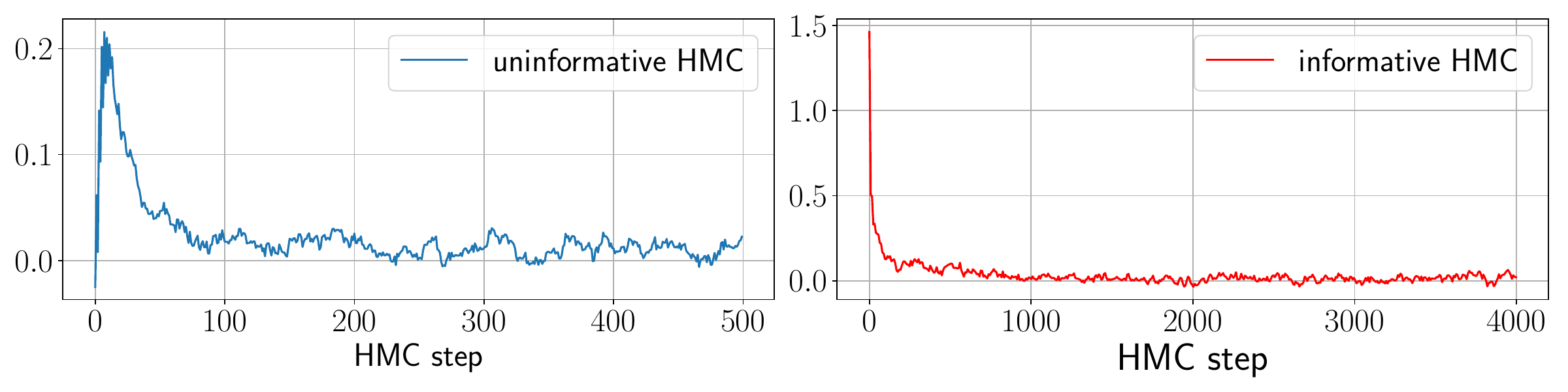}
    \caption{Trajectories of $\EE_{\bx_{\rm test}} (\lambda_t^1 - \lambda_t^2)^2/ \EE_{\bx_{\rm test}} (\lambda_t^1 - \lambda^0)^2-1$, where $\lambda_t^a:=\lambda_{\rm test}(\btheta_t^a)$ and $\lambda^{0}:=\lambda_{\rm test}(\btheta^0)$. $\btheta_t^1$ is the HMC sample at step $t$ for the first chain independently initialised from the second; $\EE_{\bx_{\rm test}}$ is an average over $5.10^4$ test samples.
    Here $L=1, d=150, \gamma=0.5, \Delta=0.1, \alpha=5, \sigma={\rm ReLU}, \bW$ is Gaussian and $\bv$ is homogenenous. HMC runs are initialised uninformatively (\textbf{left}) and informatively (\textbf{right}), in order to probe the metastable and equilibrium states, respectively. 
    This quantity approaches zero for long enough times, indicating the empirical validity of the Nishimori identity $\EE\langle\lambda^{1}\lambda^{2}\rangle= \EE\langle\lambda^{1}\rangle \lambda^{0}$ both for the posterior average $\langle\,\cdot\,\rangle$ and the average over the metastable state $\langle\,\cdot\,\rangle_{\rm meta}$ (see Remark \ref{rmk:sampling} in the main text).}
    \label{fig:NishiBreak}
\end{figure}

Let us now draw some more generic conclusions we shall need in the following. Given a generic set of OPs labelled by replica indices, say $\bQ=(Q^{ab})_{a\leq b=0,\dots,s}$, a replica symmetric ansatz for it would enforce the following form:
\begin{align}
\label{eq:generic_OP_RS_ansatz}
    \bQ=
    \begin{pmatrix}
       \rho& m\mathbf{1}^\intercal_s\\
        m\mathbf{1}_s& (Q_d-Q)I_s+Q \mathbf{1}_s\mathbf{1}^\intercal_s
    \end{pmatrix}\in\mathbb{R}^{s+1\times s+1},
\end{align}where $\mathbf{1}_s=(1,1,\dots,1)\in\mathbb{R}^s$ and $I_s\in\mathbb{R}^{s\times s}$ is the identity matrix. Under rather general terms, the Nishimori identities are actually enforcing the constraints $\rho=Q_d$, $m=Q$, yielding
\begin{align}\label{eq:generic_OP_RS_ansatz_Nishimori}
    \bQ=
    \begin{pmatrix}
        Q_d& Q\mathbf{1}^\intercal_s\\
        Q\mathbf{1}_s& (Q_d-Q)I_s+Q \mathbf{1}_s\mathbf{1}^\intercal_s
    \end{pmatrix}\in\mathbb{R}^{s+1\times s+1}.
\end{align}

As we explained in this section, the Nishimori identities are a property of posterior measures sampled at equilibrium in the Bayes optimal setting. In the numerical part of this paper exploiting Monte Carlo sampling, we checked their validity not only at equilibrium, but also whenever the algorithm is stuck in a metastable state captured by sub-optimal branches of our theory (see Remark \ref{rmk:sampling} in the main text). We report an explicit check of this fact in \figurename~\ref{fig:NishiBreak}.

\subsection{Linking free entropy and mutual information\label{app:mutual_info}}

It is possible to relate the mutual information (MI) of the inference problem to the free entropy $f_n=\E\ln \mathcal{Z}$ introduced in the main. Indeed, we can write the MI as
\begin{equation}
    \frac{I(\btheta^0;\mathcal{D})}{n} = \frac{H(\mathcal{D})}{n} - \frac{H(\mathcal{D}\mid\btheta^0)}{n},
\end{equation}
where $H(Y\mid X)$ is the conditional Shannon entropy of $Y$ given $X$. Using the chain rule for the entropy, and the definition \eqref{eq:free_entropy_def}, the free entropy can be recast as
\begin{equation}
  -f_n = \frac{H(\{ y_\mu\}_{\mu \leq n}\mid \{ \bx_\mu\}_{\mu \leq n})}{n} = \frac{H(\mathcal{D})}{n} - \frac{H(\{ \bx_\mu\}_{\mu \leq n})}{n}.
\end{equation}
On the other hand $H(\mathcal{D}\mid\btheta^0)=H(\{y_\mu\}\mid\btheta^0,\{\bx_\mu\})+H(\{\bx_\mu\})$, i.e.,
\begin{equation}
    \frac{H(\mathcal{D}\mid \btheta^0)}{n} \approx -\EE_{\lambda}  \int dy P_{\text{out}}(y\mid \lambda) \ln P_{\text{out}}(y\mid\lambda) + \frac{H(\{ \bx_\mu\}_{\mu \leq n})}{n} ,
\end{equation}
where $\lambda\sim \calN(0,K_{\rm d})$, with $K_{\rm d}$ given by \eqref{eq:covariance_vars_main} for $L=1$, while for $L=2$ and $L=3$, given our normalisation assumptions on the activation functions, $K_{\rm d}=1$ (assuming here that $\mu_0=0$, see App.~\ref{app:non-centered} if the activation $\sigma$ is non-centred). Equality holds asymptotically in the limit $\lim$.
This allows us to express the MI asymptotically as
\begin{equation}
\label{eq:MI_generic_channel}
    \frac{I(\btheta^0;\mathcal{D})}{n} = - f_n + \EE_{\lambda}  \int dy P_{\text{out}}(y|\lambda) \ln P_{\text{out}}(y|\lambda) +o_n(1).
\end{equation} 
Specialising the equation to the Gaussian channel, one obtains
\begin{equation}
\label{eq:MI_gaussian_channel}
    \frac{I(\btheta^0;\mathcal{D})}{n} = - f_n - \frac{1}{2} \ln(2\pi e \Delta).
\end{equation} 
We chose to normalise by $n$ because, in our scaling, it is always proportional to the number of total parameters, that is $\Theta(d^2)$. Hence with this choice one can interpret the parameter $\alpha$ as an effective signal-to-noise ratio.


\begin{remark}
The arguments of \cite{barbier2024phase} to show the existence of an upper bound on the mutual information per variable in the case of discrete variables, and the associated inevitable breaking of prior universality beyond a certain threshold in matrix denoising, apply to the present model too. It implies, as in the aforementioned paper, that the mutual information per variable cannot go beyond $\ln 2$ for Rademacher inner weights. Our theory is consistent with this fact; this is a direct consequence of the analysis in App.~\ref{app:large_alpha} carried out for the shallow case $L=1$ (see in particular~\eqref{eq:large_alpha_limit}) specialised to binary prior over $\bW$.
\end{remark}

\subsection{Alternative representation for \texorpdfstring{$\varepsilon^{\rm opt}$ with $L=1$}{}}\label{app:gen_err}

We recall that $\btheta^0=(\bv^0,\bW^0)$ and similarly for $\btheta^1=\btheta,\btheta^2,\ldots$ which are replicas, i.e., conditionally i.i.d. samples from $dP(\bW,\bv\mid \mathcal{D})$ (the reasoning below applies whether $\bv$ is learnable or quenched, so in general we can consider a joint posterior over both). From its definition \eqref{eq:gen_error_def}, the Bayes-optimal generalisation error can be recast as
\begin{align}
    \varepsilon^{\rm opt}= \EE_{\btheta^0,\bx_{\rm test} } \EE[y^2_{\rm test}\mid \lambda^0]
    -2\EE_{\btheta^0,\mathcal{D},\bx_{\rm test}} \EE[y_{\rm test}\mid \lambda^0]\langle\EE[y\mid \lambda]\rangle
    +  \EE_{\btheta^0,\mathcal{D},\bx_{\rm test}}\langle \EE[y\mid \lambda]\rangle^2,\label{eyStartingPoint}
\end{align}
where $\EE[y\mid \lambda]=\int dy\, y\, P_{\rm out}(y\mid \lambda)$, and  $(\lambda^a)_{a=0,\ldots,s}$ are the random variables (random due to the test input $\bx_{\rm test}$, drawn independently of the training data $\mathcal{D}$, and their respective weights $\btheta^0,\btheta$)
\begin{align}
    \lambda^a=\lambda_{\rm test}(\btheta^a) =\frac{\bv^{a\intercal}}{\sqrt{k}}\sigma\Big(\frac{\bW^a\bx_{\rm test}}{\sqrt{d}}\Big).
\end{align}
Recall that the bracket $\langle\,\cdot\,\rangle$ is the average w.r.t. to the posterior and acts on $\btheta^1=\btheta,\btheta^2,\ldots$. Notice that the last term on the r.h.s.\ of \eqref{eyStartingPoint} can be rewritten as
\begin{align*}
    \EE_{\btheta^0,\mathcal{D},\bx_{\rm test}}\langle \EE[y\mid \lambda]\rangle^2=
    \EE_{\btheta^0,\mathcal{D},\bx_{\rm test}}\langle   \EE[y\mid \lambda^1]\EE[y\mid \lambda^{2}]\rangle,
\end{align*}with superscripts being replica indices.

In order to show Result~\ref{res:gen_error} for a generic $P_{\rm out}$ we assume the joint Gaussianity of the variables $(\lambda^0,\lambda^1,\lambda^2,\ldots)$, with covariance given by $K^{ab}$ with $a,b\in\{0,1,2,\ldots\}$. Indeed, in the limit ``$\lim$'', the theory considers $(\lambda^a)_{a\ge 0}$ as jointly Gaussian under the randomness of a common input, here $\bx_{\rm test}$, conditionally on the weights $(\btheta^a)$. Their covariance depends on the weights $(\btheta^a)$ through various overlap OPs introduced in the main. In the large limit ``$\lim$'' these overlaps are assumed to concentrate under the quenched posterior average $\EE_{\btheta^0,\mathcal{D}}\langle\,\cdot\,\rangle$ towards non-random asymptotic values corresponding to the extremiser globally maximising the RS potential in Result~\ref{res:free_entropy_shallow}, with the overlaps entering $K^{ab}$ through \eqref{eq:K_RS}. Using the replica symmetric ansatz (see \eqref{eq:K-Kd-def}), and the Nishimori identities, the covariance evaluated on these overlaps shall be denoted $\bK^*$ as in the main, and its elements are written as $K^{*\,ab}=K^*+\delta_{ab}(K_{\rm d}-K^*)$. This hypothesis is then confirmed by the excellent agreement between our theoretical predictions based on this assumption and the experimental results. This implies directly \eqref{eq:Bayes_error_result} in Result~\ref{res:gen_error} from definition \eqref{eq:Bayes_error_def}. For the special case of optimal mean-square generalisation error it yields
\begin{align}\label{eq:general_gen_error_Gaussian_eq}
    \lim \,\varepsilon^{\rm opt}=\EE_{\lambda^0}\EE[y^2_{\rm test}\mid \lambda^0]-2
    \EE_{\lambda^0,\lambda^1}\EE[y_{\rm test}\mid \lambda^0]\EE[y\mid \lambda^1]
    +\EE_{\lambda^1,\lambda^2}\EE[y\mid \lambda^1]\EE[y\mid \lambda^2]
\end{align}
where, in the replica symmetric ansatz, 
\begin{align}
    \EE[(\lambda^0)^2]=K_{\rm d},\quad \EE[\lambda^0\lambda^1]= \EE[\lambda^0\lambda^2]=K^{*},\quad \EE[\lambda^1\lambda^2]=K^{*},\quad 
    \EE[(\lambda^1)^2]=\EE[(\lambda^2)^2]=K_{\rm d}.
\end{align} We thus have
\begin{align}
    \EE_{\lambda^0,\lambda^1}\EE[y_{\rm test}\mid \lambda^0]\EE[y\mid \lambda^1]
    =\EE_{\lambda^1,\lambda^2}\EE[y\mid \lambda^1]\EE[y\mid \lambda^2].
\end{align}Plugging the above in \eqref{eq:general_gen_error_Gaussian_eq} yields \eqref{eq:gen_err_result}.

\begin{figure}[t]
    \centering
    \includegraphics[width=0.5\linewidth]{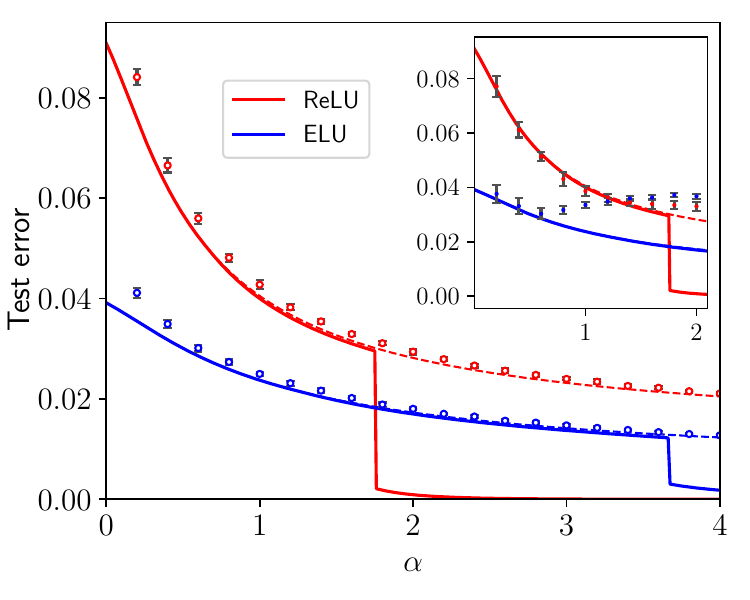}
    \caption{Theoretical prediction (solid curves) of the Bayes-optimal mean-square generalisation error for \emph{binary inner weights} and ReLU, eLU activations, with $\gamma = 0.5$, $d=150$, Gaussian label noise with $\Delta=0.1$, and fixed readouts $\bv  = \mathbf{1}$. Dashed lines are obtained from the solution of the fixed point equations \eqref{eq:NSB_equations_generic_ch} with all $\mathcal{Q}(\mathsf{v})=0$. Circles are the test error of GAMP-RIE \citep{maillard2024bayes} extended to generic activation.
    The MCMC points initialised uninformatively (inset) are obtained using \eqref{eq:general_finitesize_gen_error_noapprox}, to account for lack of equilibration due to glassiness, which prevents using \eqref{eq:simple_gen_error_for_numerics} (see Remark~\ref{rmk:Nishi_breaks}). Even in the possibly glassy region, the GAMP-RIE attains the universal branch performance. 
    Data for GAMP-RIE and MCMC are averaged over 16 data instances, with error bars representing 1 standard deviation over instances. GAMP-RIE's performance follows the universal theoretical curve even in the $\alpha$ regime where MCMC sampling experiences a computationally hard phase with worse performance, and in particular after $\alpha_{\rm sp}$.}
    \label{fig:GAMP-RIE_eLU-ReLU}
\end{figure}

Let us now prove a formula for the optimal mean-square generalisation error written in terms of the overlaps which holds for the special case of linear readout with Gaussian label noise $P_{\rm out}(y\mid \lambda)=\exp(-\frac1{2\Delta}(y-\lambda)^2)/\sqrt{2\pi \Delta}$. The following derivation is exact and does not require any Gaussianity assumption on the random variables $(\lambda^a)$. For the linear Gaussian channel the means verify $\EE[y\mid\lambda]=\lambda$ and $\EE[y^2\mid\lambda]=\lambda^2+\Delta$. Plugged into \eqref{eyStartingPoint} this yields
\begin{align}
    \varepsilon^{\rm opt}-\Delta=\EE_{\btheta^0,\bx_{\rm test} } (\lambda^0)^2
    -2\EE_{\btheta^0,\mathcal{D},\bx_{\rm test}}  \lambda^0 \langle\lambda\rangle
    +  \EE_{\btheta^0,\mathcal{D},\bx_{\rm test}}\langle  \lambda^1\lambda^2\rangle,
\end{align}whence we clearly see that the generalisation error depends only on the covariance of $\lambda^0,\lambda^1,\lambda^2$ under the randomness of the shared input $\bx_{\rm test}$ at fixed weights, regardless of the validity of the Gaussian hypothesis on the post-activations~\eqref{eq:hypothesis_gaussian_test} we assume in the replica computation. This covariance was already computed in \eqref{eq:K}; we recall it here for the reader's convenience
\begin{align}
    K(\btheta^a,\btheta^b) :=\EE_{\bx_{\rm test}} \lambda^a\lambda^b=\sum_{\ell=1}^\infty\frac{\mu_\ell^2}{\ell!} \frac{1}{k}\sum_{i,j=1}^k v_i^a(\Omega^{ab}_{ij})^\ell v^b_j=
    \sum_{\ell=1}^\infty\frac{\mu_\ell^2}{\ell!}R_\ell^{ab},
\end{align}
where $\Omega^{ab}_{ij}:= \bW_{i}^{a\intercal} \bW_{j}^b/d$, and $R_\ell^{ab}$ as introduced in \eqref{eq:K} for $a,b=0,1,2$. We stress that $K(\btheta^a,\btheta^b)$ is not the limiting covariance $K^{ab}$ whose elements are in \eqref{eq:K-Kd-def}, but rather the finite size one. $K(\btheta^a,\btheta^b)$ provides us with an efficient way to compute the generalisation error numerically, used in App.~\ref{app:GAMP}, \figurename~\ref{fig:GAMP-RIE_eLU-ReLU}, that is through the formula
\begin{align}
    \varepsilon^{\rm opt}-\Delta&=\EE_{\btheta^0}K(\btheta^0,\btheta^0)-2\EE_{\btheta^0,\mathcal{D}}\langle K(\btheta^0,\btheta^1)\rangle+
    \EE_{\btheta^0,\mathcal{D}}\langle K(\btheta^1,\btheta^2)\rangle=\sum_{\ell=1}^\infty\frac{\mu_\ell^2}{\ell!}\EE_{\btheta^0,\mathcal{D}} \langle R_{\ell}^{00}- 2R_{\ell}^{01}+ R^{12}_{\ell} \rangle.
    \label{eq:general_finitesize_gen_error_noapprox}
\end{align}
In the above, the posterior measure $\langle\,\cdot\,\rangle$ is taken care of by Monte Carlo sampling (when it equilibrates). In addition, as in the main text, we assume that in the large system limit the (numerically confirmed) identity \eqref{eq:Qell_NSB} holds. Putting all ingredients together we get
\begin{align}
    \varepsilon^{\rm opt}-\Delta
    =\EE_{\btheta^0,\mathcal{D}} &\Big\langle\mu_1^2(R_1^{00}-2R^{01}_1 +R^{12}_1)+\frac{\mu_2^2}{2}(R_2^{00}-2R^{01}_2+R^{12}_2)\nonumber\\ &+
    \EE_{v\sim P_v}v^2\big[g(\mathcal{Q}^{00}(v))- 2g(\mathcal{Q}^{01}(v))+g(\mathcal{Q}^{12} (v))\big] \Big\rangle.    \label{eq:general_finitesize_gene_error}
\end{align}
In the Bayes-optimal setting one can use again the Nishimori identities that imply $\EE_{\btheta^0,\mathcal{D}} \langle R^{12}_{1}\rangle=\EE_{\btheta^0,\mathcal{D}} \langle R^{01}_{1}\rangle$, and analogously $\EE_{\btheta^0,\mathcal{D}} \langle R^{12}_{2}\rangle=\EE_{\btheta^0,\mathcal{D}} \langle R^{01}_{2}\rangle$ and $\EE_{\btheta^0,\mathcal{D}} \langle g(\mathcal{Q}^{12}_{W}(v))\rangle=\EE_{\btheta^0,\mathcal{D}} \langle g(\mathcal{Q}^{01}_{W}(v))\rangle$. Inserting these identities into \eqref{eq:general_finitesize_gene_error} one gets
\begin{align}
    \varepsilon^{\rm opt}-\Delta&=\EE_{\btheta^0,\mathcal{D}} \Big\langle\mu_1^2(R_1^{00}-R^{01}_1 )+\frac{\mu_2^2}{2}(R_2^{00}-R^{01}_2)+
    \EE_{v\sim P_v}v^2\big[g(\mathcal{Q}^{00}(v))-g(\mathcal{Q}^{01} (v))\big] \Big\rangle.
    \label{eq:simple_gen_error_for_numerics}
\end{align}
This formula relies only on the validity of~\eqref{eq:Qell_NSB}, and makes no assumption on the law of the $\lambda$'s. That it depends only on their covariance is simply a consequence of the quadratic nature of the mean-square generalisation error.

\begin{remark}
\label{rmk:Nishi_breaks}
    Note that the derivation up to \eqref{eq:general_finitesize_gen_error_noapprox} did not assume Bayes-optimality (while \eqref{eq:simple_gen_error_for_numerics} does). Therefore, one can consider it in cases where the true posterior average $\langle \,\cdot\,\rangle$ is replaced by one which does not verify the Nishimori identities. \eqref{eq:general_finitesize_gen_error_noapprox} is the formula we use to compute the generalisation error of Monte Carlo-based estimators in the inset of \figurename~\ref{fig:GAMP-RIE_eLU-ReLU}. There, MCMC cannot equilibrate and experiences a glassy regime. This regime, at variance with the metastable states described in the main text (see Remark~\ref{rmk:sampling}), does not correspond to any sub-optimal branch of our theory: we verified numerically that indeed the Nishimori identities do not hold there.
\end{remark}

\begin{remark}\label{rem:Gibbs_error}
    Using the Nishimori identity of App.~\ref{app:nishiID} and again that, for the linear readout with Gaussian label noise $\EE[y\mid\lambda]=\lambda$ and $\EE[y^2\mid\lambda]=\lambda^2+\Delta$, it is easy to check that the so-called Gibbs error
    \begin{equation}
        \varepsilon^{\rm Gibbs} := \EE_{\bm{\theta}^0,\calD,\bx_{\rm test},y_{\rm test}}
        \big\langle (y_{\rm test} - \EE[y\mid \lambda_{\rm test}(\btheta)])^2 \big\rangle
        \label{eq:Gibbs_error}
    \end{equation}
    is related for this channel to the Bayes-optimal mean-square generalisation error through the identity
    \begin{equation}
        \varepsilon^{\rm Gibbs} - \Delta = 2(\varepsilon^{\rm opt} - \Delta).
        \label{eq:Gibbs_v_Bayes_error}
    \end{equation}
    We exploited this relationship together with the concentration of the Gibbs error w.r.t. the quenched posterior measure $\EE_{\btheta^0,\mathcal{D}}\langle\,\cdot\,\rangle$ when evaluating the numerical generalisation error of the Monte Carlo algorithms reported in the main text.
\end{remark}

\section{Shallow MLP} \label{app:shallowMLP}

\subsection{Details of the replica calculation}
\label{app:replicas}

In this section we report all the details needed to derive our results with the replica method when $L=1$. The starting point is the assumption of joint Gaussianity of the post-activations
\begin{talign*}
\big\{\lambda^a(\btheta^a):=\frac1{\sqrt k}{\bv^{0\intercal}}\sigma(\frac1{\sqrt d}{\bW^a\bx})\big\}_{a=0}^s\,
\end{talign*}
under the randomness of $\bx$, for typical $(\btheta^a)$. The covariance of this Gaussian family is
\begin{align}
    K^{ab}:=\EE_\bx\lambda^a(\btheta^a) \lambda^b(\btheta^b)=\frac{1}{k}\sum_{i,j=1}^kv^0_iv^0_j\EE_{\bx} \sigma\Big(\frac{{\bW^{a}_i\cdot\bx}}{\sqrt{d}}\Big) \sigma\Big(\frac{{\bW^{b}_j\cdot\bx}}{\sqrt{d}}\Big)\,.
\end{align}To compute it we use Mehler's formula \eqref{eq:mehler}, plus the assumption that $\|\bW^a\|^2/d$ concentrates towards $1$, which is verified by Nishimori identities:
\begin{align}
    K^{ab}=\sum_{\ell=1}^\infty\ \frac{\mu_\ell^2}{\ell!}\frac{1}{k}\sum_{i,j=1}^kv^0_iv^0_j\big(\Omega_{ij}^{ab}\big)^\ell\,,\quad \Omega_{ij}^{ab}:=\frac{\bW_i^a\cdot\bW_j^b}{d}\,.
\end{align}
Due to norm concentrations of the $\bW_i^a$'s, one can show that for a fixed $i$ only for a small number of indices $j$ $\bW_j^b$ can have an $O(d)$ projection onto $\bW_i^a$. As confirmed by our numerics, we assume this projection is large only for one index $j$, which by permutation invariance, we can assume to be $i$ itself. In other words, we are assuming that
\begin{align}
    \label{eq:diagonal_omega}\Omega_{ii}^{ab}=O(1)\,,\quad \Omega_{ij}^{ab}=O\Big(\frac{1}{\sqrt{d}}\Big)\text{ for }i\neq j\,.
\end{align}Note that this is rigorously provable for the diagonal part of the covariance $a=b$. Then our assumptions imply
\begin{align}
    \frac{1}{k}\sum_{i\neq j}^k v_i^0v_j^0
    (\Omega_{ij}^{ab})^\ell =O(k/d^{\ell/2})=O(d^{1-\ell/2}),
\end{align}which vanishes for any $\ell\geq3$. Hence, the covariance simplifies to
\begin{align}
    K^{ab}=\mu_1^2 R_1^{ab}+\frac{\mu_2^2}{2}R_2^{ab}+
    \frac{1}{k}\sum_{i=1}^k(v_i^0)^2g(\Omega_{ii}^{ab})
    +O(d^{-1/2})
\end{align}where
\begin{equation}
\label{eq:g_func_first}
    g(x) = \sum_{\ell = 3}^\infty \frac{\mu_{\ell}^2}{\ell !} x^\ell = 
     \E_{(y,z)|x} [\sigma(y)\sigma(z)] - \mu_0^2-\mu_1^2 x - \frac{\mu_2^2}{2}x^2, \qquad (y,z) \sim \calN\left((0,0),\begin{pmatrix}1&x\\x&1\end{pmatrix} \right).
\end{equation}
Here, again by permutation symmetry, we can assume that all overlaps $\Omega_{ii}^{ab}$ for $i\in\mathcal{I}_{\mathsf{v}}$ concentrate onto the same value $\mathcal{Q}^{ab}(v)$ labelled by $\mathsf{v}$, thus leading to 
\begin{align}
    K^{ab}\approx \mu_1^2 R_1^{ab}+\frac{\mu_2^2}{2}R_2^{ab}+
    \sum_{\mathsf{v}\in\mathsf{V}} \mathsf{v}^2 \frac{|\mathcal{I}_{\mathsf{v}}|}{k} \frac{1}{|\mathcal{I}_{\mathsf{v}}|} \sum_{i\in\mathcal{I}_{\mathsf{v}}}g(\Omega_{ii}^{ab})\approx \mu_1^2 +\frac{\mu_2^2}{2}R_2^{ab}+
    \sum_{\mathsf{v}\in\mathsf{V}} {P_v} (\mathsf{v})\mathsf{v}^2g(\mathcal{Q}^{ab}(\mathsf{v}))\,.
\end{align}In the above we have used the aforementioned permutation symmetry and concentration, and the fact that $ \frac{|\mathcal{I}_{\mathsf{v}}|}{k}\to P_v(\mathsf{v})$. Furthermore, we assumed that $R_1^{ab}\to1$ which is justified by the fact that $R_1^{ab}$ is a scalar overlap between two vectors $\bS_1^a,\bS_1^b$. In fact, at the present scaling of $n,k,d$, the vector $\bS_1^0$ can be retrieved exactly.

$K^{ab}$ is what governs the ``energy'' in our model, and the overlaps therein appearing thus play the role of OPs. Recall the form of the replicated partition function:
\begin{talign*}
\EE\mathcal{Z}^s
&=\EE_{\bv^0}\int \prod_{a}^{0,s}dP_W(\bW^a) \big[\EE_{\bx}\int dy \prod_{a}^{0,s} P_{\rm out}(y\mid \lambda^a(\btheta^a))\big]^n,
\end{talign*} which, after the above simplifications, reads
\begin{equation}\label{eq:replicated_with_OP}
    \EE\mathcal{Z}^s = \int d\bR_2 d\bm{\mathcal{Q}} \exp[F_S(\bR_2,\bm{\mathcal{Q}}) + nF_E(\bR_2,\bm{\mathcal{Q}})]
\end{equation}
where $\bR_2=(R_2^{ab})$ and $\bm{\mathcal{Q}}:=\{\mathcal{Q}^{ab} \mid a\le b\}$, $\mathcal{Q}^{ab}:=\{\mathcal{Q}^{ab}(\mathsf{v})\mid \mathsf{v}\in\mathsf{V}\}$. We split the discussion in the evaluation of the energetic potential $F_E$, the entropic potential $F_S$ and finally the derivation of the saddle point equations for the OPs. The calculation is performed assuming replica symmetry, as explained below.

\subsubsection{Energetic potential}\label{app:energetic_potential}
The replicated energetic term under our Gaussian assumption on the joint law of the post-activations replicas is reported here for the reader's convenience:
\begin{equation}
    F_E = \ln \int dy\int d\blambda\frac{e^{-\frac{1}{2}\blambda^\intercal\bK^{-1}\blambda}}{\sqrt{(2\pi)^{s+1}\det \bK}} \prod_{a=0}^sP_{\rm out}(y\mid\lambda^a),\quad 
\end{equation}
with
\begin{align}
\label{eq:K_RS}
    K^{ab} &= \mu_1^2+ \frac{\mu_2^2}{2}R^{ab}_2 + \EE_{v\sim P_v} v^2 g(\mathcal{Q}^{ab}(v)).
\end{align}
The energetic term $F_E$ is already expressed as a low-dimensional integral, but it simplifies considerably under the replica symmetric (RS) ansatz and after using the Nishimori identities. Let us denote $\bm{\mathcal{Q}}(\mathsf{v})=(\mathcal{Q}^{ab}(\mathsf{v}))_{a,b=0}^s$, then, using \eqref{eq:generic_OP_RS_ansatz} and \eqref{eq:generic_OP_RS_ansatz_Nishimori}
\begin{equation*}
     \bm{\mathcal{Q}}(\mathsf{v}) = \begin{pmatrix}
        1 & \calQ(\mathsf{v}) \bm{1}_s^\intercal\\
        \calQ(\mathsf{v})\bm{1}_s & (1 - \calQ(\mathsf{v})) I_s + \calQ(\mathsf{v}) \bm{1}_s \bm{1}_s^\intercal 
    \end{pmatrix} \iff
    \hat{\bm{\mathcal{Q}}}(\mathsf{v}) = \begin{pmatrix}
        0 & -\hat\calQ(\mathsf{v}) \bm{1}_s^\intercal\\
        -\hat \calQ(\mathsf{v})\bm{1}_s &  \hat\calQ(\mathsf{v}) I_s - \hat \calQ(\mathsf{v}) \bm{1}_s \bm{1}_s^\intercal 
    \end{pmatrix},   
\end{equation*}
and similarly
\begin{equation*}
     \bR_2 = \begin{pmatrix}
        R_{\rm d} & R_2 \bm{1}_s^\intercal\\
        R_2\bm{1}_s & (R_{\rm d} - R_2) I_s + R_2 \bm{1}_s \bm{1}_s^\intercal 
    \end{pmatrix} \iff
    \hat{\bR}_2 = \begin{pmatrix}
        0 & -\hat{R}_2 \bm{1}_s^\intercal\\
        -\hat{R}_2\bm{1}_s & \hat{R}_2 I_s - \hat{R}_2 \bm{1}_s \bm{1}_s^\intercal 
    \end{pmatrix}, 
    \end{equation*}
where we reported the ansatz also for the Fourier conjugates for future convenience, though not needed for the energetic potential. We are going to use repeatedly the Fourier representation of the delta function, namely $\delta(x)  = \frac1{2\pi} \int d\hat x \exp(i\hat x x)$.
Because the integrals we will end up with will always be at some point evaluated by saddle point, implying a deformation of the integration contour in the complex plane, tracking the imaginary unit $i$ in the delta functions will be irrelevant. Similarly, the normalisation $1/2\pi$ will always contribute to sub-leading terms in the integrals at hand. Therefore, we will allow ourselves to formally write $  \delta(x)  = \int d\hat x \exp( r\hat x x)$ for a convenient constant $r$, keeping in mind these considerations (again, as we evaluate the final integrals by saddle point, the choice of $r$ ends up being irrelevant).

The RS ansatz, which is equivalent to an assumption of concentration of the OPs in the high-dimensional limit, is known to be exact when analysing Bayes-optimal inference and learning, as in the present paper, see \cite{nishimori2001statistical,barbier2021overlap,barbier2022strong}. Under the RS ansatz $\bK$ acquires a similar form:
\begin{align}
\label{eq:K_matrix_RS}
    \bK=\begin{pmatrix}
        K_{\rm d} & K \bm{1}_s^\intercal\\
        K\bm{1}_s & (K_{\rm d} - K) I_s + K \bm{1}_s \bm{1}_s^\intercal 
    \end{pmatrix}
\end{align}with 
\begin{align}
\label{eq:K-Kd-def}
    &K\equiv K(R_2,\calQ)=\mu_1^2+ \frac{\mu_2^2}{2}R_2 + \EE_{v\sim P_v} v^2 g(\mathcal{Q}(v)), \quad
    &K_{\rm d}=\mu_1^2+ \frac{\mu_2^2}{2}R_{\rm d} + g(1).
\end{align}
In the RS ansatz it is thus possible to give a convenient low-dimensional representation of the multivariate Gaussian integral of $F_E$ in terms of white Gaussian random variables:
\begin{align}
&\lambda^a=\xi\sqrt{K}+u^a\sqrt{K_{\rm d}-K}\quad \text{for }a=0,1,\dots,s,
\end{align}where $\xi,(u^a)_{a=0}^s$ are i.i.d. standard Gaussian variables. Then
\begin{align}
    F_E=\ln\int dy \,\EE_{\xi,u^0}P_{\rm out}\Big(y\mid \xi\sqrt{K}+u^0\sqrt{K_{\rm d}-K} \Big)\prod_{a=1}^s\EE_{u^a}P_{\rm out}(y\mid \xi\sqrt{K}+u^a\sqrt{K_{\rm d}-K}).
\end{align}
The last product over the replica index $a$ contains identical factors thanks to the RS ansatz. Therefore, by expanding in $s\to 0^+$ we get
\begin{align}\label{eq:energetic_pot_RS_BayesOpt}
    F_E&=s\int dy\, \EE_{\xi,u^0}P_{\rm out}(y\mid \xi\sqrt{K}+u^0\sqrt{K_{\rm d}-K})\ln \EE_{u}P_{\rm out}(y\mid \xi\sqrt{K}+u\sqrt{K_{\rm d}-K})+O(s^2)\\
    &=:s \, \phi_{P_{\rm out}}(K(R_2,\mathcal{Q});K_{\rm d})+O(s^2).\label{eq:defPsiPout}
\end{align}
Notice that the energetic contribution to the free entropy has the same form as in the generalised linear model \cite{barbier2019glm}. For our running example of linear readout with Gaussian noise 
the function $\phi_{P_{\rm out}}$ reduces to
\begin{align}
\label{eq:F_E_BayesOpt_GaussChann}
     \phi_{P_{\rm out}}(K(R_2,\mathcal{Q});K_{\rm d})=-\frac{1}{2}\ln\big[2\pi e(\Delta+K_{\rm d}-K)\big].
\end{align}

\subsubsection{Entropic potential}
\label{subsec:entropic_potential}
The entropic potential is obtained by counting the degeneracy of configurations yielding the same values of OPs appearing in $\bK$:
\begin{align}
    e^{F_S}&=\int \prod_{a=0}^sd\bS^a_2\int \prod_{a=0}^{s}dP_W(\bW^a)\delta\Big(\bS^a_2-\frac{
    \bW^{a\intercal}{\rm diag}(\bv^0)\bW^a}{\sqrt{k}}\Big)\nonumber\\
    &\times \prod_{a\leq b}^{0,s}\prod_{\mathsf{v}\in \mathsf{V}}\delta(d|\mathcal{I}_\mathsf{v}|\,\mathcal{Q}^{ab}(\mathsf{v})-\sum_{i\in\mathcal{I}_\mathsf{v}}{\bW^{a\intercal}_i\bW_i^{b}}) \prod_{a\leq b}^{0,s}\delta(d^2R_2^{ab}-{\Tr \,\bS_{2}^{a} \bS_{2}^b})\,,
\end{align}where we have introduced the integral over the Hermitian matrices $d\bS_2^a=\prod_{\alpha_1\leq \alpha_2}dS^a_{2;\alpha_1\alpha_2}$. Defining
\begin{align}
    V_W^{kd}(\bm{\mathcal{Q}}):=\int \prod_{a=0}^{s}dP_W(\bW^a)\prod_{a\leq b}^{0,s}\prod_{\mathsf{v}\in \mathsf{V}}\delta(d|\mathcal{I}_\mathsf{v}|\,\mathcal{Q}^{ab}(\mathsf{v})-\sum_{i\in\mathcal{I}_\mathsf{v}}{\bW^{a\intercal}_i\bW_i^{b}})
\end{align}the entropic potential can be conveniently recast in terms of the following conditional measure
\begin{align}
    P((\bS_2^a)\mid\bm{\mathcal{Q}} ) &=
    V_W^{kd}(\bm{\mathcal{Q}})^{-1}\int \prod_{a}^{0,s}dP_W(\bW^a) \delta(\bS^a_2-
    \bW^{a\intercal}\diag(\bv )\bW^a/\sqrt{k})
    \prod_{a\leq b}^{0,s}\prod_{\mathsf{v}\in \mathsf{V}}\delta(d|\mathcal{I}_\mathsf{v}|\,\mathcal{Q}^{ab}(\mathsf{v})-\sum_{i\in\mathcal{I}_\mathsf{v}}{\bW^{a\intercal}_i\bW_i^{b}}),
    \label{eq:truePS2_app}
\end{align}
as 
\begin{align}\label{eq:eFS}
    e^{F_S}:= 
    V_W^{kd}(\bm{\mathcal{Q}}) \int dP((\bS_2^a)\mid\bm{\mathcal{Q}}) \prod_{a\leq b}^{0,s} \delta(d^2R_2^{ab}-{\Tr \,\bS_{2}^{a} \bS_{2}^b}).
\end{align}

Recall $\mathsf{V}$ is the support of $P_v$ (assumed discrete for the moment). Recall also that we have quenched the readout weights to the ground truth. This is a measure over different replicas of the random matrices $\bS^a_2$, defined in terms of the distribution of the matrices $\bW^a$ by the first delta function in \eqref{eq:truePS2_app}, coupled through the term $\mathcal{Q}^{ab}$ in the second delta function. This coupling between replicas marks a difference with the computation of \cite{maillard2024bayes}: to proceed, we need to relax this measure to something more manageable. We thus first evaluate the exact asymptotic of its trace second moment, to eventually write a relaxation in a moment-matching scheme.

\subsubsection{Exact second moment of \texorpdfstring{$P((\bS_2^a)\mid\bm{\mathcal{Q}})$}{P(S|QW)}\label{app:second_moment_P(S2|QW)}}

In this measure, one can compute the asymptotic of its second moment
\begin{align}
    \int dP((\bS_2^a)\mid \bm{\mathcal{Q}} ) \frac{1}{d^2}\Tr\,\bS_2^a\bS_2^{b}&=
    V_W^{kd}(\bm{\mathcal{Q}})^{-1}\int \prod_{a}^{0,s}dP_W(\bW^a)\frac{1}{kd^2}\Tr[\bW^{a\intercal}\diag(\bv )\bW^{a}\bW^{b\intercal}\diag(\bv )\bW^{b}]\nonumber\\
    &\qquad\qquad \times\prod_{a\leq b}^{0,s}\prod_{\mathsf{v}\in \mathsf{V}}\delta(d|\mathcal{I}_\mathsf{v}|\,\mathcal{Q}^{ab}(\mathsf{v})-\sum_{i\in\mathcal{I}_\mathsf{v}}{\bW^{a\intercal}_i\bW_i^{b}}).
\end{align}
The measure is coupled only through the last $\delta$'s. We can decouple the measure at the cost of introducing Fourier conjugates whose values will then be fixed by a saddle point computation. The second moment computed will not affect the saddle point; hence, it is sufficient to determine the value of the Fourier conjugates through the computation of $V_W^{kd}(\bm{\mathcal{Q}})$, which rewrites as
\begin{align}
    V_W^{kd}(\bm{\mathcal{Q}})&=\int \prod_{a}^{0,s}dP_W(\bW^a)\prod_{a\leq b}^{0,s}\prod_{\mathsf{v}\in \mathsf{V}} d\hat B^{ab}(\mathsf{v})\exp\big[-\hat B^{ab}(\mathsf{v})
    (d|\mathcal{I}_\mathsf{v} |\mathcal{Q}^{ab}(\mathsf{v})-\sum_{i\in\mathcal{I}_\mathsf{v}}{\bW^{a\intercal}_i\bW_i^{b}})
    \big]\nonumber\\
    &\approx\prod_{\mathsf{v}\in \mathsf{V}}\exp\Big(d|\mathcal{I}_\mathsf{v}|\,\extr_{(\hat B^{ab}(\mathsf{v}))}\Big[-\sum_{a\le b,0}^s\hat B^{ab}(\mathsf{v}) \mathcal{Q}^{ab}(\mathsf{v})+\ln\int \prod_{a=0}^sdP_W(w_a)e^{\sum_{a\leq b,0}^s \hat B^{ab}(\mathsf{v})w_a w_b}\Big]\Big).
    \label{eq:VW_app}
\end{align}In the last line we have used saddle point integration over $\hat B^{ab}(\mathsf{v})$ and the approximate equality is up to a multiplicative $\exp(o(n))$ constant. 
From the above, it is clear that the stationary $\hat B^{ab}(\mathsf{v})$ are such that
\begin{align}
    \mathcal{Q}^{ab}(\mathsf{v})=\frac{\int \prod_{r=0}^sdP_W(w_r) w_aw_b \prod_{r\leq t,0}^s e^{ \hat B^{rt}(\mathsf{v}) w_{r}w_{t}}}{
    \int \prod_{r=0}^sdP_W(w_r) \prod_{r\leq t,0}^s e^{ \hat B^{rt}(\mathsf{v}) w_{r}w_{t}}
    }=:\langle w_aw_b\rangle_{\hat \bB(\mathsf{v})}.
\end{align}Using these notations, the asymptotic trace moment of the $\bS_2$'s at leading order becomes
\begin{align}
    \int dP((\bS_2^a)\mid \bm{\mathcal{Q}} ) &\frac{1}{d^2}\Tr\,\bS_2^a\bS_2^{b}=\frac{1}{kd^2}\sum_{i,l=1}^k\sum_{j,p=1}^d \langle W_{ij}^{a}v^0_iW_{ip}^{a} W_{lj}^{b}v^0_l
    W_{lp}^{b}\rangle_{\{\hat \bB(\mathsf{v})\}_{\mathsf{v}\in\mathsf{V}}}\nonumber\\
    &=
    \frac{1}{k}\sum_{\mathsf{v}\in\mathsf{V}} \mathsf{v}^2 \sum_{i\in\mathcal{I}_{\mathsf{v}}}\Big\langle \Big(\frac{1}{d}\sum_{j=1}^d W_{ij}^{a}W_{ij}^{b} \Big)^2\Big\rangle_{\hat \bB(\mathsf{v})}+
    \frac{1}{k} \sum_{j=1}^d \Big\langle 
    \sum_{i=1}^k \frac{v^0_i(W_{ij}^{a})^2}{d} \sum_{l\neq i,1}^k \frac{v^0_l(W_{lj}^{b})^2}{d}
    \Big\rangle_{\{\hat \bB(\mathsf{v})\}_{\mathsf{v}\in \mathsf{V}}}.
\end{align}We have used the fact that $\smash{\langle\,\cdot\,\rangle_{\hat \bB(\mathsf{v})}}$ is symmetric if the prior $P_W$ is, thus forcing us to match $j$ with $p$ if $i\neq l$. Considering that by the Nishimori identities $\mathcal{Q}^{aa}(\mathsf{v})=1$, it implies $\hat B^{aa}(\mathsf{v})=0$ for any $a=0,1,\dots,s$ and $\mathsf{v}\in\mathsf{V}$. Furthermore, the measure $\langle\,\cdot\,\rangle_{\hat \bB(\mathsf{v})}$ is completely factorised over neuron and input indices. Hence, every normalised sum can be assumed to concentrate to its expectation by the law of large numbers. Specifically, we can write that with high probability as $d,k\to \infty$,
\begin{align}
  \frac{1}{d}\sum_{i\in\mathcal{I}_\mathsf{v}}
  \sum_{j=1}^d W_{ij}^{a}W_{ij}^{b} \xrightarrow{}|\mathcal{I}_\mathsf{v}| \mathcal{Q}^{ab}(\mathsf{v}),\qquad \frac{1}{k} \sum_{\mathsf{v},\mathsf{v}'\in \mathsf{V}} \mathsf{v}\mathsf{v}' \sum_{j=1}^d  
    \sum_{i\in \mathcal{I}_{\mathsf{v}}} \frac{(W_{ij}^{a})^2}{d} \sum_{l\in \mathcal{I}_{\mathsf{v}'},l\neq i} \frac{(W_{lj}^{b})^2}{d}
    \approx \gamma  \sum_{\mathsf{v},\mathsf{v}'\in \mathsf{V}} \frac{|\mathcal{I}_{\mathsf{v}}||\mathcal{I}_{\mathsf{v}'}|}{k^2} \mathsf{v}\mathsf{v}'\to \gamma \bar v^2,
\end{align}
where we used $|\mathcal{I}_\mathsf{v}|/k\to P_v(\mathsf{v})$ as $k$ diverges. Consequently, the second moment at leading order appears as claimed:
\begin{align}
\label{eq:truePS2_second_moment}
    \int dP((\bS_2^a)\mid \bm{\mathcal{Q}}) &\frac{1}{d^2}\Tr\,\bS_2^a\bS_2^{b}=\sum_{\mathsf{v}\in\mathsf{V}}P_v(\mathsf{v}) \mathsf{v}^2 \mathcal{Q}^{ab}(\mathsf{v})^2+\gamma\bar v^2=\EE_{v\sim P_v} v^2 \mathcal{Q}^{ab}(v)^2+\gamma\bar v^2.
\end{align} 

\subsubsection{Relaxation of \texorpdfstring{$P((\bS_2^a)\mid\bm{\mathcal{Q}})$}{P(S|QW)} via maximum entropy with moment matching}
We now show how to obtain the relaxation $\tilde P((\bS_2^a)\mid \bm{\mathcal{Q}}) $ in \eqref{eq:effectivePS2}, which we report here for the reader's convenience:
\begin{align}
 \tilde P((\bS_2^a)\mid \bm{\mathcal{Q}}) := \tilde{V}_W^{-kd}(\bm{\mathcal{Q}}) \prod_a^{0,s} P_{S}(\bS_2^a) \prod_{a<b}^{0,s} e^{\frac 12 \tau(\mathcal{Q}^{ab}) \Tr\,\bS^a_2\bS^b_2}
\end{align}
where $P_{S}$ is the probability density of a generalised Wishart random matrix, i.e., of $\tilde \bW^{\intercal}{\rm diag}(\bv)\tilde\bW/\sqrt{k}$ with $\tilde \bW \in \mathbb{R}^{k\times d}$ made of i.i.d. standard \emph{Gaussian} entries, $\tilde V_W^{kd}(\bm{\mathcal{Q}})$ is the proper normalisation constant, and $\tau(\mathcal{Q}^{ab})$ is such that
\begin{align}
\int d\tilde P((\bS_2^a)\mid \bm{\mathcal{Q}}) &\frac{1}{d^2}\Tr\,\bS_2^a\bS_2^{b}=\EE_{v\sim P_v} v^2 \mathcal{Q}^{ab}(v)^2+\gamma\bar v^2\,.
\end{align}We shall see that the above equation converts into a convenient relation involving the inverse function ${\rm mmse}_S^{-1}$ when taking the replica symmetric ansatz. The effective law $\tilde P((\bS_2^a)\mid \bm{\mathcal{Q}}) $ is the least restrictive choice among the Wishart-type distributions with a trace moment fixed precisely to the one above. In more specific terms, it is the solution of the following maximum entropy problem:
\begin{align}
    \inf_{P,\tau}\Big\{D_{\rm KL}(P\,\|\,P_S^{\otimes s+1})+\sum_{a\leq b,0}^s\tau^{ab} \Big(\EE_P\frac{1}{d^2}\Tr\,\bS_2^a\bS_2^b-\gamma\bar v^2-\EE_{v\sim P_v} v^2 \mathcal{Q}^{ab}(v)^2\Big)\Big\},
\end{align}where $P_S$ is a generalised Wishart distribution (as defined above \eqref{eq:effectivePS2}), and $P$ is in the space of joint probability distributions over $s+1$ symmetric matrices of dimension $d\times d$. The rationale behind the choice of $P_S$ as a base measure is that, in absence of any other information, a statistician can always use a generalised Wishart measure for the $\bS_2$'s if they assume universality in the law of the inner weights. This ansatz would still yield a non-trivial performance, achieved by our adaptation of GAMP-RIE in App.~\ref{app:GAMP} for generic activations.

Note that if $a=b$ then, by the Nishimori identities, the second moment above matches precisely $R_{\rm d}=1+\gamma\bar v^2$. This entails directly $\tau^{aa}=0$, as the generalised Wishart prior $P_S$ already imposes this constraint.

\subsubsection{Entropic potential with the relaxed measure}\label{app:entropic_contribution}
We now use the results from the previous paragraphs to compute the entropic contribution $F_S$ to the free entropy, \eqref{eq:eFS}. Indeed, let us proceed with the relaxation of the measure $P((\bS_2^a)\mid \bm{\mathcal{Q}})$ by replacing it with $\tilde P((\bS_2^a)\mid \bm{\mathcal{Q}})$ derived above:
\begin{align}\label{eq:F_S_step1_appendix}
    e^{F_S}= 
    V_W^{kd}(\bm{\mathcal{Q}}) \int d\hat{\bR}_2 \exp\Big(-\frac{d^2}{2}\sum_{a\leq b,0}^s\hat R^{ab}_2R^{ab}_2\Big)\frac{1}{\tilde V^{kd}_W(\bm{\mathcal{Q}})}\int \prod_{a=0}^s dP_S(\bS_2^a) \exp\Big(\sum_{a\leq b,0}^s\frac{\tau_{ab}+\hat R_2^{ab}}{2}\Tr\,\bS_2^a\bS_2^b\Big)
\end{align}where we have introduced another set of Fourier conjugates $\hat\bR_2$ for $\bR_2$. 
The factor $V_W^{kd}(\bm{\mathcal{Q}})$ was already treated in \eqref{eq:VW_app}. However, here it will contribute as a tilt of the overall entropic contribution, and the Fourier conjugates $\hat{\mathcal{Q}}^{ab}(\mathsf{v})$ will appear in the final variational principle.

As usual, the Nishimori identities impose $R_2^{aa}=R_{\rm d}=1+\gamma\bar v^2$ without the need of any Fourier conjugate. Hence, similarly to $\tau^{aa}$, $\hat R_2^{aa}=0$ too. Furthermore, in the hypothesis of replica symmetry, we set  $\tau^{ab}=\tau$ and $\hat R_2^{ab}=\hat R_2$ for all $0\leq a<b\leq s$. Then, when the number of replicas $s$ tends to $0^+$, we can recognise the free entropy of a matrix denoising problem. More specifically, using the Hubbard-Stratonovich transformation (i.e.,  $\E_{\bZ} \exp(\frac d 2 \Tr\,\bM \bZ) = \exp(\frac d 4 \Tr\,\bM^2)$
for a $d\times d$ symmetric matrix $\bM$ with $\bZ$ a standard GOE matrix) we get
\begin{align}
\label{eq:J_replicas}
    J_n(\tau,\hat R_2)&:=\lim_{s\to 0^+}\frac{1}{ns}\ln\int \prod_{a=0}^s dP_S(\bS_2^a) \exp\Big(\frac{\tau+\hat R_2}{2}\sum_{a< b,0}^s\Tr\,\bS_2^a\bS_2^b\Big)\nonumber\\
    &=\frac{1}{n}\EE\ln\int dP_{\tilde S}(\tilde \bS_2)\exp\frac{d}{2}\Tr\Big(\sqrt{\tau+\hat R_2}\bY\tilde\bS_2-(\tau+\hat R_2)\frac{\tilde\bS_2^2}{2}\Big),
\end{align}where $\bY=\bY(\tau+\hat R_2)=\sqrt{\tau+\hat R_2}\tilde\bS_2^0+\bxi$ with $\bxi$ a standard GOE matrix, $\tilde\bS_2=\bS_2/\sqrt{d}$ and analgously for the ground truth matrix, and the outer expectation is w.r.t.\ $\bY$ (or $\tilde \bS^0,\bxi$). Thanks to the fact that the base measure $P_{\tilde S}$ is rotationally invariant, the above can be solved exactly in the limit $n\to\infty,\,n/d^2\to\alpha$ (see e.g. \cite{matrix_inference_Barbier}):
\begin{align}
    J(\tau,\hat R_2)=\lim  J_n(\tau,\hat R_2)=\frac{1}{\alpha}\Big(\frac{(\tau+\hat R_2)R_{\rm d}}{4}-\iota(\tau+\hat R_2)\Big), \quad \text{with} \quad \iota(x):=\frac{1}{8}+\frac{1}{2}\Sigma(\rho_{\bY(x)}).
\end{align}Here $\iota(x)=\lim I(\bY(x);\tilde \bS^0_2)/d^2$ is the limiting mutual information between data $\bY(x)$ and signal $\tilde \bS^0_2$ for the channel $\bY(x)=\sqrt{x}\tilde \bS^0_2+\bxi$, the measure $\rho_{\bY(x)}$ is the asymptotic spectral law of the observation matrix $\bY(x)$, and $\Sigma(\mu):=\int \ln|x-y|d\mu(x)d\mu(y)$. Using free probability, the law $\rho_{\bY(x)}$ can be obtained as the free convolution of a generalised Marchenko-Pastur distribution (the asymptotic spectral law of $\tilde\bS^0_2=\tilde\bW^{0\intercal}{\rm diag}(\bv^0)\tilde\bW^0/\sqrt{kd}$, which is a generalised Wishart random matrix) and the semicircular distribution (the asymptotic spectral law of $\bxi$), see \cite{potters2020first}. We provide the code to obtain this distribution numerically in the attached repository. The function ${\rm mmse}_S(x)$ is obtained through a derivative of $\iota$, using the so-called I-MMSE relation \cite{guo2005mutual,matrix_inference_Barbier}:
\begin{align}
\label{eq:def_mmse_function}
4\frac{d}{dx}\iota(x)={\rm mmse}_S(x)=\frac{1}{x}\Big(1-\frac{4\pi^2}{3}\int \mu^3_{\bY(x)}(y)dy\Big).
\end{align}
The normalisation $\frac1{ns}\ln\tilde V_W^{kd}(\bm{\mathcal{Q}})$ in the limit $n\to \infty,s\to0^+$ can be simply computed as $J(\tau,0)$.

For the other normalisation, following the same steps as in the previous section, we can simplify $V^{kd}_W(\bm\calQ)$ as follows:
\begin{align}
    \frac{1}{ns}\ln V_W^{kd}(\bm{\mathcal{Q}})\approx \frac{\gamma}{\alpha s} \sum_{\mathsf{v}\in\mathsf{V}}\frac{1}{k}
    {\rm extr}|\mathcal{I}_\mathsf{v}|\Big[-\sum_{a\leq b,0}^s\hat {\mathcal{Q}}^{ab}_{W}(\mathsf{v}) \mathcal{Q}^{ab}(\mathsf{v})+
    \ln\int \prod_{a=0}^sdP_W(w_a)e^{\sum_{a\leq b,0}^s\hat {\mathcal{Q}}^{ab}_{W}(\mathsf{v})w_aw_b}\Big],
\end{align} as $n$ grows,
where extremisation is w.r.t.\ the hatted variables only. Thanks to the Nishimori identities we have that at the saddle point $\hat{\mathcal{Q}}^{aa}(\mathsf{v})=0$ and ${\mathcal{Q}}^{aa}(\mathsf{v})=1$. This, together with standard steps and the RS ansatz, allows to write the $d\to\infty,s\to0^+$ limit of the above as
\begin{align}
    \lim_{s\to0^+}\lim\frac{1}{ns}
    \ln V_W^{kd}(\bm{\mathcal{Q}})= \frac{\gamma}{\alpha} \EE_{v\sim P_v}
    {\rm extr}\Big[-\frac{\hat{\mathcal{Q}}(v) \mathcal{Q}(v)}{2} +
    \psi_{P_W}(\hat{\mathcal{Q}}(v))\Big]
\end{align}with $\psi_{P_W}(\,\cdot\,)$ as in the main. Gathering all these results yields directly
\begin{align}
    \lim_{s\to0^+}\lim\frac{F_S}{ns}= {\rm extr}\Big\{&\frac{\hat R_2(R_{\rm d}-R_2)}{4\alpha}-\frac{1}{\alpha}\big[\iota(\tau+\hat R_2)-\iota(\tau)\big] +\frac{\gamma}{\alpha}\EE_{v\sim P_v}\Big[\psi_{P_W}(\hat{\mathcal{Q}}(v))-\frac{\hat{\mathcal{Q}}(v) \mathcal{Q}(v)}{2}\Big]
    \Big\}.
    \label{eq:app_entropic_final}
\end{align} Extremisation is w.r.t.\ $ \hat R_2,\hat{\mathcal{Q}}$. $\tau$ has to be intended as a function of $\mathcal{Q}=\{ \calQ(\mathsf{v})\mid \mathsf{v}\in\mathsf{V}\}$ through the moment matching condition:
\begin{align}
    4\alpha\, \partial_\tau J(\tau,0)=R_{\rm d}- 4\iota'(\tau)=\EE_{v\sim P_v} v^2 \mathcal{Q}(v)^2 +\gamma\bar v^2,
\end{align}which is the $s\to 0^+$ limit of the moment matching condition between $P((\bS_2^a)\mid \bm{\mathcal{Q}})$ and $\tilde P((\bS_2^a)\mid \bm{\mathcal{Q}})$. Simplifying using the value of $R_{\rm d}=1+\gamma\bar v^2$ according to the Nishimori identities, and using the I-MMSE relation between $\iota(\tau)$ and ${\rm mmse}_S(\tau)$, we get
\begin{align}
\label{eq:moment_matching}
    {\rm mmse}_S(\tau)=1-\EE_{v\sim P_v} v^2 \mathcal{Q}(v)^2\quad\iff\quad \tau={\rm mmse}_S^{-1}\big(
    1-\EE_{v\sim P_v} v^2 \mathcal{Q}(v)^2
    \big).
\end{align} Since ${\rm mmse}_S$ is a monotonic decreasing function of its argument (and thus invertible), the above always has a solution, and it is unique for a given collection $\mathcal{Q}$.

\subsubsection{RS free entropy and saddle point equations}
Putting the energetic~\eqref{eq:defPsiPout} and entropic~\eqref{eq:app_entropic_final} contributions together we obtain the variational replica symmetric free entropy potential:
\begin{align}
     f^{(1)}_{\rm RS}&:= \phi_{P_{\rm out}}(K(R_2,\mathcal{Q});K_{\rm d}) + \frac{1}{4\alpha}(1 + \gamma \bar{v}^2-R_2) \hat{R}_2 + \frac{\gamma}{\alpha}\EE_{ v\sim P_v}\big[
     \psi_{P_W}(\hat{\mathcal{Q}}(v))-\frac{1}{2}\mathcal{Q}(v) \hat{\mathcal{Q}}(v) \big]\nonumber \\
     &\qquad +\frac{1}\alpha\big[\iota(\tau(\mathcal{Q})) - \iota(\hat R_2 + \tau(\mathcal{Q}))\big],
     \label{eq:free_ent_appendix}
\end{align}
which is then extremised w.r.t. $\{\hat{\mathcal{Q}}(\mathsf{v}),\mathcal{Q}(\mathsf{v})\mid \mathsf{v}\in\mathsf{V}\}$, $\hat R_2$, $R_2$, while $\tau$ is a function of ${\mathcal{Q}}$ through the moment matching condition \eqref{eq:moment_matching}.  
The saddle point equations are then 
\begin{equation}
\label{eq:NSB_equations_generic_ch}
\begin{sqcases}
& \calQ(\mathsf{v}) = \EE_{w^0,\xi} [ w^0 \thav{w}_{\Hat\calQ(\mathsf{v})} ], \\
& P_v(\mathsf{v})\hat \calQ(\mathsf{v}) = \frac{1}{2\gamma}(R_2 - \gamma\bar v^2-\EE_{v\sim P_v}v^{2} \mathcal{Q}(v)^2)\partial_{\calQ(\mathsf{v})} \tau(\mathcal{Q}) + 2\frac{\alpha}{\gamma} \partial_{\calQ(\mathsf{v})} \phi_{P_{\rm out}}(K(R_2,\mathcal{Q});K_{\rm d}), \\
& R_2= R_{\rm d}-\frac{1}{\hat R_2 + \tau(\mathcal{Q})}(1-\frac{4\pi^2}{3}\int \mu^3_{\bY(\hat R_2 + \tau(\mathcal{Q}))}(y)dy) , \\
& \hat R_2 = 4\alpha \,\partial_{R_2} \phi_{P_{\rm out}}(K(R_2,\mathcal{Q});K_{\rm d}) ,
\end{sqcases}   
\end{equation}
where $w^0\sim P_W,\xi\sim \mathcal{N}(0,1)$ and we define the measure
\begin{align}\label{eq:bracket_FP_equations}
    \langle\,\cdot\,\rangle_{x}=\langle\,\cdot\,\rangle_{x}(w^0,\xi):=
    \frac{\int dP_W(w)(\,\cdot\,)e^{
    (\sqrt{x}\xi+x w^0)w-\frac{1}{2}xw^2
    }}{\int dP_W(w)e^{
    (\sqrt{x}\xi+xw^0)w-\frac{1}{2}xw^2
    }}   . 
\end{align}

All the above formulae are easily specialised for the linear readout with Gaussian label noise using \eqref{eq:F_E_BayesOpt_GaussChann}. We report here the saddle point equations in this case (recalling that $g$ is defined in \eqref{eq:g_func_first}):
\begin{equation}
\begin{sqcases}
\label{NSB_equations_gaussian_ch}
& \calQ(\mathsf{v}) = \EE_{w^0,\xi} [ w^0 \thav{w}_{\Hat\calQ(v)} ], \\
& \hat \calQ(\mathsf{v}) = \frac{1}{2\gamma P_v(\mathsf{v})}(R_2 -\gamma\bar v^2- \EE_{v\sim P_v}v^2 \mathcal{Q}(v)^2)\partial_{\calQ(\mathsf{v})} \tau(\mathcal{Q}) + \frac{\alpha}{\gamma} \frac{ \mathsf{v}^2 \, g'(\mathcal{Q}(\mathsf{v}))}{\Delta + \frac{1}{2}\mu_2^2(R_{\rm d}-R_2) + g(1) - \EE_{v\sim P_v} {v}^2 g(\mathcal{Q}(v))}, \\
& R_2= R_{\rm d}-\frac{1}{\hat R_2 + \tau(\calQ)}(1-\frac{4\pi^2}{3}\int \mu^3_{\bY(\hat R_2 + \tau(\mathcal{Q}))}(y)dy) ,\\
& \hat R_2 =  \frac{\alpha\mu_2^2}{\Delta +\frac{1}{2}\mu_2^2(R_{\rm d}-R_2) + g(1) - \EE_{v\sim P_v}  v^2g(\mathcal{Q}(v))}.
\end{sqcases}    
\end{equation}

If one assumes that the overlaps appearing in \eqref{eq:simple_gen_error_for_numerics} are self-averaging around the values that solve the saddle point equations (and maximise the RS potential), that is $R^{00}_1,R_1^{01}\to1$ (as assumed in this scaling), $R_2^{00}\to R_{\rm d}, R_2^{01}\to R_2^*$, and $\calQ^{00}(\mathsf{v})\to1,\calQ^{01}(\mathsf{v})\to \calQ^*(\mathsf{v})$, then the limiting Bayes-optimal mean-square generalisation error for the linear readout with Gaussian noise case appears as
\begin{equation}
\begin{aligned}
    \varepsilon^{\rm opt}-\Delta = K_{\rm d} -K^*=\frac{\mu_2^2}{2}(R_{\rm d}-R_2^*) +  g(1) - \EE_{v\sim P_v} v^2g(\mathcal{Q}^*(v)) .
\end{aligned}    \label{eq:Gaussian_output_channel_generror}
\end{equation}

\subsubsection{Non-centred activations}\label{app:non-centered}
Consider a non-centred activation function, i.e., $\mu_0\neq 0$ in \eqref{eq:sigma_hermite}. This reflects on the law of the post-activations, which will still be Gaussian, centred at
\begin{align}
    \EE_\bx\lambda^a=\frac{\mu_0}{\sqrt
    k}\sum_{i=1}^kv_i=:\mu_0
    \Lambda,
\end{align}and with the covariance given by \eqref{eq:K} (we are assuming $ \|\bW_i^a\|^2/d\to 1$). In the above, we have introduced the new mean parameter $\Lambda$. Notice that, if the $\bv $'s have a $\bar v=O(1)$ mean, then $\Lambda$ scales as $\sqrt{k}$ due to our choice of normalisation.

One can carry out the replica computation for a fixed $\Lambda$. This new parameter, being quenched, does not affect the entropic term. It will only appear in the energetic term as a shift to the means, yielding
    \begin{equation}
    F_E=F_E(\bK,\Lambda) = \ln \int dy\int d\blambda\frac{e^{-\frac{1}{2}\blambda^\intercal\bK^{-1}\blambda}}{\sqrt{(2\pi)^{s+1}\det \bK}} \prod_{a=0}^s P_{\rm{out}}(y\mid \lambda^a+\mu_0 \Lambda).
\end{equation}
Within the replica symmetric ansatz, the above turns into
\begin{align*}
    e^{F_E}=\int dy \,\EE_{\xi,u^0}P_{\rm out}\Big(y\mid \mu_0 \Lambda + \xi\sqrt{\frac{m_K^2}{K}}+u^0\sqrt{\rho_K-\frac{m_K^2}{K}}\Big)\prod_{a=1}^s\EE_{u^a}P_{\rm out}(y\mid \mu_0 \Lambda + \xi\sqrt{K}+u^a\sqrt{K_{\rm d}-K}).
\end{align*}
Therefore, the simplification of the potential $F_E$ proceeds as in the centred activation case, yielding at leading order in the number $s$ of replicas
\begin{align*}
    \frac{F_E(K_{\rm d},K,\Lambda)}{s}\!=\!\int dy\, \EE_{\xi,u^0}P_{\rm out}\Big(y\mid \mu_0 \Lambda + \xi\sqrt{K}+u^0\sqrt{K_{\rm d}-K}\Big)\ln \EE_{u}P_{\rm out}(y\mid\mu_0 \Lambda + \xi\sqrt{K}+u\sqrt{K_{\rm d}-K}) + O(s)
\end{align*}
in the Bayes-optimal setting. In the case when $P_{\rm out}(y\mid \lambda)=f(y-\lambda)$ then one can verify that the contributions due to the means, containing $\mu_0$, cancel each other. This is verified in our running example where $P_{\rm out}$ is the Gaussian channel:
\begin{equation}
    \frac{F_E(K_{\rm d},K,\Lambda)}{s} = 
    -\frac{1}{2}\ln\big[2\pi(\Delta+K_{\rm d}-K)\big]-\frac{1}{2} - \frac{\mu_0^2}{2}\frac{(\Lambda - \Lambda)^2}{\Delta+K_{\rm d}-K} + O(s)=-\frac{1}{2}\ln\big[2\pi(\Delta+K_{\rm d}-K)\big]-\frac{1}{2} + O(s).
\end{equation}

\subsection{Alternative simplifications of \texorpdfstring{$P((\bS^a_2)\mid \bm{\calQ})$}{TEXT} through moment matching}\label{app:alternatives}

A crucial step that allowed us to obtain a closed-form expression for the model's free entropy is the relaxation $\tilde{P}((\bS^a_2)\mid \bm{\calQ})$~\eqref{eq:effectivePS2} of the true measure $P((\bS^a_2)\mid \bm{\calQ})$~\eqref{eq:truePS2} entering the replicated partition function. The specific form we chose (tilted Wishart distribution with a matching second moment) has the advantage of capturing crucial features of the true measure, such as the fact that the matrices $\bS^a_2$ are generalised Wishart matrices with coupled replicas, while keeping the problem solvable with techniques derived from random matrix theory of rotationally invariant ensembles. In this appendix, we report some alternative routes one can take to simplify, or potentially improve the theory.

\subsubsection{A factorised simplified distribution}\label{app:simpler_ansatz}

In the specialisation phase, one can assume that the only crucial feature to keep track in relaxing $P((\bS^a_2)\mid\bm{\calQ})$~\eqref{eq:truePS2} is the coupling between different replicas, becoming more and more relevant as $\alpha$ increases. In this case, inspired by \cite{sakata2013,kabashima2016phase}, in order to relax~\eqref{eq:truePS2} we can propose the Gaussian ansatz
\begin{align}
    &d\bar{P}((\bS_2^a)\mid\bm{\calQ}) = \prod_{a=0}^s d \bS^a_{2}\prod_{\alpha=1}^d  \delta(S^a_{2;\alpha \alpha}  - \sqrt{k} \bar{v})\times  \prod_{\alpha_1<\alpha_2}^d  
    \frac{e^{-\frac{1}{2}\sum_{a,b=0}^s S^{a}_{2;\alpha_1\alpha_2} \bar{\tau}^{ab}(\bm{\calQ}) S^{b}_{2;\alpha_1\alpha_2}}}{\sqrt{(2\pi)^{s+1}\det(\bar{\bm{\tau}}(\bm{\calQ})^{-1})}},\label{eq:gaussPS2}
\end{align}
where $\bar{v}$ is the mean of the readout prior $P_v$, and $\bar{\bm{\tau}}(\bm{\calQ}):= (\bar{\tau}^{ab}(\bm{\calQ}))_{a,b}$ is fixed by 
\begin{equation*}
    [\bar{\bm{\tau}}(\bm{\calQ})^{-1}]_{ab} = \E_{v\sim P_v} v^2 \calQ^{ab}(v)^2.
\end{equation*}
In words, first, the diagonal elements of $\bS_2^a$ are $d$ random variables whose $O(1)$ fluctuations cannot affect the free entropy in the asymptotic regime we are considering, being too few compared to $n=\Theta(d^2)$. Hence, we assume they concentrate to their mean. Concerning the $d(d-1)/2$ off-diagonal elements of the matrices $(\bS_2^a)_a$, they are zero-mean variables whose distribution at given $\bm{\calQ}$ is assumed to be factorised over the input indices. The definition of $\bar{\bm{\tau}}(\bm{\calQ})$ ensures matching with the true second moment~\eqref{eq:truePS2_second_moment}.

\eqref{eq:gaussPS2} is considerably simpler than~\eqref{eq:effectivePS2}:
following this ansatz, the entropic contribution to the free entropy gives

\begin{align}
    &e^{\bar{F}_S}:= \int \prod_{a\le b,0}^s d \hat{R}_2^{ab}\,
     e^{kd \ln V_W(\bm{\mathcal{Q}})+\frac{d^2}{4}\Tr \hat \bR^{\intercal}_2 \bR_2}
    \Big[\int \prod_{a=0}^s dS^a_2\,\frac{e^{-\frac{1}{2}\sum_{a,b=0}^s S^{a}_2  [\bar{\tau}^{ab}(\bm{\calQ}) +\hat R_2^{ab} ]S^{b}_2}}{\sqrt{(2\pi )^{s+1}\det( \bar{\bm{\tau}}(\bm{\calQ})^{-1})}}
    \Big]^{d(d-1)/2}\nonumber\\
    &\qquad\qquad\qquad\times \int\prod_{a=0}^s\prod_{\alpha=1}^d dS^a_{2;\alpha\alpha} \delta(S^a_{2;\alpha\alpha}-\sqrt{k}\bar{v})\,e^{-\frac{1}{4}\sum_{a,b=0}^s\hat R_2^{ab}\sum_{\alpha=1}^dS_{2;\alpha\alpha}^aS_{2;\alpha\alpha}^b},
\end{align}
instead of~\eqref{eq:F_S_step1_appendix}. Integration over the diagonal elements $(S_{2;\alpha\alpha}^a)_{\alpha}$ can be done straightforwardly, yielding
\begin{align}
    e^{\bar{F}_S} &=\int \prod_{a\leq b,0}^s d\hat{R}_2^{ab} \,e^{kd\ln V_W(\bm{\mathcal{Q}}) +\frac{d^2}{4}\Tr\hat \bR_2^\intercal(\bR_2-\gamma\mathbf{1}\mathbf{1}^\intercal \bar{v}^2)}\Big[\int \prod_{a=0}^s dS^a_2\,\frac{e^{-\frac{1}{2}\sum_{a,b=0}^s S^{a}_2  [\bar{\tau}^{ab}(\bm{\calQ}) +\hat R_2^{ab} ]S^{b}_2}}{\sqrt{(2\pi )^{s+1}\det( \bar{\bm{\tau}}(\bm{\calQ})^{-1})}}
    \Big]^{d(d-1)/2}.
\end{align}
The remaining Gaussian integral over the off-diagonal elements of $\bS_2$ can be performed exactly, leading to
\begin{align}
\label{eq:exp_Z^n}
    e^{\bar{F}_S} &=\int \prod_{a\leq b,0}^s d\hat{R}_2^{ab} \,e^{kd\ln V_W(\bm{\mathcal{Q}}) +\frac{d^2}{4}\Tr\hat \bR_2^\intercal(\bR_2-\gamma\mathbf{1}\mathbf{1}^\intercal\bar{v}^2)-\frac{d(d-1)}{4}\ln\det[I_{s+1}+\hat\bR_2 \bar{\bm{\tau}}(\bm{\calQ})^{-1}]}.
\end{align}
In order to proceed and perform the $s\to 0^+$ limit, we use the RS ansatz for the overlap matrices, combined with the Nishimori identities, as explained above. The only difference w.r.t. the approach detailed in App.~\ref{app:replicas} is the determinant in the exponent of the integrand of~\eqref{eq:exp_Z^n}, which reads
\begin{align}
\ln\det[I_{s+1}+\hat\bR_2\bar{\bm{\tau}}(\bm{\calQ})^{-1}] = 
s\ln [1+\hat{R}_2 (1-\E_{v\sim P_v}v^2 \mathcal{Q}(v)^2)]-s\hat{R}_2
+O(s^2).
\end{align}
After taking the replica and high-dimensional limits, the resulting free entropy is
\begin{equation}
\begin{aligned}
\label{freeEntSpecialisation-Bayes-Opt}
    f_{\rm sp}^{(1)} = {}&
    \phi_{P_{\rm out}}(K(R_2,\calQ);K_{\rm d})
    + \frac{(1+\gamma \bar{v}^2 -R_2) \hat{R}_2}{4\alpha}
    + \frac{\gamma}{\alpha} \E_{v\sim P_v} \big[ \psi_{P_W}(\hat \calQ (v))  
    -\frac{1}{2} \mathcal{Q}(v) \hat{\calQ}(v) \big] \\
    &\qquad- \frac{1}{4\alpha}\ln \big[ 1+\hat R_2   (1 - \E_{v\sim P_v} v^2 \mathcal{Q}(v)^2) \big],
\end{aligned}    
\end{equation}
to be extremised w.r.t. $R_2, \hat{R}_2, \{\calQ (\mathsf{v}), \hat{\calQ} (\mathsf{v})\}$.
The main advantage of this expression over~\eqref{eq:free_ent_appendix} is its simplicity: the moment-matching condition fixing $\bar{\bm{\tau}}(\bm{\calQ})$ is straightforward (and has been solved explicitly in the final formula) and the result does not depend on the non-trivial (and difficult to numerically evaluate) function $\iota(x)$, which is the mutual information of the associated matrix denoising problem (which has been effectively replaced by the much simpler denoising problem of independent Gaussian variables under Gaussian noise). Moreover, one can show, in the same fashion as done in App.~\ref{app:large_alpha}, that the generalisation error predicted from this expression has the same large-$\alpha$ behaviour than the one obtained from~\eqref{eq:free_ent_appendix}. However, not surprisingly, being derived from an ansatz ignoring the Wishart-like nature of the matrices $\bS_2^a$, this expression does not reproduce the expected behaviour of the model in the universal phase, i.e. for $\alpha < \alpha_{\rm sp}(\gamma)$. 

\begin{figure}[t!!!]
    \centering
    \includegraphics[width=0.485\linewidth,clip]{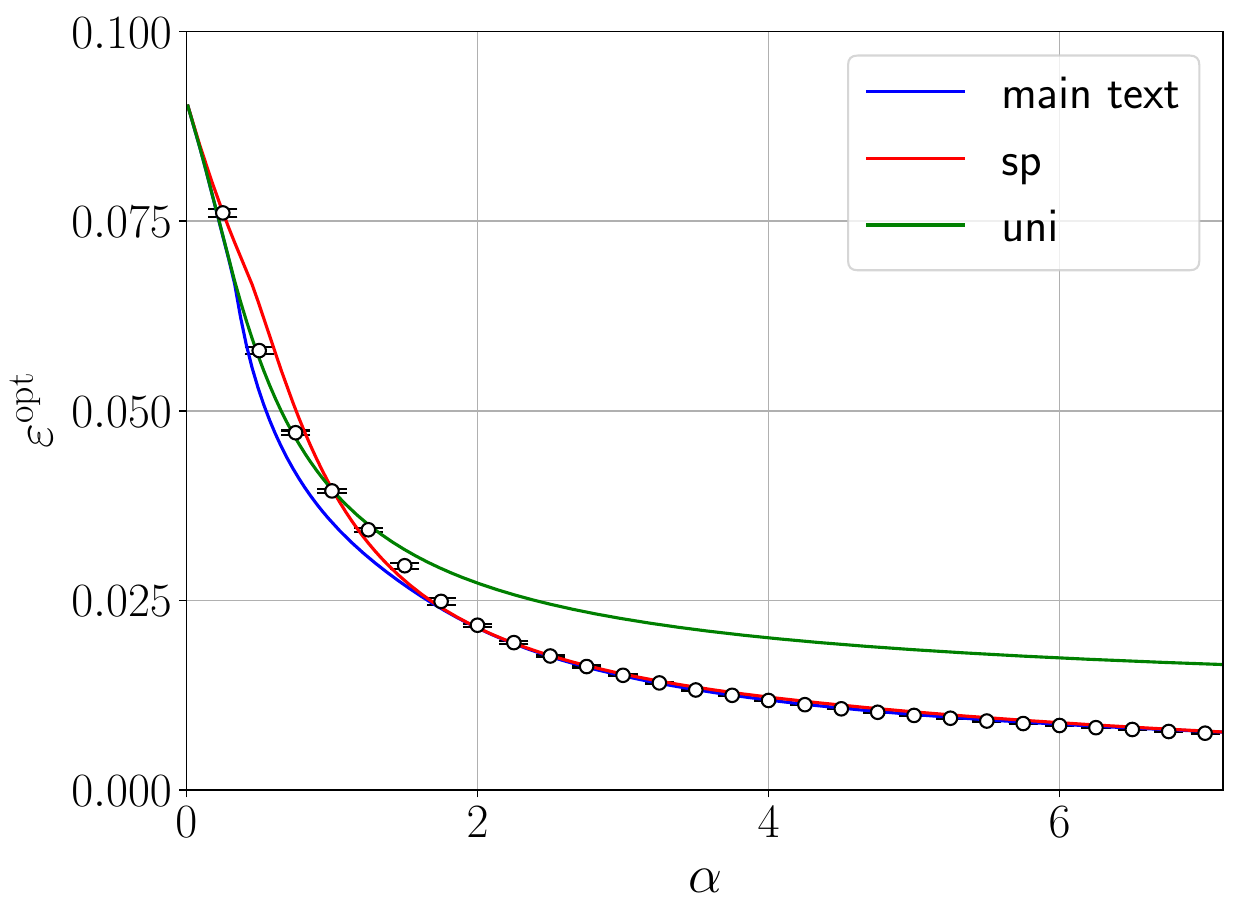}
    \includegraphics[width=0.485\linewidth,clip]{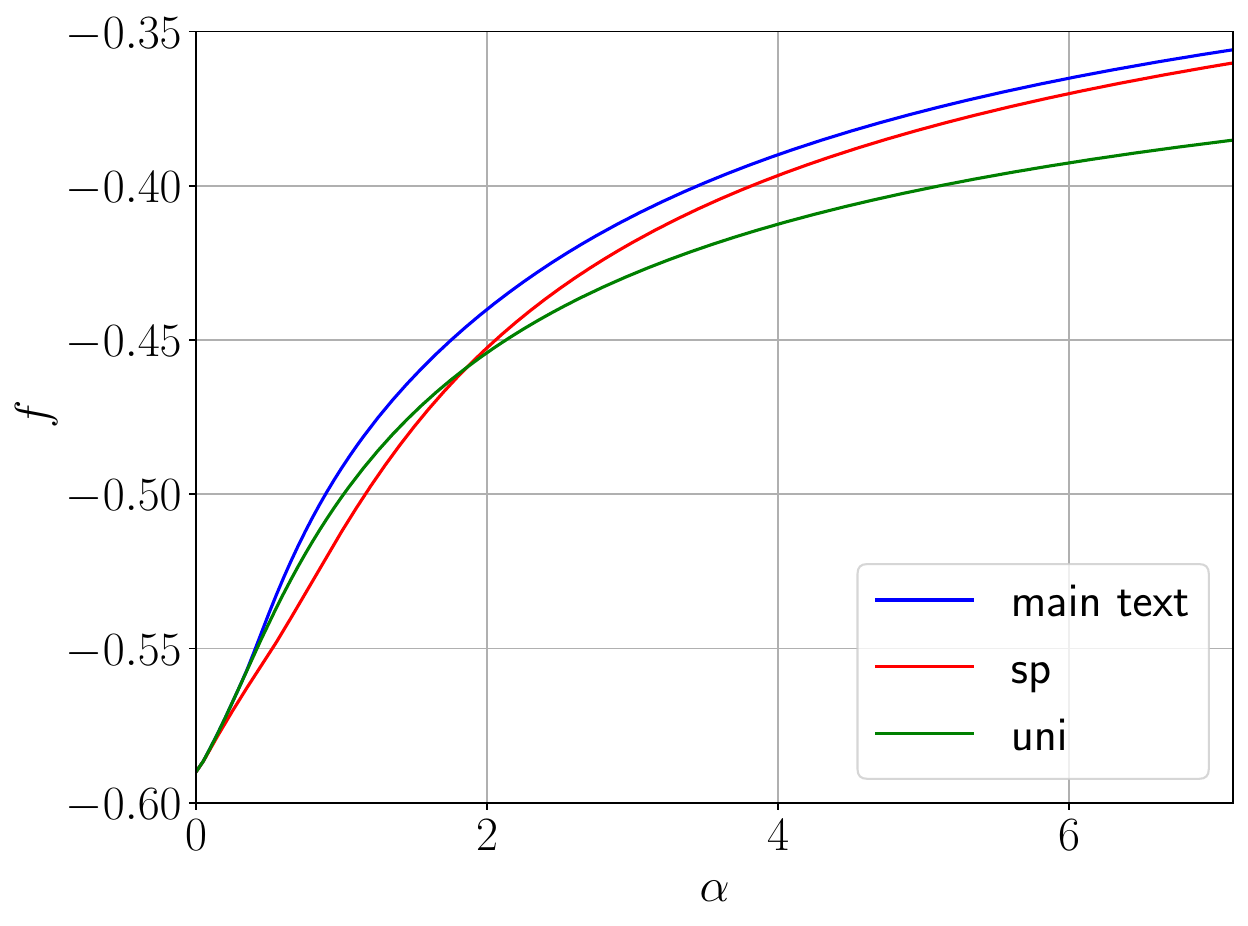}
    \includegraphics[width=0.485\linewidth,clip]{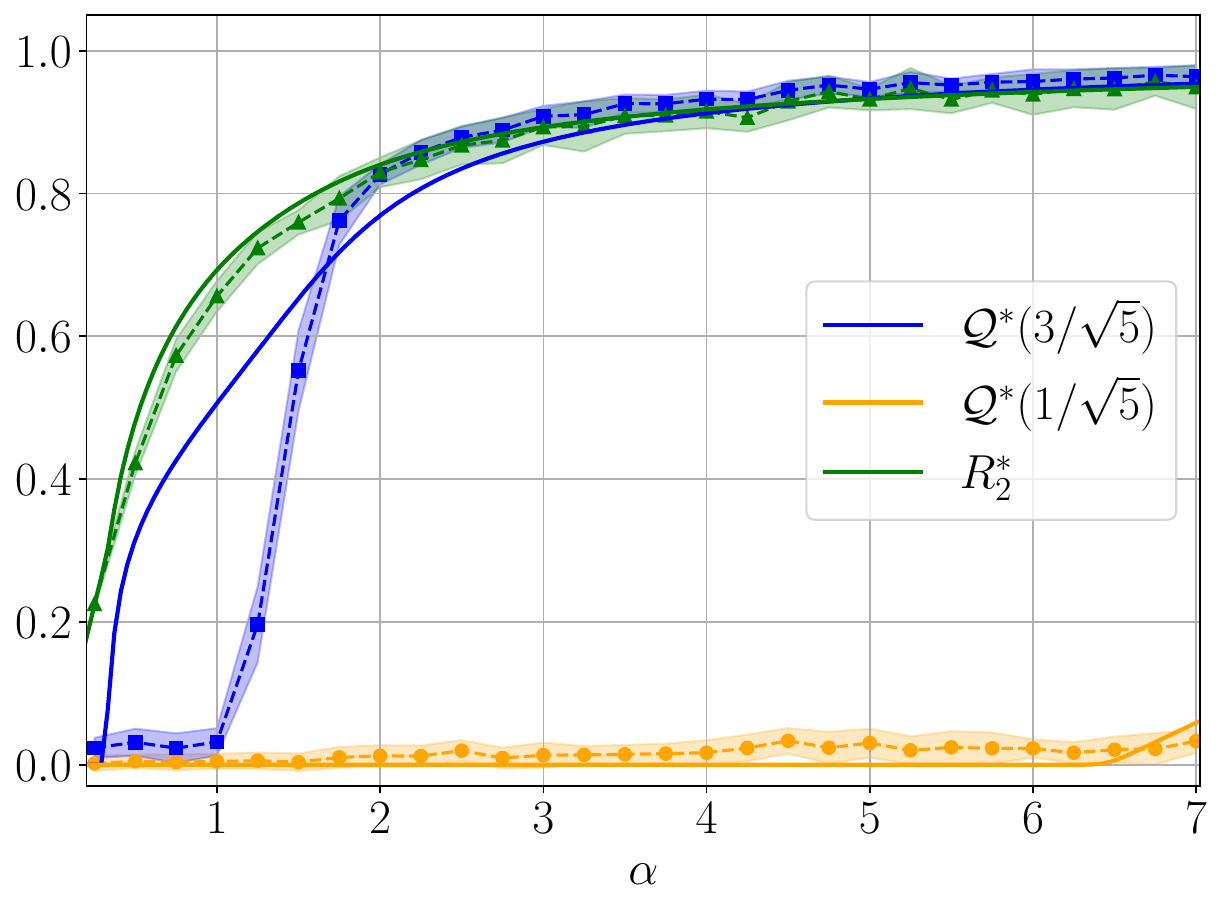}
    \includegraphics[width=0.485\linewidth,clip]{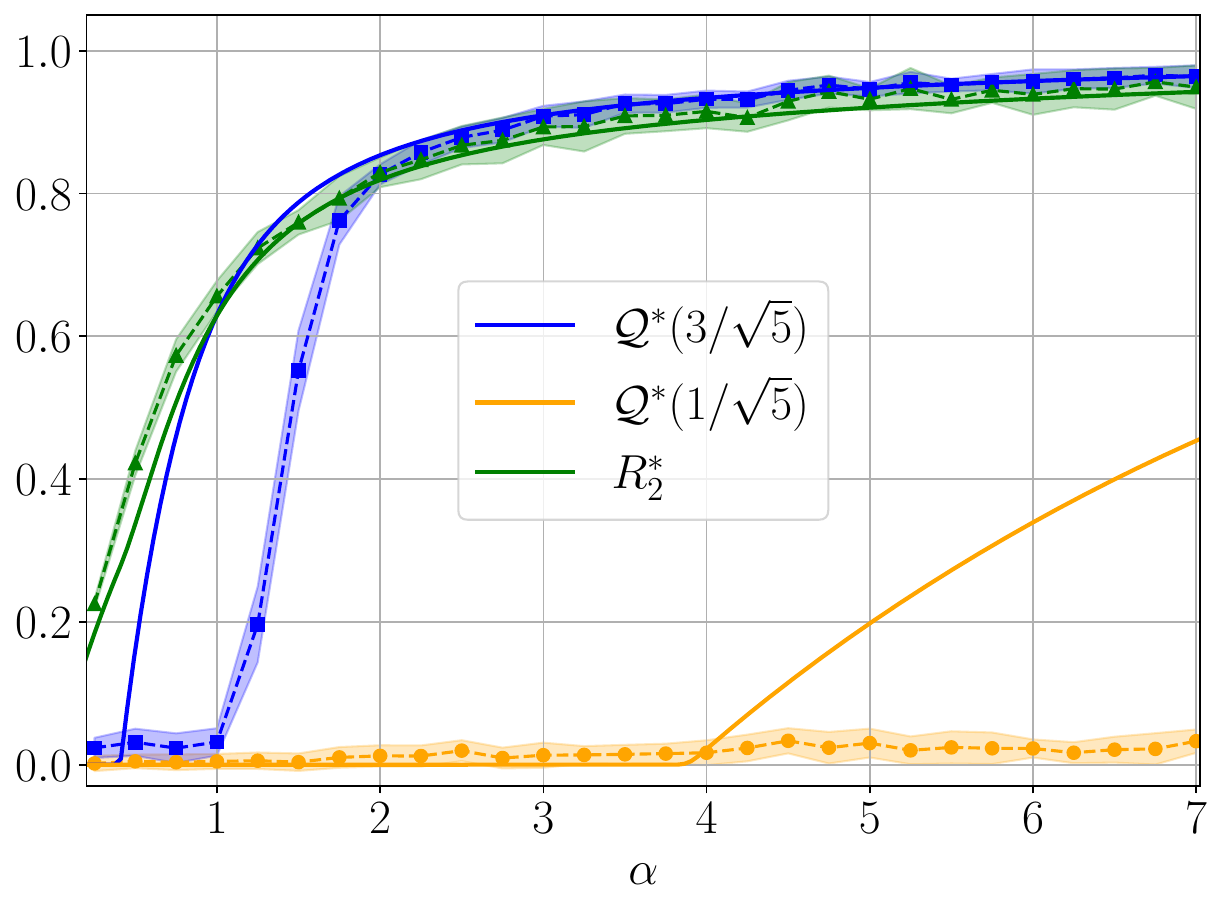}
    
    \caption{Different theoretical curves and numerical results for ${\rm ReLU}(x)$ activation, $P_v = \frac{1}{4}(\delta_{-3/\sqrt{5}} + \delta_{-1/\sqrt{5}} + \delta_{1/\sqrt{5}} + \delta_{3/\sqrt{5}})$ , $d=200$, $\gamma=0.5$, with linear readout with Gaussian noise of variance $\Delta=0.1$
    \textbf{Top left}: Optimal mean-square generalisation error predicted by the theory reported in the main text (solid blue) versus the branch obtained from the simplified ansatz~\eqref{eq:gaussPS2} (solid red); the green solid line shows the universal branch corresponding to $\mathcal{Q} \equiv 0$, and empty circles are HMC results with informative initialisation and homogeneous quenched readouts. 
    \textbf{Top right}: Theoretical free entropy curves (colors and linestyles as top left). 
    \textbf{Bottom}: Predictions for the overlaps $\mathcal{Q}(\mathsf{v})$ and $R_2$ from the theory devised in the main text (\textbf{left}) and in App.~\ref{app:simpler_ansatz} (\textbf{right}).
    }
    \label{fig:Simplified_vs_Unified}
\end{figure}

To fix this issue, one can compare the predictions of the theory derived from this ansatz, with the ones obtained by plugging $\calQ(\mathsf{v}) = 0 \ \forall \ \mathsf{v}$ (denoted $\calQ\equiv 0$) into the theory devised in the main text~\eqref{eq:fRS},
\begin{equation}
    f_{\rm uni}^{(1)} :=  \phi_{P_{\rm out}}(K(R_2,\mathcal{Q}\equiv 0);K_{\rm d}) + \frac{1}{4\alpha}(1 + \gamma \bar{v}^2-R_2) \hat{R}_2
     -\frac{1}{\alpha} \iota(\hat R_2 )  ,
\end{equation}
to be extremised now only w.r.t. the scalar parameters $R_2$, $\hat{R}_2$ (one can easily verify that, for $\calQ\equiv 0$, $\tau(\calQ) = 0$ and the extremisation w.r.t. $\hat{\calQ}$ in~\eqref{eq:fRS} gives $\hat{\calQ} \equiv 0$). Notice that $f_{\rm uni}^{(1)}$ is not depending on the prior over the inner weights, which is the reason why we are calling it ``universal''. For consistency, the two free entropies $f_{\rm sp}^{(1)}$, $ f_{\rm uni}^{(1)}$ should be compared through a discrete variational principle, that is the free entropy of the model is predicted to be
\begin{equation}
    \bar{f}^{(1)}_{\rm  RS} := \max \{\extr f_{\rm uni}^{(1)}, \extr f_{\rm sp}^{(1)}\},
    \label{eq:f_meta}
\end{equation}
instead of the unified variational form~\eqref{eq:fRS}. Quite generally, $\extr f_{\rm uni}^{(1)} > \extr f_{\rm sp}^{(1)}$ for low values of $\alpha$, so that the behaviour of the model in the universal phase is correctly predicted. The curves cross at a critical value
\begin{equation}
    \bar{\alpha}_{\rm sp} (\gamma) = \sup \{ \alpha \mid \extr f_{\rm uni}^{(1)} > \extr f_{\rm sp}^{(1)} \},
    \label{eq:alpha_c_meta}
\end{equation}
instead of the value $\alpha_{\rm sp} (\gamma)$ reported in the main. This approach has been profitably adopted in~\cite{barbier2024phase} in the context of matrix denoising
, a problem sharing some of the challenges presented in this paper. In this respect, it provides a heuristic solution that quantitatively predicts the behaviour of the model in most of its phase diagram. Moreover, for any activation $\sigma$ with a second Hermite coefficient $\mu_2=0$ (e.g., all odd activations) the ansatz \eqref{eq:gaussPS2} yields the same theory as the one devised in the main text, as in this case $K(R_2, \calQ)$ entering the energetic part of the free entropy does not depend on $R_2$, so that the extremisation selects $R_2=\hat{R}_2=0$ and the remaining parts of~\eqref{freeEntSpecialisation-Bayes-Opt} match the ones of~\eqref{eq:fRS}. Finally, \eqref{eq:gaussPS2} is consistent with the observation that specialisation never arises in the case of quadratic activation and Gaussian prior over the inner weights: in this case, one can check that the universal branch $\extr f_{\rm uni}^{(1)}$ is always higher than $\extr f_{\rm sp}^{(1)}$, and thus never selected by~\eqref{eq:f_meta}. For a convincing check on the validity of this approach, and a comparison with the theory devised in the main text and numerical results, see \figurename~\ref{fig:Simplified_vs_Unified}, top left panel.

\begin{figure}[t!!!]
    \centering
    \includegraphics[width=0.5\linewidth,clip]{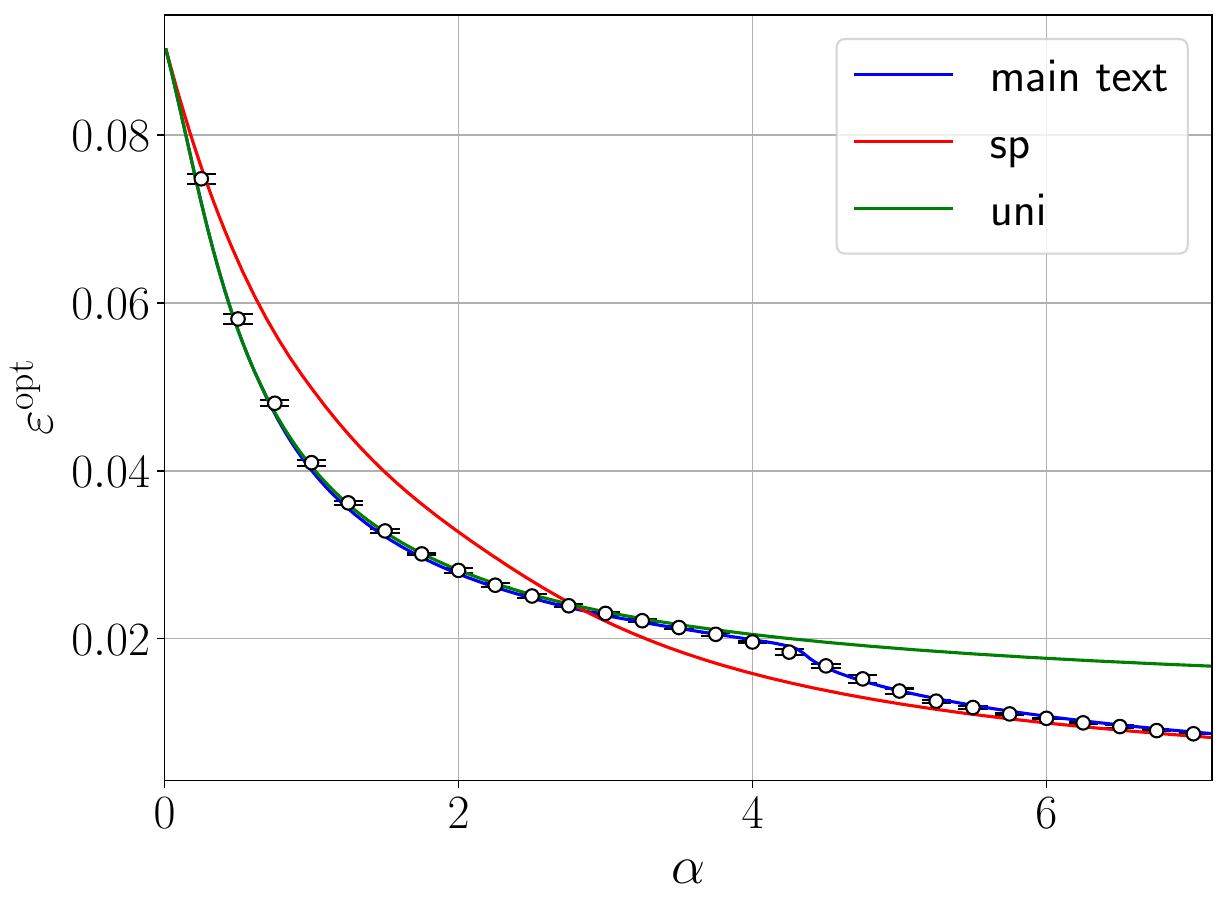}
    
    \caption{Generalisation error for ${\rm ReLU}$ activation and Rademacher readout prior $P_v$ of the theory reported in the main text (solid blue) versus the branch obtained from the simplified ansatz~\eqref{eq:gaussPS2} (solid red); the green solid line shows $\mathcal{Q} \equiv 0$ (universal branch), and empty circles are HMC results with informative initialisation and homogeneous quenched readouts. All hyperparameters are the same as in~\figurename~\ref{fig:Simplified_vs_Unified}.}
    \label{fig:Simplified_vs_Unified_2}
\end{figure}

However, despite its merits listed above, this Appendix's approach presents some issues, both from the theoretical and practical points of view:
\begin{enumerate}
    \item[$(i)$] the final free entropy of the model is obtained by comparing curves derived from completely different ans\"atze for the distribution $P((\bS^a_2)\mid\bm{\calQ})$ (Gaussian with coupled replicas, leading to $f_{\rm sp}$, vs. pure generalised Wishart with independent replicas, leading to $f_{\rm uni}$), rather than within a unified theory as in the main text;
    \item[$(ii)$] the predicted critical value $\bar{\alpha}_{\rm sp} (\gamma)$ seems to be systematically larger than the one observed in experiments (see \figurename~\ref{fig:Simplified_vs_Unified}, top right panel, and compare the crossing point of the ``sp'' and ``uni'' free entropies with the actual transition where the numerical points depart from the universal branch in the top left panel); 
    \item[$(iii)$] predictions for the functional overlap $\calQ^*$ from this approach are in much worse agreement with experimental data w.r.t. the ones from the theory presented in the main text (see \figurename~\ref{fig:Simplified_vs_Unified}, bottom panel, and compare with \figurename~\ref{fig:v_non-constant} in the main text);
    \item[$(iv)$] in the cases we tested, the prediction for the generalisation error from the theory devised in the main text are in much better agreement with numerical simulations than the one from this Appendix (see \figurename~\ref{fig:Simplified_vs_Unified_2} for a comparison).
\end{enumerate}
Therefore, the more elaborate theory presented in the main is not only more meaningful from the theoretical viewpoint, but also in overall better agreement with simulations.

\subsubsection{Possible refined analyses with structured \texorpdfstring{$\bS_2$}{S2} matrices}\label{app:structured_S2}

In the main text, we kept track of the inhomogeneous profile of the readouts induced by the non-trivial distribution $P_v$, which is ultimately responsible for the sequence of specialisation phase transitions occurring at increasing $\alpha$, thanks to a functional OP $\calQ(\mathsf{v})$ measuring how much the student's hidden weights corresponding to all the readout elements equal to $\mathsf{v}$ have aligned with the teacher's. However, when writing $\tilde{P}((\bS_2^a)\mid \bm{\calQ})$ we treated the tensor $\bS_2^a$ as a whole, without considering the possibility that its ``components'' 
\begin{align}
S_{2;\alpha_1\alpha_2}^a(\mathsf{v}) :=  \frac{\mathsf{v}}{\sqrt{| \mathcal{I}_{\mathsf{v}}|}}\sum_{i\in \mathcal{I}_{\mathsf{v}}} W^a_{i\alpha_1} W^a_{i\alpha_2}\label{eq:S2_components}    
\end{align}
could follow different laws for different $\mathsf{v}\in \mathsf{V}$. 
To do so, let us define
\begin{align}\label{eq:Q2_general_relation}
   R_2^{ab}=\frac{1}{k} \sum_{\mathsf{v},\mathsf{v}'} \mathsf{v}\,\mathsf{v}' \sum_{i\in \mathcal{I}_{\mathsf{v}}, j \in \mathcal{I}_{\mathsf{v'}}} (\Omega_{ij}^{ab})^2  = \sum_{\mathsf{v},\mathsf{v}'} \frac{\sqrt{|\mathcal{I}_{\mathsf{v}} | |\mathcal{I}_{\mathsf{v'}}|}}{k} \calQ_2^{ab}(\mathsf{v},\mathsf{v}'), \quad \text{where} \quad \calQ_2^{ab}(\mathsf{v},\mathsf{v}') := \frac1{d^2}\Tr \,\bS_2^a(\mathsf{v})\bS_2^{b}(\mathsf{v}')^\intercal.
\end{align}
The generalisation of \eqref{eq:truePS2_second_moment} then reads
\begin{align}
    \int dP((\bS_2^a)\mid \bm{\mathcal{Q}}) &\frac{1}{d^2}\Tr\,\bS_2^a(\mathsf{v})\bS_2^{b}(\mathsf{v}')^\intercal=\delta_{\mathsf{v}\mathsf{v}'} \mathsf{v}^2 \mathcal{Q}^{ab}(\mathsf{v})^2+\gamma \,\mathsf{v}\mathsf{v}' \sqrt{P_v(\mathsf{v}) P_v(\mathsf{v}')} 
    \label{eq:moment_matching_v}
\end{align}
w.r.t. the true distribution $P((\bS_2^a)\mid \bm{\mathcal{Q}})$ reported in~\eqref{eq:truePS2}. Despite the already good match of the theory in the main with the numerics, taking into account this additional level of structure thanks to a refined simplified measure could potentially lead to further improvements. The simplified measure able to enforce this moment-matching while taking into account the Wishart form \eqref{eq:S2_components} of the matrices $(\bS_2^a(\mathsf{v}))$ is
\begin{align}\label{eq:unsolvable_relaxation}
    d\bar{P}((\bS_2^a)\mid\bm{\calQ}) \propto \prod_{\mathsf{v} \in \mathsf{V}} \prod_{a} dP_S^{\mathsf{v}}(\bS_2^a(\mathsf{v}))  \times\prod_{\mathsf{v} \in \mathsf{V}} \prod_{a < b}e^{\frac{1}{2}  \bar{\tau}^{ab}_\mathsf{v}(\bm{\calQ})\Tr \bS_2^a(\mathsf{v}) \bS_2^b(\mathsf{v})},
\end{align}
where $P_S^\mathsf{v}$ is the law of a random matrix $\mathsf{v} \bar{\bW} \bar{\bW}^\intercal |\mathcal{I}_{\mathsf{v}}|^{-1/2}$ 
with $\bar\bW\in\mathbb{R}^{d\times|\mathcal{I}_{\mathsf{v}}|}$ 
having i.i.d. standard Gaussian entries. For properly 
chosen $(\bar{\tau}_\mathsf{v}^{ab})$, \eqref{eq:moment_matching_v} is verified for this simplified measure. 

However, the OPs $(\calQ_2^{ab}(\mathsf{v},\mathsf{v}'))$ are difficult to deal with if keeping a general form, as they not only imply coupled replicas $(\bS_2^a(\mathsf{v}))_a$ for a given $\mathsf{v}$ (a kind of coupling that is easily linearised with a single Hubbard-Stratonovich transformation, within the replica symmetric treatment justified in Bayes-optimal learning), but also a coupling for different values of the variable $\mathsf{v}$. Linearising it would yield a more complicated matrix model than the integral reported in~\eqref{eq:J_replicas}, because the resulting coupling field would break rotational invariance and therefore the model does not have a form which is known to be solvable, see \cite{kazakov2000solvable}.



A first idea to simplify $P((\bS^a_2)\mid\bm{\calQ})$~\eqref{eq:truePS2} while taking into account the additional structure induced by \eqref{eq:Q2_general_relation}, \eqref{eq:moment_matching_v} and keeping the model solvable, is to consider a generalisation of the relaxation \eqref{eq:gaussPS2}. This entails dropping entirely the dependences among matrix entries, induced by their Wishart-like form \eqref{eq:S2_components}, for each $\bS_2^a(\mathsf{v})$. In this case, the moment constraints \eqref{eq:moment_matching_v} can be exactly enforced by choosing the simplified measure
\begin{align}
    &d\bar{P}((\bS_2^a)\mid\bm{\calQ}) = \prod_{\mathsf{v}\in \mathsf{V}}\prod_{a=0}^s d \bS^a_{2}(\mathsf{v})\prod_{\alpha=1}^d  \delta(S^a_{2;\alpha \alpha}(\mathsf{v}) - \mathsf{v} \sqrt{|\mathcal{I}_\mathsf{v}|})\times  \prod_{\mathsf{v}\in \mathsf{V}}\prod_{\alpha_1<\alpha_2}^d  
    \frac{e^{-\frac{1}{2}\sum_{a,b=0}^s S^{a}_{2;\alpha_1\alpha_2}(\mathsf{v}) \bar{\tau}_\mathsf{v}^{ab}(\bm{\calQ}) S^{b}_{2;\alpha_1\alpha_2}(\mathsf{v})}}{\sqrt{(2\pi)^{s+1}\det(\bar{\bm{\tau}}_\mathsf{v}(\bm{\calQ})^{-1})}}.
\end{align}
The parameters $(\bar\tau^{ab}_{\mathsf{v}}(\bm{\calQ}))$ are then properly chosen to enforce \eqref{eq:moment_matching_v} for all $0\le a\le b\le s$ and $\mathsf{v}, \mathsf{v}'\in \mathsf{V}$. Using this measure, the resulting entropic term, taking into account the degeneracy of the OPs $(\calQ_2^{ab}(\mathsf{v},\mathsf{v}'))$ and $(\calQ^{ab}(\mathsf{v}))$, remains tractable through Gaussian integrals (the energetic term is obviously unchanged once we express $(R_2^{ab})$ entering it using these new OPs through the identity \eqref{eq:Q2_general_relation}, and keeping in mind that nothing changes for higher order overlaps compared to the theory in the main). We leave for future work the analysis of this Gaussian relaxation and other possible simplifications of \eqref{eq:unsolvable_relaxation} leading to solvable models.

\subsection{Large sample rate limit of \texorpdfstring{$f_{\rm RS}^{(1)}$}{fRS}}\label{app:large_alpha}
In this section we show that when the prior over the weights is discrete the MI can never exceed the entropy of the prior itself. As for the main, we consider the readouts quenched to the ground truth ones, since they cannot affect the MI between weights and data at this scaling. For this appendix we restrict to $L=1$, but the argument can be generalised to an arbitrary number of layers. 

We first need to control the function $\rm mmse$ when its argument is large. By a saddle point argument, one can show that the leading term for ${\rm mmse}_S(\tau)$ when $\tau\to\infty$ is of the type $C(\gamma)/\tau$ for a proper constant $C$ depending at most on $\gamma$. 

We now notice that the equation for $\hat{\mathcal{Q}}(v)$ in \eqref{eq:NSB_equations_generic_ch} can be rewritten as
\begin{align}
    P_v(\mathsf{v})\hat \calQ(\mathsf{v}) = \frac{1}{2\gamma}[{\rm mmse}_S(\tau)-{\rm mmse}_S(\tau+\hat R_2)]\partial_{\calQ(\mathsf{v})} \tau  + 2\frac{\alpha}{\gamma} \partial_{\calQ(\mathsf{v})} \phi_{P_{\rm out}}(K(R_2,\mathcal{Q});K_{\rm d}).
\end{align}
For $\alpha\to\infty$ we make the self-consistent ansatz $\mathcal{Q}(\mathsf{v})=1-o_\alpha(1)$. As a consequence, using the aforementioned scaling of the $\rm mmse(\tau)$, $1/\tau$ has to vanish by the moment matching condition \eqref{eq:moment_matching} as $o_\alpha(1)$ too. Using the very same equation, we are also able to evaluate $\partial_{\mathcal{Q}(\mathsf{v})}\tau$ as follows:
\begin{align}
    \partial_{\mathcal{Q}(\mathsf{v})}\tau=\frac{-2P_v(\mathsf{v})\mathsf{v}^2 \mathcal{Q}(\mathsf{v})}{{\rm mmse'}(\tau)} \approx \frac{2P_v(\mathsf{v})\mathsf{v}^2 \mathcal{Q}(\mathsf{v})\tau^2}{C(\gamma)} 
\end{align}
as $\alpha\to\infty$, where we have used ${\rm mmse}_S(\tau)\approx C(\gamma)/\tau$ to estimate the derivative. We use the same approximation for the two $\rm mmse$'s appearing in the fixed point equation for $\hat{\mathcal{Q}}(\mathsf{v})$:
\begin{align}
    \hat\calQ(\mathsf{v}) \approx \frac{\mathsf{v}^2 \mathcal{Q}(\mathsf{v})}{\gamma C(\gamma)} \frac{\tau^2}{\tau(\tau+\hat R_2)}\hat R_2 + 2\frac{\alpha}{P_v(\mathsf{v})\gamma} \partial_{\calQ(\mathsf{v})} \phi_{P_{\rm out}}(K(R_2,\mathcal{Q});K_{\rm d}),
\end{align}where we are neglecting multiplicative constants for brevity.
From the last equation in \eqref{eq:NSB_equations_generic_ch} we see that $\hat R_2$ cannot diverge more than $O(\alpha)$. Thanks to the above approximation and the first equation of \eqref{eq:NSB_equations_generic_ch} this entails that $\mathcal{Q}(\mathsf{v})$ is approaching $1$ exponentially fast in $\alpha$ ($1-\calQ(\mathsf{v})=O(e^{-c\alpha})$ for some $c>0$) due to the discreteness of its prior, which in turn implies $\tau$ is diverging exponentially in $\alpha$. As a consequence
\begin{align}
    \frac{\tau^2}{\tau(\tau+\hat R_2)}\approx 1.
\end{align}
Furthermore, one also has
\begin{align}
    \frac{1}{\alpha}[\iota(\tau)-\iota(\tau+\hat R_2)]=-\frac{1}{4\alpha}\int_\tau^{\tau+\hat R_2}  {\rm mmse}_S(t)\,dt\approx-\frac{C(\gamma)}{4\alpha}\ln (1+\frac{\hat R_2}{\tau})\xrightarrow[]{\alpha\to\infty}0,
\end{align}as $\frac{\hat R_2}{\tau}$ vanishes with exponential speed in $\alpha$.

Concerning the function $\psi_{P_W}$, given that it is related to a Bayes-optimal scalar Gaussian channel, and its SNRs $\hat{\mathcal{Q}}(\mathsf{v})$ are all diverging one can compute the integral by saddle point, which is inevitably attained at the ground truth:
\begin{align}
    \psi_{P_W}(\hat{\mathcal{Q}}(v))&-\frac{\hat{\mathcal{Q}}(v)\mathcal{Q}(v)}{2}\approx \EE_{w^0}\ln\int dP_W(w)\mathbbm{1}(w=w^0)\nonumber\\
    &+\EE\Big[
    (\sqrt{\hat{\mathcal{Q}}(v)}\xi+\hat{\mathcal{Q}}(v)w^0)w^0-\frac{\hat{\mathcal{Q}}(v)}{2}(w^0)^2
    \Big]-\frac{\hat{\mathcal{Q}}(v)(1-O(e^{-c\alpha}))}{2}=-
    H(W)+o_\alpha(1).
\end{align}
Considering that $\phi_{P_{\rm out}}(K(R_2,\mathcal{Q});K_{\rm d})\xrightarrow[]{\alpha\to\infty}\phi_{P_{\rm out}}(K_{\rm d};K_{\rm d})$, and using \eqref{eq:MI_generic_channel}, it is then straightforward to check that our RS version of the MI saturates to the entropy of the prior $P_W$ when $\alpha\to\infty$:
\begin{align}
\label{eq:large_alpha_limit}
    -\frac{\alpha}{\gamma}\text{extr} \,f_{\rm RS}^{(1)} +\frac{\alpha}{\gamma} \EE_{\lambda}  \int dy P_{\text{out}}(y|\lambda) \ln P_{\text{out}}(y|\lambda)\xrightarrow[]{\alpha\to\infty}
    H(W),
\end{align}where the factor $\alpha/\gamma$ is due to the fact that the components of $\bW$ are $kd$, and not $n$.

\subsection{Extension of GAMP-RIE to arbitrary activation for \texorpdfstring{$L=1$}{}} \label{app:GAMP}
For simplicity, let us consider $P_{\rm out}(y\mid\lambda)=\exp(-\frac1{2\Delta}(y-\lambda)^2)/\sqrt{2\pi\Delta}$, which entails:
\begin{align}
y_{\mu} \mid (\btheta^0,\bx_\mu) \overset{\rm{d}}{=} \frac{\bv^{\intercal}}{\sqrt k} \sigma \Big( \frac{\bW^0 \bx_\mu}{\sqrt d} \Big) + \sqrt{\Delta} \,z_\mu, \quad \mu =1\dots,n,
\end{align}
where $z_\mu$ are i.i.d. standard Gaussian random variables and $\overset{\rm d}{{}={}}$ means equality in law. Expanding $\sigma$ in the Hermite polynomial basis we have
\begin{align}\label{eq:hexpan}
y_\mu \mid (\btheta^0,\bx_\mu) \overset{\rm{d}}{=} \mu_0\frac{\bv^{\intercal} \bm{1}_k}{\sqrt{k}}+ \mu_1 \frac{\bv^{\intercal} \bW^0 \bx_\mu}{\sqrt{kd}} + \frac{\mu_2}{2} \frac{\bv^{\intercal}}{\sqrt k} \He_2 \Big( \frac{\bW^0 \bx_\mu}{\sqrt d} \Big) + \dots + \sqrt{\Delta} z_\mu 
\end{align}
where $\dots$ represents the terms beyond second order. 
Without loss of generality, for this choice of output channel we can set $\mu_0 = 0$ as discussed in App.~\ref{app:non-centered}. For low enough $\alpha$ it is reasonable to assume that higher order terms in $\dots$ cannot be learnt given quadratically many samples and, as a result, play the role of effective noise, which we assume independent of the first three terms. We shall see that this reasoning actually applies to the extension of the GAMP-RIE we derive, which plays the role of a ``smart'' spectral algorithm, regardless of the value of $\alpha$.
Therefore, these terms accumulate in an asymptotically Gaussian noise thanks to the central limit theorem (it is a projection of a centred function applied entry-wise to a vector with i.i.d.\ entries), with variance $g(1)$. We thus obtain the effective model 
\begin{align}
y_\mu \mid (\btheta^0,\bx_\mu) \overset{\rm{d}}{=}  \mu_1 \frac{\bv^{\intercal} \bW^0 \bx_\mu}{\sqrt{kd}} + \frac{\mu_2}{2} \frac{\bv^{\intercal}}{\sqrt k} \He_2 \Big( \frac{\bW^0 \bx_\mu}{\sqrt d} \Big) + \sqrt{\Delta + g(1) } \, z_\mu .
\end{align}
The first term in this expression can be learnt with vanishing error given quadratically many samples (Remark \ref{rem:linear}), hence it can be ignored. This consideration allows to further simplify the model to
\begin{align}
\bar y_\mu :=  y_\mu - \mu_1 \frac{\bv^{\intercal} \bW^0 \bx_\mu}{\sqrt{kd}}\overset{\rm d}{{}={}} \frac{\mu_2}{2} \frac{\bv^{\intercal}}{\sqrt k} \He_2 \Big( \frac{\bW^0 \bx_\mu}{\sqrt d} \Big) + \sqrt{\Delta + g(1) } \, z_\mu,
\end{align}
where $\bar y_\mu$ is $ y_\mu$ with the (asymptotically) perfectly learnt linear term removed, and the last equality in distribution is again conditional on $(\btheta^0,\bx_\mu)$. From the formula
\begin{align}
\frac{\bv^{\intercal}}{\sqrt{k}} \He_2 \Big( \frac{\bW^0 \bx_\mu}{\sqrt d} \Big) = \Tr \frac{\bW^{0\intercal} \diag(\bv ) \bW^0}{d\sqrt{k}} \bx_\mu \bx_\mu^\intercal - \frac{\bv^{\intercal} \bm{1}_k}{\sqrt{k}}\approx \frac{1}{\sqrt{k}d} \Tr[( \bx_\mu \bx_\mu^\intercal - I_d)\bW^{0\intercal} \diag(\bv ) \bW^0 ],
\end{align}
where $\approx$ is exploiting the concentration $\Tr \bW^{0\intercal} \diag(\bv ) \bW^0 /(d\sqrt{k}) \to \bv^{\intercal} \bm{1}_k/\sqrt{k}$,
and the Gaussian equivalence property that $\bM_\mu:=(\bx_\mu \bx_\mu^\intercal - I_d)/\sqrt{d}$ behaves like a GOE sensing matrix, i.e., a symmetric matrix whose upper triangular part has i.i.d. entries from $\mathcal{N}(0,(1+\delta_{ij})/d)$ \cite{maillard2024bayes}, the model can be seen as a GLM with signal $\bar\bS^0_2 := \bW^{0\intercal} \diag(\bv ) \bW^0/\sqrt{kd}$:
\begin{align}\label{eq:Maillard_starting_point}
    y^{\rm GLM}_\mu=\frac{\mu_2}{2}\Tr [\bM_\mu \bar\bS^0_2] +\sqrt{\Delta+g(1)}\,z_\mu.
\end{align}
Starting from this equation, the arguments of App.~\ref{app:replicas} and \cite{maillard2024bayes}, based on known results on the GLM \cite{barbier2019glm} and matrix denoising \cite{barbier2022statistical,maillard2022perturbative,matrix_inference_Barbier}, allow us to obtain the free entropy of such matrix sensing problem. The result is consistent with the $\mathcal{Q} \equiv 0$ solution of the saddle point equations obtained from the replica method in App.~\ref{app:replicas}, which, as anticipated, corresponds to the case where the Hermite polynomial combinations of the signal following the second one are not learnt.

Note that, as supported by the numerics, the model actually admits specialisation when $\alpha$ is big enough, hence the above equivalence cannot hold on the whole phase diagram at the information theoretic level. In fact, if specialisation occurs one cannot consider the $\dots$ terms in \eqref{eq:hexpan} as noise uncorrelated with the first ones, as the model is aligning with the actual teacher's weights, such that it learns all the successive terms at once.

\begin{algorithm}[t]
   \caption{GAMP-RIE for training shallow neural networks with arbitrary activation}
   \label{alg:gamp}
\begin{algorithmic}\label{algo}
    \STATE {\bfseries Input:} Fresh data point $\bx_{\text{test}}$ with unknown associated response $y_{\text{test}}$, dataset $\mathcal{D}=\{(\bx_\mu, y_\mu)\}_{\mu=1}^n$. 
    \STATE {\bfseries Output:} Estimator $\hat y_{\text{test}}$ of $y_{\text{test}}$.
    \STATE Estimate $y^{(0)} := \mu_0 \bv^{\intercal} \bm{1}/\sqrt{k}$ as 
    \begin{equation*}
        \hat y^{(0)} = \frac{1}{n}\sum_{\mu} y_\mu ;
    \end{equation*}
    \STATE  Estimate $\langle \bW^{\intercal} \bv \rangle/\sqrt k$ using (\ref{eq:mmse-S1}).
    \STATE Estimate the $\mu_1$ term in the Hermite expansion (\ref{eq:hexpan}) as
    \begin{align*}
        \hat y_\mu^{(1)} &= \mu_1 \frac{ \langle \bv^\intercal \bW \rangle \bx_\mu }{\sqrt{kd}} ;
    \end{align*}
    \STATE  Compute 
    \begin{align*}
        \tilde y_\mu &=  \frac{y_\mu - \hat y_\mu^{(0)} - \hat y_\mu^{(1)}}{\mu_2/2} ; \qquad \tilde \Delta = \frac{\Delta + g(1)}{\mu_2^2/4} ;
    \end{align*}
    \STATE  Input $\{(\bx_\mu, \tilde y_\mu)\}_{\mu=1}^n$ and $\tilde \Delta$ into Algorithm 1 in \citep{maillard2024bayes} to estimate $\langle \bW^\intercal \diag(\bv) \bW \rangle$;
    \STATE  Output 
    \begin{align} \label{eq:output_GAMP_RIE}
        \hat y_{\text{test}} = \hat y^{(0)} + \mu_1 \frac{ \langle \bv^\intercal \bW \rangle \bx_{\text{test}}}{\sqrt{kd}} + \frac{\mu_2}{2} \frac{1}{d\sqrt k} \Tr[ (\bx_{\text{test}} \bx_{\text{test}}^\intercal - \I ) \langle \bW^\intercal \diag(\bv) \bW \rangle ].
    \end{align}
\end{algorithmic}
\end{algorithm}

We now assume that this mapping holds at the algorithmic level, namely, that we can process the data algorithmically as if they were coming from the identified GLM, and thus try to infer the signal $\bar\bS_2^0 = \bW^{0\intercal} \diag(\bv ) \bW^0/\sqrt {kd}$ and construct a predictor from it. Based on this idea, we propose Algorithm~\ref{alg:gamp} that can indeed reach the performance predicted by the $\mathcal{Q} \equiv 0$ solution of our replica theory. 

\begin{remark}\label{rem:linear}
In the linear data regime, where $n/d$ converges to a fixed constant $\alpha_1$, only the first term in (\ref{eq:hexpan}) can be learnt while the rest behaves like noise. By the same argument as above, the model is equivalent to 
\begin{align}\label{eq:linear-model}
     y_\mu = \mu_1 \frac{\bv^{\intercal} \bW^0 \bx_\mu}{\sqrt{kd}} + \sqrt{\Delta + \nu - \mu_0^2 - \mu_1^2} \, z_\mu,
\end{align}
where $\nu = \EE_{z\sim\calN(0,1)}\sigma(z)^2$.
The above identifies again a GLM with signal $\bS_1^0 = \bW^{0\intercal} \bv /\sqrt k$ and Gaussian sensing vectors $\bx_{\mu}$. Define $q_1$ as the limit of $\bS_1^{a\intercal}\bS_1^b /d$ where $\bS_1^a, \bS_1^b$ are drawn independently from the posterior. With $k \rightarrow \infty$, the signal converges in law to a standard Gaussian vector. Using known results on GLMs with Gaussian signal \cite{barbier2019glm}, we obtain the following  equations characterising $q_1$:
\begin{align*}
q_1 & = \frac{\hat q_1}{\hat q_1 + 1}, \qquad
\hat q_1 = \frac{\alpha_1}{1 + \Delta_1 - q_1},\quad \text{where} \quad \Delta_1 = \frac{\Delta + \nu - \mu_0^2 - \mu_1^2}{\mu_1^2}.
\end{align*}
In the quadratic data regime, as $\alpha_1=n/d$ goes to infinity, the overlap $q_1$ converges to $1$ and the first term in (\ref{eq:hexpan}) is learnt with vanishing error.

Moreover, since $\bS_1^0$ is asymptotically Gaussian, the linear problem (\ref{eq:linear-model}) is equivalent to denoising the Gaussian vector $(\bv^{\intercal} \bW^0 \bx_\mu/\sqrt{kd})_{\mu=0}^n$ whose covariance is known as a function of $\bX = (\bx_1 , \dots, \bx_n) \in \R^{d\times n}$. This leads to the following simple MMSE estimator for $\bS_1^0$:
\begin{align}\label{eq:mmse-S1}
    \langle \bS_1^0 \rangle = \frac{1}{\sqrt{d \Delta_1}}\left(\mathbf I + \frac{1}{d \Delta_1} \bX \bX^\intercal \right)^{-1} \bX \by
\end{align}
where $\by = (y_1, \dots, y_n)$. Note that the derivation of this estimator does not assume the Gaussianity of $\bx_\mu$.
\end{remark}

\begin{remark}
The same argument can be easily generalised for general $P_{\text{out}}$, leading to the following equivalent GLM in the universal $\calQ^*\equiv 0$ phase of the quadratic data regime:
\begin{align}
    y_\mu^{\rm GLM} \sim \tilde P_{\text{out}}(\,\cdot\mid \Tr [\bM_\mu \bar\bS^0_2] ), \quad \text{where} \quad  \tilde P_{\text{out}}(y|x) := \mathbb E_{z \sim \mathcal N(0,1)} P_{\text{out}}\Big(y \mid  \frac{\mu_2}{2} x + z\sqrt{g(1)} \Big),
\end{align}
and $\bM_\mu$ are independent GOE sensing matrices.
\end{remark}

\begin{remark}
One can show that the system of equations in \eqref{NSB_equations_gaussian_ch} with $\mathcal{Q}(\mathsf{v})$ all set to $0$ (and consequently $\tau=0$) can be mapped onto the fixed point of the state evolution equations (92), (94) of the GAMP-RIE in \cite{maillard2024bayes} up to changes of variables. This confirms that when such a system has a unique solution, which is the case in all our tests, the GAMP-RIE asymptotically matches our universal solution. Assuming the validity of the aforementioned effective GLM, a potential improvement for discrete weights could come from a generalisation of GAMP which, in the denoising step, would correctly exploit the discrete prior over inner weights rather than using the RIE (which is prior independent). However, the results of \cite{barbier2024phase} suggest that optimally denoising matrices with discrete entries is hard, and the RIE is the best efficient procedure to do so. Consequently, we tend to believe that improving GAMP-RIE in the case of discrete weights is out of reach without strong side information about the teacher, or exploiting non-polynomial-time algorithms (see App.~\ref{app:hardness}).
\end{remark}

\subsection{Algorithmic complexity of finding the specialisation solution for \texorpdfstring{$L=1$}{} \label{app:hardness}}

We now provide empirical evidence concerning the computational complexity to attain specialisation, namely to have one of the $\mathcal{Q}(\mathsf{v})>0$, or equivalently to beat the ``universal'' performance ($\mathcal{Q}(\mathsf{v})=0$ for all $\mathsf{v}\in\mathsf{V}$) in terms of generalisation error. We tested two algorithms that can find it in affordable computational time: ADAM with optimised batch size for every dimension tested (the learning rate is automatically tuned), and Hamiltonian Monte Carlo (HMC), both trying to infer a two-layer teacher network with Gaussian inner weights. Both algorithms were tested with readout weights frozen to the teacher ones. We will later on discuss the case of learnable readouts.

\begin{figure}[htp!]
\centering
\centerline{
\includegraphics[width=.49\linewidth,trim={0 0 0 0},clip]{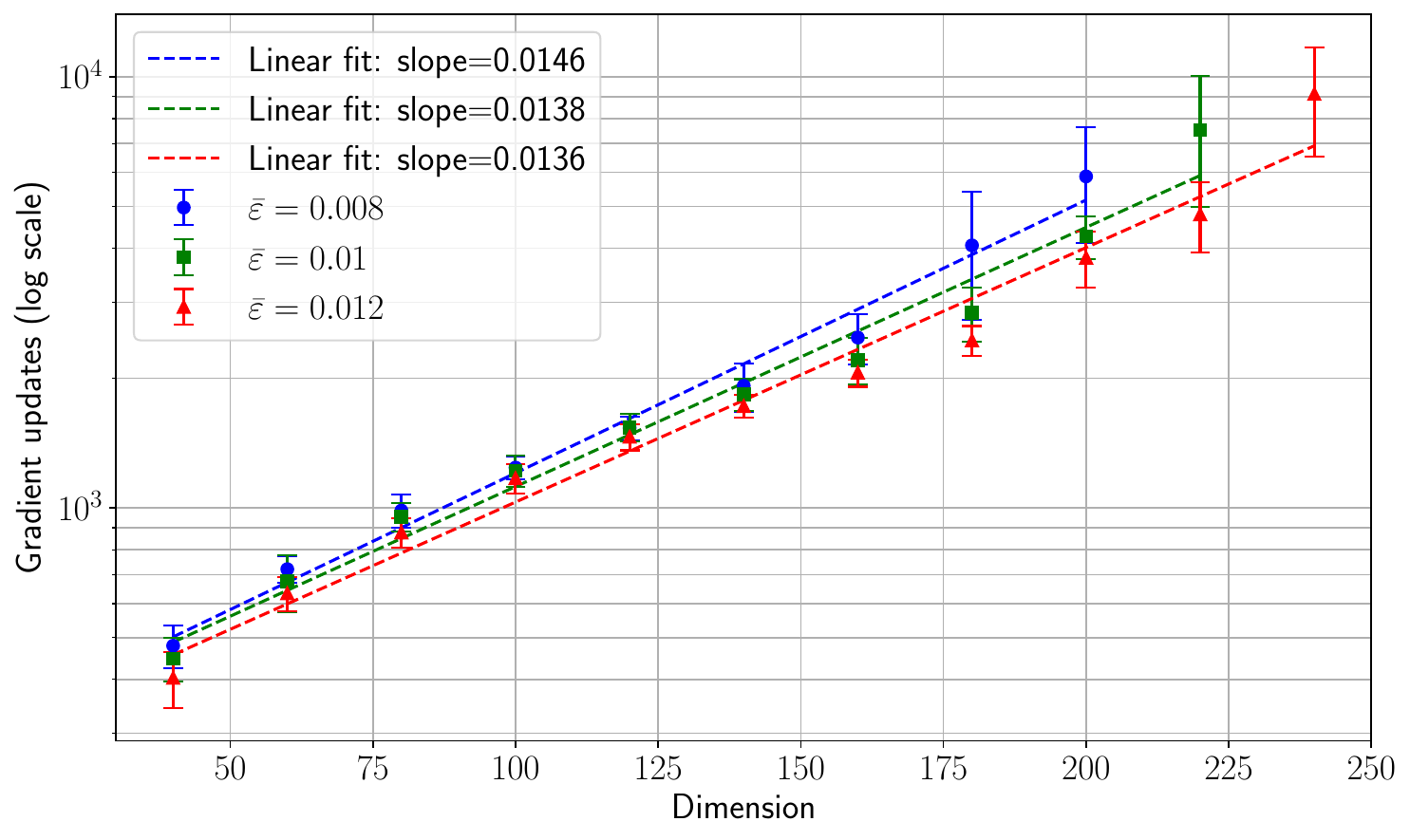}
\includegraphics[width=.49\linewidth,trim={0 0 0 0},clip]{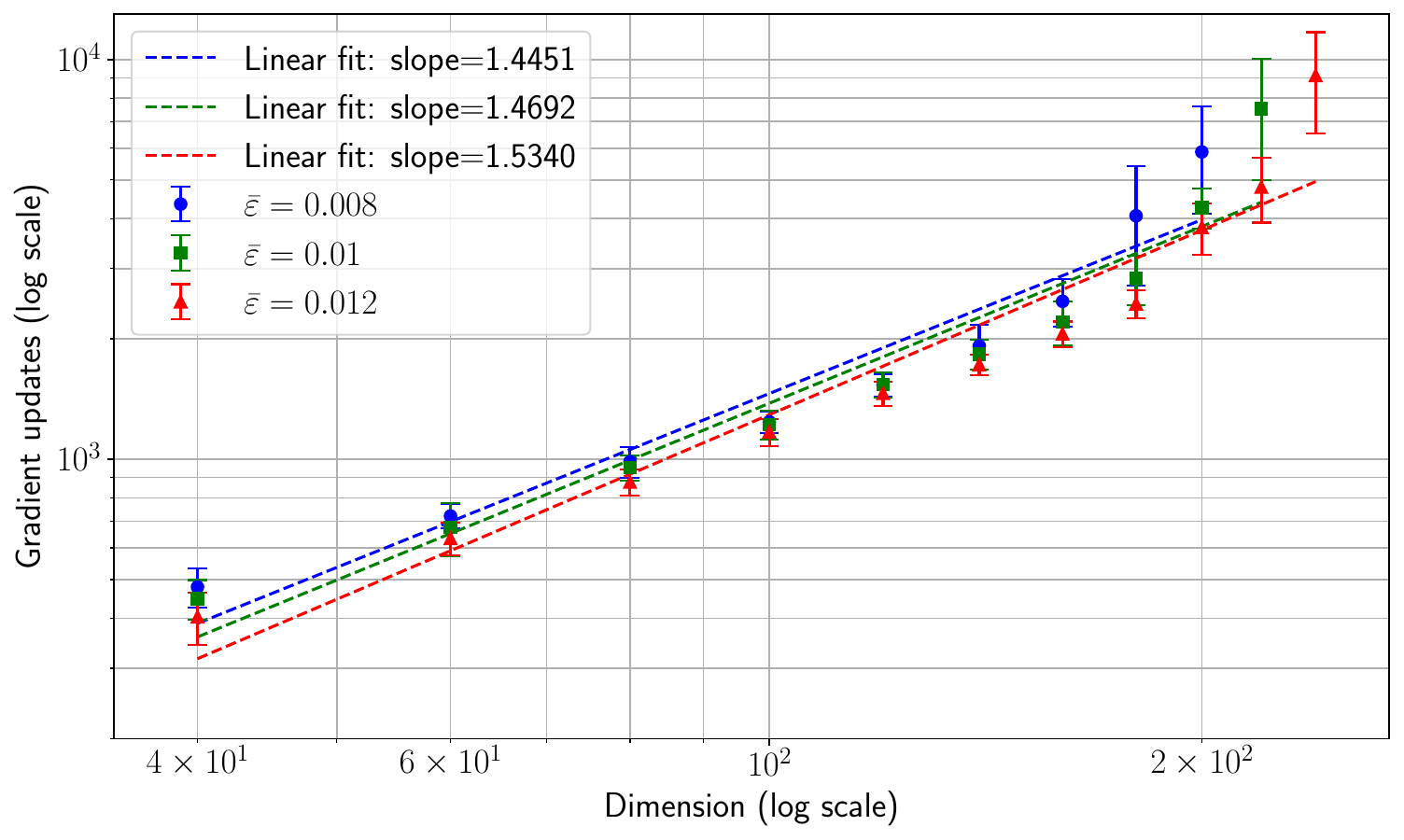}
}
\vspace{1pt} 
\centerline{
\includegraphics[width=.49\linewidth,trim={0 0 0 0},clip]{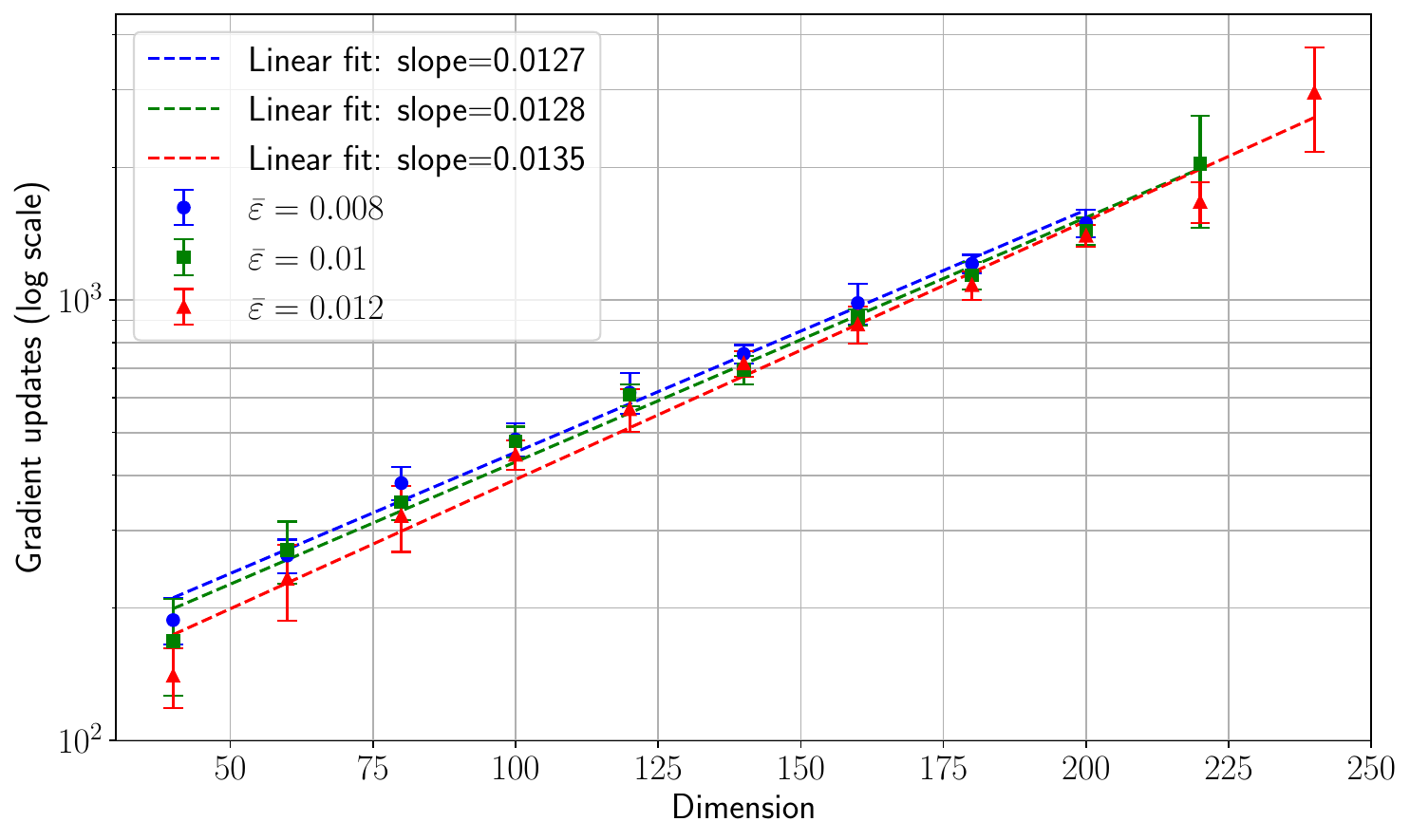}
\includegraphics[width=.49\linewidth,trim={0 0 0 0},clip]{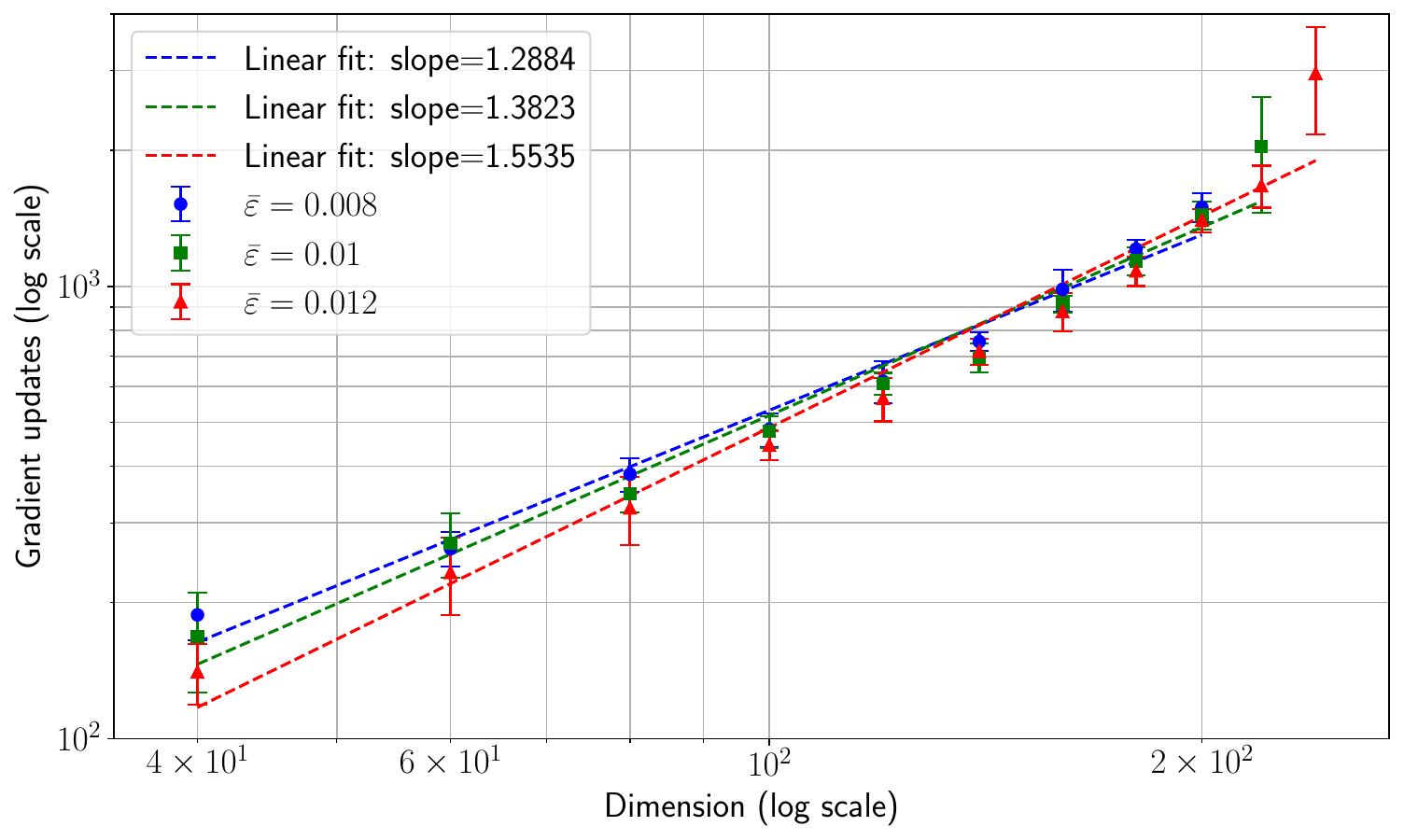}
}
\vspace{1pt} 
\centerline{
\includegraphics[width=.49\linewidth,trim={0 0 0 0},clip]{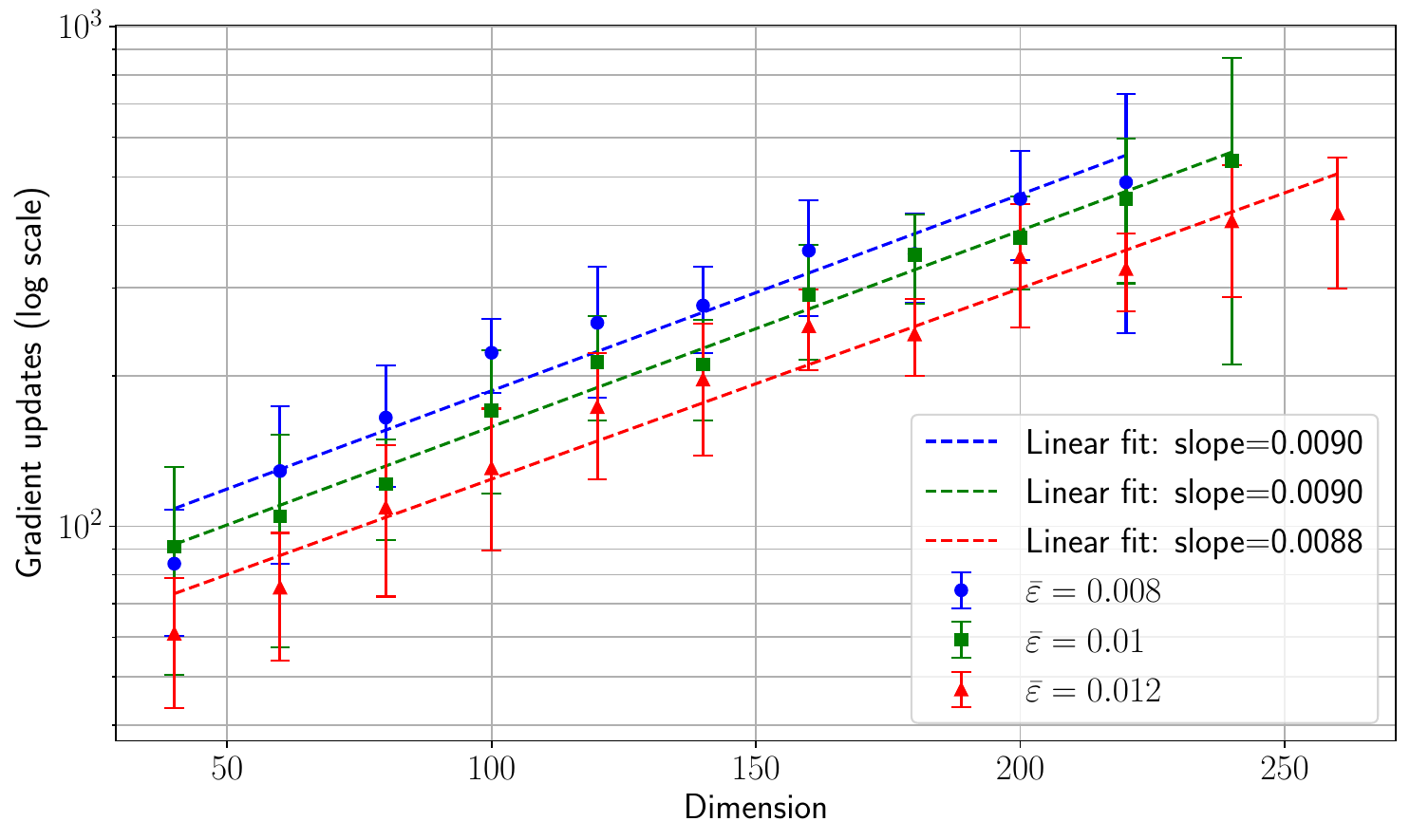}
\includegraphics[width=.49\linewidth,trim={0 0 0 0},clip]{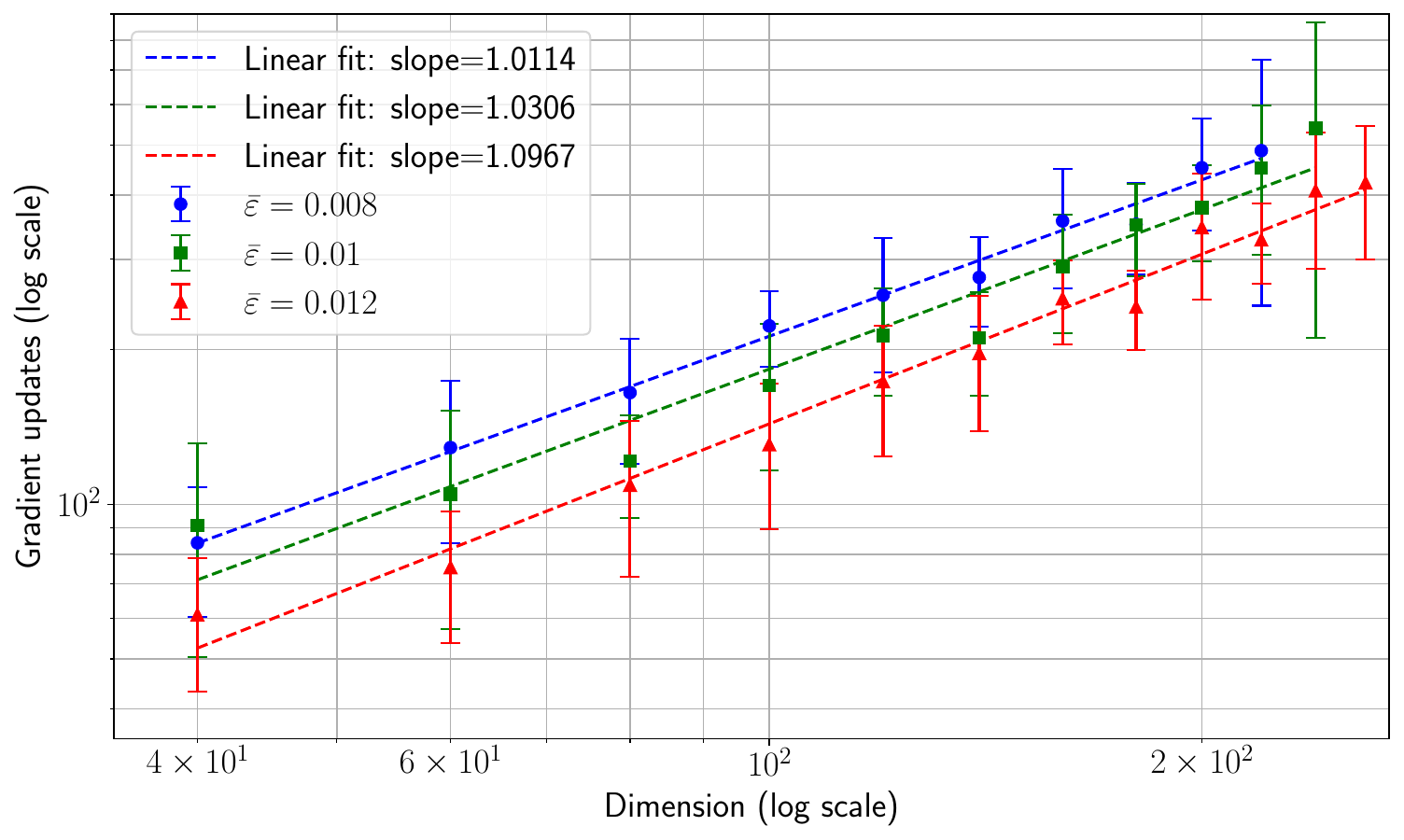}
}
\vspace{-10pt}
    \captionof{figure}{
    Semilog (\textbf{Left}) and log-log (\textbf{Right}) plots of the number of gradient updates needed to achieve a test loss below the threshold $\bar{\varepsilon}< \varepsilon^{\rm uni}$. Student network trained with ADAM with optimised batch size for each point. The dataset was generated from a teacher network with ${\rm ReLU}(x)$ activation and parameters $\Delta = 10^{-4}$ for the Gaussian noise variance of the linear readout, $\gamma=0.5$ and $\alpha=5.0$ for which $\varepsilon^{\rm opt}-\Delta=1.115 \times 10^{-5}$.
     Points are obtained averaging over 10 teacher/data instances with error bars representing the standard deviation.
    Each row corresponds to a different distribution of the readouts, kept fixed during training. \textbf{Top}: homogeneous readouts, for which the error of the universal branch is $\varepsilon^{\rm uni}-\Delta=1.217\times 10^{-2}$. \textbf{Centre}: Rademacher readouts, for which $\varepsilon^{\rm uni}-\Delta=1.218\times 10^{-2}$.
    \textbf{Bottom}: Gaussian readouts, for which $\varepsilon^{\rm uni}-\Delta=1.210\times 10^{-2}$. The quality of the fits can be read from Table~\ref{tab:adam}.
}
    \label{fig:hardness_adam}
\vskip .1in

    \begin{tabular}{lc|c|c|c|c|c|c|}
            &        & \multicolumn{3}{c|}{$\chi^2$ exponential fit} & \multicolumn{3}{c|}{$\chi^2$ power law fit}\\
        \hline
        Readouts    &$\bar{\varepsilon}=$& $0.008$ & $0.010$ & $0.012$ & $0.008$ & $0.010$ & $0.012$ \\
        \hline
         Homogeneous&&$\bm{5.57}$ & $\bm{9.00}$ & $\bm{21.1}$ &$32.3$&$26.5$&$61.1$ \\
         Rademacher&& $\bm{4.51}$ & $\bm{6.84}$ & $\bm{12.7}$ &$12.0$&$17.4$ &$16.0$\\
         Uniform $[-\sqrt{3},\sqrt{3}]$&& $\bm{5.08}$ & $\bm{1.44}$ & ${4.21}$ &$8.26$ &$8.57$ &$\bm{3.82}$\\
         Gaussian&&$2.66$&$\bm{0.76}$&$3.02$ & $\bm{0.55}$&$2.31$ &$\bm{1.36}$\\
    \end{tabular}
    \captionof{table}{$\chi^2$ test for exponential and power-law fits for the time needed by ADAM to reach the thresholds $\bar{\varepsilon}$, for various priors on the readouts. Fits are displayed in \figurename~\ref{fig:hardness_adam}. Smaller values of $\chi^2$ (in bold, for given threshold and readouts) indicate a better compatibility with the hypothesis.}
    \label{tab:adam}
\end{figure}

\paragraph{ADAM}
We focus on ${\rm ReLU}(x)$ activation, with $\gamma=0.5$, Gaussian output channel with low label noise ($\Delta=10^{-4}$) and $\alpha=5.0>\alpha_{\rm sp}$ ($= 0.22,0.12,0.02$ for homogeneous, Rademacher and Gaussian readouts respectively, thus we are deep in the specialisation phase in all the cases we report), so that the specialisation solution exhibits a very low generalisation error. We test the learnt model at each gradient update measuring the generalisation error with a moving average of 10 steps to smoothen the curves. Let us define $\varepsilon^{\rm uni}$ as the generalisation error associated with the overlap $\mathcal{Q} \equiv 0$, then fixing a threshold $ \varepsilon^{\rm opt} < \bar{\varepsilon} < \varepsilon^{\rm uni}$, we define $\bar{t}(d)$ the time (in gradient updates) needed for the algorithm to cross the threshold for the first time. We optimise over different batch sizes $B_p$ as follows: we define them as $B_p = \left\lfloor \frac{n}{2^p} \right\rfloor,\quad p = 2,3,\dots,\lfloor \log_2(n) \rfloor-1$. Then for each batch size, the student network is trained until the moving average of the test loss drops below $\bar{\varepsilon}$ and thus outperforms the universal solution; we have checked that in such a scenario, the student ultimately gets close to the performance of the specialisation solution. The batch size that requires the fewest gradient updates is selected. We used the ADAM routine implemented in PyTorch.

\begin{figure}[hb!]
    \centering
    \includegraphics[width=0.45\linewidth]{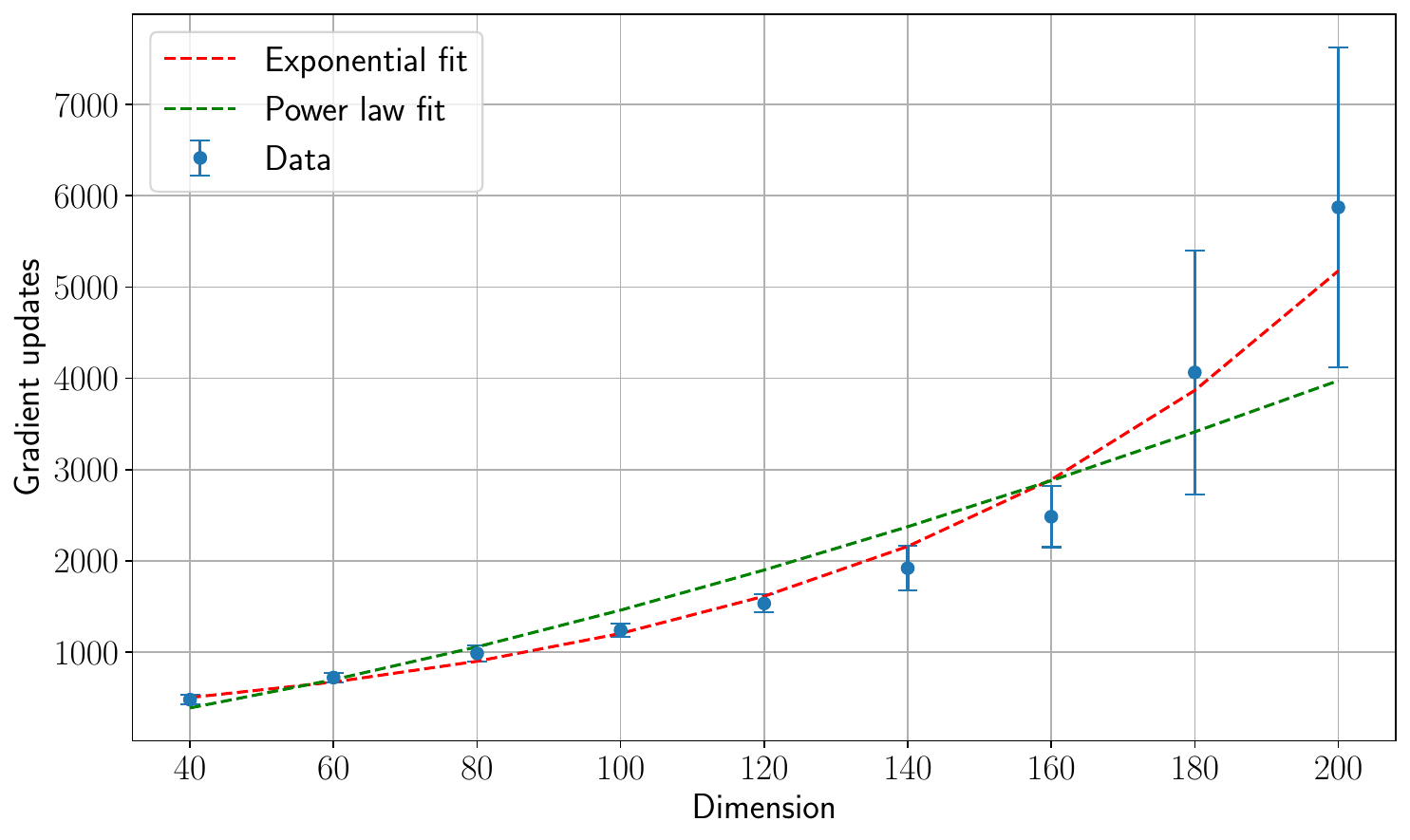}
    \hspace{2ex}
    \includegraphics[width=0.45\linewidth]{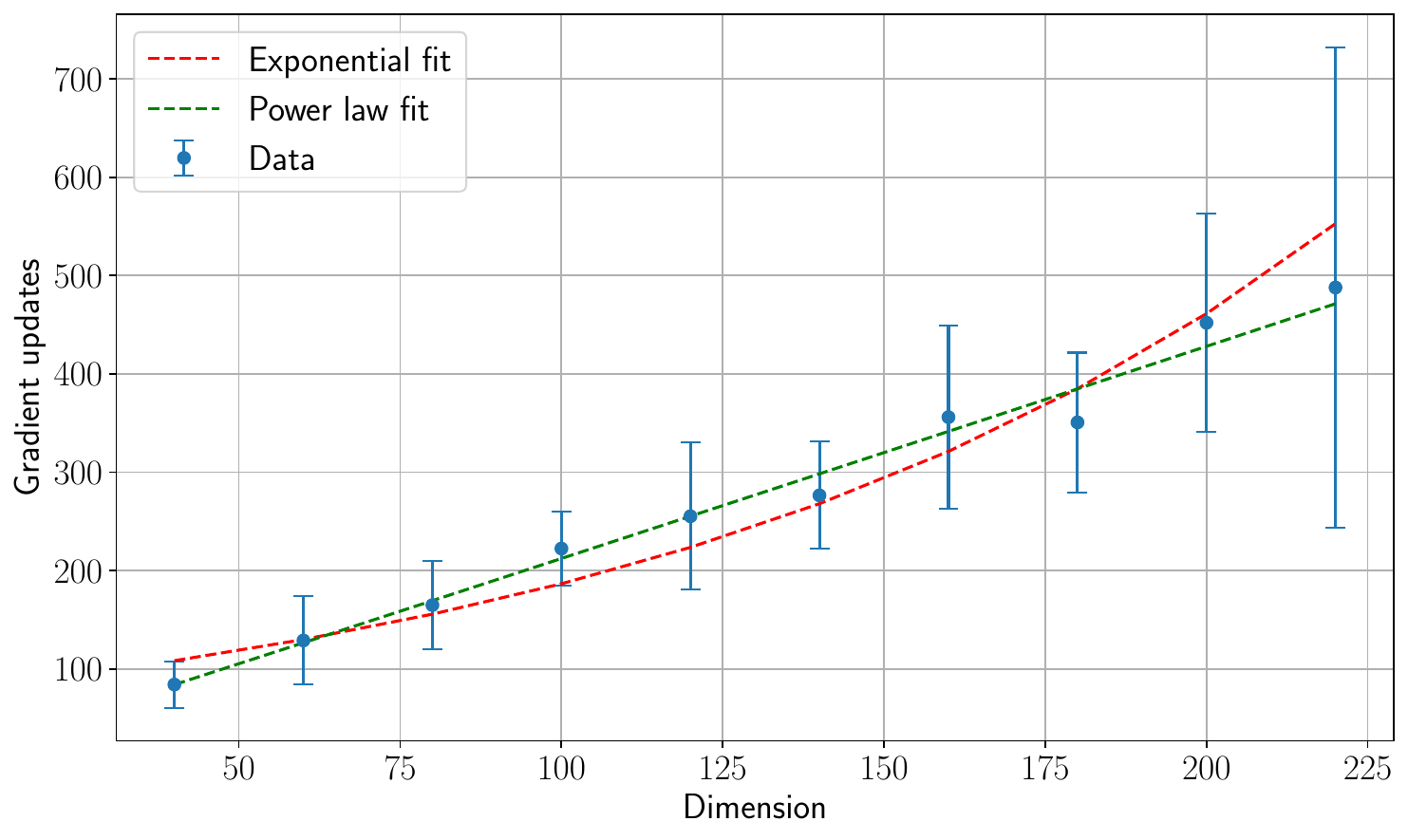}
    \caption{Same as in \figurename~\ref{fig:hardness_adam}, but in linear scale for better visualisation, for homogeneous readouts (\textbf{Left}) and Gaussian readouts (\textbf{Right}), with threshold $\bar{\varepsilon}=0.008$.}
    \label{fig:hardness_adam_linear}
\end{figure}

We test different distributions for the readout weights (kept fixed to $\bv $ during training of the inner weights). We report all the values of $\bar{t}(d)$ in \figurename~\ref{fig:hardness_adam} and \ref{fig:hardness_adam_linear} for various dimensions $d$ at fixed $(\alpha,\gamma)$, providing an exponential fit $\bar{t}(d) = \exp(a d + b)$ (left panel) and a power-law fit $\bar{t}(d) = a d^b $ (right panel). We report the $\chi^2$ test for the fits in Table~\ref{tab:adam}.
We observe that for homogeneous and Rademacher readouts, the exponential fit is more compatible with the experiments, while for Gaussian readouts the comparison is inconclusive.

\begin{figure}[hb!]
    \centering
    \includegraphics[width=0.34\linewidth,trim={0 0 0 0},clip]{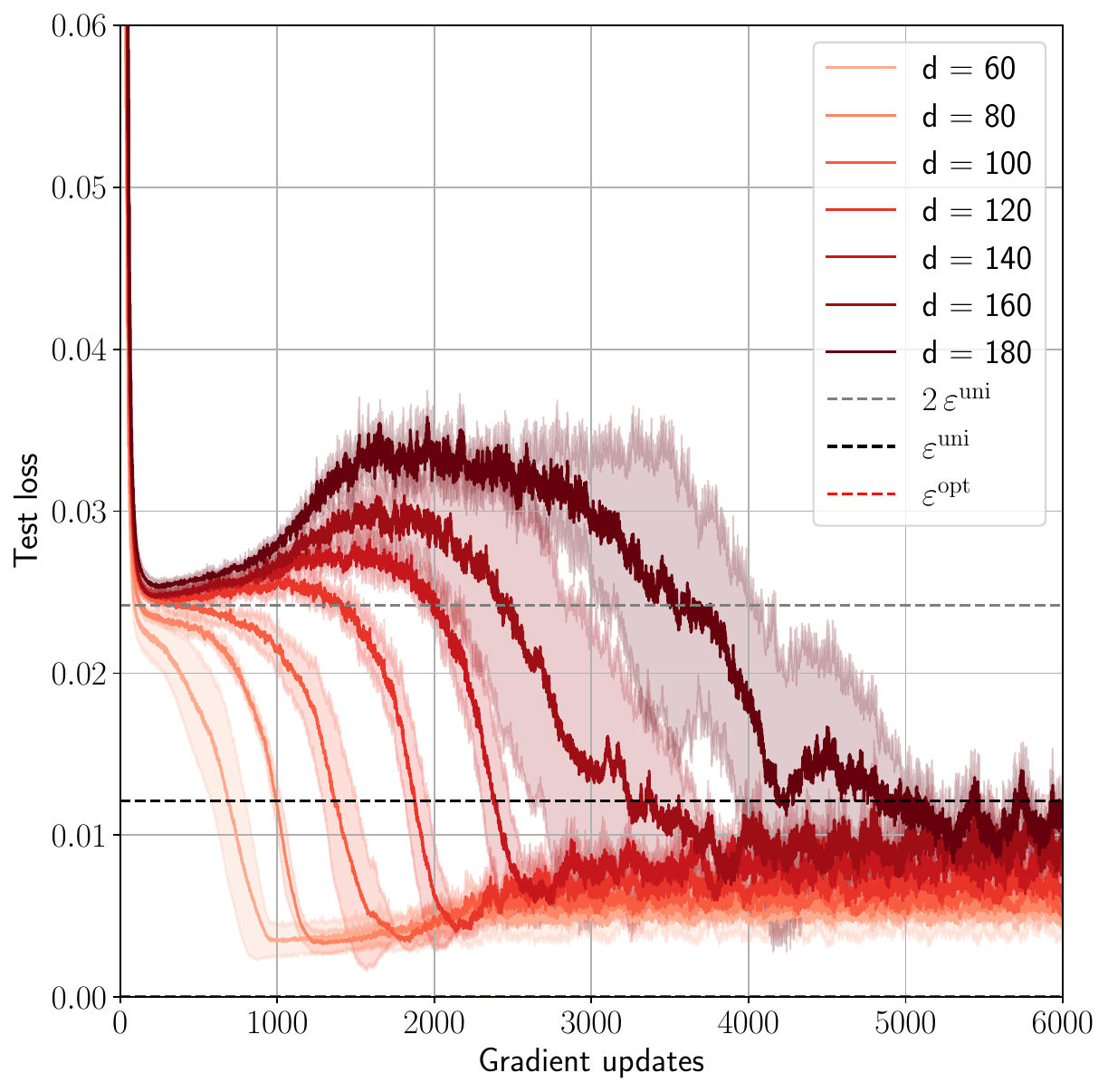}
    \includegraphics[width=0.325\linewidth,trim={0.9cm 0 0 0},clip]{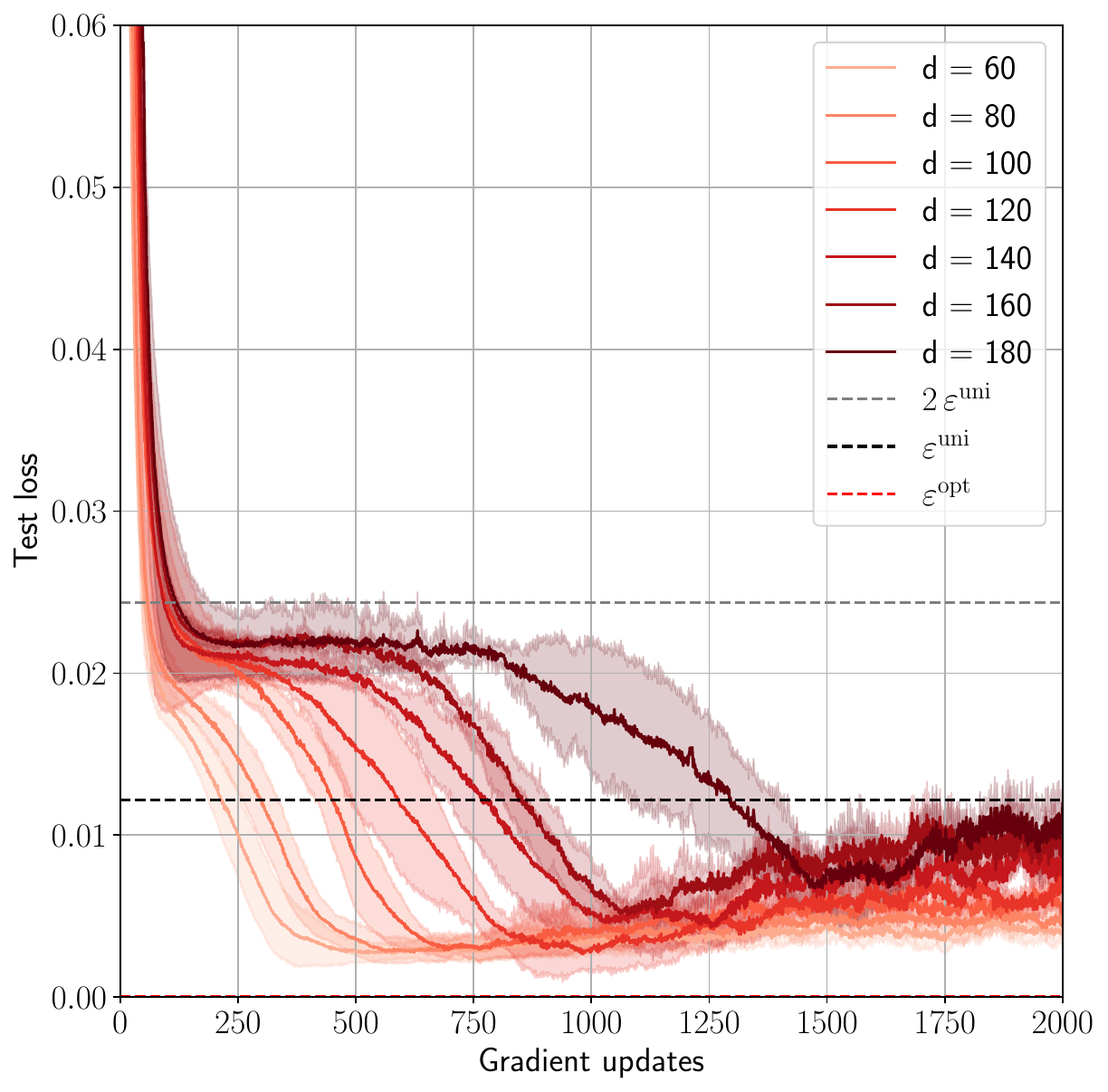}
    \includegraphics[width=0.32\linewidth,trim={0.9cm 0 0 0},clip]{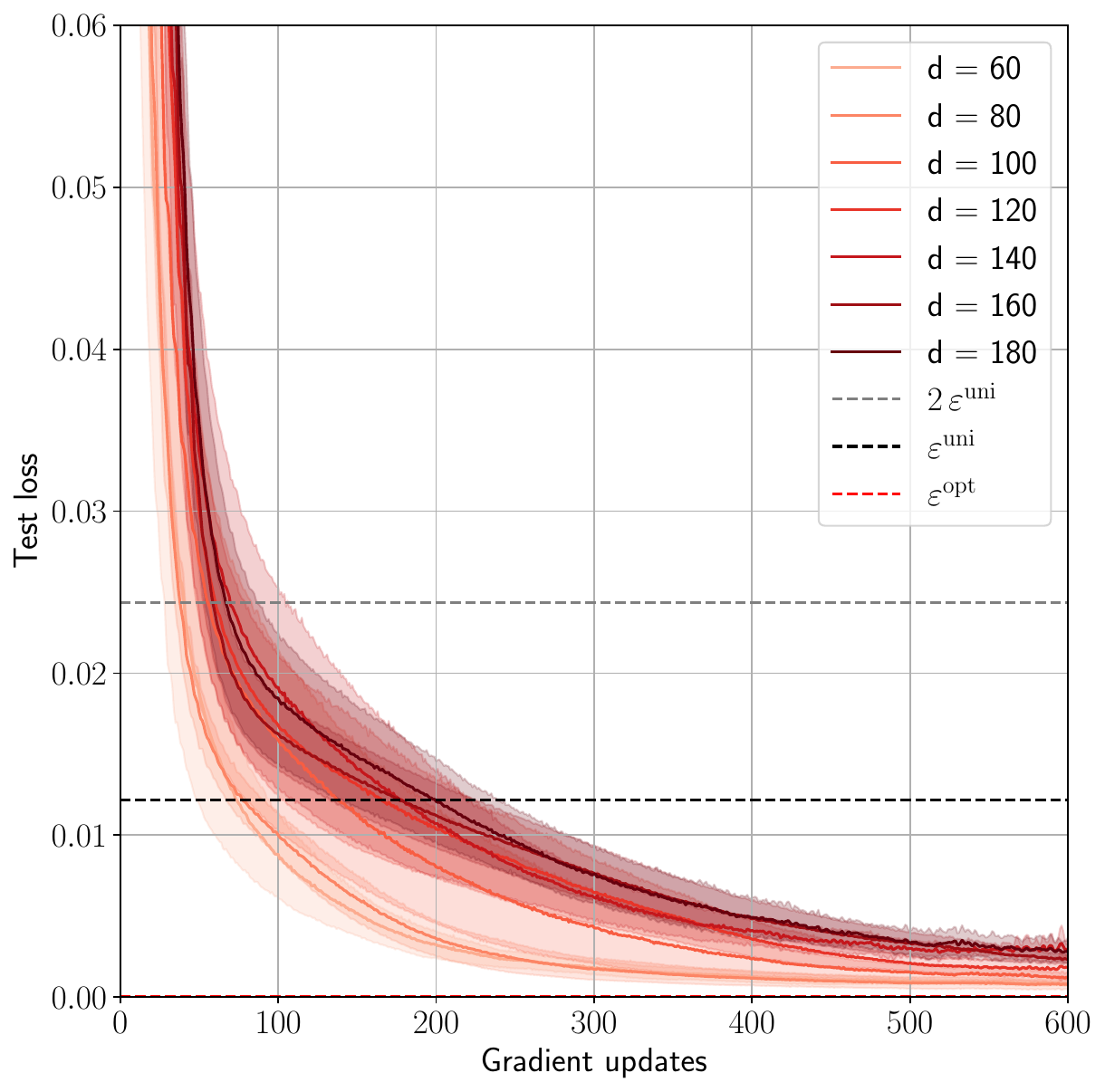}
    \caption{Trajectories of the generalisation error of neural networks trained with ADAM at fixed batch size $B=\lfloor n/4\rfloor$, learning rate 0.05, for ${\rm ReLU}(x)$ activation with parameters $\Delta = 10^{-4}$ for the linear readout, $\gamma=0.5$ and $\alpha=5.0 > \alpha_{\rm sp}$ ($= 0.22,0.12,0.02$ for homogeneous, Rademacher and Gaussian readouts respectively). The error $\varepsilon^{\rm uni}$ is the mean-square generalisation error associated with the universal solution with overlap $\mathcal{Q} \equiv 0$.
    \textbf{Left}: Homogeneous readouts.
    \textbf{Centre}: Rademacher readouts.
    \textbf{Right}: Gaussian readouts. Readouts are kept fixed (and equal to the teacher's) in all cases during training. Points on the solid lines are obtained by averaging over 5 teacher/data instances, and shaded regions around them correspond to 1 standard deviation.
    }
    \label{fig:hardness_adam_runs}
\end{figure}

In \figurename~\ref{fig:hardness_adam_runs}, we report the test loss of ADAM as a function of the gradient updates used for training, for various dimensions and choice of the readout distribution (as before, the readouts are not learnt but fixed to the teacher's). Here, we fix a batch size for simplicity. For both the cases of homogeneous ($\bv=\bm{1}$) and Rademacher readouts (left and centre panels), the model experiences plateaux in performance increasing with the system size, in accordance with the observation of exponential complexity we reported above. The plateaux happen at values of the test loss comparable with twice the value for the Bayes error predicted by the universal branch of the theory (remember the relationship between Gibbs and Bayes errors reported in App.~\ref{app:gen_err}). The curves are smoother for the case of Gaussian readouts.

\paragraph{Hamiltonian Monte Carlo}
The experiment is performed for the polynomial activation $\sigma_3 = \He_2/\sqrt 2 + \He_3/6$ with parameters $\Delta = 0.1$ for the Gaussian noise in the linear readout, $\gamma=0.5$ and $\alpha=1.0 >\alpha_{\rm sp}$ ($=0.26,0.30,0.02$ for homogeneous, Rademacher and Gaussian readouts respectively). Our HMC consists of $4000$ iterations for homogeneous readouts, or $2000$ iterations for Rademacher and Gaussian readouts. Each iteration is adaptive (with initial step size of $0.01$) and uses $10$ leapfrog steps. Instead of measuring the Gibbs error, whose relationship with $\varepsilon^{\rm opt}$ holds only at equilibrium (see the last remark in App.~\ref{app:gen_err}), we measured the teacher-student $R_2$-overlap which is meaningful at any HMC step and is informative about the learning. For a fixed threshold $\bar{R}_2$ and dimension $d$, we measure $\bar{t}(d)$ as the number of HMC iterations needed for the $R_2$-overlap between the HMC sample (obtained from uninformative initialisation) and the teacher weights $\bW^0$ to cross the threshold. This criterion is again enough to assess that the student outperforms the universal solution.

\begin{figure}[b!]
    \centering
    \includegraphics[width=0.32\linewidth,trim={1.1cm 0 0.3cm 0},clip]{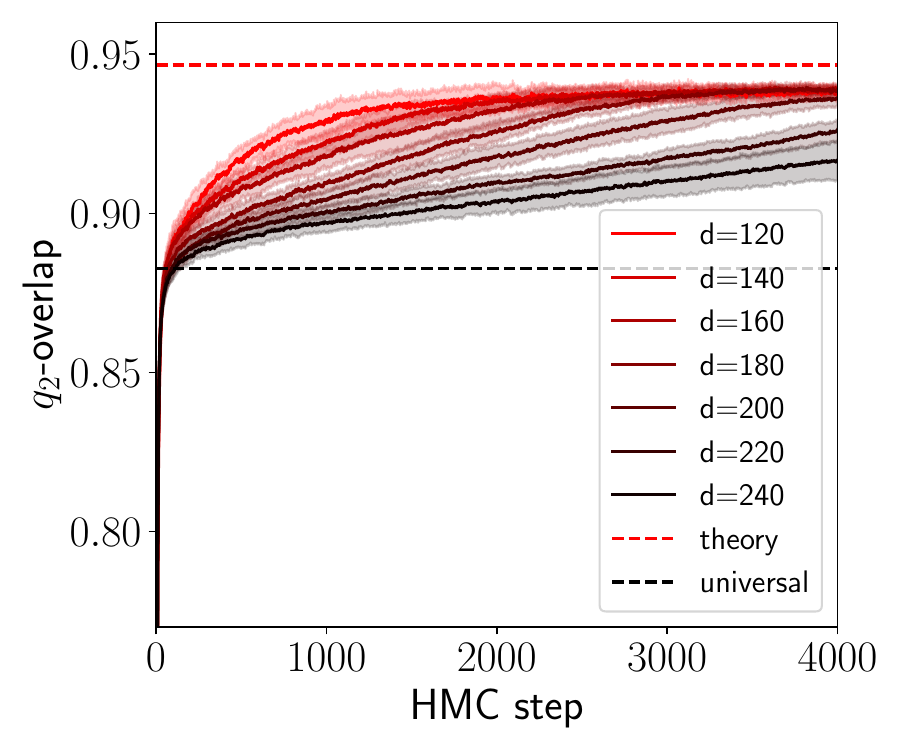}
    \includegraphics[width=0.32\linewidth,trim={1.1cm 0 0.3cm 0},clip]{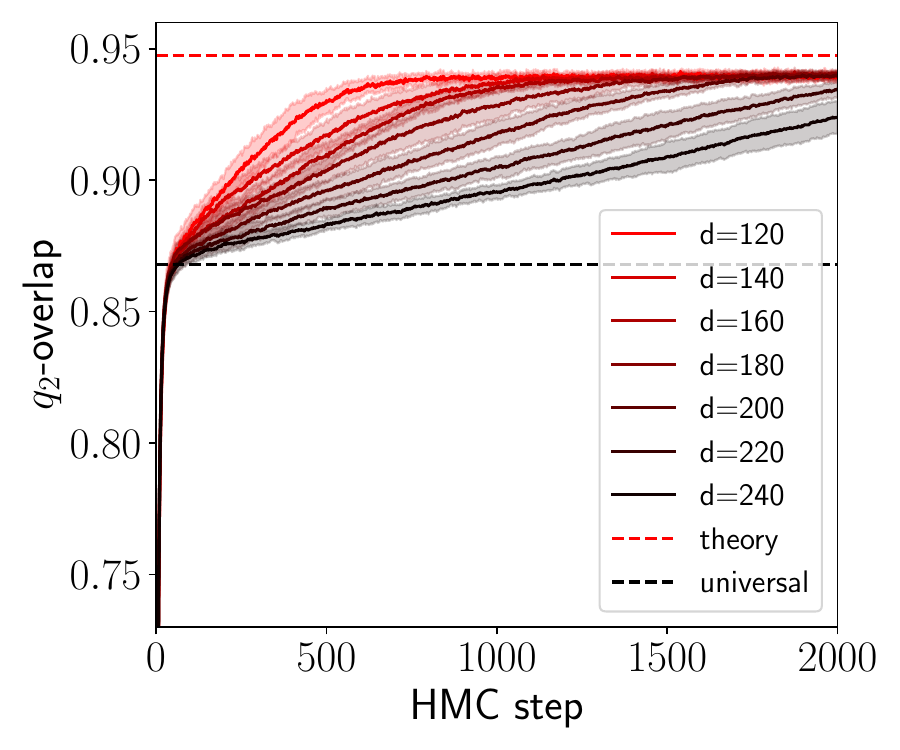}
    \includegraphics[width=0.32\linewidth,trim={1.1cm 0 0.3cm 0},clip]{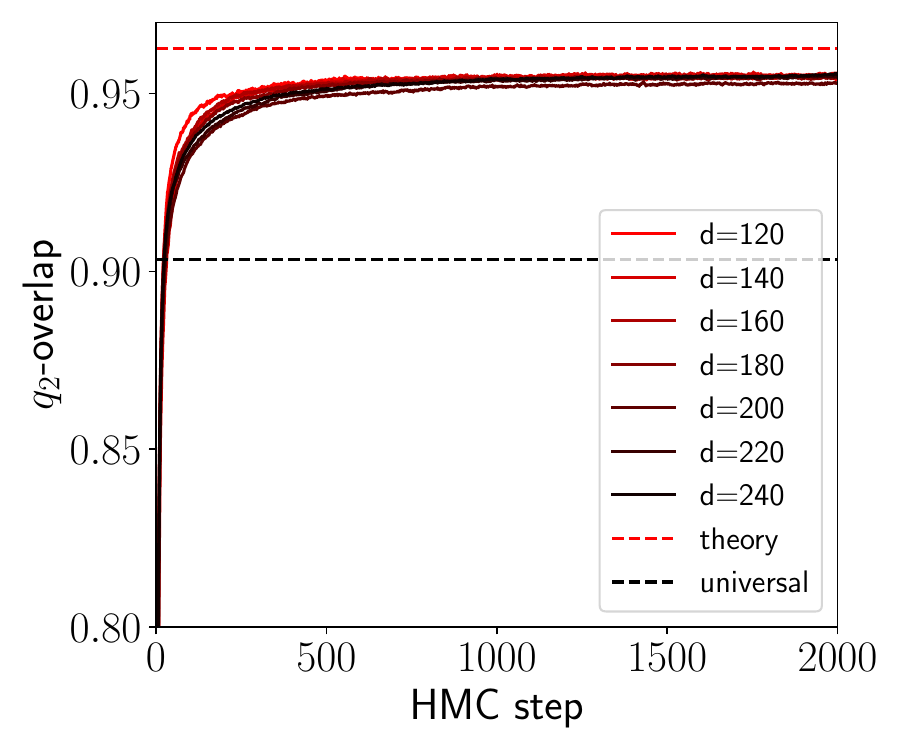}
    \caption{Trajectories of the overlap $R_2$ in HMC runs initialised uninformatively for the polynomial activation $\sigma_3= \He_2/\sqrt 2 + \He_3/6$ with parameters $\Delta = 0.1$ for the linear readout, $\gamma=0.5$ and $\alpha=1.0$. 
    \textbf{Left}: Homogeneous readouts.
    \textbf{Centre}: Rademacher readouts.
    \textbf{Right}: Gaussian readouts.
    Points on the solid lines are obtained by averaging over 10 teacher/data instances, and shaded regions around them correspond to 1 standard deviation.
    Notice that the $y$-axes are limited for better visualisation. For the left and centre plot, any threshold (horizontal line in the plot) between the prediction of the $ \mathcal{Q} \equiv 0$ branch of the theory (black dashed line) and its prediction for the $R_2^{\rm sp}$ (red dashed line, obtained with informative initialisation) crosses the curves in points $\bar{t}(d)$ more compatible with an exponential fit (see \figurename~\ref{fig:hardness_HMC} and Table~\ref{tab:HMC}, where these fits are reported and $\chi^2$-tested). For the cases of homogeneous and Rademacher readouts, the value of the overlap at which the dynamics slows down (predicted by the $\mathcal{Q}\equiv 0$ branch) is in quantitative agreement with the theoretical predictions (lower dashed line). The theory is instead off by $\approx 1\%$ for the values $R_2$ at which the runs ultimately converge.
    }
    \label{fig:hardness_HMC_runs}
\end{figure}

\begin{figure}[htp!]
\begin{center}
\centerline{
\includegraphics[width=.49\linewidth,trim={0 0 0 0},clip]{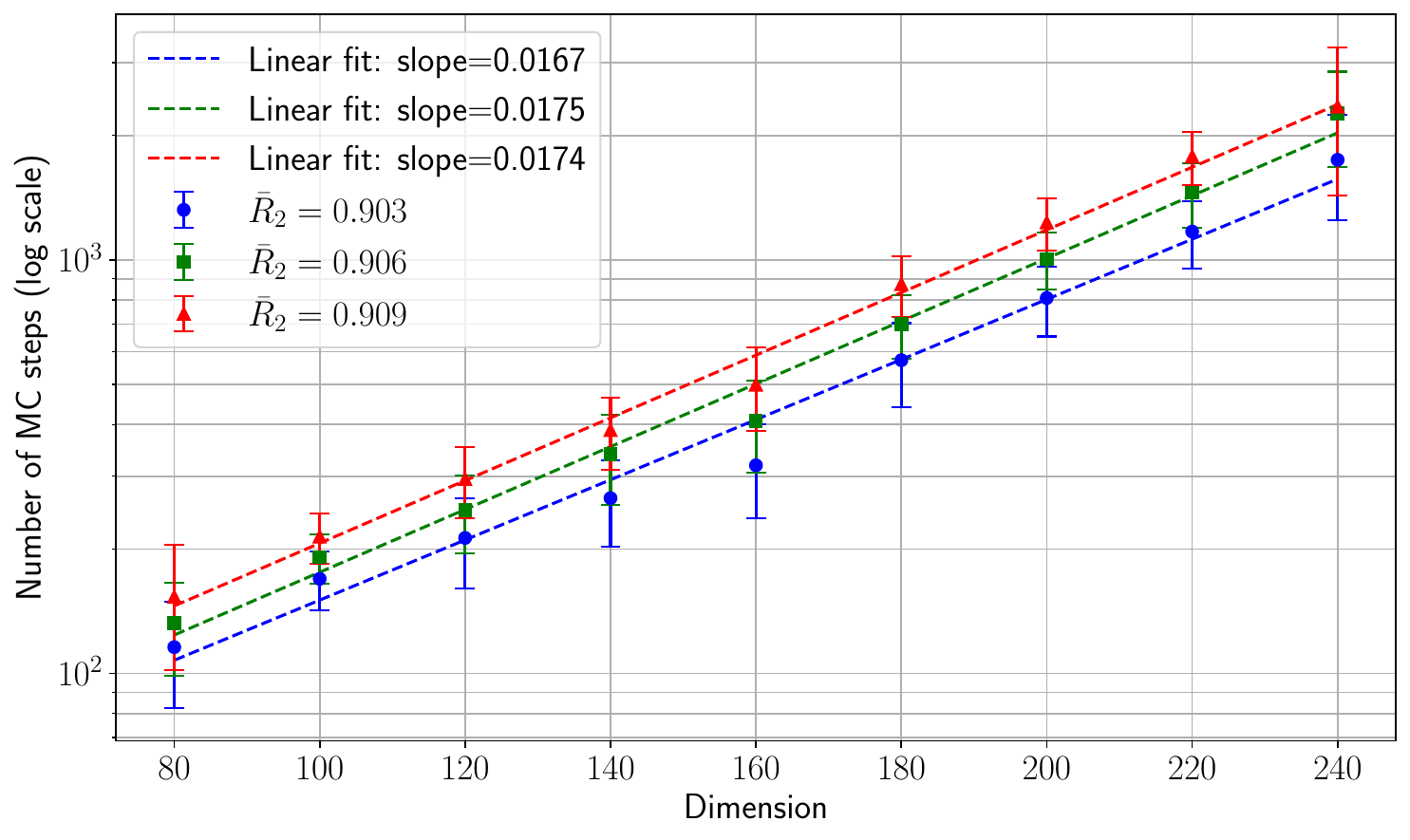}
\includegraphics[width=.49\linewidth,trim={0 0 0 0},clip]{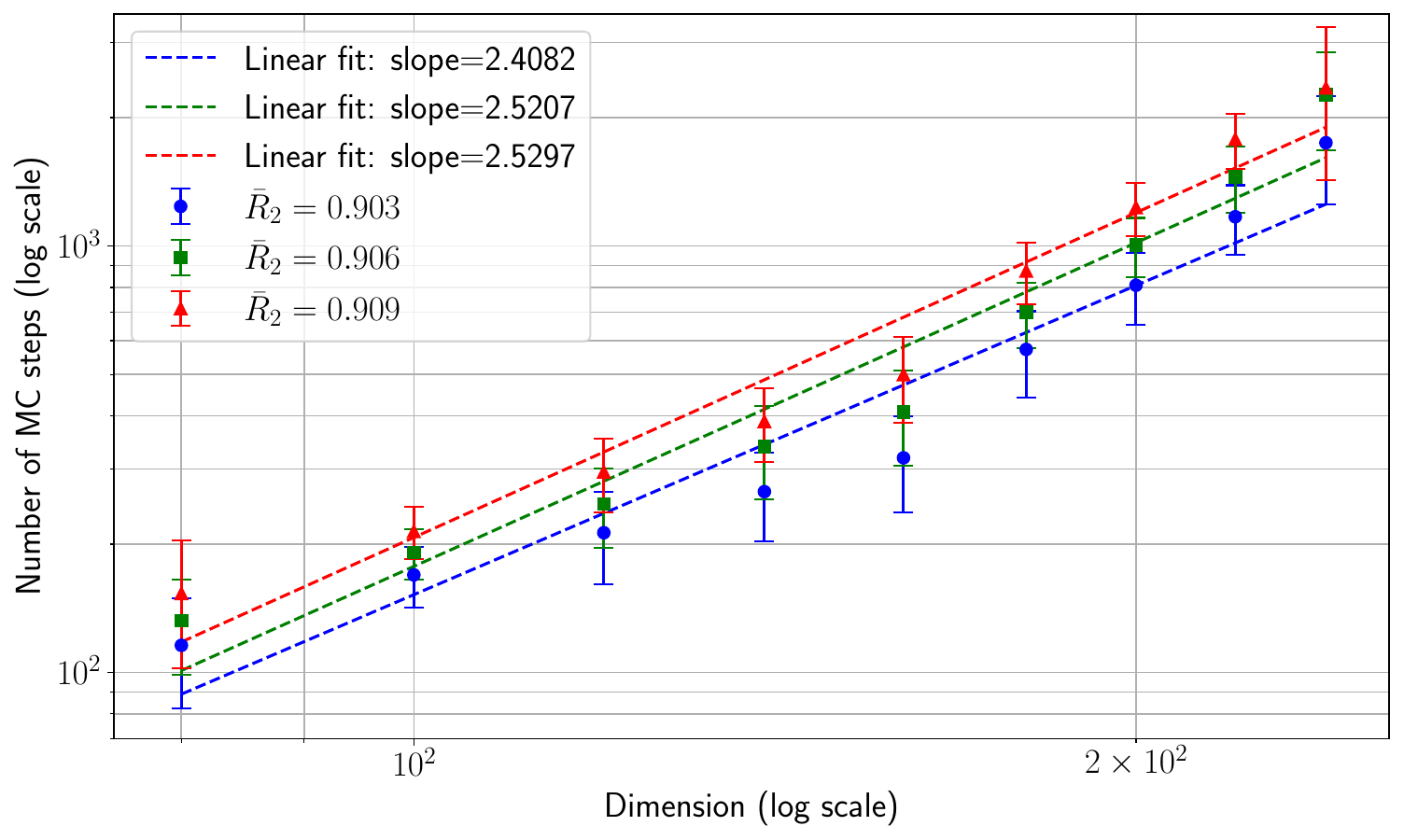}
}
\vspace{1pt} 
\centerline{
\includegraphics[width=.49\linewidth,trim={0 0 0 0},clip]{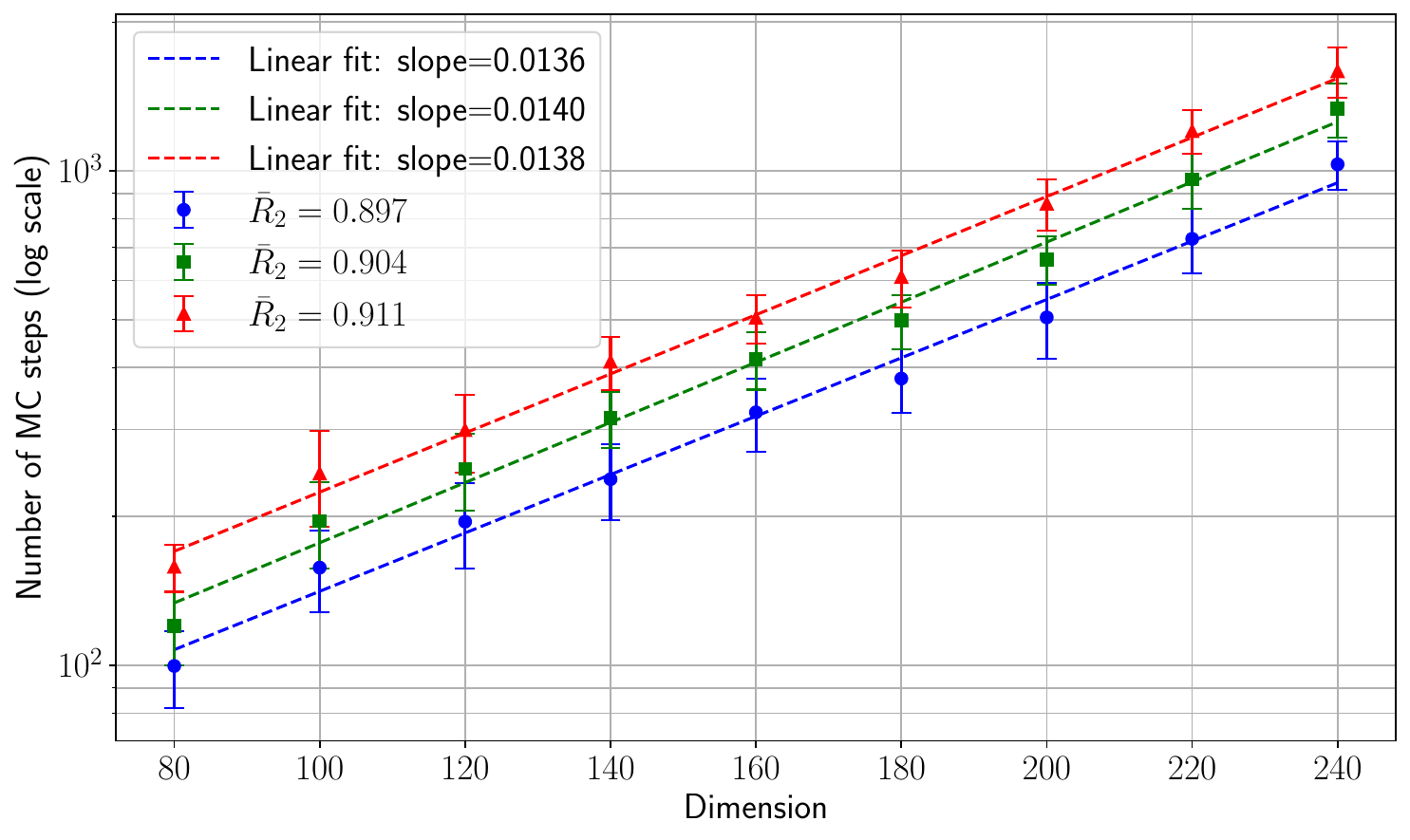}
\includegraphics[width=.49\linewidth,trim={0 0 0 0},clip]{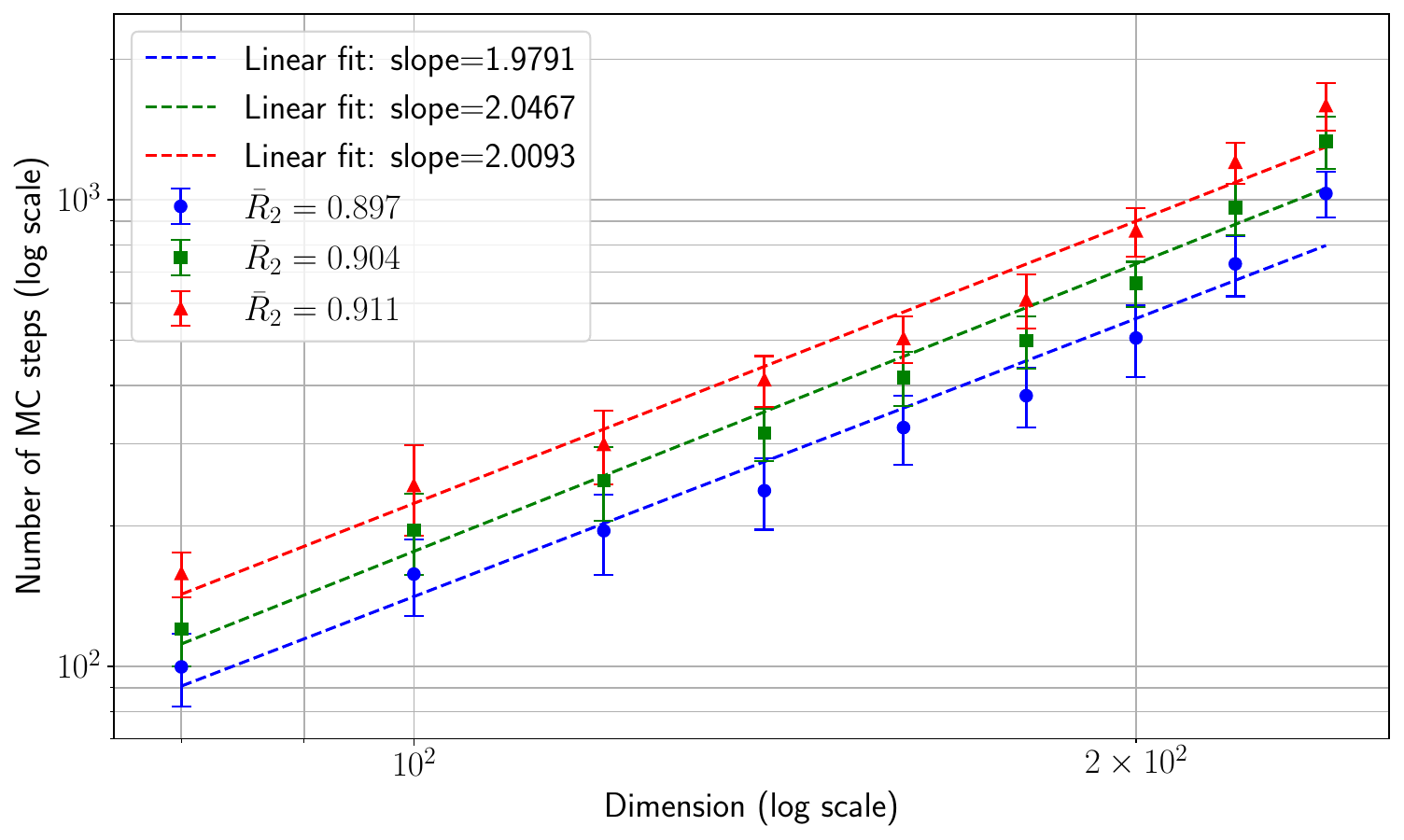}
}
\vspace{1pt} 
\centerline{
\includegraphics[width=.49\linewidth,trim={0 0 0 0},clip]{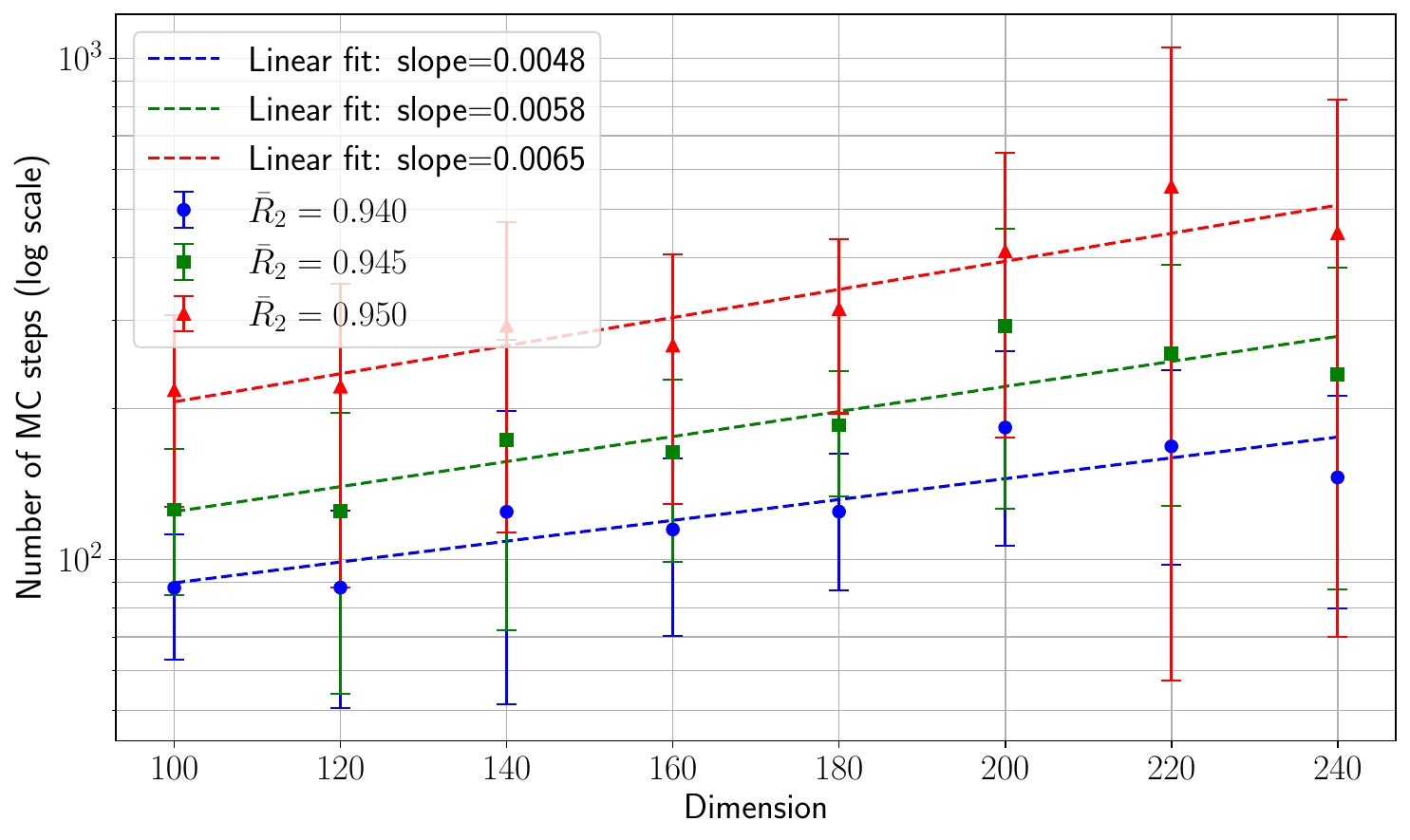}
\includegraphics[width=.49\linewidth,trim={0 0 0 0},clip]{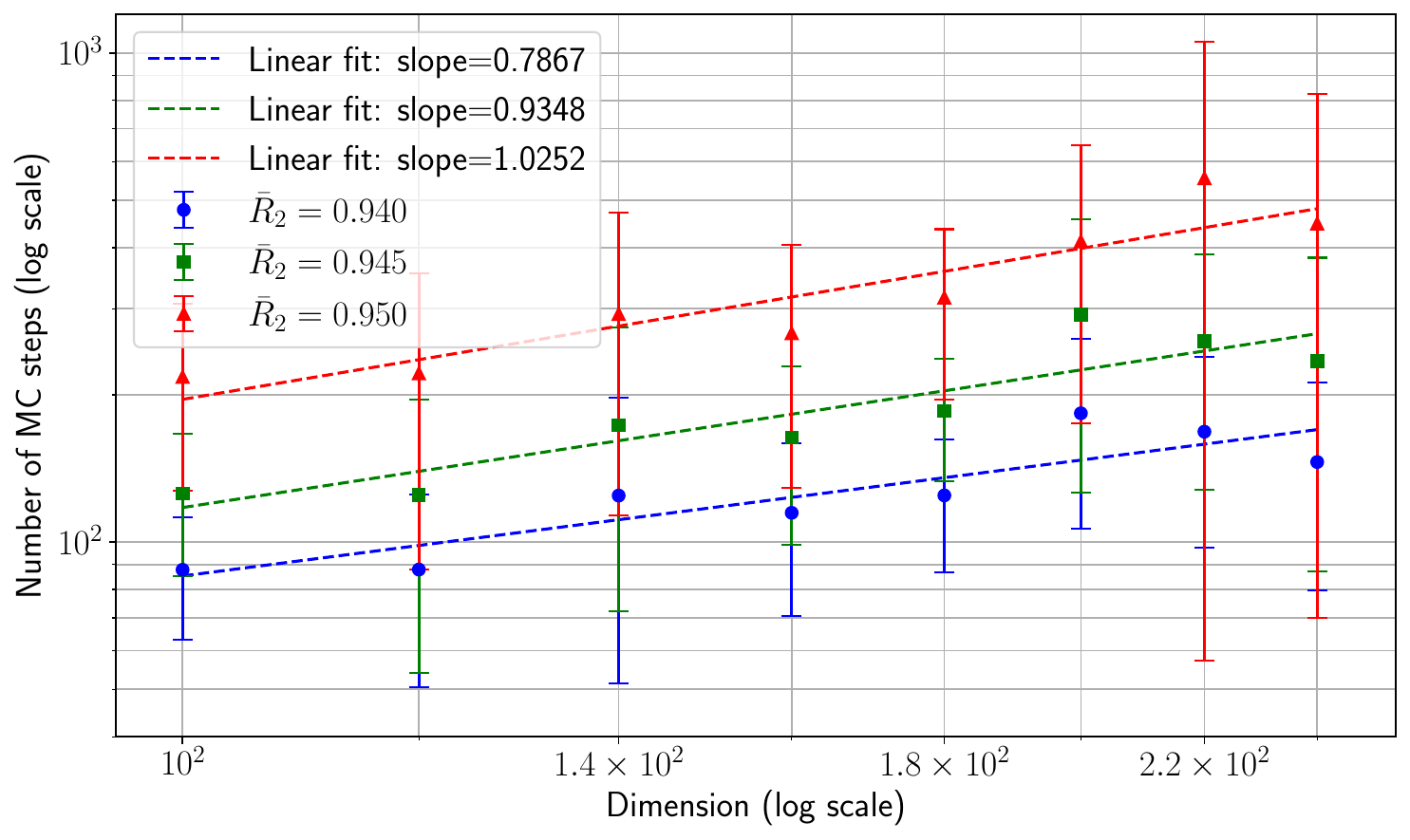}
}
\vspace{-10pt}
    \captionof{figure}{
    Semilog (\textbf{Left}) and log-log (\textbf{Right}) plots of the number of Hamiltonian Monte Carlo steps needed to achieve an overlap $\bar{R}_2>R_2^{\rm uni}$, that certifies the universal solution is outperformed. 
    The dataset was generated from a teacher with polynomial activation $\sigma_3 = \He_2/\sqrt 2 + \He_3/6$ and parameters $\Delta = 0.1$ for the linear readout, $\gamma=0.5$ and $\alpha=1.0>\alpha_{\rm sp}$ ($=0.26,0.30,0.02$ for homogeneous, Rademacher and Gaussian readouts respectively).
    Student weights are sampled using HMC (initialised uninformatively) with $4000$ iterations for homogeneous readouts (\textbf{Top row}, for which $R_2^{\rm uni}=0.883$), or $2000$ iterations for Rademacher (\textbf{Centre row}, with $R_2^{\rm uni}=0.868$) and Gaussian readouts (\textbf{Bottom row}, for which $R_2^{\rm uni}=0.903$). Each iteration is adaptative (with initial step size of $0.01$) and uses $10$ leapfrog steps. $R_2^{\rm sp}=0.941, 0.948, 0.963$ in the three cases. The readouts are kept fixed during training. 
    Points are obtained averaging over 10 teacher/data instances with error bars representing the standard deviation.
    }
    \label{fig:hardness_HMC}
\end{center}
\vskip -.25in

    \centering
    \begin{tabular}{lc|c|c|c|c|c|c|}
            &        & \multicolumn{3}{c|}{$\chi^2$ exponential fit} & \multicolumn{3}{c|}{$\chi^2$ power law fit}\\
        Readouts    & & \multicolumn{3}{c|}{} & \multicolumn{3}{c|}{} \\
        \hline
         Homogeneous & ($\bar{R}_2\in \{ 0.903, 0.906, 0.909\}$) &   $\bm{2.22}$ & $\bm{1.47}$ & $\bm{1.14}$ &$8.01$ &$7.25$ &$6.35$  \\
         Rademacher& ($\bar{R}_2\in \{0.897 ,0.904 ,0.911 \}$)& $\bm{1.88}$ & $\bm{2.12}$ & $\bm{1.70}$ &$8.10$ &$7.70$ &$8.57$ \\
         Gaussian& ($\bar{R}_2\in \{0.940 ,0.945 ,0.950 \}$)& $0.66$&$\bm{0.44}$&$\bm{0.26}$  & $\bm{0.62}$&$0.53$ &$0.39$\\
    \end{tabular}
    \captionof{table}{$\chi^2$ test for exponential and power-law fits for the time needed by Hamiltonian Monte Carlo to reach the thresholds $\bar{R}_2$, for various priors on the readouts. For a given row, we report three values of the $\chi^2$ test per hypothesis, corresponding with the thresholds $\bar{R}_2$ on the left, in the order given. Fits are displayed in \figurename~\ref{fig:hardness_HMC}. Smaller values of $\chi^2$ (in bold, for given threshold and readouts) indicate a better compatibility with the hypothesis.}
    \label{tab:HMC}
\end{figure}

As before, we test homogeneous, Rademacher and Gaussian readouts, getting to the same conclusions: while for homogeneous and Rademacher readouts exponential time is more compatible with the observations, the experiments remain inconclusive for Gaussian readouts (see \figurename~\ref{fig:hardness_HMC}). We report in \figurename~\ref{fig:hardness_HMC_runs} the values of the overlap $R_2$ measured along the HMC runs for different dimensions. Note that, with HMC steps, all $R_2$ curves saturate to a value that is off by $\approx 1\%$ w.r.t.\ that predicted by our theory for the selected values of $\alpha,\gamma$ and $\Delta$. Whether this is a finite size effect, or an effect not taken into account by the current theory is an interesting question requiring further investigation, see App.~\ref{app:structured_S2} for possible directions.

\paragraph{Learnable readouts}
As discussed in the main text, the static properties of the model remain unchanged whether the readout weights are quenched to the teacher values or learned during training. However, the dynamics can differ when the readouts are learnable. We verified that, for ADAM, the results regarding hardness are qualitatively unchanged when the readouts are learned. Although ADAM can achieve a lower test error in this case, the convergence time required to reach this solution increases substantially.

Specifically, when the readouts are fixed, specialisation occurs after approximately $10^4 - 10^5$ gradient updates for homogeneous priors (see \figurename~\ref{fig:hardness_adam_runs}, left). In contrast, as shown in \figurename~\ref{fig:HMC_hardness}, learning the readouts increases the number of gradient updates required for specialisation by at least an order of magnitude.

For HMC, which is constrained to sample according to the prior over both the inner and readout weights, the behaviour is essentially identical whether the readouts are fixed or learnable. The reasoning in Remark~\ref{rmk:readouts} therefore applies equally to HMC, as it is a posterior sampler.

\subsection{A potential route for a proof for \texorpdfstring{$L=1$}{}}
\label{sec:partial_proof}
Here we provide an argument for a potential proof of our results based the adaptive interpolation technique introduced in \cite{barbier2019adaptive}, used as in \cite{barbier2019glm}. In order to make the model more amenable to rigorous treatment, we select activation functions with $\mu_0=\mu_1=\mu_2=0$ and all-ones readouts $\mathbf{v}=\mathbf{1}$. While assumptions on $\mu_0,\mu_1$ are not that restrictive, $\mu_2=0$ is what induces the main simplifications as it erases from the analyses the role of the overlaps $R_2$. It is also useful to write the replica potential we are targeting, which in the hypotheses listed above and with a little abuse of notation reads
\begin{align}
\label{eq:RS_freeent_mu2_0}
    f_{\rm RS}^{(1)}(\calQ,\hat\calQ) = \frac{\gamma}{\alpha} \psi_{P_W}(\hat\calQ) 
    +\phi_{P_{\rm out}}\big(g(\calQ),g(1)\big) 
    -\frac{\gamma}{2\alpha} \calQ\hat\calQ\,.
\end{align}
Apart from the presence of $g(\calQ)$ inside $\phi_{P_{\rm out}}$ the above formula looks like the one of a standard generalised linear model \cite{barbier2019glm}, we shall thus use a similar formalism. Let us define an interpolating model:
\begin{align}
    S_{t\mu}&:=\frac{\sqrt{1-t}}{\sqrt{k}}\sum_{i=1}^k \varphi\Big( \frac{1}{\sqrt{d}} \bW_i^{* \intercal} \bx_\mu \Big)+ \sqrt{G(t)} \, V_\mu + \sqrt{g(1)t - G(t)} \, U_\mu^*
\end{align}where $t\in[0,1]$, $G(t)$ is a non-negative interpolating function, and $V_\mu,U^*_\mu\iid\mathcal{N}(0,1)$. Note that $U_\mu^*$ is a starred variable, thus we consider it as learnable. The labels for this interpolating model are then given by
\begin{align}
    Y_{t\mu}\sim P_{\rm out}(\cdot\mid S_{t\mu})
\end{align}
To complete the interpolation, we also need another Gaussian observation channel about $\bW^*$:
\begin{align}
    {Y}^G_{t;ij}=\sqrt{R(t)}W_{ij}^*+Z_{ij}
\end{align}where $Z_{ij}\iid\mathcal{N}(0,1)$. The interpolating functions $G(t),R(t)$ have to be appropriately chosen later, in order to make some remainders vanish. The only requirement for now is that $G(0)=R(0)=0$. Define $u_{y}(x):=\ln P_{\rm out}(y\mid x)$, and denote for brevity $\bY_t=(Y_{t\mu})_{\mu\leq n}$, $\bY^G_t=(Y^G_{t;ij})_{i\leq k,j\leq d}$, $\bU=(U_\mu)_{\mu\leq n}$, $\bV=(V_\mu)_{\mu\leq n}$ and
\begin{align}
    s_{t\mu}&:=\frac{\sqrt{1-t}}{\sqrt{k}}\sum_{i=1}^k \varphi\Big( \frac{1}{\sqrt{d}} \bW_i^{ \intercal} \bx_\mu \Big)+ \sqrt{G(t)} \, V_\mu + \sqrt{g(1)t - G(t)} \, U_\mu
\end{align}
Then the above interpolating model induces a Hamiltonian that reads
\begin{align}
    -\mathcal{H}_t(\bX,\bY_t,
    \bY^G_{t},&\bV,\bU,\bW)=\sum_{\mu=1}^n u_{Y_{t\mu}}(s_{t\mu})-\frac{1}{2}\|\bY^G_t-\sqrt{R(t)}\bW\|^2,
\end{align}and the corresponding quenched free entropy is
\begin{align}
    f_n(t)=\frac{1}{n}\EE_{(t)}\ln\mathcal{Z}_t
\end{align}with
\begin{align}
    \EE_{(t)}[\cdot]&= \EE_{\bX,\bV,\bU^*,\bW^*}\int d\bY_t d\bY^G_t e^{-\mathcal{H}_t(\bX,\bY_t,
    \bY^G_{t},\bV,\bU^*,\bW^*)} [\cdot]\,,\quad 
    \mathcal{Z}_t&=\int dP_W(\bW)D\bU 
    e^{-\mathcal{H}_t(\bX,\bY_t,
    \bY^G_{t},\bV,\bU,\bW)}\,.
\end{align}

At the extrema of interpolation we have
\begin{align}
    f_n(0)&=f_n-\frac{\gamma}{2\alpha}\nonumber\\
    f_n(1)&=\frac{\gamma}{\alpha}\psi_{P_W}(R(1))+\phi_{P_{\rm out}}(G(1),g(1))-\frac{\gamma}{2\alpha}(1+R(1))\,.
\end{align}
The role of the interpolation is that of decoupling continuously the quenched disorder in the $\bx_\mu$'s from the weights $\bW^*$, and simultaneously linearising the non-linearity $\varphi$. We shall now convince the reader that there are choices for the functions $G,R$ that produce \eqref{eq:RS_freeent_mu2_0}. To begin with, we need to control the $t$-derivative of the interpolating free entropy:
\begin{align}
    \frac{d}{dt}f_n(t)=&-\frac{1}{n}\EE_{(t)}\frac{d}{dt}\mathcal{H}_t(\bX,\bY_t,
    \bY^G_{t},\bV,\bU^*,\bW^*)\ln \mathcal{Z}_t-\frac{1}{n}\EE_{(t)}\langle \frac{d}{dt}\mathcal{H}_t(\bX,\bY_t,
    \bY^G_{t},\bV,\bU,\bW)\rangle_t
\end{align}where $\langle\cdot\rangle_t$ is the Gibbs measure associated with the Hamiltonian $\mathcal{H}_t$. 

Let us focus first on the second term on the r.h.s. Using the Nishimori identities on it we readily get
\begin{align}
    {\rm II}=\frac{1}{n}\EE_{(t)}\frac{d}{dt}\mathcal{H}_t(\bX,\bY_t,
    \bY^G_{t},\bV,\bU^*,\bW^*)=
    \frac{1}{n}\EE_{(t)}\Big[\sum_{\mu=1}^n u'_{Y_{t\mu}}(S_{t\mu})\dot{S}_{t\mu}+\frac{\dot{R}(t)}{2\sqrt{R(t)}}\sum_{i,j}^{k,d}W^*_{ij}(Y^G_{t;ij}-\sqrt{R(t)}W^*_{ij})\Big]\,.
\end{align}Considering that $Y^G_{t;ij}-\sqrt{R(t)}W^*_{ij}=Z_{ij}$, which is independent of $W_{ij}^*$, and that 
\begin{align*}
    \int d Y_{t\mu} P_{\rm out}(Y_{t\mu}\mid S_{t\mu}) u'_{Y_{t\mu}}(S_{t\mu})=\partial_x 
    \int d y P_{\rm out}(y\mid x)
    \Big|_{x=S_{t\mu}}=0
\end{align*}we have that II$=0$ identically.
Concerning the first term instead:
\begin{align}
    {\rm I}= \EE_{(t)}\ln \mathcal{Z}_t\frac{1}{n}\Big[\sum_{\mu=1}^n
    u'_{Y_{t\mu}}(S_{t\mu})\dot{S}_{t\mu}+\sum_{i,j}^{k,d}(Y_{t;ij}^G-\sqrt{R(t)} W^*_{ij})\frac{\dot{R}(t)}{2\sqrt{R(t)}}W^*_{ij}\Big]\,.
\end{align}
We start by the first term on the r.h.s.\ which requires the most care. After replacing $\dot{S}_{t\mu}$ with its expression we have
\begin{align}\label{eq:intermediate_2}
    \EE_{(t)}\ln \mathcal{Z}_t\frac{1}{2n} \sum_{\mu=1}^n
    u'_{Y_{t\mu}}(S_{t\mu})\Big[-\frac{1}{\sqrt{(1-t)}}\lambda^*_\mu+\frac{\dot{G}(t)}{\sqrt{G(t)}}V_\mu+\frac{g(1)-\dot{G}(t)}{\sqrt{g(1)t-G(t)}}U_\mu^*\Big]
\end{align}where $\lambda_\mu^*=\frac{1}{\sqrt{k}}\sum_{i\leq k}\varphi\Big( \frac{1}{\sqrt{d}} \bW_i^{ *\intercal} \bx_\mu \Big)$. In the GLM one aims at integrating $\bx_\mu$ by parts, but here it is not possible due to the presence of the non-linearity. Hence we need to make a Gaussian assumption to treat it. 

\begin{assumption}
Defining $\lambda_\mu=\frac{1}{\sqrt{k}}\sum_{i\leq k}\varphi\Big( \frac{1}{\sqrt{d}} \bW_i^{ \intercal} \bx_\mu \Big)$ and $\lambda_\mu^*$ as above, the following holds under the randomness of $\bx_\mu$:
\begin{align}\label{eq:GEP_appendix}
    &\lambda_\mu,\lambda_\mu^*\iid\mathcal{N}(\mathbf{0},C(\bW,\bW^*))\,,\quad C_{ab}\equiv C_{ab}(\bW,\bW^*)= \sum_{\ell\geq 3} \frac{\mu_\ell^2}{\ell!}\frac{1}{k}\sum_{i,j=1}^{k}\Big(\frac{\bW^a\bW^{b\intercal}}{d}\Big)_{ij}^{\circ \ell}
\end{align}where $a,b=\cdot,\ast$, and $\cdot$ labels a posterior sample.

\end{assumption}

When integrating by parts, one needs to take into account that the probability weights hidden in $\EE_{(t)}$ also depend on $S_{t\mu}$. Bearing this in mind, integration by parts of $\lambda_\mu^*,U_\mu^*,V_\mu$ in \eqref{eq:intermediate_2} yields
\begin{align}
    &-\EE_{(t)}\ln \mathcal{Z}_t\frac{1}{2n} \sum_{\mu=1}^n\big[
    u''_{Y_{t\mu}}(S_{t\mu})+(u'_{Y_{t\mu}}(S_{t\mu}))^2\big]C_{**}-\frac{1}{2}\EE_{(t)}\langle \frac{1}{n}\sum_{
    \mu=1}^n u'_{Y_{t\mu}}(S_{t\mu}) u'_{Y_{t\mu}}(s_{t\mu}) C_{\cdot*}\rangle_t\nonumber\\
    &+\EE_{(t)}\ln \mathcal{Z}_t\frac{1}{2n} \sum_{\mu=1}^n\big[
    u''_{Y_{t\mu}}(S_{t\mu})+(u'_{Y_{t\mu}}(S_{t\mu}))^2\big]\dot{G}(t)+\frac{\dot{G}(t)}{2}\EE_{(t)}\langle \frac{1}{n}\sum_{
    \mu=1}^n u'_{Y_{t\mu}}(S_{t\mu}) u'_{Y_{t\mu}}(s_{t\mu})\rangle_t\nonumber\\
    &+\EE_{(t)}\ln \mathcal{Z}_t\frac{1}{2n} \sum_{\mu=1}^n\big[
    u''_{Y_{t\mu}}(S_{t\mu})+(u'_{Y_{t\mu}}(S_{t\mu}))^2\big](g(1)-\dot{G}(t))\,.
\end{align}
Considering that $u''_{y}(x)+(u'_{y}(x))^2=\partial_x^2 P_{\rm out}(y\mid x)/P_{\rm out}(y\mid x)$, by gathering all the previous terms together \eqref{eq:intermediate_2} becomes
\begin{align}
    \EE_{(t)}\ln \mathcal{Z}_t\frac{1}{2n} \sum_{\mu=1}^n\frac{\partial_x^2 P_{\rm out}(Y_{t\mu}\mid S_{t\mu})}{P_{\rm out}(Y_{t\mu}\mid S_{t\mu})}(g(1)-C_{**})+ \frac{1}{2}\EE_{(t)}\langle \frac{1}{n}\sum_{\mu=1}^n u'_{Y_{t\mu}}(S_{t\mu}) u'_{Y_{t\mu}}(s_{t\mu}) (
    \dot{G}(t)-C_{\cdot*})\rangle_t
\end{align}

Concerning instead the second term on the r.h.s. of I, it can be simplified via a standard integration by parts of the Gaussian random variable $Y^G_{t;ij}-\sqrt{R(t)}W^*_{ij}=Z_{ij}$. We thus report just the final result for I:
\begin{align}
    {\rm I}&=\EE_{(t)}\ln \mathcal{Z}_t\frac{1}{2n} \sum_{\mu=1}^n\frac{\partial_x^2 P_{\rm out}(Y_{t\mu}\mid S_{t\mu})}{P_{\rm out}(Y_{t\mu}\mid S_{t\mu})}(g(1)-C_{**})+ \frac{1}{2}\EE_{(t)}\langle \frac{1}{n}\sum_{\mu=1}^n u'_{Y_{t\mu}}(S_{t\mu}) u'_{Y_{t\mu}}(s_{t\mu}) (
    \dot{G}(t)-C_{\cdot*})\rangle_t\nonumber\\
    &-\frac{\gamma}{2\alpha}\dot{R}(t)(1-\calQ(t))-\frac{\gamma}{2\alpha}\dot{R}(t)\Big[\calQ(t)-\frac{1}{kd}\EE_{(t)}\langle\Tr\bW^*\bW^\intercal\rangle_t\Big]
\end{align}
where we have added and subtracted the term containing $\calQ(t)$ in the second line. $\calQ(t)$ here is an arbitrary non-negative function for the moment.

By a simple application of the fundamental theorem of integral calculus we have thus proved the following \emph{sum rule}:
\begin{proposition}[Sum rule] Assume the GEP in \eqref{eq:GEP_appendix} holds. Then:
\begin{align}
    f_n=f_n(0)+\frac{\gamma}{2\alpha}=f_n(1)+\frac{\gamma}{2\alpha}-
    \int_0^1 I(t)dt=\frac{\gamma}{\alpha}\psi_{P_W}(R(1))+\phi_{P_{\rm out}}(G(1),g(1))-\frac{\gamma}{2\alpha}R(1)-\int_0^1 {\rm I}(t)dt
\end{align}
    where we have stressed the $t$-dependence of I.
\end{proposition}
It is now time to make some choices about our interpolating functions. Firstly, we link $\calQ(t)$ and $G(t)$ as follows: $G(t)=\int_0^t g(\calQ(s))ds$. Then, out of convenience we call $\hat\calQ(t)=\dot{R}(t)$. Secondly, we need the following
\begin{assumption}\label{ass:concentrations_adaptive}
    The equation
    \begin{align}\label{eq:ODE_appendix}
        \calQ(t)=\frac{1}{kd}\EE_{(t)}\langle\Tr[\bW^*\bW^\intercal]\rangle_t
    \end{align}has a solution. Furthermore, assume
    \begin{align}\label{eq:off_diag_**}
        &\EE_{(t)}(g(1)-C_{**})^2=o_n(1)\\
        \label{eq:off_diag_*.} &\EE_{(t)}\langle(g(\calQ(t))-C_{.*})^2\rangle_t=o_n(1)
    \end{align}uniformly in $t$.
\end{assumption}
Assumption \eqref{eq:ODE_appendix} is not trivial, as $G(t)$ and $\calQ(t)$ are now linked and thus $\calQ(t)$ appears on both sides of the above equality (it is contained in the definition of $\EE_{(t)}\langle\cdot\rangle_t$). A formal proof of \eqref{eq:off_diag_**} is at reach with standard concentration of measure tools, whereas \eqref{eq:off_diag_*.} requires much more care. The proofs of \eqref{eq:off_diag_**} and \eqref{eq:off_diag_*.} are both left for future work. Both of them are enforcing that
\begin{align*}
    \frac{1}{k}\sum_{i,j=1}^{k}\Big(\frac{\bW^a\bW^{b\intercal}}{d}\Big)_{ij}^{\circ \ell}\approx \frac{1}{k}\sum_{i=1}^{k}\Big(\frac{\bW^a_i\cdot\bW^{b}_i}{d}\Big)^{ \ell}
\end{align*}under the $\EE_{(t)}\langle\cdot\rangle_t$ measure for $\ell\geq3$. Since there is permutation symmetry over the readout neurons when $\mathbf{v}=\mathbf{1}$, all the terms in the above equation are basically assumed to concentrate onto \eqref{eq:ODE_appendix}.

Under Assumption \ref{ass:concentrations_adaptive} the sum rule reads
\begin{align}
    f_n=\frac{\gamma}{\alpha}\psi_{P_W}(\int_0^1 \hat\calQ(t)dt)+\phi_{P_{\rm out}}(\int_0^1 g(\calQ(t))dt,g(1))-\frac{\gamma}{2\alpha}\int_0^1 \hat\calQ(t)\calQ(t)dt+o_n(1)\,.
\end{align}
Observe that $\phi_{P_{\rm out}}$, $g$ and $\psi_{P_W}$ are all non-decreasing and convex functions of their arguments. Furthermore, the above estimate holds for \emph{any} function $\hat\calQ(t)$, whereas $\calQ(t)$ has been fixed as the solution of \eqref{eq:ODE_appendix}. We start by choosing $\hat\calQ(t)=\hat\calQ=$const, and we use Jensen's inequality on $g$:
\begin{align}
    f_n&\geq \frac{\gamma}{\alpha}\psi_{P_W}(\hat\calQ)+\phi_{P_{\rm out}}(g(Q),g(1))-\frac{\gamma}{2\alpha}\hat\calQ Q+o_n(1)\geq \inf_\calQ f_{\rm RS}^{(1)}(\calQ,\hat\calQ)+o_n(1)
\end{align}with $Q=\int_0^1\calQ(t) dt$. The above is then made tight by taking the $\sup_r$.

The converse bound is instead obtained by using Jensen's inequality to take the $\int_0^1\,dt$ out of the $\psi$ functions, which yields
\begin{align}
    f_n&\leq \int_0^1 f_{\rm RS}^{(1)}(\calQ(t),\hat\calQ(t)) dt+o_n(1)\,.
\end{align}
In order to make the bound tight, we now choose $\hat\calQ(t)$ as the solution of the optimisation $\inf_{\hat\calQ} f_{\rm RS}^{(1)}(\calQ(t),\hat\calQ)$, which is unique by convexity. Therefore:
\begin{align}
    f_n&\leq \int_0^1 \inf_{\hat\calQ} f_{\rm RS}^{(1)}(\calQ(t),\hat\calQ) dt+o_n(1)\leq \sup_\calQ \inf_{\hat\calQ} f_{\rm RS}^{(1)}(\calQ,\hat\calQ) +o_n(1)\,.
\end{align}
To summarise
\begin{align}
    \sup_\calQ \inf_{\hat\calQ} f_{\rm RS}^{(1)}(\calQ,\hat\calQ)+o_n(1)\leq f_n\leq \sup_\calQ\inf_{\hat\calQ}f_{\rm RS}^{(1)}(\calQ,\hat\calQ) +o_n(1)\,.
\end{align}
Strictly speaking, the two variational principles on the two sides of these bounds are different, but for sure they have the same stationary points. Under suitable conditions, see for instance Corollary 7 in the Supplementary Information of \cite{barbier2019glm}, they actually yield the same value, which would close the proof.

\begin{figure}[b!]
    \centering
    \includegraphics[width=0.478\linewidth,clip]{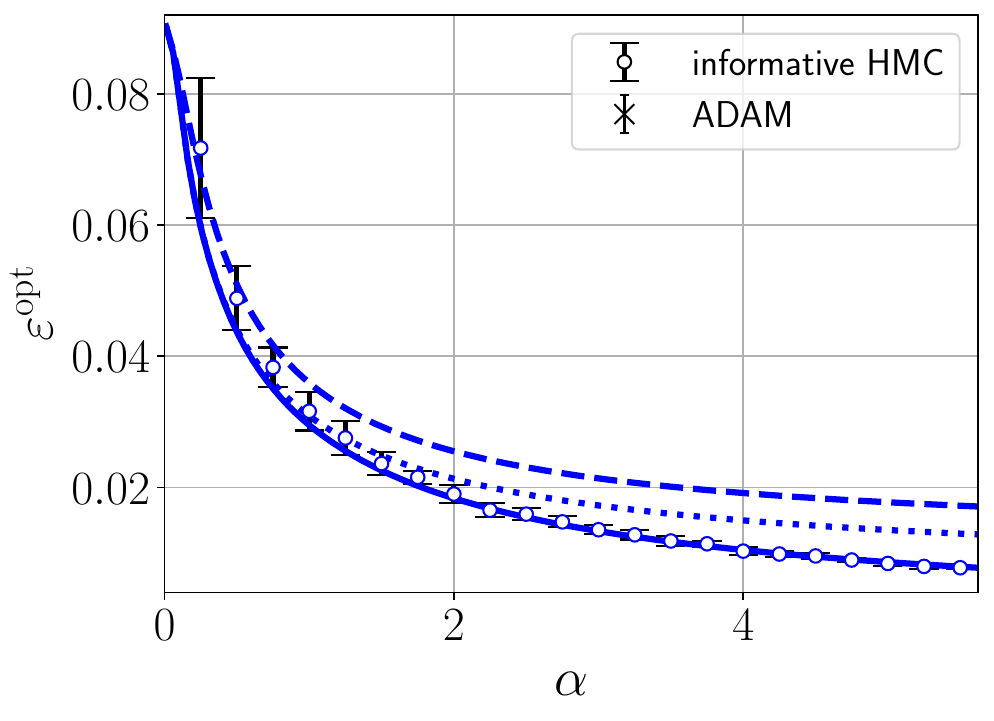}
    \includegraphics[width=0.485\linewidth,clip]{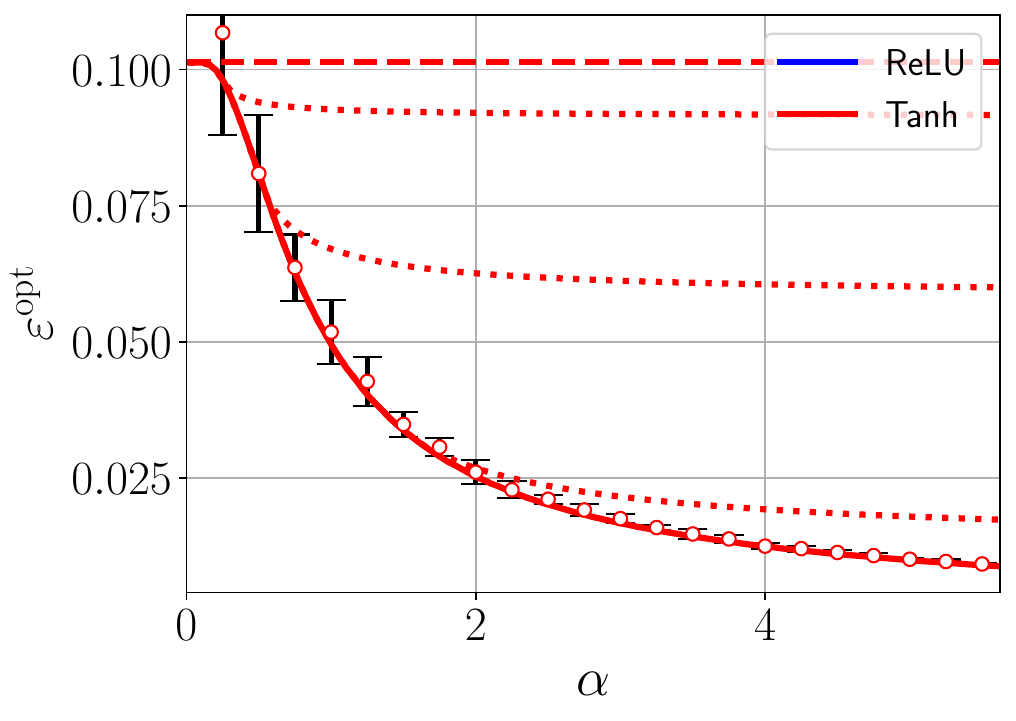}
    \includegraphics[width=0.6\linewidth,trim={0.3cm 0 0 0},clip]{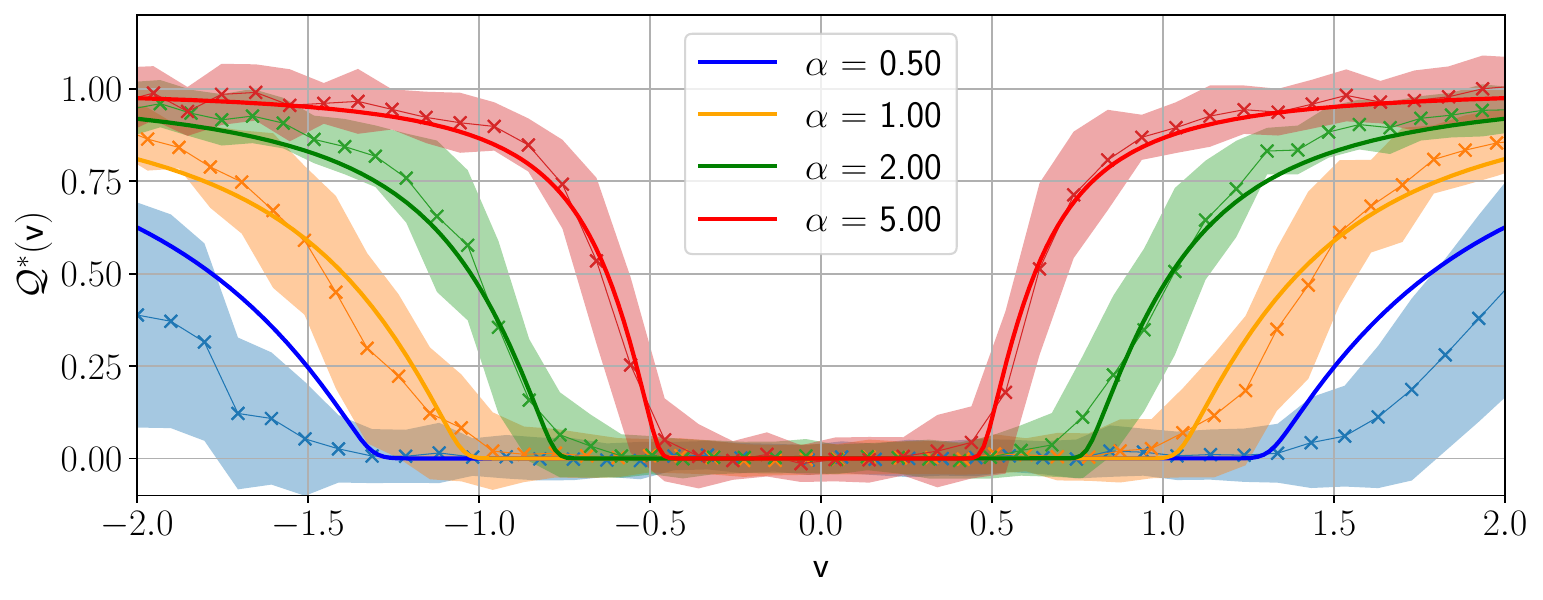}\hspace*{-0.1cm}
  \vspace{-5pt}

    \caption{\textbf{Top:} Theoretical prediction (solid curves) of the specialisation mean-square generalisation error $\varepsilon^{\rm sp}$ for \emph{Gaussian inner weights} with ${\rm ReLU}(x)$ activation (blue curves) and $\tanh(2x)$ activation (red curves), $d=200$, $\gamma=0.5$, with linear readout and Gaussian label noise of variance $\Delta=0.1$. The dashed lines show the theoretical prediction associated with the universal branch of our theory, $\varepsilon^{\rm uni}$. Markers are for Hamiltonian Monte Carlo with informative initialisation on the target (empty circles). Each point is averaged over 12 teacher/training-set instances; error bars denote the sample standard deviation across instances. Generalisation errors are numerically evaluated as half Gibbs errors, assuming the validity of Nishimori identities on metastable states as in the main (see also App.~\ref{app:gen_err} and \eqref{eq:Gibbs_v_Bayes_error}). The empirical average over test inputs is computed from $10^5$ i.i.d.\ test samples.  \textbf{Bottom:} Theoretical prediction (solid curves) of the overlap for different sampling ratios $\alpha$ for Gaussian inner weights, $\sigma(x)={\rm ReLU}(x), d=200, \gamma=0.5,\Delta=0.1$ and Gaussian readouts. The shaded curves were obtained from informed HMC. Using a single posterior sample $\bW$ (per $\alpha$ and data instance), $\mathcal{Q}(\mathsf{v})$ is evaluated numerically by dividing the interval $[-2,2]$ into bins and then computing the value of the overlap associated with the readout value in that bin. Each point has been averaged over 100 instances of the training set, and shaded regions around them correspond to 1 standard deviation. \textbf{Note:} in these plots the readouts are \emph{learnable} and drawn from a Gaussian prior, $P_v=\mathcal N(0,1)$.
    }
    \label{fig:gen_error_learnable_readouts}
\end{figure}

\begin{figure}[b!]
    \centering
    \includegraphics[width=0.496\linewidth,clip]{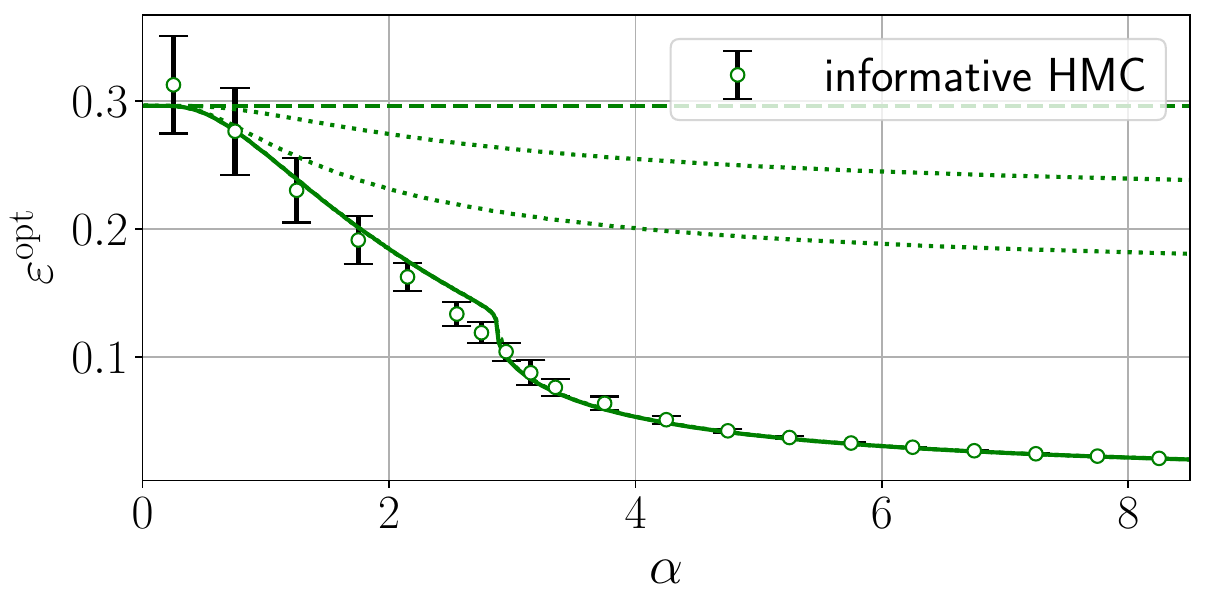}
    \includegraphics[width=0.463\linewidth,clip]{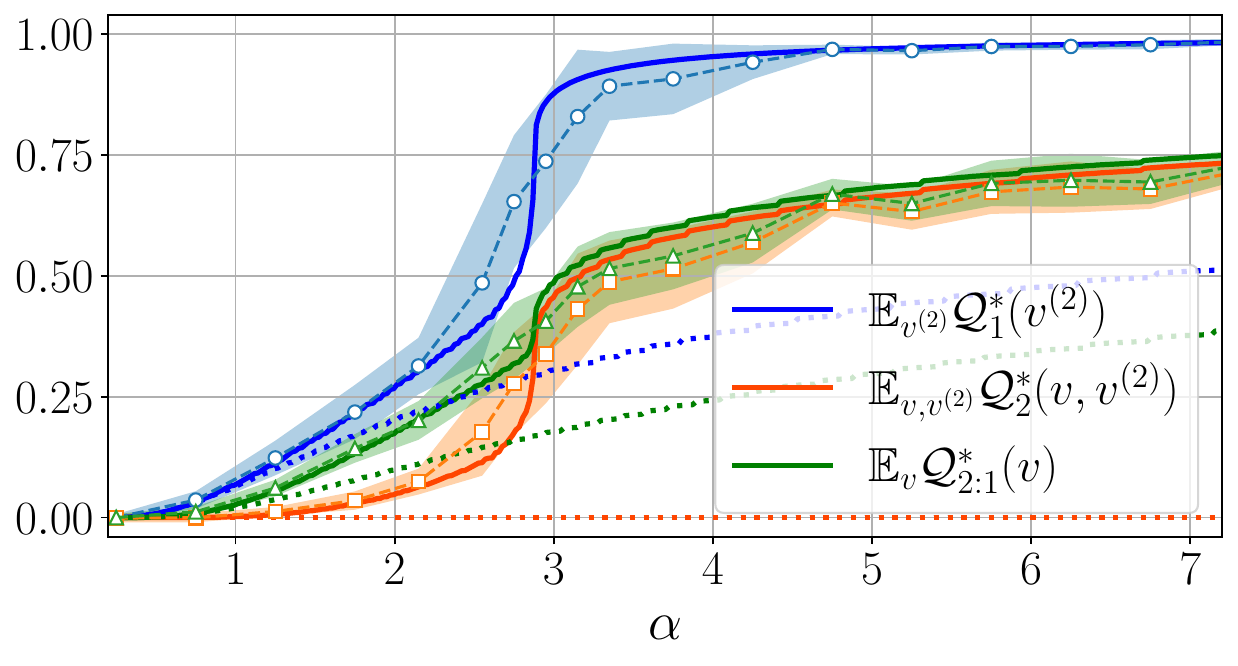}

    \caption{ \textbf{Left}: Theoretical prediction (green solid curve) of the Bayes-optimal mean-square generalisation error for $L=2$ with Gaussian inner weights, $\sigma(x)=\tanh(2x)/\sigma_{\tanh}, d=200, \gamma_1=\gamma_2=0.5,\Delta=0.2$ and different $P_v$ laws. The dashed and dotted lines have the same meaning as in \figurename~\ref{fig:3LNNgen_error_gauss}. Points are obtained with Hamiltonian Monte Carlo with informative initialisation. Each point has been averaged over 20 instances of the data, with error bars representing 1 standard deviation. The generalisation error is computed empirically from $10^4$ i.i.d. test samples. \textbf{Right}: Solid and dotted curves represent, respectively, the mean of different overlaps at equilibrium and in metastable specialised states, as function of the sampling ratio $\alpha$ for $L=2$ with Gaussian inner weights, $\sigma(x)=\tanh(2x)/\sigma_{\tanh},d=200, \gamma_1=\gamma_2=0.5,\Delta=0.2$. The shaded curves were obtained from informed HMC. Each point has been averaged over 20 instances of the training set, with 1 standard deviation depicted. \textbf{Note:} in these plots the readouts are \emph{learnable} and drawn from a Gaussian prior, $P_v=\mathcal N(0,1)$.
    }
    \label{fig:gen_error_learnable_readouts_L=2}
\end{figure}

\subsection{Generalisation errors for learnable readouts}
\label{subsec:learnable_readouts}
In the main we prove that, from an information theoretical point of view, having the readouts learnable or fixed to those of the teacher does not alter the problem. In particular, the generalisation errors predicted by our theory should be the same for both cases.

This is indeed numerically verified. In \figurename~\ref{fig:gen_error_learnable_readouts}, \ref{fig:gen_error_learnable_readouts_L=2} we show that HMC posterior samples, in the case of learnable readouts, yield the generalisation error predicted by our theory.

\section{Deep MLP} \label{app:deep}

\subsection{Details of the replica calculation}
\label{app:replicas_deep}

Let us take a multi-layer perceptron with $L = O(1)$ hidden layers. In order to be more general let us take at each layer a different activation function $$\sigma^{(l)}(z) = \sum_{\ell \neq 0,2} \frac{\mu^{(l)}_\ell}{\ell!} \He_\ell(z),$$ so that the MLP we will study is
\begin{align*}
        \mathcal{F}^{(L)}_{\btheta^0}(\bx):=\frac{\bv^{0\intercal}}{\sqrt{k^L}} \sigma^{(L)}\Big(\frac{\bW^{0(L)}}{\sqrt{k_{L-1}}} \sigma^{(L-1)}\Big(\frac{\bW^{0(L-1)}}{\sqrt{k_{L-2}}}\cdots \sigma^{(1)}\Big(\frac{\bW^{0(1)}}{\sqrt{k_0}}\bx\Big)\cdots\Big)\Big),
\end{align*}with $\btheta^0$ denoting the whole collection of the teacher's parameters.
To make the equations lighter we take centred and normalised activations $\EE_z \, \sigma^{(l)}(z)^2 = 1$, $\mu^{(l)}_0=0$, and we allow for different priors over the inner weights $\bW^{(l)} \sim P_{W_l}$. Importantly, we assume $\mu_2^{(l)}=0$. Treating terms associated with the second Hermite coefficient would require spherical integration and a measure relaxation analogous to the approach used in the shallow case. We leave this extension for future work.

Let us define the pre- and post-activations respectively as
\begin{align*}
\Big\{\bh^{(l)a}:=\frac1{\sqrt{k_{l-1}}}{\bW^{(l)a}\bx^{ (l-1)a}}\,,\quad
\bx^{(l)a}:=\sigma^{(l)}(\bh^{(l)a})\Big\}_{a=0}^s\,  ,
\end{align*}
where $\bx^{a (0)} := \bx, \, \forall \, a=0,\dots,s$ represents the input data and 
\begin{align*}
    \Big\{\lambda^a(\btheta^a):=\frac1{\sqrt{k_L}}{\bv^{0\intercal}}\sigma^{(L)}(\bh^{(L)a})\Big\}_{a=0}^s\,
\end{align*}    
are the replicated readouts. 

As for the shallow case, the key assumption is the joint Gaussianity of $\{ \lambda^a(\btheta^a)\}_{0\leq a\leq s}$ under the the common input randomness $\bx$. Since they are centred (recall $\mu_0^{(l)}=0$), in order to characterise their distribution it suffices to evaluate their covariance, which for analogy with the shallow case shall be denoted as
\begin{align}
    K^{ab}:=\EE_\bx\lambda^a(\btheta^a)
    \lambda^b(\btheta^b)=\frac{1}{k_L}\sum_{i,j=1}^{k_L}v^0_iv^0_j\;\EE_{\bx} \sigma^{(L)}(h_i^{(L)a})\sigma^{(L)}(h_j^{(L)b})\,.
\end{align}
To further simplify the above expectation, we need to use recursively Mehler's formula (see App.~\ref{app:hermite}) from the first pre-activation on, as follows. To begin with, define
\begin{align}
    \Omega^{(1)ab}_{i_1j_1}:= \EE_{\bx}h^{(1)a}_{i_1}h^{(1)b}_{j_1}
    =\EE_{\bx} \Big(\frac{\bW^{(1)a}_{i_1}\cdot\bx}{\sqrt{d}}\Big) \Big(\frac{\bW^{(1)b}_{j_1}\cdot\bx}{\sqrt{d}}\Big)=\frac{\bW_{i_1}^a\cdot\bW_{j_1}^b}{d}\,.
\end{align}This allows  us to compute the covariance of the second layer pre-activations under the same randomness:
\begin{align}
    \Omega_{i_2j_2}^{(2)ab}:= \EE_{\bx}h^{(2)a}_{i_2}h^{(2)b}_{j_2}
    = \EE_{\bx} \Big(\frac{\bW^{(2)a}_{i_2}\cdot\bx^{(1)a}}{\sqrt{k_1}}\Big) \Big(\frac{\bW^{(2)b}_{j_2}\cdot\bx^{(1)b}}{\sqrt{k_1}}\Big)\,.
\end{align} The expectation is resolved once one computes the covariance of the first layer post-activations by means of Mehler's formula, yielding
\begin{align}
    \Omega_{i_2j_2}^{(2)ab} =\frac{1}{k_1}\bW^{(2)a\intercal}_{i_2}\,\big((\mu_1^{(1)})^2\bOmega^{(1)ab}+g^{(1)}(\bOmega^{(1)ab})\big)\,\bW^{(2)b}_{j_2}
\end{align}where the function $g^{(1)}$, defined as in \eqref{eq:g_func_first}, is applied element-wise to the matrix argument. From this moment on we assume that pre-activations are Gaussian at each layer under the common randomness $\bx$, as they are always expressed in terms of rescaled sums. Under this assumption we can thus infer a generic recursion for the pre-activation covariances:
\begin{align}\label{eq:recursion_covar_pre-act}
    \Omega^{(l+1)ab}_{i_{l+1}j_{l+1}}:= \EE_{\bx}h^{(l+1)a}_{i_{l+1}}h^{(l+1)b}_{j_{l+1}}=\frac{1}{k_l}\bW^{(l+1)a\intercal}_{i_{l+1}}\,\big(
    (\mu_1^{(l)})^2\bOmega^{(l)ab} +g^{(l)}(\bOmega^{(l)ab})
    \big)\,\bW^{(l+1)b}_{j_{l+1}}
\end{align}
which naturally leads to the covariance we actually need for $K^{ab}$, i.e. $\Omega^{(L)ab}_{ij}=\EE_{\bx}h^{(L)a}_i h^{(L)b}_j$:
\begin{align}
    K^{ab} = \frac{(\mu_1^{(L)})^2}{k_{L}} \bv^{0\intercal} \,\big(\bOmega^{(L)ab}+g^{(L)}(\bOmega^{(L)ab})\big)\, \bv^0\,.
\end{align}

Let us define the following set of vectors and matrices for future convenience
\begin{equation}\label{eq:effective_readouts_wieghts}
\begin{aligned}
    \bW^{(l':l)a} &= \frac{1}{\sqrt{k_{l'-1}k_{l'-2}\dots k_l}}\bW^{(l')a} \bW^{(l'-1)a}\dots \bW^{(l)a}\, ,\\
    \bv^{(l)a} &= \frac{1}{\sqrt{k_l k_{l+1}\dots k_L}} \bW^{(l) a \intercal}\bW^{(l+1) a \intercal} \dots\bW^{(L) a \intercal}\bv^0 \, .
\end{aligned}    
\end{equation}
They will emerge from the computation due to the linear term in the Hermite expansion of the activation functions. The objects in \eqref{eq:effective_readouts_wieghts} represent effective readout vectors and weight matrices, that the student can learn independently from the actual weights and readouts. With this notation the post-activation covariance reads (recall $k_0=d$)
\begin{equation}
\begin{aligned}
    K^{ab} &= \frac{(\mu_1^{(L)}\mu_1^{(L-1)}\dots \mu_1^{(1)})^2}{k_0} \bv^{a (1) \intercal } \bv^{b (1)} 
    + \frac{(\mu_1^{(L-1)}\mu_1^{(L-2)}\dots \mu_1^{(1)})^2}{k_1} \bv^{a (2) \intercal }  g^{(1)} (\bOmega^{ab(1)}) \bv^{b (2)} \\
    &\quad + \dots + \frac{( \mu_1^{(1)})^2}{k_{L-1}} \bv^{a (L) \intercal }g^{(L-1)} (\bOmega^{ab(L-1)}) \bv^{b (L)} 
    + \frac{1}{k_{L}} \bv^{0 \intercal }g^{(L)} (\bOmega^{ab(L)}) \bv^{0}
\end{aligned}    
\end{equation}
As in \eqref{eq:diagonal_omega} we will assume that for all $l = 1,\dots, L$
\begin{align}
    \Omega^{ab(l)}_{i_li_l}=O(1)\,,\quad \Omega^{ab(l)}_{i_lj_l}=O\Big(\frac{1}{\sqrt{k_{l-1}}}\Big)\text{ for }i_l\neq j_l\,.
\end{align}
Therefore, only the diagonal elements of the matrix $g^{(l)}(\bOmega^{ab(l)})$ will contribute in the thermodynamic limit. 

Note that the overlap $\frac{1}{k_0}\bv^{(1)a\intercal } \bv^{(1)b}$ is analogous to the OP described by \cite{cui2023bayes} in the deep setting; $\bv^{(1)0}$ is the only feature of the target function that is learnable in the $n \propto d$ regime, which we will consider as known, namely all the overlaps between $\bv^{(1)a}$'s are set to $1$. Analogously to what happens in the shallow case, the components of the other $\bv^{(l)a}$'s enter trivially the energetic term. Specifically, only those components of $\bv^{(l)a}$ that are perfectly reconstructible by the student would enter the energy, namely those for which the associated $\Omega_{i_li_l}^{(l)ab}=O(1)$ and not smaller. Hence without loss of generality one can assume that all the $\bv^{(l)a}$ are set to $\bv^{(l)0}$ as in the shallow case. This allows to considerably simplify the equations and leads to
\begin{equation}
\begin{aligned}
    K^{ab} &=(\mu_1^{(L)}\mu_1^{(L-1)}\dots \mu_1^{(1)})^2 
    + \frac{(\mu_1^{(L-1)}\mu_1^{(L-2)}\dots \mu_1^{(1)})^2}{k_1}\sum_{i_1}^{k_1} (v^{0 (2)}_{i_1})^2 g^{(1)} \Big(\Omega^{ab(1)}_{i_1 i_1} \Big) \\
    &\quad + \dots + \frac{( \mu_1^{(1)})^2}{k_{L-1}} \sum_{i_{L-1}}^{k_{L-1}} (v^{0 (L)}_{i_{L-1}})^2 g^{(L-1)} \Big(\Omega^{ab(L-1)}_{i_{L-1} i_{L-1}} \Big)  
    + \frac{1}{k_{L}} \sum_{i_L}^{k_L} (v^{0}_{i_L})^2 g^{(L)} \Big(\Omega^{ab(L)}_{i_L i_L} \Big) 
\end{aligned} 
\label{eq:K_deep_app_final}
\end{equation}
with, additionally
\begin{equation}
\begin{aligned}
    \Omega^{ab(l+1)}_{i_{l+1} i_{l+1}}
    &\approx \frac{(\mu_1^{(l)})^2}{k_{l}} \bW^{a (l+1) \intercal }_{i_{l+1}} \bOmega^{ab(l)} \bW^{b(l+1)}_{i_{l+1}}+ \frac{1}{k_{l}}\sum_{i_l}^{k_l} W^{(l+1)a}_{i_{l+1}i_l} W^{(l+1)b}_{i_{l+1}i_l} g^{(l)} \Big(\Omega^{(l)ab}_{i_l i_l} \Big)\,.
\end{aligned}    
\end{equation}
Here is a difference with the shallow case: the way the replicas $\bW^{(l+1)a}$ align with one another may depend not only on the ``left'' indices, $i_{l+1}$ in the formula above, but can be affected also by the reconstruction performance from previous layers, encoded in $\Omega_{i_li_l}^{(l)ab}$. The values of $\Omega_{i_li_l}^{(l)ab}$ are themselves driven by the values of the associated $v^{(l+1)0}_{i_l}$ that appear coupled to them in the energy through $K^{ab}$ as above. Hence, we can choose to label the values of $\Omega_{i_li_l}^{(l)ab}$ through those of $v_{i_l}^{(l+1)0}$. We denote these values as $\mathsf{v}^{(l+1)}$, collected in the sets $\mathsf{V}^{(l+1)}$, and $\mathcal{I}_{\mathsf{v}^{(l+1)}}=\{i_l\mid v_{i_l}^{(l+1)0}=\mathsf{v}^{(l+1)}\}$.

This brings us to defining the following overlaps:
\begin{align}
    \label{eq:direct_layer_overlaps}
    \mathcal{Q}^{ab}_{l}(i_{l},\mathsf{v}^{(l)}) &= \frac{1}{|\mathcal{I}_{\mathsf{v}^{(l)}}|}\sum_{i_{l-1}\in \mathcal{I}_{\mathsf{v}^{(l)}}}W_{i_{l}i_{l-1}}^{(l)a} W_{i_{l}i_{l-1}}^{(l)b}\, , \\
    \label{eq:mixed_layer_overlaps} \mathcal{Q}^{ab}_{{l':l}}(i_{l'},\mathsf{v}^{(l)}) &=\frac{1}{|\mathcal{I}_{\mathsf{v}^{(l)}}|}\sum_{i_{l-1}\in \mathcal{I}_{\mathsf{v}^{(l)}}}W_{i_{l'} i_{l-1}}^{(l':l)a} W_{i_{l'} i_{l-1}}^{(l':l)b} \, .
\end{align}
Using \eqref{eq:recursion_covar_pre-act}, the diagonal elements of each $\bOmega^{(l)ab}$ in terms of those overlaps read
\begin{equation}
\begin{aligned}
    \bOmega^{ab(l+1)}_{i_{l+1} i_{l+1}}
    &\approx \frac{(\mu_1^{(l)})^2}{k_{l}} \bW^{a (l+1) \intercal }_{i_{l+1}} \bOmega^{ab(l)} \bW^{b(l+1)}_{i_{l+1}} +  \sum_{\mathsf{v}^{(l+1)}\in\mathsf{V}^{(l+1)}}\frac{|\mathcal{I}_{\mathsf{v}^{(l+1)}}|}{k_l}\mathcal{Q}_{{l+1}}^{ab}({i_{l+1}},\mathsf{v}^{(l+1)}) \frac{1}{|\mathcal{I}_{\mathsf{v}^{(l+1)}}|}\sum_{i_l\in \mathcal{I}_{\mathsf{v}^{(l+1)}}} g^{(l)} \Big(\Omega^{ab(l)}_{i_l i_l} \Big)\,.
\end{aligned}    
\end{equation}
This is the first step of the recursion; in order to express everything in terms of the overlaps $\mathcal{Q}$, we need to express the first term as well in terms of diagonal elements, keeping in mind that $\Omega_{i_1i_1}^{(1)ab}=\bW^{(1)a\intercal}_{i_1} I_d\bW^{(1)b}_{i_1}/d=:\mathcal{Q}_{1}^{ab}(i_1)$. In other words, the first overlap is not labelled by any other index, as the $\bW^{(1)}$'s here do not sandwich any matrix other than the identity. Hence no inhomogeneity can arise. An analogous reasoning holds for $\mathcal{Q}^{ab}_{{l+1:1}}(i_{l+1})$.

At a generic step of the recursion, we have
\begin{equation}
\begin{aligned}
    \Omega^{ab(l+1)}_{i_{l+1} i_{l+1}} &\approx (\mu_1^{(l)}\dots \mu_1^{(1)})^2 \mathcal{Q}^{ab}_{{l+1:1}}(i_{l+1}) + (\mu_1^{(l)}\dots \mu_1^{(2)})^2
    \sum_{\mathsf{v}^{(2)} \in\mathsf{V}^{(2)}}\frac{|\mathcal{I}_{\mathsf{v}^{(2)}}|}{k_1}\mathcal{Q}_{{l+1:2}}^{ab}({i_{l+1}},\mathsf{v}^{(2)}) \frac{1}{|\mathcal{I}_{\mathsf{v}^{(2)}}|}\sum_{i_l\in \mathcal{I}_{\mathsf{v}^{(2)}}}
    g^{(1)} \Big(\Omega^{(1)ab}_{i_1 i_1} \Big) \\
    &+ \dots + (\mu_1^{(l)})^2 
    \sum_{\mathsf{v}^{(l)} \in\mathsf{V}^{(l)}}\frac{|\mathcal{I}_{\mathsf{v}^{(l)}}|}{k_{l-1}}
    \mathcal{Q}_{{l+1:l}}^{ab}({i_{l+1}},\mathsf{v}^{(l)}) \frac{1}{|\mathcal{I}_{\mathsf{v}^{(l)}}|}\sum_{i_{l-1}\in \mathcal{I}_{\mathsf{v}^{(l)}}} g^{(l-1)} \Big(\Omega^{(l-1)ab}_{i_{l-1} i_{l-1}} \Big) \\
    &+ \sum_{\mathsf{v}^{(l+1)} \in\mathsf{V}^{(l+1)}}\frac{|\mathcal{I}_{\mathsf{v}^{(l+1)}}|}{k_{l}}\mathcal{Q}_{{l+1}}^{ab}({i_{l+1}},\mathsf{v}^{(l+1)}) \frac{1}{|\mathcal{I}_{\mathsf{v}^{(l+1)}}|}\sum_{i_l\in \mathcal{I}_{\mathsf{v}^{(l+1)}}} g^{(l)} \Big(\Omega^{(l)ab}_{i_l i_l} \Big) \, ,
\end{aligned}
\label{eq:Omega_deep_app_final}
\end{equation}
where $\bOmega^{ab(1)}_{i_1 i_1} = \mathcal{Q}_{1}(i_1)$. This defines the full recursion, which allows one to compute the covariance $K^{ab}$ only in terms of the above mentioned overlaps.     

As the derivation may become very cumbersome, we specialise to the $L=2$ setting in the following section.

\subsubsection{Two hidden layers \texorpdfstring{$L=2$}{}}
In the case of two hidden layers networks the equation for the covariance is
\begin{equation}
\begin{aligned}
    K^{ab} &= (\mu_1^{(2)} \mu_1^{(1)})^2 
    + \frac{( \mu_1^{(2)})^2}{k_1}\sum_{i_1}^{k_1} (v^{0 (2)}_{i_1})^2 g^{(1)} \Big(\mathcal{Q}_{1}^{ab}(i_1) \Big)  \\
    & \quad
    + \frac{1}{k_{2}} \sum_{i_2}^{k_2} (v^{0}_{i_2})^2 g^{(2)} \Big((\mu_1^{(1)})^2 \mathcal{Q}^{ab}_{2:1}(i_2) + \sum_{\mathsf{v}^{(2)} \in\mathsf{V}^{(2)}}\frac{|\mathcal{I}_{\mathsf{v}^{(2)}}|}{k_{1}}\mathcal{Q}_{2}^{ab}({i_{2}},\mathsf{v}^{(2)}) \frac{1}{|\mathcal{I}_{\mathsf{v}^{(2)}}|}\sum_{i_1\in \mathcal{I}_{\mathsf{v}^{(2)}}} g^{(1)} (\mathcal{Q}_{1}^{ab}(i_1))\Big) \, .
\end{aligned}    
\end{equation}
Importantly, the index $i_1$ is linked only to the vector $\bv^{0(2)}$ and the overlap $\mathcal{Q}_{1}^{ab}$, while the index $i_2$ to the vector $\bv^{0}$ and the overlaps $\mathcal{Q}_{2}^{ab}$ and $\mathcal{Q}_{2:1}^{ab}$. We can relabel the values of $\calQ_{1}$ with those of $\mathsf{v}^{(2)}$: $ \mathcal{Q}_{1}(\mathsf{v^{(2)}})=\calQ_{1}(i_1)$ for all $i_1\in\mathcal{I}_{\mathsf{v}^{(2)}}$. An analogous relabelling can be carried out for $\calQ_2(i_2,\mathsf{v}^{(2)})$ in the index $i_2$, based on the values of $v^0_{i_2}$. After the relabelling, one could also redefine the overlaps through partial traces, in order to mimic the notation from the main, as follows:
\begin{align}
    \label{eq:direct_layer_overlaps_traces}
    \mathcal{Q}^{ab}_{l}(\mathsf{v}^{(l+1)},\mathsf{v}^{(l)}) &= \frac{1}{|\mathcal{I}_{\mathsf{v}^{(l+1)}}||\mathcal{I}_{\mathsf{v}^{(l)}}|} \sum_{i_{l}\in \mathcal{I}_{\mathsf{v}^{(l+1)}}}\sum_{i_{l-1}\in \mathcal{I}_{\mathsf{v}^{(l)}}}W_{i_{l}i_{l-1}}^{(l)a} W_{i_{l}i_{l-1}}^{(l)b}\, , \\
    \label{eq:mixed_layer_overlaps_traces} 
    \mathcal{Q}^{ab}_{{l':l}}(\mathsf{v}^{(l'+1)},\mathsf{v}^{(l)}) &=\frac{1}{|\mathcal{I}_{\mathsf{v}^{(l'+1)}}||\mathcal{I}_{\mathsf{v}^{(l)}}|}
    \sum_{i_{l'}\in \mathcal{I}_{\mathsf{v}^{(l'+1)}}}
    \sum_{i_{l-1}\in \mathcal{I}_{\mathsf{v}^{(l)}}}W_{i_{l'} i_{l-1}}^{(l':l)a} W_{i_{l'} i_{l-1}}^{(l':l)b} \, .
\end{align}
Consider also that $\frac{|\mathcal{I}_{\mathsf{v}^{(2)}}|}{k_{1}}\to P_{v^{(2)}}(\mathsf{v}^{(2)})$, and $\frac{|\mathcal{I}_{\mathsf{v}}|}{k_{2}}\to P_v(\mathsf{v})$. This allows us to recast $K^{ab}$ in the asymptotic limit as
\begin{equation}
\begin{aligned}
    K^{ab} &= (\mu_1^{(2)} \mu_1^{(1)})^2 
    + ( \mu_1^{(2)})^2 \EE_{v^{(2)} \sim P_{v^{(2)}}} (v^{(2)})^2 g^{(1)} \Big(\mathcal{Q}_{1}^{ab}(v^{(2)}) \Big)  \\
    & \quad
    + \EE_{v \sim P_{v}} (v)^2 g^{(2)} \Big((\mu_1^{(1)})^2 \mathcal{Q}^{ab}_{2:1}(v) + \EE_{v^{(2)} \sim P_{v^{(2)}}}\mathcal{Q}^{ab}_{2}(v,v^{(2)}) g^{(1)}\big(\mathcal{Q}^{ab}_{1}(v^{(2)})\big)\Big) \, .
\end{aligned}    
\end{equation}

Notice that as soon as the covariance of the post-activation is written in terms of overlaps, this fully determines the energetic part appearing in the free entropy. Indeed, in the RS ansatz, $\bK$ appears as in \eqref{eq:K_matrix_RS}, where
\begin{equation}
\begin{aligned}
\label{eq:Kd-K-deep}
    K^{(2)}(\bar\calQ) &= (\mu_1^{(2)} \mu_1^{(1)})^2 
    + ( \mu_1^{(2)})^2 \EE_{v^{(2)} \sim P_{v^{(2)}}} (v^{(2)})^2 g^{(1)} \Big(\mathcal{Q}_{1}(v^{(2)}) \Big)  \\
    & \quad
    + \EE_{v \sim P_{v}} (v)^2 g^{(2)} \Big((\mu_1^{(1)})^2 \mathcal{Q}_{2:1}(v) + \EE_{v^{(2)} \sim P_{v^{(2)}}}\mathcal{Q}_{2}(v,v^{(2)}) g^{(1)}\big(\mathcal{Q}_{1}(v^{(2)})\big)\Big), \\
    K_{\rm d}&=1.
\end{aligned}
\end{equation}
while $\rho_K=K_{\rm d}$ and $m_K=K$ using the Nishimori identities. Therefore, the energetic term will be equal to \eqref{eq:energetic_pot_RS_BayesOpt}, with $K=K^{(2)}(\bar\calQ)$ and $K_{\rm d}=1$ as defined in the last equations.

Let us now discuss the entropic contribution, associated with the OPs defined above.
This can be written as
\begin{align}
    e^{F_S}&=\int d\bW^{a (2:1)} \int \prod_{a=0}^{s}\prod_{l=1}^{2}dP_{W_l}(\bW^{a (l)})\delta\Big(\bW^{a (2:1)}-\frac{
    \bW^{a (2)}\bW^{a(1)}}{\sqrt{k_1}}\Big)\nonumber\\
    &\times \prod_{a\leq b}^{0,s}\prod_{\mathsf{v}\in \mathsf{V}} \prod_{\mathsf{v^{(2)}}\in \mathsf{V^{(2)}}}\delta(\,|\mathcal{I}_\mathsf{v}||\mathcal{I}_\mathsf{v^{(2)}}|\mathcal{Q}_{2}^{ab}(\mathsf{v}, \mathsf{v^{(2)}})-\sum_{{i_2}\in\mathcal{I}_\mathsf{v}}\sum_{i_1\in\mathcal{I}_\mathsf{v^{(2)}}}{W^{(2)a}_{i_2 i_1}W_{i_2 i_1}^{(2)b}})\nonumber\\
    & \times \prod_{a\leq b}^{0,s}\prod_{\mathsf{v^{(2)}}\in \mathsf{V^{(2)}}} \delta(d|\mathcal{I}_\mathsf{v^{(2)}}|\,\mathcal{Q}_{1}^{ab}(\mathsf{v^{(2)}})-\sum_{i_1\in\mathcal{I}_\mathsf{v^{(2)}}}{\bW^{a(1)\intercal}_{i_1}\bW_{i_1}^{b (1)}}) \nonumber  \\
    &\times\prod_{a\leq b}^{0,s}\prod_{\mathsf{v}\in \mathsf{V}}\delta(d|\mathcal{I}_\mathsf{v}|\,\mathcal{Q}_{2:1}^{ab}(\mathsf{v})-\sum_{{i_2}\in\mathcal{I}_\mathsf{v}} {\bW^{a(2:1)\intercal}_{i_2}\bW_{i_2}^{b(2:1)}}).
\end{align}
Besides the labelling of $\calQ_2$ in terms of two indices $\mathsf{v},\mathsf{v}^{(2)}$, another important difference with respect to the shallow case is the presence of the overlap $\mathcal{Q}_{2:1}(v)$ between the replicated matrices $\bW^{(2:1)a}=\bW^{a(2)}\bW^{a(1)}/\sqrt{k_1}$. This overlap will depend on the alignment between the first and second layer weights. In order to encode this dependence we use a similar relaxation of the measure as the one described in Section \ref{subsec:entropic_potential}. We thus define
\begin{align}
    dP((\bW^{(2:1)a})\mid \bm{\calQ}_{1},\bm{\calQ}_2)&\propto \prod_{a=0}^s d\bW^{(2:1)a}\int\prod_{a=0}^s\prod_{l=1}^2 dP_{W_l}(\bW^{(l)a})\delta\Big(\bW^{ (2:1)a}-\frac{
    \bW^{ (2)a}\bW^{(1)a}}{\sqrt{k_1}}\Big) \nonumber\\
    &\times \prod_{a\leq b}^{0,s}\prod_{\mathsf{v}\in \mathsf{V}} \prod_{\mathsf{v^{(2)}}\in \mathsf{V^{(2)}}}\delta(\,|\mathcal{I}_\mathsf{v}||\mathcal{I}_\mathsf{v^{(2)}}|\mathcal{Q}_{2}^{ab}(\mathsf{v}, \mathsf{v^{(2)}})-\sum_{{i_2}\in\mathcal{I}_\mathsf{v}}\sum_{i_1\in\mathcal{I}_\mathsf{v^{(2)}}}{W^{(2)a}_{i_2 i_1}W_{i_2 i_1}^{(2)b}})\nonumber\\
    & \times \prod_{a\leq b}^{0,s}\prod_{\mathsf{v^{(2)}}\in \mathsf{V^{(2)}}} \delta(d|\mathcal{I}_\mathsf{v^{(2)}}|\,\mathcal{Q}_{1}^{ab}(\mathsf{v^{(2)}})-\sum_{i_1\in\mathcal{I}_\mathsf{v^{(2)}}}{\bW^{a(1)\intercal}_{i_1}\bW_{i_1}^{b (1)}})
    \label{eq:dPW21_app}
\end{align}where the normalisation constant is implicit. The aim is now to relax this measure to the one of a product of two Ginibre matrices with a proper tilt, given by the coupling between replicas. The relaxation we choose is the one with a matched second moment $\frac{1}{d|\mathcal{I}_\mathsf{v}|}\sum_{{i_2}\in\mathcal{I}_\mathsf{v}}\EE[ \bW^{(2:1)a\intercal}_{i_2} \bW^{(2:1)b}_{i_2} \mid \bm{\calQ}_{1},\bm{\calQ}_2]$ where $\EE[\cdot \mid \bm{\calQ}_{1},\bm{\calQ}_2]$ denotes the expectation w.r.t.\ the conditional measure defined above. As done in App.~\ref{app:second_moment_P(S2|QW)}, by rewriting Dirac deltas in Fourier form, the measure decouples and the calculation goes through, yielding asymptotically
\begin{align}\label{eq:moment_matching_W21}
    \frac{1}{d|\mathcal{I}_{\mathsf{v}}|}\sum_{i_2\in\mathcal{I}_{\mathsf{v}}}\EE[ \bW^{(2:1)a\intercal}_{i_2} \bW^{(2:1)b}_{i_2} \mid \bm{\calQ}_{1},\bm{\calQ}_2]\approx 
    \EE_{v^{(2)}\sim P_{v^{(2)}}}\calQ^{ab}_2(\mathsf{v},v^{(2)})\calQ^{ab}_{1}(v^{(2)})\,.
\end{align}In order to fix this moment in our relaxation we thus need a Lagrange multiplier for each value $\mathsf{v}\in\mathsf{V}$:
\begin{align}
    d\bar{P}((\bW^{(2:1)a})\mid\bm{\calQ}_{1},\bm{\calQ}_2)= 
    \prod_{\mathsf{v}\in\mathsf{V}} V(\bm{\tau}_\mathsf{v})^{-1} \prod_{a=0}^sd\calN(\bU^a_\mathsf{v})d\calN(\bV^a)
    e^{
    \sum_{a<b,0}^s \tau^{ab}_\mathsf{v}\Tr\bU^a_\mathsf{v}\bV^a(\bU^b_\mathsf{v}\bV^b)^\intercal
    }
\end{align}where $\bU^a_\mathsf{v}\in\mathbb{R}^{|\mathcal{I}_\mathsf{v}|\times k_1}$, $\bV^a\in\mathbb{R}^{k_1\times d}$ are matrices with i.i.d. Gaussian elements (their factorised measure being synthetically denoted by $\calN$), $\bW^{(2:1)a} = \bU^a_\mathsf{v} \bV^a$ and $\bm{\tau}_\mathsf{v}=(\tau^{ab}_\mathsf{v})_{a< b=0,\ldots,s}$.

With this relaxation we have that
\begin{align}
    e^{F_S}&=V_{W_1}^{k_1d}(\bm{\calQ}_{1})V_{W_2}^{k_1k_2}(\bm{\calQ}_2)\int d \hat{\bm{\calQ}}_{2:1}\int 
\prod_{\mathsf{v}\in\mathsf{V}} V(\bm{\tau}_\mathsf{v})^{-1} \prod_{a=0}^sd\calN(\bU^a_\mathsf{v})d\calN(\bV^a)
    e^{
    \sum_{a<b,0}^s (\tau^{ab}_\mathsf{v}+\hat\calQ^{ab}_{2:1}(\mathsf{v}))\Tr\bU^a_\mathsf{v}\bV^a(\bU^b_\mathsf{v}\bV^b)^\intercal
    }
    \nonumber\\
    &\times e^{-d\sum_{a<b,0}^s\sum_{\mathsf{v}\in\mathsf{V}}|\mathcal{I}_\mathsf{v}|\hat{\calQ}^{ab}_{2:1}(\mathsf{v}){\calQ}^{ab}_{2:1}(\mathsf{v}) }\,.
\end{align}
Where not specified, integrals over OPs and their Fourier conjugates run over all replica indices and $\mathsf{v}$ values. Standard steps as the ones in App.~\ref{app:replicas}, after taking the $0$ replica limit, yield
\begin{align}\label{eq:f_RS_2layers}
    f^{(2)}_{\rm RS}&:=  \phi_{P_{\rm out}}(K^{(2)}(\bar\calQ);1)+\frac{\gamma_1}{\alpha} \EE_{v^{(2)} \sim P_{v^{(2)}}} \Big[\psi_{P_{1}}(\hat{\mathcal{Q}}_{1}(v^{(2)})) -\frac{1}{2} \mathcal{Q}_{1}(v^{(2)}) \hat{\mathcal{Q}}_{1}(v^{(2)}) \Big] \nonumber\\
    &+\frac{\gamma_1 \gamma_2}{\alpha} \EE_{v \sim P_v, v^{(2)} \sim P_{v^{(2)}}} \Big[\psi_{P_{2}}(\hat{\mathcal{Q}}_{2}(v, v^{(2)})) -\frac{1}{2} \mathcal{Q}_{2}(v,v^{(2)}) \hat{\mathcal{Q}}_{2}(v,v^{(2)}) \Big] \nonumber\\
    &+\frac{\gamma_2}{\alpha}\EE_{v\sim P_v}\big[\frac{1}{2}\hat{\calQ}_{2:1}(v)(1-{\calQ}_{2:1}(v))-\iota_v(\tau_v+\hat{\calQ}_{2:1}(v))+\iota_v(\tau_v)\big]
\end{align}
where $\iota_v(x)$ is the MI of the following matrix denoising problem:
\begin{align}\label{eq:RLE_deep_relaxation}
    \bY_\mathsf{v}(x) =\sqrt{x}\frac{\bU^0_\mathsf{v}\bV^0}{\sqrt{k_1}}+\bZ_\mathsf{v}\  \in\mathbb{R}^{|\mathcal{I}_\mathsf{v}|\times d}
\end{align}with $\bU^0_\mathsf{v}\in\mathbb{R}^{|\mathcal{I}_\mathsf{v}|\times k_1}$, $\bV^0\in\R^{k_1\times d}$ and $\bZ_\mathsf{v}\in\R^{|\mathcal{I}_\mathsf{v}|\times d}$ three Ginibre matrices. Furthermore we assume $|\mathcal{I}_\mathsf{v}|/k_2\to P_v(\mathsf{v})$, $k_2/d\to \gamma_2$, $k_1/d\to\gamma_1$ and $n/d^2\to\alpha$. Hence $|\mathcal{I}_\mathsf{v}|/d\to P_v(\mathsf{v})\gamma_2$, and 
\begin{align}\label{eq:iota_v_deep}
    \iota_\mathsf{v}(x):=\lim_{d\to\infty}\frac{x\EE\|\bU^0_\mathsf{v}\bV^0\|^2}{2k_1|\mathcal{I}_\mathsf{v}|d}-\frac{1}{|\mathcal{I}_\mathsf{v}|d}\EE\ln\int_{\mathbb{R}^{|\mathcal{I}_\mathsf{v}|\times k_1}} d\mathcal{N}(\bU_\mathsf{v})
    \int_{\mathbb{R}^{k_1\times d}} d\mathcal{N}(\bV)\exp \Tr\Big(
    \sqrt{\frac{x}{k_1}}\bY_\mathsf{v}(x)(\bU\bV)^\intercal-\frac{x}{2k_1}\bU\bV(\bU\bV)^\intercal
    \Big).
\end{align}The above matrix integral can be solved by means of the rectangular spherical integral, whose asymptotics is studied in \cite{GUIONNET2023_rect}.
Since we will not be needing explicitly this expression, we report only the one for the associated mmse$(x)$ function, as derived in \cite{pourkamali2023rectangular}:
\begin{align}\label{eq:asymmetric_Mdenoising_MMSE}
    {\rm mmse}_\mathsf{v}(x)&:=2\frac{d}{dx}\iota_\mathsf{v}(x)=\lim_{d\to\infty}\frac{1}{k_1|\mathcal{I}_\mathsf{v}|d}\EE\|\bU^0_\mathsf{v}\bV^0-\langle\bU_\mathsf{v}\bV\rangle\|^2\nonumber\\
    &\,=\frac{1}{x}\Big[1-P_v(\mathsf{v})\gamma_2\Big(\frac{1}{P_v(\mathsf{v})\gamma_2}-1\Big)^2\int \frac{\rho_{\bY_\mathsf{v}(x)}(y)}{y^2}dy-P_v(\mathsf{v})\gamma_2\frac{\pi^2}{3}\int \rho_{\bY_\mathsf{v}(x)}^3(y)dy\Big]\,.
\end{align}Here $\rho_{\bY_\mathsf{v}(x)}$ is the singular value density density of $\frac{\bY_\mathsf{v}(x)}{\sqrt{|\mathcal{I}_\mathsf{v}|}}$ obtained from a rectangular free convolution as described in the main. The moment matching condition \eqref{eq:moment_matching_W21} in its replica symmetric version thus reads
\begin{align}
    {\rm mmse}_\mathsf{v}(\tau_\mathsf{v})=1-\EE_{v^{(2)}\sim P_{v^{(2)}}}\calQ_2(\mathsf{v},v^{(2)})\calQ_{1}(v^{(2)})\,.
\end{align}

Recalling that $\tau_\mathsf{v}=\tau_\mathsf{v}(\mathcal{Q}_{2},\calQ_{1})$, the saddle point equations are obtained by equating the gradient of $f^{(2)}_{\rm RS}$ w.r.t.\ the order parameters to $0$:
\begin{equation}
\label{eq:deep_Saddlepoint_eq}
\begin{sqcases}
&{\mathcal{Q}}_{1}(\mathsf{v^{(2)}}) = \EE_{w_1^0\sim P_{W_1},\xi\sim \mathcal{
N}(0,1)} [ w_1^0 \thav{w_1}_{\Hat{\mathcal{Q}}_{1}(\mathsf{v^{(2)}})} ], \\
& P_{v^{(2)}}(\mathsf{v}^{(2)})\hat{\mathcal{Q}}_{1}(\mathsf{v^{(2)}}) = 2\frac{\alpha}{\gamma_1} \partial_{\mathcal{Q}_{1}(\mathsf{v}^{(2)})} \phi_{P_{\rm out}}(K^{(2)}(\bar\calQ);1)+\frac{\gamma_2}{\gamma_1}\EE_{v \sim P_v}\big[\calQ_{2:1}(v)\!-\!\EE_{v^{(2)}\sim P_{v^{(2)}}}\calQ_2(v,v^{(2)})\calQ_{1}(v^{(2)})\big] \partial_{\calQ_{1}(\mathsf{v}^{(2)})}\tau_{v}\\
& \mathcal{Q}_{2}(\mathsf{v},\mathsf{v^{(2)}}) = \EE_{w_2^0\sim P_{W_2},\xi\sim \mathcal{
N}(0,1)} [ w_2^0 \thav{w_2}_{\Hat{\mathcal{Q}}_{2}(\mathsf{v},\mathsf{v^{(2)}})} ],\\
& P_{v}(\mathsf{v})P_{v^{(2)}}(\mathsf{v}^{(2)})\hat{\mathcal{Q}}_{2}(\mathsf{v},\mathsf{v^{(2)}}) = 2\frac{\alpha}{\gamma_1 \gamma_2} \partial_{\mathcal{Q}_{2}(\mathsf{v},\mathsf{v^{(2)}})} \phi_{P_{\rm out}}(K^{(2)}(\bar\calQ);1)\\
&\qquad\qquad\qquad\qquad\qquad\qquad\qquad\qquad\qquad+\frac{P_{v}(\mathsf{v})}{\gamma_1}\big[\calQ_{2:1}(\mathsf{v})-\EE_{\mathsf{v}^{(2)}\sim P_{v^{(2)}}}\calQ_2(\mathsf{v},v^{(2)})\calQ_{1}(v^{(2)})\big]\partial_{\calQ_{2}(\mathsf{v},\mathsf{v}^{(2)})}\tau_\mathsf{v}\\
&{\mathcal{Q}}_{2:1}(\mathsf{v}) = 1-{\rm mmse}_\mathsf{v}(\tau_\mathsf{v}+\hat\calQ_{2:1}(\mathsf{v}))\\
&P_{v}(\mathsf{v})\hat{\mathcal{Q}}_{2:1}(\mathsf{v}) = 2\frac{\alpha}{\gamma_2} \partial_{\mathcal{Q}_{2:1}(\mathsf{v})} \phi_{P_{\rm out}}(K^{(2)}(\bar\calQ);1).
\end{sqcases}
\end{equation}

\subsubsection{Three or more hidden layers}

To extend the derivations to an arbitrary number of layers, one has to find a way to write the entropic contributions of the overlaps entering the energetic part, see for example~\eqref{eq:Omega_deep_app_final}. The challenging part is due to the overlaps $\calQ_{l':l}$, defined in~\eqref{eq:mixed_layer_overlaps}. Indeed, the analogous of the measure~\eqref{eq:dPW21_app} over the matrices $(\bW^{(l':l)a})$ should be conditioned over all the overlaps defined from subsets of the indices $\{l',l'-1,\cdots,l\}$, which encode the possibility of all possible partial reconstructions (of $(\bW^{(l'':l''')0})$). We leave this challenge for future works. Here, we focus on the case of activations with $\mu_1^{(l)} = 0$ (in addition to $\mu_0^{(l)}=\mu_2^{(l)}=0$). In this case, the post-activation covariance~\eqref{eq:K_deep_app_final} is easy to write:
\begin{equation}
\begin{aligned}
    K^{ab} &=\frac{1}{k_{L}} \sum_{i_L}^{k_L} (v^{0}_{i_L})^2 g^{(L)} \Big(\Omega^{ab(L)}_{i_L i_L} \Big) \\
    \Omega^{ab(l+1)}_{i_{l+1} i_{l+1}} &\approx \sum_{\mathsf{v}^{(l+1)} \in\mathsf{V}^{(l+1)}}\frac{|\mathcal{I}_{\mathsf{v}^{(l+1)}}|}{k_{l}}\mathcal{Q}_{{l+1}}^{ab}({i_{l+1}},\mathsf{v}^{(l+1)}) \frac{1}{|\mathcal{I}_{\mathsf{v}^{(l+1)}}|}\sum_{i_l\in \mathcal{I}_{\mathsf{v}^{(l+1)}}} g^{(l)} \Big(\Omega^{(l)ab}_{i_l i_l} \Big) \, .
\end{aligned} 
\end{equation}
In this recursion, only single-layer overlaps are entering, which have a simple entropic contribution. Moreover, no effective readout $\bv^{(l)}$ is entering this equation, which means that nothing distinguishes neurons in hidden layers $l<L$. By our exchangeability hypothesis on neurons connected to readouts with the same amplitude, the order parameters can then be written as
\begin{align}
    Q_l^{ab} &= \frac{1}{k_l k_{l-1}} \Tr \bW^{(l)a} \bW^{(l)b\intercal} \quad\quad \text{for } l=1,\ldots,L-1,\\
    \calQ^{ab}_L(\mathsf{v}) &= \frac{1}{| \calI_{\mathsf{v}}|k_{L-1}} \sum_{i\in \calI_{\mathsf{v}}} (\bW^{(L)a} \bW^{(L)b\intercal})_{ii} ,
\end{align}
where we used a non-calligraphic symbol for $Q_l$ to emphasise that these are just scalars, not functions of readout values. In terms of the order parameters, the recursion can be solved as
\begin{align}
    K^{ab} = \EE_{v\sim P_v} v^2 g^{(L)} \Big(\calQ^{ab}_L(v)g^{(L-1)}\big(Q_{L-1}^{ab} g^{(L-2)}(\cdots Q_{2}^{ab} g^{(1)}(Q_1^{ab}) \cdots ) \big) \Big).
\end{align}
The energetic term follows as before. For the entropic part, we notice that the contributions for each order parameter is factorised,
\begin{align}
    e^{F_S}&=V_{W_L}^{k_L k_{L-1}}(\bm{\calQ}_{L}) \prod_{l=1}^{L-1} V_{W_l}^{k_l k_{l-1}}(\bm{Q}_l),
\end{align}
where, as before,
\begin{align}
    &\lim_{s\to0^+}\lim\frac{1}{ns}
    \ln V_{W_l}^{k_l k_{l-1}}(\bm{Q}_l) = \frac{\gamma_{l}\gamma_{l-1}}{\alpha} 
    {\rm extr}\Big[-\frac{\hat{Q}_l Q_l}{2} +
    \psi_{P_{W}}(\hat{Q}_l)\Big]\\
    &\lim_{s\to0^+}\lim\frac{1}{ns}
    \ln  V_{W_L}^{k_L k_{L-1}}(\bm{\calQ}_{L}) = \frac{\gamma_{L}\gamma_{L-1}}{\alpha} \EE_{v\sim P_v}
    {\rm extr}\Big[-\frac{\hat{\mathcal{Q}}_L(v) \mathcal{Q}_L(v)}{2} +
    \psi_{P_{W}}(\hat{\mathcal{Q}}_L(v))\Big]
\end{align}
Denote by $K^{(L)}(\bar{\calQ})$ the off-diagonal of the matrix $\bK=K^{ab})_{a,b=0}^s$ in the RS ansatz. The free entropy follows:
\begin{align}
    f^{(L)}_{\rm RS}=  \phi_{P_{\rm out}}( K^{(L)}(\bar{\calQ});1)
    +\frac{\gamma_L \gamma_{L-1} }{\alpha} \EE_{v\sim P_v}\big[\psi_{P_{W_L}}(\hat{\calQ}_{L}(v)) -\frac{1}{2} \calQ_{L}(v) \hat{\calQ}_{L}(v) \big]
     +\sum_{l=1}^{L-1}\frac{ \gamma_l \gamma_{l-1}}{\alpha} \big[\psi_{P_{W_l}}\!(\hat{Q}_{l}) -\frac{1}{2} Q_{l} \hat{Q}_{l} \big] .
\end{align}

\subsection{Structured data: quenching the first layer weights}\label{app:structured_data}
In this subsection we show consistency between the computations for the $L=2$ case and the structured data setting. In fact, the latter is equivalent to taking a 2 hidden layer NN where the first activation is $\sigma^{(1)}={\rm Id}$ and the first set of weights are quenched and given to the student. In the notation of the previous section, this implies directly $\calQ_{1}(v^{(2)})=1$ for all $v^{(2)}$'s. Furthermore, since $g^{(1)}=0$ in the definition \eqref{eq:Kd-K-deep} of $K^{(2)}$, $\calQ_2$ disappears from the energetic part, letting entropy win. We thus conclude right away that $\calQ_2(v,v^{(2)})=0$ for all $v, v^{(2)}$'s. This implies in turn that $\tau_v=0$ for all $v$'s. The formula for the free entropy thus simplifies to
\begin{align}
    f^{(2)}_{\rm RS}&:=  \phi_{P_{\rm out}}(K^{(2)}(\bar\calQ);1)+\frac{\gamma_2}{\alpha}\EE_{v\sim P_v}\big[\frac{1}{2}\hat{\calQ}_{2:1}(v)(1-{\calQ}_{2:1}(v))-\iota_v(\hat{\calQ}_{2:1}(v))\big]
\end{align}
The correct way to think of $\iota_v$ is now that of \eqref{eq:iota_v_deep} where the annealed variables $\bV$, mimicking the original $\bW^{(1)}$, are fixed to the ground truth value $\bV^0$, as they are given to the Statistician. In that case, \eqref{eq:RLE_deep_relaxation} reduces to a set of $|\mathcal{I}_v|$ decoupled random linear estimation problems with a Gaussian prior on $\bU_\mathsf{v}$. The latter be integrated via Gaussian integration yielding precisely:
\begin{align}
    \iota_v(x)=\frac{1}{2}\int  \ln(1+xs)\rho_{\rm MP}(s;1/\gamma_1) ds
\end{align}with $\rho_{\rm MP}(s;1/\gamma_1)$ the asymptotic spectral density of the Wishart matrix $\bV^{0\intercal}\bV^0/k_1$, namely a Marchenko-Pastur of parameter $d/k_1=1/\gamma_1$. Note that $\int s\,\rho_{\rm MP}(s;1/\gamma_1) ds=1$. Hence
\begin{align}
    f^{(2)}_{\rm RS}&:=  \phi_{P_{\rm out}}(K^{(2)}(\bar\calQ);1)+\frac{\gamma_2}{2\alpha}\EE_{v\sim P_v}\big[-\hat{\calQ}_{2:1}(v){\calQ}_{2:1}(v)+\int (\hat{\calQ}_{2:1}(v)s-\ln(1+\hat{\calQ}_{2:1}(v)s)d\mu_C(s)\big]\,,
\end{align}the last integral being exactly $\psi_{P_{\mathbf{w}}}(\hat{\calQ}_{2:1}(v))$ in \eqref{eq:psi_P_W_structureddata}.

\section{Details on the numerical procedures} \label{app:numericsdetails}

In this appendix, we detail the implementation of the various numerical experiments involving algorithms such as Hamiltonian Monte Carlo (HMC), Markov Chain Monte Carlo (MCMC) and ADAM. Most of these algorithms were employed through their standard implementations available in \texttt{numpy}, \texttt{tensorflow} and \texttt{pytorch} Python libraries. As already discussed, the GAMP-RIE algorithm introduced in~\cite{maillard2024bayes} and publicly released in~\cite{maillard2024github} was adapted to accommodate generic activation functions and inhomogeneous readouts. The only algorithm implemented entirely from scratch is the MCMC procedure used to sample from the posterior distribution with a Rademacher prior on the inner weights.

\subsection*{Sampling algorithms }
Depending on the setting of each experiment, different algorithms and libraries are used to sample from the posterior distribution (see TABLE~\ref{tab:exp})
\begin{itemize}
    \item \textbf{Markov Chain Monte-Carlo (MCMC)} for Rademacher prior.
    \item \textbf{Hamiltonian Monte Carlo (HMC)} for Gaussian prior. HMC augments the parameter space with auxiliary momenta and simulates Hamiltonian dynamics to propose distant moves with high acceptance probability. HMC is implemented in different Python libraries
    \begin{itemize}
        \item HMC package in \texttt{tensorflow.probability} \cite{tensorflow2015-whitepaper},
        \item No-U-Turn Sampler (NUTS) \cite{hoffman2014nuts} implemented in \texttt{NumPyro} \cite{bingham2019pyro}. This is an advanced version of HMC that  automatically adapts the trajectory length to avoid redundant retracing.
    \end{itemize}
\end{itemize}

Let $\btheta^a_t$ be the parameter sample obtained by running one of these sampling algorithm in $t$ steps . The experimental Bayes-optimal error evaluated at $\btheta_t^a$ is
\begin{align*}
    \varepsilon^{\rm exp}_t = \frac{1}{n_{\rm test}} \sum_{\mu=1}^{n_{\rm test}} \frac{1}{2} [ \lambda( \btheta_t^a, \bx_{\mu}) - \lambda( \btheta^0 , \bx_{\mu}) ]^2
\end{align*}
where $\bx_{\mu}$ are data from a test set of size $10^4$-$10^5$. Important experimental parameters include:
\begin{itemize}
    \item \textbf{Burn-in steps.} During the burn-in period, the sampler is run for a sufficiently large number of steps to reach a stationary state, either from informative or uninformative initialisation. Stationarity can be assessed from the plot of $\varepsilon^{\rm exp}_t$ versus $t$, where it fluctuates around a constant value. This differs from the case where the sampler gets stuck, in which $\varepsilon^{\rm exp}_t$ remains constant after a certain number of steps.
    \item \textbf{Sampling steps.} In the sampling period, the sampler continues to run after the burn-in period. The Bayes-optimal error is computed as the average of $\varepsilon^{\rm exp}_t$ for each value of time step $t$ in this period. This averaging helps reduce the effect of dynamical fluctuations in the trajectory of $\varepsilon^{\rm exp}_t$. Without it, the estimated Bayes-optimal errors obtained from half of Gibbs errors exhibit larger standard deviations. 
\end{itemize}
The following parameters are library-specific:
\begin{itemize}
    \item \textbf{Acceptance rate} for NUTS \texttt{NumPyro}. This parameter was set between 0.6 and 0.7 in all experiments.
    \item \textbf{Tree depth} for NUTs \texttt{NumPyro}. This parameter specifies the maximum number of binary-doubling expansions of the Hamiltonian trajectory, corresponding to a maximum of $2^{\rm depth} - 1$ leapfrog steps per iteration. The depth was chosen between 7 and 8 in all experiments.
    \item \textbf{Initial step size} for HMC \texttt{tensorflow}. This parameter is fixed to $0.01$ in all experiments.
    \item \textbf{Number of adaptation steps} for HMC \texttt{tensorflow}. This parameter is set to be the total number of steps (burn-in steps plus sampling steps). In other words, every HMC step is adaptive, so the initial step size matters little: it will automatically adjust during the HMC trajectory to optimize sampling efficiency.
    \item \textbf{Number of leapfrog steps} for HMC \texttt{tensorflow}. Leapfrog steps control how long HMC simulates Hamiltonian dynamics before making a proposal. This parameter fixed to $10$ in all experiments
\end{itemize}
The following techniques are used are used to reduce finite-dimensional effects:
\begin{itemize}
    \item \textbf{Averaging over sampling steps.} This is discussed in the bullet point on sampling steps above.
    \item \textbf{Reducing readout fluctuations.} For experiments with fixed readouts, as $k$ is typically of the order $10^2$, the empirical readout density can considerably differ from the true one. This finite-size effect increases the variance of the Bayes-optimal error estimate. We reduce this variance as follows. For instance, binary readouts are generated with equal numbers of $1$ and $-1$. The same idea applies to other discrete readouts. For readouts with a continuous density such as Gaussian, we generate many ($10^2$ - $10^4$) readout samples, sort their entries in increasing order, and average over the sorted vectors. This way of generating readouts yields more accurate estimates for Bayes-optimal error with fewer teacher instances.
\end{itemize}

\begin{table}
\centering
\begin{tabular}{|c|c|c|c|c|c|}
\hline
Figure & Tool & Burn-in steps & Sampling steps & No. of instances\\
\hline 
\ref{fig:kernel_vs_NN}, \ref{fig:gen_error_gauss} & NUTS \texttt{NumPyro} & 5000-8000 & 1 & 12 \\
\hline
\ref{fig:gen_errors_univ_spec} & MCMC  & highly varied \footnote{To reduce the computational cost, we use a different number of MCMC steps for each $\alpha$, based on how long it takes to reach the equilibrium; for some values of $\alpha$, this number can be up to 500 times higher than for others. These data are provided in a spreadsheet on the GitHub repository.} & 1/10 no. of burn-in steps & 16 \\
\hline
\ref{fig:v_non-constant} & NUTS \texttt{NumPyro} & 1000-8000 & 1-20 & 12-100 \\
\hline
\ref{fig:gaussian-vector-input}, \ref{fig:mnist-input} & HMC \texttt{tensorflow} & 4000 & 500 &  9 \\
\hline
\ref{fig:3LNNgen_error_gauss} & NUTS \texttt{NumPyro} & 7000-25000 & 500 & 20 \\
\hline
\ref{fig:v_non-constant_2}, \ref{fig:3LNN_NSB_FSB} & NUTS \texttt{NumPyro} & 7000-25000 & 20 & 20 \\
\hline
\ref{fig:local_averages_overlaps} & HMC \texttt{tensorflow} & 50000, 150000, 25000 \footnote{respectively for $\alpha=1.75, 2.75, 3.75$.} & 1 &  100 \\
\hline
\ref{fig:deep-tanh} & HMC \texttt{tensorflow} & 2500-8000 & 500 &  9 \\
\hline
\ref{fig:gen_error_learnable_readouts} & NUTS \texttt{NumPyro} & 1000-8000 & 1-500 &  12-100 \\
\hline
\ref{fig:gen_error_learnable_readouts_L=2} & NUTS \texttt{NumPyro} & 7000-25000 & 20-500 &  20 \\
\hline
\end{tabular}
\caption{Parameters for the experiments.}
\label{tab:exp}
\end{table}

\subsection*{ADAM-based optimisation}

ADAM is a first-order stochastic optimiser that adapts per-parameter learning rates using running estimates of the first and second moments of the gradients. In contrast to HMC, which requires a fully specified probabilistic model and prior, ADAM is a practical optimisation algorithm widely used to train large neural networks. We therefore employ it to estimate the generalisation error achieved by student networks trained with standard optimisation methods on datasets generated by teachers with the same architecture.

In our experiments, we examine the generalisation error of student networks trained with ADAM as a function of the number of gradient updates. Networks are initialised randomly from independent standard-normal draws. Typical optimiser settings are no weight decay, learning rates in the range $10^{-2}-10^{-3}$, large mini-batches (typically between $\lfloor n/4 \rfloor$ and $\lfloor n/8 \rfloor $), and up to $3\times10^{5}$ gradient steps. During optimisation we record the predictive performance (mean squared error) at regular intervals as a function of gradient steps; these test-loss trajectories are averaged across independent teacher runs and reported in \figurename~\ref{fig:hardness_main}, \ref{fig:ADAM_HMC_hardness_L=2} and \ref{fig:hardness_adam_runs}.

\subsection*{Random feature model trained by ridge regression}

We also study student networks trained as random feature models (RFMs), where the student does not learn its hidden weights but instead fixes them at random and trains only a linear readout via ridge regression. In this setting, the student network builds its feature matrix $\mathbf{\Phi}^{\rm RF} = \sigma(\bW^{\rm RF}\bX/\sqrt{d})/\sqrt{\beta k d}$ using randomly drawn standard normal weights $\bW^{\rm RF} \in \mathbb{R}^{\beta kd \times d}$ with $\beta = O(1)$, which are independent of the teacher, and then learns only the readout weights $\ba$ by solving the ridge-regularised least-squares problem $\min_\ba \|\mathbf{\Phi}^{\rm RF} \, \ba - \by\|^2 + t \|\ba\|^2$. In \figurename~\ref{fig:kernel_vs_NN}, we fix $\beta=3$ and sweep over different dataset sizes, drawing independent training and test sets for each realisation. The regularisation strength $t$ is selected through a lightweight validation procedure over a small set of candidate values, while large-scale problems are solved efficiently with standard conjugate gradient procedures.

\end{document}